%% file: pdf.tex
\newcommand\SEKImasterusepackages{\usepackage[T1]{fontenc}}
\newcommand\SEKIusersusepackages
{%
\usepackage{named,makeidx}
\bibliographystyle{named}
\usepackage{url,amssymb}
\usepackage{epic,eepic}
\usepackage{pstricks}
\usepackage{graphicx}
\usepackage{makeidx}
\usepackage{mathdots}
\makeindex
\input headerhot
\input headernamesrest
\input headerforformulas
\input headersugarterms

\input quotation

\input namedhelper

\input hbdictionary

\input headerunits
\usepackage{macros}
\setcounter{secnumdepth}{3}
\setcounter{tocdepth}{2}
\renewcommand\namefont{\sc}
}

\newcommand\SEKIREVIEWSTATUS{1}

\newcommand\SEKIREVIEWER
{{\bundynamenoindex} 
\\University of Edinburgh, 
Centre for Intelligent Systems and their Applications,
School of Informatics,
\\Informatics Forum, 10 Crichton Street, \EB, EH8 9AB, Scotland 
\\E-mail: {\tt  a.bundy@ed.ac.uk}
\\WWW: \url{http://homepages.inf.ed.ac.uk/bundy}
\\\mbox{}\\
{\gramlichnamenoindex} 
\\Vienna University of Technology, Faculty of Informatics, 
  Theory and Logic Group,
\\Favoritenstr. 9, E185/2, A--1040 Wien, Austria
\\E-mail: {\tt gramlich@logic.at}
\\WWW: \url{http://www.logic.at/staff/gramlich}
}

\newcommand\SEKIDOCUMENTCLASS{0}

\newcommand\SEKIYEAR{2013}

\newcommand\SEKINUMBEROFYEAR{02}

\newcommand\SEKITYPE{0}

\newcommand\SEKIPUBLISHEDTYPE
{0}
\title
{Automation of Mathematical Induction\\as part of the History of Logic%
}
\author
{\moorenamenoindex
 \\\Dept\ \CS, Gates Dell C., 2317 Speedway,
 \\\unitexasaustin, Austin, TX 78701 
 \\\url{moore@cs.utexas.edu}
 \\\mbox{}
 \\\wirthnamenoindex
 \\\Institute
 \\\emailcp
}
\date{\small Searchable Online Edition,
\\Submitted \May\,2, 2013,
\\Revision Accepted \May\,31, 2013,
\\Thoroughly \Rev\ and \Extd\ \Jul\,28, 2014
}

\input seki-deckblatt-5

\newif\ifaux
\IfFileExists{\jobname.aux}{\auxtrue}{\auxfalse}%

\newcommand\arXivfootnotemarkref
[1]{\footnotemark[\ifaux\ref{#1}\else 666\fi]}%

\input specialfonts

\begin{document}
\makecover
\maketitle
\tableofcontents
\cleardoublepage
\input body
\end{document}

%% file: headernamesrest.tex
%
%

\newcommand\namefont{}

\usepackage{url}

\hyphenation{
ab-brevi-a-tion
ab-brevi-a-tions
ab-ge-leitete
ab-ge-leitetem
ab-ge-leiteten
ab-ge-leiteter
ab-ge-leitetes
ab-weicht
accord-ing
Acker-mann
actual
actual-ly
Akten-zeichen
aktiv
aktive
aktivem
aktiven
aktiver
aktives
although
always
ana-logous-ly
analysis
an-ge-mie-te-ten
annota-tions
ante-cedent
ante-cedents
antici-pa-tion
appli-ca-tion
appli-ca-tions
area
areas
argu-ment
aspect
asser-tion
asser-tions
assump-tion
assump-tions
Aus-wahl-funktion
Aus-spra-che
author
authors
auto-ma-ted
auto-ma-tion
auto-ma-tisch
auto-ma-ti-sche
auto-ma-ti-schem
auto-ma-ti-schen
auto-ma-ti-scher
auto-ma-ti-sches
Autoren
axio-ma-tics
axiom
Axiome
axioms
basic
basis
be-ein-flusst
be-endet
beharr-lich
being
be-kannte
be-kanntem
be-kannten
be-kannter
be-kanntes
be-rech-tig-te
Bereich
Be-rei-chen
beside
besides
be-zeich-net
Be-zeich-nung
bietet
bis-heri-gen
certain
che-mi-sche
che-mi-schem
che-mi-schen
che-mi-sches
che-mi-scher
com-pati-bi-lity
classi-fi-ca-tion
concept
con-struc-tor
co-rol-lary
counter-example
counter-examples
creati-vity
custom-ary
Dag-stuhl
define
defi-niert
defi-nierte
defi-niertem
defi-nierten
defi-nierter
defi-niertes
defi-nit
defi-nite
defi-nitem
defi-niten
defi-niter
defi-nites
defi-ni-tion
defi-ni-tions
de-monstra-tion
denen
der-arti-ge
der-arti-gen
der-arti-gem
der-arti-ger
der-arti-ges
De-signs
des-cente
descent
deter-mined
deutsche
deutschem
deutschen
deutscher
deutsches
develop-ing
deviate
diese
direc-tion
dis-equality
dis-equali-ties
dis-respected
domain
durch-aus
Effi-zienz
easily
effec-tive
ehren-amt-licher
ehren-amt-liches
ehren-amt-liche
ehren-amt-lichen
ehren-amt-lichem
eigent-lichen
Ein-arbei-tung
eine
einem
einen
einer
eines
either
elek-tro-ni-scher
elek-tro-ni-sches
elek-tro-ni-sche
elek-tro-ni-schen
elek-tro-ni-schem
element
elements
elemen-tary
em-pfohlen
em-pfun-dener
em-pfun-denes
em-pfun-dene
em-pfun-denen
em-pfun-denem
enforce
enforced
enforces
eng-lisch
Engli-scher
Engli-sches
Engli-sche
Engli-schen
Engli-schem
englisch-sprachig
englisch-sprachi-ge
englisch-sprachi-gem
englisch-sprachi-gen
englisch-sprachi-ger
englisch-sprachi-ges
English
entitled
epsi-lon
equi-valent
equi-valence
especial-ly
essay
essays
essen-tial
estab-lish
evaluation
every
exist-ence
existie-ren
existiert
existier-te
existier-ten
explain
explained
ex-plicit-ly
Ent-deckung
ent-schul-di-gen
Ent-wick-lung
Ent-wick-lun-gen
Ent-wurf
etab-lier-ter
etwas
every-where
exami-na-tion
exist
ex-peri-ence
Ex-peri-mente
existen-tial
extra-pola-tion
fach-wissen-schaft-liche
fach-wissen-schaft-lichen
Fach-be-reich
Fahr-zeu-gen
Falsi-fi-ka-tion
father
figure
final
finally
finite
finitis-tic
focus
forma-tion
formula
founda-tions
frequent
frequent-ly
further-more
general
ge-sicher-te
ge-sicher-tem
ge-sicher-ten
ge-sicher-ter
ge-sicher-tes
Ge-werk-schaft
glei-chen
Glei-chung
Glei-chun-gen
geo-metry
Geo-metrie
Gesamt-komplex
ge-sehen
Gestalt
grie-chi-sche
grie-chi-schem
grie-chi-schen
grie-chi-scher
grie-chi-sches
groun-ded-ness
glo-bal-ly 
Grund-lage
Grund-lagen
Grund-lagen-wissen-schaft 
Grund-lagen-wissen-schaf-ten 
guaran-tee-ing
Haltung
haupt-ver-ant-wort-lich
Haupt-ver-ant-wort-liche
Haupt-ver-ant-wort-lichem
Haupt-ver-ant-wort-lichen
Haupt-ver-ant-wort-licher
Haupt-ver-ant-wort-liches
have
Herbrand
hin-rei-chend
hin-rei-chen-de
hin-rei-chen-dem
hin-rei-chen-den
hin-rei-chen-der
hin-rei-chen-des
human
humans
hyper-text
Hyper-texte
hypo-theses
hypo-thesis
hypo-the-size
hypo-the-sizing
idealis-ti-sche
idealis-ti-schem
idealis-ti-schen
idealis-ti-scher
idealis-ti-sches
ideali-za-tion
ihre
ihrem
ihren
ihrer
ihres
indeed
in-definite
index-ed
Induk-tions-be-wei-ser
Induk-tions-ord-nun-gen
Inferenz-systeme
in-finie
in-finite
in-formatics
inner-most
ins-ge-samt
inter-action
inter-connec-tion
inter-connec-tions
interest
inter-pre-ta-tion
intro-duc-tion
intro-duc-tions
irreflexive
Jahres-tref-fen
je-weils
Kai-sers-lau-tern
Kaplan
Kenn-zeichen
kenn-zeich-net
knowledge
knowledge-able
klassi-sche
klassi-schen
klassi-schem
klassi-scher
konnte
Kritik
kriti-sche
kriti-schen
kriti-schem
kriti-scher
kriti-sches
Kumu-la-tion
latter
leistet
lesen
little
logic
logics
logi-sche
logi-schen
logi-schem
logi-scher
logi-sches
magni-tude
mani-fold
mani-folds
mani-pu-la-tion
mathe-ma-tica
mathe-ma-tical
mathe-ma-tician
mathe-ma-ticians
mathe-ma-tics
mathe-ma-ti-cal-ly
Mathe-ma-tik
mathe-ma-ti-sche
mathe-ma-ti-schem
mathe-ma-ti-schen
mathe-ma-ti-scher
mathe-ma-ti-sches
method
methods
metho-di-cal
methodo-logi-sche
methodo-logi-schen
methodo-logi-schem
methodo-logi-scher
methodo-logi-sches
micro-processor
micro-processors
Mie-ter
Modell-klasse
model
models
modern
moder-ne
moder-nem
moder-nen
moder-ner
moder-nes
mono-toni-city
mora-lisch
mora-li-sche
mora-li-schen
mora-li-schem
mora-li-scher
mora-li-sches
micro-pro-cessor
Mieter-sache
Miet-gebrauch
Miet-vertrag
Miet-vertrages
mini-mum
mis-anthropic
mis-under-stood
mis-under-standing
mitt-lere
Musik
Nach-mieterin
natu-ral
Natur-wissen-schaft
Natur-wissen-schaf-ten
Neben-kosten
nega-tiv
nega-tive
neo-logism
never-the-less
Nomina-list
Nomina-listen
notable
nothing
notion
numeral
numer-i-cal
Ober-begriff
object
objects
Objekt
occur
occur-rence
occur-ren-ces
ohne
onto-logisch
onto-logi-sche
onto-logi-schem
onto-logi-schen
onto-logi-scher
onto-logi-sches
operationali-zation
order
Origi-nal
Origi-nals
Origi-nal-zitate
organi-sa-to-risch
organi-sa-to-ri-sche
organi-sa-to-ri-schem
organi-sa-to-ri-schen
organi-sa-to-ri-scher
organi-sa-to-ri-sches
other-wise
para-digm
Para-digma
Para-digmas
para-digms
Parallelen-axiom
para-meter
para-meters
parti-ci-pa-tion
philo-so-phisch
philo-so-phi-sche
philo-so-phi-schem
philo-so-phi-schen
philo-so-phi-scher
philo-so-phi-sches
philo-sophy
parallel
partial-i-ty
poly-nomial
poly-nomials
posi-tiv
posi-tive
post-humously
product
pro-posi-tion
pro-posi-tions
pro-posi-tio-nal
ratio-nalis-mus
ratio-nalis-tisch 
ratio-nalis-tische 
ratio-nalis-tischem 
ratio-nalis-tischen 
ratio-nalis-tischer 
ratio-nalis-tisches 
really
Real-wissen-schaf-ten
recht-eckige
recht-eckigem
recht-eckigen
recht-eckiger
recht-eckiges
re-cog-nition
re-duci-bi-lity
Re-duk-tions-re-la-tion
regard-less
Regel
relation
relation-ship
relation-ships
re-organi-za-tion
repeti-tion
replace
represent
represents
represen-ted
represen-ting
repre-sen-ta-tion
require
requires
reso-lu-tion
restric-tive
result
Rhein-steig
Sammel-bezeich-nung 
schrift-lich
schwerere
Schwierig-keiten
second
seiten-gleich
seiten-gleiche
seiten-gleichem
seiten-gleichen
seiten-gleiches
seiten-gleicher
seman-tic
seman-tics
separa-tion
sequence
sequen-ces
several
sinn-voller
sitt-lich
sitt-liche
sitt-lichem
sitt-lichen
sitt-liches
sitt-licher
speaking
solu-tion
solu-tions
sophis-ti-ca-tion
Spezi-fi-ka-tio-nen
Spezi-fi-ka-tions-sprache
Sko-lem-i-za-tion
simpli-fi-cation
speci-fi-ca-tion
speci-fi-ca-tions
speci-fier
speci-fied
stable
sub-sequent
sym-metrical
sym-metrical-ly
tauto-logy
taking
tertium
theorem
theorems
Theorie
Theorien
theory
theo-re-ti-sche
theo-re-ti-schen
theo-re-ti-schem
theo-re-ti-sches
theo-re-ti-scher
thesis
topo-logy
total
trans-cend
typischer-weise
un-answer-ed
un-bestimmt
un-bestimmte
un-bestimm-tem
un-bestimm-ten
un-bestimm-ter
un-bestimm-tes
under
under-stand
under-stands
under-stood
unless
Un-par-tei-lich-keit
unter-bewusst
unter-schied
until
ur-element
ur-elements
Ur-sache
usage
valid-i-ty
values
ver-an-lasst
Ver-an-stal-tung
Ver-bin-dung
ver-brei-ter-tem
Ver-mieters
Ver-mie-terin
ver-pflich-te-te
viel-leicht
Vien-na
Vor-annahme
Vor-annahmen
vor-kom-men-den
vor-para-dig-ma-ti-schen
vor-para-dig-ma-ti-scher
Vor-schrif-ten
wahr-ge-nom-men
weder
wei-te-ren
Weni-ger-ver-brauchs
wesent-liche
wesent-lichem
wesent-lichen
wesent-liches
wesent-licher
Win-kel
Wissen-schaft
Wissen-schaften
Wissen-schaft-ler
Wissen-schaft-lern
wissen-schaft-liche
wissen-schaft-lichen
wissen-schaft-licher
wissen-schaft-liches
Wissen-schafts-ent-wick-lung
wurde
zahl-reiche
zahl-reichen
zahl-reichem
zahl-reicher
zahl-reiches
zu-neh-men-den
zu-gleich
Zeich-nun-gen
zwei-sprachi-ge
zwei-sprachi-gem
zwei-sprachi-gen
zwei-sprachi-ger
zwei-sprachi-ges
Zweit-woh-nungs-steu-er
zwischen
}

\newcommand\majorfootroom{\raisebox{-1.9ex}{\rule{0ex}{.5ex}}}

\newcommand\footroom{\raisebox{-1.5ex}{\rule{0ex}{.5ex}}}
\newcommand\mediumfootroom{\raisebox{-1.25ex}{\rule{0ex}{3ex}}}
\newcommand\smallfootroom{\raisebox{-1.0ex}{\rule{0ex}{3ex}}}
\newcommand\tinyfootroom{\raisebox{-.5ex}{\rule{0ex}{1ex}}}

\newcommand\majorheadroom{\rule{0ex}{3.2ex}}
\newcommand\headroom{\rule{0ex}{2.8ex}}
\newcommand\mediumheadroom{\rule{0ex}{2.4ex}}
\newcommand\smallheadroom{\rule{0ex}{2.0ex}}



\newcommand\alan     {Alan}

\newcommand\andrew   {An\-drew}
\newcommand\anne     {Anne}
\newcommand\bernd    {Bernd}
\newcommand\bernard  {Bernard}
\newcommand\bernhard {Bern\-hard}

\newcommand\catherine{{\namefont Catherine}}
\newcommand\charles  {Charle\es}

\newcommand\christoph{Christoph}

\newcommand\claus    {Clau\es}
\newcommand\david    {{\namefont David}}
\newcommand\donald   {Donald}
\newcommand\dieter   {Dieter}

\newcommand\emmy     {{\namefont Emmy}}

\newcommand\frank    {Frank}

\newcommand\john     {John}

\newcommand\klaus    {Klau\es}

\newcommand\maxchristiannamenonamefont{Max} 
\newcommand\maxchristianname{{\namefont\maxchristiannamenonamefont}}

\newcommand\martin   {{\namefont Mar\-tin}}
\newcommand\marvin   {{\namefont Mar\-vin}}


\newcommand\peter    {Peter}
\newcommand\pierre   {Pierre}
\newcommand\raymond  {Raymond}

\newcommand\richard  {Ri\-chard}
\newcommand\robert   {Ro\-bert}
\newcommand\robin    {Robin}

\newcommand\tobias   {{\namefont To\-bi\-a\es}}

\newcommand\ulrich   {Ul\-rich}

\newcommand\warren   {Warren}

\newcommand\wilhelm  {\mbox{Wilhelm}}


\newcommand\acerbiindex     {\index{Acerbi, Fabio}}
\newcommand\acerbi          {{\namefont Acerbi}}
\newcommand\acerbiname      {\acerbiindex{\namefont Fabio \acerbi}}
\newcommand\ackermannindex  {\index{Ackermann, Wilhelm (1896--1962)}}

\newcommand\ackermann
                   {{\namefont A\nolinebreak\hskip-.07em\nolinebreak cker\-mann}}
\newcommand\ackermannname   {\ackermannindex{\namefont \wilhelm\ \ackermann}}
\newcommand\ackermannlifetime{(1896--1962)}
\newcommand\ackermannfunction{\index{Ackermann function}\ackermann\ function}

\newcommand\aristotleindex  {\index{Aristotle (384--322\,\BC)}}
\newcommand\aristotle       {{\namefont Ari\esi totle}}

\newcommand\aristotelian    {{\namefont Ari\esi totelian}}
\newcommand\aristotelianlogicindex{\index{Aristotelian logic}}
\newcommand\aristotelianlogic{\aristotelianlogicindex\aristotelian\ logic}

\newcommand\aubinindex      {\index{Aubin, Raymond}}

\newcommand\aubin           {\mbox{\namefont Aubin}}
\newcommand\aubinname       {\aubinindex{\namefont\raymond\ \aubin}}

\newcommand\barnerindex     {\index{Barner, Klaus}}

\newcommand\barner          {{\namefont Barner}}
\newcommand\barnername      {\barnerindex{\namefont\klaus\ \barner}}

\newcommand\basinindex      {\index{Basin, David}}

\newcommand\basin           {{\namefont Basin}}
\newcommand\basinname       {\basinindex{\namefont\david\ \basin}}

\newcommand\bendix          {Bendix}


\newcommand\bledsoeindex    {\index{Bledsoe, W. W. (1921--1995)}} 
\newcommand\bledsoe         {{\namefont Bledsoe}}
\newcommand\bledsoename     {\bledsoeindex{\namefont W.\,W.\,\,\bledsoe}}
\newcommand\bledsoelifetime {(1921--1995)}

\newcommand\boole           {{\namefont Boole}}

\newcommand\myboolean       {\boole an}
\newcommand\myBoolean       {\boole an}

\newcommand\boyerindex      {\index{Boyer, Robert S. (*1946)}}
\newcommand\boyer           {{\namefont Boyer}}
\newcommand\boyername       {\boyerindex{\namefont\robert\ S. \boyer}}
\newcommand\boyerswifename  {{\namefont\anne\ O. \boyer}}
\newcommand\boyerlifetime   {(*1946)}
\newcommand\boyermoore      {\boyerindex\mooreindex\boyer--\moore}
\newcommand\boyermooretheoremproversindex{\index{Boyer--Moore theorem provers}}
\newcommand\boyermooretheoremprover
                      {\boyermooretheoremproversindex\boyermoore\ theorem prover}
\newcommand\boyermooreTheoremProver
                      {\boyermooretheoremproversindex\boyermoore\ Theorem Prover}
\newcommand\boyermooretheoremprovers{\boyermooretheoremprover s}
\newcommand\boyermoorewaterfallindex{\index{Boyer--Moore waterfall}}
\newcommand\boyermoorewaterfall{\boyermoorewaterfallindex\boyermoore\ waterfall}
\newcommand\boyermooreWaterfall{\boyermoorewaterfallindex\boyermoore\ Waterfall}
\newcommand\boyermooremachines
                            {\index{Boyer--Moore machines}\boyermoore\ machines}




\newcommand\bundyindex      {\index{Bundy, Alan (*1947)}}

\newcommand\bundy           {{\namefont Bundy}}
\newcommand\bundynamenoindex{{\namefont\alan\ \bundy}}
\newcommand\bundyname       {\bundyindex\bundynamenoindex}
\newcommand\bundylifetime   {(*1947)}

\newcommand\burstallindex   {\index{Burstall, Rod M. (*1934)}}

\newcommand\burstall        {{\namefont Burstall}}
\newcommand\burstallname    {\burstallindex{\namefont Rod M. \burstall}}
\newcommand\burstalllifetime{(*1934)}

\newcommand\church          {{\namefont Church}}

\newcommand\churchrosser    {\church--\rosser}
\newcommand\churchrosserproperty{\index
                                {Church--Rosser property}\churchrosser\ property}
\newcommand\churchrossertheorem{\index
                                {Church--Rosser Theorem}\churchrosser\ Theorem}

\newcommand\dedekindindex   {\index{Dedekind, Richard (1831--1916)}}
\newcommand\dedekind        {{\namefont Dedekind}}
\newcommand\dedekindname    {\dedekindindex{\namefont\richard\ \dedekind}}
\newcommand\dedekindlifetime{(1831--1916)}




\newcommand\euclidindex     {\index{Euclid}}
\newcommand\euclid          {{\namefont Euclid}}

\newcommand\euclidname      {\euclidindex\euclid}


\newcommand\fermatindex     {\index{Fermat, Pierre (160?--1665)}}
\newcommand\fermat          {\mbox{\namefont Fermat}}
\newcommand\fermatname      {\fermatindex{\namefont \pierre\ \fermat}}

\newcommand\fermatbirthyear 
{160?}
\newcommand\fermatdeathyear {1665}
\newcommand\fermatlifetime  {(\fermatbirthyear--\fermatdeathyear)}







\newcommand\michaelgabbayindex
{}

\newcommand\gentzen         {{\namefont Gentzen}}

\newcommand\gersonindex     {\index{Gerson, Levi ben (1288--1344)}}
\newcommand\gerson          {{\namefont Gerson}}
\newcommand\gersonname      {\gersonindex{\namefont Levi ben \gerson}}
\newcommand\gersonlifetime  {(1288--1344)}

\newcommand\goedel          {{\namefont G\oe del}}

\newcommand\secondincompletenesstheorem
{second \incompletenesstheorem}
\newcommand\secondIncompletenessTheorem
{Second \IncompletenessTheorem}
\newcommand\goedelsfirstincompletenesstheorem
{\goedel'\es\ \firstincompletenesstheorem}

\newcommand\firstincompletenesstheorem
{first \incompletenesstheorem}
\newcommand\firstIncompletenessTheorem
{First \IncompletenessTheorem}
\newcommand\incompletenesstheorem{incompleteness theorem}
\newcommand\IncompletenessTheorem{Incompleteness Theorem}

\newcommand\goldsteinindex  {\index{Goldstein, Catherine (*1958)}}

\newcommand\goldstein       {{\namefont Gold\-stein}}
\newcommand\goldsteinname   {\goldsteinindex{\namefont\catherine\ \goldstein}}

\newcommand\gordonindex     {\index{Gordon, Mike J. C. (*1948)}}

\newcommand\gordon          {{\namefont Gordon}}
\newcommand\gordonname      {\gordonindex{\namefont Mike J. C. \gordon}}
\newcommand\gordonlifetime  {(*1948)}

\newcommand\gramlichindex           {\index{Gramlich, Bernhard (1959--2014)}}

\newcommand\gramlich        {{\namefont Gram\-lich}}
\newcommand\gramlichnamenoindex{{\namefont\bernhard\ \gramlich}}
\newcommand\gramlichname    {\gramlichindex\gramlichnamenoindex}
\newcommand\gramlichlifetime{(1959--2014)}


\newcommand\harmelenindex   {\index{Harmelen, Frank van (*1960)}}

\newcommand\harmelen        {{\namefont Harmelen}}
\newcommand\harmelenname    {\harmelenindex{\namefont\frank\ van \harmelen}}

\newcommand\hayesindex      {\index{Hayes, Pat(rick) J. (*1944)}}

\newcommand\hayes           {{\namefont Hayes}}
\newcommand\hayesname       {\hayesindex{\namefont Pat \hayes}}
\newcommand\hayeslifetime   {(*1944)}









\newcommand\hilbert         {\mbox{\namefont Hilbert}}

\newcommand\hilbertsepsilonlongindex{\index{Hilbert's epsilon}}

\newcommand\hilbertsepsilon
                     {\hilbertsepsilonlongindex\hilbert'\es\ \nlbmath\varepsilon}

\newcommand\hippasosindex   {\index{Hippasus of Metapontum (\ca\,550\,\BC)}}
\newcommand\hippasos        {{\namefont Hippasu\es}}
\newcommand\hippasosname    {\hippasosindex{\namefont\hippasos\ of Metapontum}}
\newcommand\hippasoslifetime{(\ca\,550\,\BC)}

\newcommand\huntindex       {\index{Hunt, Warren A.}}

\newcommand\hunt            {{\namefont Hunt}}
\newcommand\huntname        {\huntindex{\namefont\warren~A. \hunt}}

\newcommand\hutterindex     {\index{Hutter, Dieter (*1959)}}

\newcommand\hutter          {{\namefont Hutter}}
\newcommand\huttername      {\hutterindex{\namefont\dieter\ \hutter}}
\newcommand\hutterlifetime  {(*1959)}
\newcommand\huygensindex    {\index{Huygens, Christiaan (1629--1695)}}
\newcommand\huygens         {{\namefont Huygen\es}}
\newcommand\huygensname     {\huygensindex{\namefont Christiaan \huygens}}
\newcommand\huygenslifetime {(1629--1695)}
\newcommand\irelandindex    {\index{Ireland, Andrew}}

\newcommand\ireland         {{\namefont Ireland}}
\newcommand\irelandname     {\irelandindex{\namefont\andrew\ \ireland}}

\newcommand\kantindex                 {\index{Kant, Immanuel (1724--1804)}}

\newcommand\kant            {{\namefont Kant}}

\newcommand\kaufmannindex   {\index{Kaufmann, Matt (*1952)}}

\newcommand\kaufmann        {{\namefont Kaufmann}}
\newcommand\kaufmannname    {\kaufmannindex{\namefont Matt \kaufmann}}
\newcommand\kaufmannlifetime{(*1952)}


\newcommand\knuth           {{\namefont Knuth}}


\newcommand\kowalskiindex   {\index{Kowalski, Robert A. (*1941)}}

\newcommand\kowalski        {{\namefont Kowalski}}
\newcommand\kowalskiname    {\kowalskiindex{\namefont \robert\ A. \kowalski}}
\newcommand\kowalskilifetime{(*1941)}

\newcommand\kuehlerindex    {\index
                            {Kuehler, Ulrich (*1964)@{K\"uhler, Ulrich (*1964)}}}

\newcommand\kuehler         {{\namefont K\ue h\-ler}}
\newcommand\kuehlername     {\kuehlerindex{\namefont\ulrich\ \kuehler}}
\newcommand\kuehlerlifetime {(*1964)}




\newcommand\loechnerindex   {\index
                           {Loechner, Bernd (*1967)@{L\"ochner, Bernd (*1967)}}}

\newcommand\loechner        {{\namefont L\oe chner}}
\newcommand\loechnername    {\loechnerindex{\namefont\bernd\ \loechner}}
\newcommand\loechnerlifetime{(*1967)}

\newcommand\loef            {{\namefont\martin-L\oe f}}

\newcommand\madlener        {{\namefont Mad\-lener}}
\newcommand\madlenername    {{\namefont\klaus\ \madlener}}

\newcommand\maurolycusindex {\index{Maurolico, Francesco (1494--1575)}} 
\newcommand\maurolycus      {{\namefont Maurolico}}

\newcommand\maurolycusname  {\maurolycusindex{\namefont Francesco \maurolycus}}
\newcommand\maurolycuslifeplace{Messina}
\newcommand\maurolycuslifetime{(1494--1575)}
\newcommand\mccarthyindex   {\index{McCarthy, John (1927--2011)}}
\newcommand\mccarthy        {\mbox{\namefont McCarthy}}
\newcommand\mccarthyname    {\mccarthyindex{\namefont\john\ \mccarthy}}

\newcommand\meltzerindex    {\index{Meltzer, Bernard (1916?--2008)}}

\newcommand\meltzer         {{\namefont Meltzer}}
\newcommand\meltzername     {\meltzerindex{\namefont \bernard\ \meltzer}}
\newcommand\meltzerlifetime {(1916?--2008)}

\newcommand\michieindex     {\index{Michie, Donald (1923--2007)}}

\newcommand\michie          {{\namefont Michie}}
\newcommand\michiename      {\michieindex{\namefont\donald\ \michie}}
\newcommand\michielifetime  {(1923--2007)}

\newcommand\milnerindex     {\index{Milner, Robin (1934--2010)}}
\newcommand\milner          {{\namefont Milner}}
\newcommand\milnername      {\milnerindex{\namefont\robin\ \milner}} 
\newcommand\milnerlifetime  {(1934--2010)}
\newcommand\minsky          {{\namefont Minsky}}
\newcommand\minskyname      {{\namefont\marvin\ \minsky}}



\newcommand\mooreindex      {\index{Moore, J Strother (*1947)}}
\newcommand\moore           {{\namefont Moore}}
\newcommand\moorenamenoindex{{\namefont J~Strother \moore}}
\newcommand\moorename       {\mooreindex\moorenamenoindex}
\newcommand\moorelifetime   {(*1947)}
\newcommand\mooreswifename  {{\namefont Jo \moore}}



\newcommand\newmanindex     {\index{Newman, Max(well) H. A. (1897--1984)}}

\newcommand\newman          {{\namefont Newman}}
\newcommand\newmanname      {\newmanindex{\namefont Max~H.~A. \newman}}
\newcommand\newmanlemma     {\index{Newman Lemma}\newman\ Lemma}

\newcommand\noetherindex                 {\index{Noether, Emmy (1882--1935)}}

\newcommand\noether         {{\namefont Noether}}
\newcommand\noethername     {\noetherindex{\namefont    \emmy\ \noether}}
\newcommand\noetherfathername{\maxchristianname\ \noether}
\newcommand\noetherbirthyear{1882}
\newcommand\noetherdeathyear{1935}
\newcommand\noetherlifetime {(\noetherbirthyear--\noetherdeathyear)}
\newcommand\noetherfatherlifetime{(1844--1921)}
\newcommand\noetherian      {\noether\-ian}
\newcommand\theoremofnoetherianinduction{\index
{induction!Noetherian}Theorem of\/ \noetherian\ Induction}
\newcommand\structuralinductionindex{\index{induction!structural}}
\newcommand\axiomofstructuralinductionfirsthalf{Axiom}
\newcommand\axiomofstructuralinductionsecondhalf
                            {\structuralinductionindex of Structural Induction}
\newcommand\axiomofstructuralinduction{\axiomofstructuralinductionfirsthalf\ 
\axiomofstructuralinductionsecondhalf}
\newcommand\axiomofchoice   {\index{choice!Axiom of Choice}Axiom of Choice}

\newcommand\padawitz        {{\namefont Pada\-witz}}
\newcommand\padawitzname    {{\namefont\peter\ \padawitz}}

\newcommand\pascalindex     {\index{Pascal, Blaise (1623--1662)}}
\newcommand\pascal          {{\namefont Pascal}}
\newcommand\pascalname      {\pascalindex{\namefont Blaise \pascal}}
\newcommand\pascalbirthyear {1623}
\newcommand\pascaldeathyear {1662}
\newcommand\pascallifetime  {(\pascalbirthyear--\pascaldeathyear)}

\newcommand\peanoindex      {\index{Peano, Guiseppe (1858--1932)}}
\newcommand\peanoaxiomsindex{\index{Peano axioms}}
\newcommand\peano           {{\namefont Peano}}
\newcommand\peanoname       {\peanoindex{\namefont Guiseppe \peano}}
\newcommand\peanolifetime   {(1858--1932)}
\newcommand\peanoaxiom      {\peanoaxiomsindex\peano\ axiom}


\newcommand\peterindex      {\index{P\'eter, R\'osza (1905--1977)}}

\newcommand\peterrosza      {{\namefont P\'eter}}
\newcommand\petername       {\peterindex{\namefont R\'osza \peterrosza}}
\newcommand\peterlifetime   {(1905--1977)}

\newcommand\pieriindex      {\index{Pieri, Mario (1860--1913)}}
\newcommand\pieri           {{\namefont Pieri}}
\newcommand\pieriname       {\pieriindex{\namefont Mario \pieri}}
\newcommand\pierilifetime   {(1860--1913)}


\newcommand\platoindex      {\index{Plato (427--347\,\BC)}}
\newcommand\plato           {{\namefont Plato}}
\newcommand\platoname       {\platoindex\plato} 
\newcommand\platolifetime   {(427--347\,\BC)}
\newcommand\plotkin         {{\namefont Plotkin}}
\newcommand\plotkinname     {{\namefont Gordon \plotkin}}
\newcommand\plotkinlifetime {(*1946)}

\newcommand\popplestone     {{\namefont Popplestone}}
\newcommand\popplestonename {{\namefont\robin\ J. \popplestone}}
\newcommand\popplestonelifetime{(1938--2004)}


\newcommand\presburgerindex {\index{Presburger, Moj{\.z}es{}z (1904--1943?)}}

\newcommand\presburger      {{\namefont Pre\esi burger}} 
\newcommand\presburgername  
               {\presburgerindex{\namefont\mbox{Moj\zwithdot es{}z} \presburger}}
\newcommand\presburgerlifetime{(1904--1943?)}
\newcommand\presburgerarithmetic{\presburger\ Arithmetic}
\newcommand\presburgertruename{Prez\-burger}

\newcommand\protzenindex    {\index{Protzen, Martin (*1962)}}

\newcommand\protzen         {{\namefont Protzen}}
\newcommand\protzenname     {\protzenindex{\namefont\martin\ \protzen}}
\newcommand\protzenlifetime {(*1962)}



\newcommand\rezaei          {{\namefont Rezaei}}
\newcommand\rezaeinamenoindex{{\namefont Marianeh \rezaei}}


\newcommand\robinsonindex   {\index{Robinson, J. Alan (*1930?)}}

\newcommand\robinson        {{\namefont Robinson}}
\newcommand\robinsonname    {\robinsonindex{\namefont J.\hskip.25em\alan\ \robinson}}
\newcommand\robinsonlifetime{(*1930?)}

\newcommand\rosser          {{\namefont Rosser}}


\newcommand\russellsparadoxindex{\index{Russell's Paradox}}
\newcommand\russell         {{\namefont Russell}}


\newcommand\russellsparadox {\russellsparadoxindex\russell'\es\ Paradox}

\newcommand\samoaindex      {\index{Schmidt-Samoa, Tobias (*1973)}}

\newcommand\samoa           {{\namefont Schmidt-Sa\-moa}}
\newcommand\samoaname       {\samoaindex{\namefont\tobias\ \samoa}}
\newcommand\samoalifetime   {(*1973)}

\newcommand\scottindex      {\index{Scott, Dana S. (*1932)}}

\newcommand\scott           {{\namefont Scott}}
\newcommand\scottname       {\scottindex{\namefont Dana S. \scott}}

\newcommand\shankar         {{\namefont Shankar}}


\newcommand\simonyiindex    {\index{Simonyi, Charles}}
\newcommand\simonyi         {{\namefont Simonyi}}
\newcommand\simonyiname     {\simonyiindex{\namefont\charles\ \simonyi}}
\newcommand\simonyilifetime {(*1948)}

\newcommand\skolem          {{\namefont Skolem}}

\newcommand\skolemization   {\skolem\-ization}


\newcommand\tait            {{\namefont Tait}}


\newcommand\waltherindex    {\index{Walther, Christoph (*1950)}}

\newcommand\walther         {{\namefont Walther}}
\newcommand\walthername     {\waltherindex{\namefont\christoph\ \walther}} 
\newcommand\waltherlifetime {(*1950)}



\newcommand\principleofdependentchoice{\principleofdependentchoiceindex\
                                                  Principle of Dependent Choice}
\newcommand\principleofdependentchoiceindex{\index
{choice!Principle of Dependent Choice}}

\newcommand\wirthindex                 {\index{Wirth, Claus-Peter (*1963)}}

\newcommand\wirth           {{\namefont Wirth}}
\newcommand\wirthnamenoindex{{\namefont\claus-\peter\ \wirth}}
\newcommand\wirthname       {\wirthindex\wirthnamenoindex}
\newcommand\wirthlifetime   {(*1963)}












\newcommand\ca   {ca.}

\newcommand\FB   {FB}

\newcommand\FBautinfveryshort{\FB\ AI}

\newcommand\f    {\mbox{}{f.}}   
\newcommand\ff   {\mbox{}{ff.}}  


\newcommand\eMAIL{e-mail}
\newcommand\ifandonlyif{if \nolinebreak and \onlyif}
\newcommand\Int  {Int.}
\newcommand\Inc  {Inc.}

\newcommand\maslong{mathematics assistance system}

\newcommand\onlyif{only \nolinebreak if}


\newcommand\qedabbrev{q.e.d.}

\newcommand\getittotheright[1]  
{\hfill\mbox{}\penalty 100\mbox{\ \,}\nolinebreak
\hfill\hfill\hfill\nolinebreak\mbox{#1}\ignorespaces}

\newcommand\role{r\^ole}


\newcommand\Univ {Univ.}
\newcommand\Vol  {Vol.}
\newcommand\WWW  {WWW}


\newcommand\AI   {Artificial Intelligence}

\newcommand\aswell{as \nolinebreak well}
\newcommand\aswellas{\aswell\ \nolinebreak as}
\newcommand\BC   {{\sc b.c.}} 

\newcommand\Cf   {Cf.}
\newcommand\cf   {cf.}
\def\nlbcite{\nolinebreak\cite}
\newcommand\Cfnlb{\Cf\nolinebreak}
\newcommand\cfnlb{\cf\nolinebreak}

\newcommand\Conf {Conf.}

\newcommand\CS   {Computer \Sci}

\newcommand\Dec  {Dec.}
\newcommand\Dept {Dept.}
\newcommand\descenteinfinieindex{\index{descente infinie}}
\newcommand\descenteinfinienoindex{descente infinie}
\newcommand\descenteinfinie{\descenteinfinieindex\descenteinfinienoindex}

\newcommand\DescenteInfinie{\descenteinfinieindex Descente Infinie}

\newcommand\theinnermostpartofW
{\math{\forall u\tight<v\stopq\app P  u}}

\newcommand\edn  {edn.}

\newcommand\eg   {e.g.}

\newcommand\etc  {\&c.}

\newcommand\Extd {Extd.}
\newcommand\reductioadabsurdum{{\it argumentum ad absurdum}}
\newcommand\areductioadabsurdum{an \reductioadabsurdum}

\newcommand\firstorder{first-order}

\newcommand\ie   {i.e.}

\newcommand\incl {incl.}

\newcommand\Jul  {July}

\newcommand\May  {May}


\def\note{Note}
\newcommand\notes{Notes}
\newcommand\Oct  {Oct.}
\newcommand\p    {p.}
\newcommand\pp   {pp.}
\newcommand\PP[2]{\pp\,\ignorespaces#1--\ignorespaces#2}

\newcommand\PhD  {PhD}
\newcommand\PhDthesis{\PhD\ thesis}

\newcommand\pnc  {posi\-tive/ne\-ga\-tive-condi\-tional}

\newcommand\rev  {rev.}
\newcommand\Rev  {Rev.}

\newcommand\sect {\myparagraphsymbol} 
\newcommand\sects{\myparagraphsymbol\myparagraphsymbol}

\newcommand\Sci  {Sci.}



\newcommand\wellfounded{well-founded}

\newcommand\wellfoundedness{well-founded\-ness}
\newcommand\Wellfoundedness{Well-founded\-ness}
\newcommand\WellFoundedness{Well-Founded\-ness}

\newcommand\wrog {w.l.o.g.} 
\newcommand\wrt  {w.r.t.}





\newcommand\thewordand{and}
\newcommand\litspageref[1]{Page\,#1}

\newcommand\litspagefromtoref[2]{Pages \nolinebreak #1--#2}

\newcommand\litnoteref[1]{\note\,#1}

\newcommand\itemname{item}

\newcommand\lititemref[1]{\itemname\,#1}
\newcommand\litfiguref[1]{Figure\,#1}

\newcommand\litsectref[1]{\sect\,#1} 

\newcommand\litchapref[1]{Chapter\,#1} 

\newcommand\litchapfromtoref[2]{Chapters\,#1--#2}

\newcommand\Examplename{Ex\-am\-ple}
\newcommand\litexamref[1]{\Examplename\,#1}

\newcommand\litlemmref[1]{Lem\-ma\,#1}

\newcommand\litcororef[1]{Co\-rollary\,#1}
\newcommand\litpropref[1]{Pro\-po\-si\-tion\,#1}

\newcommand\litexamrefs[2]{Examples~#1~\thewordand~#2}

\newcommand\litnoterefs[2]
{\notes\         \nolinebreak #1 \thewordand\ \nolinebreak #2}
\newcommand\litsectrefs[2]
{\sects\         \nolinebreak #1 \thewordand\ \nolinebreak #2}

\newcommand\litexamrefss
[3]{Examples \nolinebreak #1, #2, \thewordand\ \nolinebreak #3}

\newcommand\litnoterefss
[3]{\notes\ \nolinebreak #1, #2, \thewordand\ \nolinebreak #3}

\newcommand\lemmref[1]{\litlemmref{\ref{#1}}}

\newcommand\examref[1]{\litexamref{\ref{#1}}}

\newcommand\figuref[1]{\litfiguref{\ref{#1}}}
\newcommand\noteref[1]{\litnoteref{\ref{#1}}}
\newcommand\sectref[1]{\litsectref{\ref{#1}}}
\newcommand\nlbsectref[1]{\nolinebreak\sectref{#1}}

\newcommand\examrefs[2]{\litexamrefs{\ref{#1}}{\ref{#2}}}

\newcommand\noterefs[2]{\litnoterefs{\ref{#1}}{\ref{#2}}}
\newcommand\sectrefs[2]{\litsectrefs{\ref{#1}}{\ref{#2}}}

\newcommand\examrefss[3]{\litexamrefss{\ref{#1}}{\ref{#2}}{\ref{#3}}}

\newcommand\noterefss[3]{\litnoterefss{\ref{#1}}{\ref{#2}}{\ref{#3}}}

\newcommand\nthpositioner[2]
{#1\raisebox{0.52ex}{\scriptsize\hspace{0.07em}#2}}
\newcommand\nth[1]{\nthtinypositioner{#1}{\nthstring{#1}}}
\newcommand\nthtinypositioner[2]{#1\raisebox{0.52ex}{\tiny\hspace{0.07em}#2}}

\newcommand\mthpositioner[2]
{\math{#1}\raisebox{0.52ex}{\scriptsize\hspace{0.07em}#2}}
\newcommand\modulointocountzero[2]
{\count1=#1
\count2=#2
\count0=\count1
\divide  \count0 by \count2
\multiply\count0 by-\count2
\advance \count0 by \count1}
\newcommand\absolutevalueintocountzero[1]
{\count0=#1
\ifnum\count0<0\multiply\count0 by -1\fi}
\newcommand\nthstring[1]
{\def\myargone{#1}\ifcat a\myargone th\else\nthstringnochar{#1}\fi}
\newcommand\nthstringnochar[1]
{\absolutevalueintocountzero{#1}%
\modulointocountzero{\count0}{100} 
\ifnum\count0>9\ifnum\count0<20 th\else\stupidnthstring\fi
                                  \else\stupidnthstring\fi}
\newcommand\stupidnthstring
{\modulointocountzero{\count0}{10}
\ifnum\count0=1 \hskip-0.2em st\else
\ifnum\count0=2 nd\else
\ifnum\count0=3 rd\else 
                th\fi\fi\fi}

\newcommand\writeascents
[1]{\count4=#1
\ifnum\count4<0 
-\multiply\count4 by -1\fi
\modulointocountzero{\count4}{10}
\divide\count4 by 10
\count3=\the\count0
\modulointocountzero{\count4}{10}
\divide\count4 by 10
\the\count4
.\the\count0
\the\count3
}

\newcommand\frenchnthstring[1]
{\def\myargone{#1}\ifcat a\myargone th\else\frenchnthstringnochar{#1}\fi}
\newcommand\frenchnthstringnochar[1]
{\absolutevalueintocountzero{#1}%
\modulointocountzero{\count0}{100} 
\ifnum\count0>9\ifnum\count0<20 th\else\frenchstupidnthstring\fi
                                  \else\frenchstupidnthstring\fi}
\newcommand\frenchstupidnthstring
{\modulointocountzero{\count0}{10}
\ifnum\count0=1 \hskip-0.2em re\else
\ifnum\count0=2 me\else
\ifnum\count0=3 rd\else 
                th\fi\fi\fi}


\newcommand\ACLTWO    {\index{ACL2@{\sc ACL2}}{\sc ACL2}}

\newcommand\BAROQUE   {\index{Baroque@{\sc Baroque}}{\sc Baroque}}

\newcommand\CLAM      {{\rm CL\kern-.36em\raise.39ex\hbox{\sc a}\kern-.15emM}}
\newcommand\COQ       {\index{Coq@{\sc Coq}}{\sc Coq}}

\newcommand\INKA      {\index{Inka@{\sc Inka}}{\sc Inka}}

\newcommand\RRL       {\index{Rrl@{\sc Rrl}}{\sc Rrl}}

\newcommand\THMindexstart{\index{Thm@{\sc Thm}|(}}
\newcommand\THMindexend{\index{Thm@{\sc Thm}|)}}
\newcommand\THM       {\index{Thm@{\sc Thm}}{\sc Thm}}
\newcommand\QTHM      {\index{Qthm@{\sc Qthm}}{\sc Qthm}}
\newcommand\NQTHM     {\index{Nqthm@{\sc Nqthm}}{\sc Nqthm}}
\newcommand\NUPRL     {\index{Nuprl@{\sc Nuprl}}{\sc Nuprl}}
\newcommand\PROLOG    {\index{Prolog@{\sc Prolog}}{\sc Prolog}}

\newcommand\OYSTER    {{\sc Oyster}}
\newcommand\OYSTERCLAM{\index{Oyster/CLAM@{\OYSTER/\CLAM}}\OYSTER/\CLAM}
\newcommand\LCF       {\index{LCF@{\sc LCF}}{\sc LCF}}
\newcommand\LEOII     {\index{Leo-II@{\sc Leo-II}}\mbox{\sc Leo-II}}

\newcommand\ISAPLANNER{\index{IsaPlanner@{\sc IsaPlanner}}{\sc Isa\-Plan\-ner}}
 
\newcommand\LISPnoindex{{\sc LISP}}
\newcommand\LISP      {\index{LISP@\LISPnoindex}\LISPnoindex}
\newcommand\COMMONLISP{\index{Common Lisp@{\sc Common Lisp}}{\sc Common Lisp}}
 
\newcommand\HASKELL   {\index{Haskell@{\sc Haskell}}{\sc Haskell}}

\newcommand\PURELISPTP{\index
          {Pure LISP Theorem Prover@{\sc Pure \LISPnoindex\ Theo\-rem Pro\-ver}}%
                                            {\sc Pure \LISP\ Theo\-rem Pro\-ver}}
 
\newcommand\QUODLIBET {\index{QuodLibet@{\sc Quod\-Libet}}{\sc Quod\-Libet}}
\newcommand\QML       {{\sc Qml}} 
 
\newcommand\TEXMACS   {{\sc T\kern-.1667em\lower.5ex\hbox{E}\kern-.125emX\kern-.1em\lower.5ex\hbox{\textsc{m\kern-.05ema\kern-.125emc\kern-.05ems}}}}

\newcommand\UNICOM    {\index{Unicom@{\sc Unicom}}{\sc Unicom}}
\newcommand\KNUTHBENDIX{\knuth--\bendix}

\newcommand\ml        {\index{ml@{\sc ml}}{\sc ml}}

\newcommand\ISABELLE  {\index{Isabelle}{\sc Isa\-belle}}
\newcommand\HOL       {\index{HOL@{\sc HOL}}{\sc HOL}}
\newcommand\ISABELLEHOL{\index{Isabelle/HOL}{\sc Isa\-belle}/\HOL}
\newcommand\Pascal{{\sc Pascal}}
\newcommand\PVS{\index{PVS@{\sc PVS}}{\sc PVS}}
\newcommand\SATALLAX{\index{Satallax@{\sc Satallax}}{\sc Satallax}}

\newcommand\VAMPIRE{{\sc Vampire}}

\newcommand\WALDMEISTER{\index
                            {Waldmeister@{\sc Wald\-Meister}}{\sc Wald\-Meister}}

\newcommand\MIT{\mbox{MIT}}
\newcommand\theMIT{\MIT}


\newcommand\Avignon        {Avignon}

\newcommand\Budapest       {Budapest}

\newcommand\EB             {Edin\-burgh}
\newcommand\ForrestHillEB  {Forrest Hill} 

\newcommand\StanfordCA     {Stan\-ford (CA)}

\newcommand\Wernigerode    {Werni\-gerode}


\newcommand\plzwernigerode{\mbox{38855}}

\newcommand\uniEB       {University of \EB}

\newcommand\unitexasaustin{The University of Texas at Austin}

\newcommand\uniEBshort{\Univ\ \EB}

\def       \emailcp      {{\tt wirth@logic.at}}

\newcommand\Institutedept
{\FBautinfveryshort}
\newcommand\Instituteinst
{\mbox{Hochschule Harz}}
\newcommand\Instituteplac
{\plzwernigerode\,\Wernigerode}
\newcommand\Institutecoun{Germany}

\newcommand\Institute
{\Institutedept, \Instituteinst, \Instituteplac, \Institutecoun}

\newcommand\SFB        {Collaborative Research Center}

\newcommand\oldsfbshort{SFB\,314}








\newcommand\academicpress{Academic Press (\elsevier)}

\newcommand\elsevier{Elsevier}








\newcommand\CADEname{\Int\ \Conf\ on Automated Deduction}
\newcommand\CADEshort{CADE}
\newcommand\thelocalCADE[3]{\nth{#1} \CADEname\ (\CADEshort), #3, #2}

\newcommand\thesixteenthCADEninetynine              
{\thelocalCADE  {16}{1999}{Trento (Italy)}}

\newcommand\thenineteenthCADEthree                   
{\thelocalCADE  {19}{2003}{Miami Beach (FL)}}






\newcommand\CTRSname{\Int\ Workshop on Conditional Term Rewriting Systems (CTRS)}

\newcommand\theCTRS[3]{\nth{#1} \CTRSname, #3, #2}

\newcommand\thefirstCTRSeightyseven       {\theCTRS{1}{1987}{Orsay (France)}}








\newcommand\IJCAI    {\mbox{IJCAI}}














\newcommand\newspaperreference[5]
{\def\nameofjournalpress{#2}#1, #4 #5, #3\if?\nameofjournalpress
\else, #2\fi}


\newcommand\dateinjournal[1]{}

\newcommand\journalreference[6]
{\def\nameofjournalpress{#2}#1\nolinebreak\hskip.2em%
\dateinjournal{(#3) }{\mbox{\bf #4}}, \PP{#5}{#6}\if?\nameofjournalpress
\else, #2\fi}

\newcommand\journalreferenceprintyear[6]
{\def\nameofjournalpress{#2}#1 
#4:#5--#6, #3%
\if?\nameofjournalpress
\else, #2\fi}

\newcommand\journalreferenceprintyearaspartofnumber[6]
{\def\nameofjournalpress{#2}#1 
{#4/#3}, \PP{#5}{#6}\if?\nameofjournalpress
\else, #2\fi}

\newcommand\jscname
{J. Symbolic Computation}

\newcommand\jscprintyear
{\journalreferenceprintyear{\jscname}\academicpress}

\newcommand\tcsname{Theoretical \CS}
\newcommand\tcsjournal
{\journalreference\tcsname\elsevier}
\newcommand\tcsjournalprintyear
{\journalreferenceprintyear\tcsname\elsevier}


\urldef\wirthkuhnurl\url
{http://wirth.bplaced.net/SEKI/welcome.html#SWP-2007-01}

\urldef\urlsrninetythreedashzerofive\url
{http://wirth.bplaced.net/SEKI/welcome.html#SR-93-05}

\urldef\urlsreightyeighttwelve\url
{http://wirth.bplaced.net/SEKI/welcome.html#SR-88-12}

%% file: headerforformulas.tex
\mathcommand\ident[1]{\mathsf{#1}}
\newcommand\plussymbol  {\ident{+}}
\newcommand\minussymbol {\ident{-}}
\newcommand\dividesymbol{\ident{/}}
\newcommand\timessymbol {\ident{*}}


\newcommand\nat     {\ident{nat}}

\newcommand\lists   {\ident{list}}
\newcommand\set     {\ident{set}}

\newcommand\bool    {\ident{bool}}

\newcommand\ortreesort{\ident{or\mbox-tree}}
\newcommand\andtreesort{\ident{and\mbox-tree}}

\newcommand\naturalssymbol{\ident{naturals}}
\newcommand\gensymsymbol{\ident{gensym}}
\mathcommand\mbpsymbol{\ident{m\hspace{-0.055em}b\hspace{-0.045em}p}}

\newcommand\csymbol     {\ident c}
\newcommand\esymbol     {\ident e}
\newcommand\fsymbol     {\ident f}
\newcommand\gsymbol     {\ident g}
\newcommand\hsymbol     {\ident h}
\newcommand\ksymbol     {\ident k}
\newcommand\psymbol     {\ident p}
\newcommand\ssymbol     {\ident s}
\newcommand\Everysymbol {\ident{Every}}
\newcommand\Permsymbol {\ident{Perm}}
\newcommand\RExistssymbol{\ident{Rexists}}
\newcommand\invertsymbol{\ident{invert}}
\newcommand\invsymbol{\ident{inv}}
\newcommand\abssymbol   {\ident{abs}}
\newcommand\cnssymbol   {\ident{cons}}
\mathcommand\cnsindexsymbol[1]{\ident{cons}_{#1}}
\newcommand\carsymbol   {\ident{car}}

\newcommand\cdrsymbol   {\ident{cdr}}
\newcommand\lengthsymbol{\ident{length}}
\newcommand\sizesymbol{\ident{size}}
\newcommand\dlsymbol    {\ident{dl}}
\newcommand\dloncesymbol{\ident{delfirst}}
\newcommand\rcsymbol    {\ident{rc}}
\newcommand\brsymbol    {\ident{br}}
\newcommand\revtailsymbol{\ident{revtail}}
\newcommand\revsymbol{\ident{rev}}
\newcommand\appendsymbol {\ident{append}}
\newcommand\zeropredicatesymbol{\ident{zerop}}
\newcommand\eqsymbol        {\ident{eq}}
\newcommand\ifthensymbol    {\mbox{\ident{If{}Then}}}
\newcommand\ifthenelsesymbol{\mbox{\ident{If{}ThenElse}}}
\mathcommand\eqindexsymbol        [1]{\eqsymbol        _{#1}}
\mathcommand\ifthenindexsymbol    [1]{\ifthensymbol    _{#1}}
\mathcommand\ifthenelseindexsymbol[1]{\ifthenelsesymbol_{#1}}
\newcommand\orsymbol    {\ident{or}}
\newcommand\andsymbol   {\ident{and}}
\newcommand\leqsymbol   {\ident{leq}}
\newcommand\lessymbol   {\ident{less}}
\newcommand\lexlessymbol{\ident{lexless}}
\newcommand\lexlimlessymbol{\ident{lexlimless}}
\newcommand\lexsymbol   {\ident{lex}}
\newcommand\acksymbol   {\ident{ack}}
\newcommand\switchsymbol{\ident{switch}}
\newcommand\swatchsymbol{\ident{swatch}}
\newcommand\diveinssymbol{\ident{div1}}
\newcommand\divzweisymbol{\ident{div2}}
\newcommand\divrestsymbol{\ident{divrest}}
\newcommand\diveinstailsymbol{\ident{div1tail}}
\newcommand\divzweitailsymbol{\ident{div2tail}}
\newcommand\remsymbol{\ident{rem}}
\newcommand\divsymbol{\ident{div}}

\newcommand\turingmachinesymbol{\ident T}
\newcommand\terminatespsymbol  {\ident{terminatesp}}
\newcommand\statesymbol        {\ident{state}}
\newcommand\cmdsymbol          {\ident{cmd}}
\newcommand\nthsymbol          {\ident{nth}}
\newcommand\doublesymbol       {\ident{double}}

\newcommand\ppsymbol           {\ident{p}}
\newcommand\qpsymbol           {\ident{q}}
\newcommand\Epsymbol           {\ident{E}}
\newcommand\Ppsymbol           {\ident{P}}
\newcommand\Qpsymbol           {\ident{Q}}
\newcommand\Marriessymbol      {\ident{Marries}}
\newcommand\Lovessymbol        {\ident{Loves}}
\newcommand\StolenBysymbol     {\ident{StolenBy}}
\newcommand\Humansymbol        {\ident{Human}}
\newcommand\Evensymbol         {\ident{Even}}
\newcommand\Oddsymbol          {\ident{Odd}}
\newcommand\Primesymbol        {\ident{Prime}}
\newcommand\EveryPairsymbol   {\ident{EveryPair}}
\newcommand\Givesymbol         {\ident{Give}}
\newcommand\Fathersymbol       {\ident{Father}}
\newcommand\Elephantpsymbol    {\ident{Elephant}}
\newcommand\Flowerpsymbol    {\ident{Flower}}
\newcommand\Germanpsymbol      {\ident{German}}
\newcommand\Bicyclepsymbol     {\ident{Bicycle}}
\newcommand\Hugepsymbol        {\ident{Huge}}
\newcommand\Animalpsymbol      {\ident{Animal}}
\newcommand\Malepsymbol        {\ident{Male}}
\newcommand\Boypsymbol         {\ident{Boy}}
\newcommand\Girlpsymbol        {\ident{Girl}}
\newcommand\Femalepsymbol      {\ident{Female}}
\newcommand\Roundpsymbol       {\ident{Round}}
\newcommand\Quadrangularpsymbol{\ident{Quadrangular}}
\newcommand\Metpsymbol         {\ident{Met}}
\newcommand\Kissedpsymbol      {\ident{Kissed}}
\newcommand\Bishopsymbol       {\ident{Bishop}}
\newcommand\mindexsymbol[1]{\existsvari w{#1}}

\newcommand\nonnegpsymbol      {\ident{nonnegp}}
\newcommand\wellsymbol         {\ident{well}}
\newcommand\welltailsymbol     {\ident{welltail}}
\newcommand\varsymbol          {\ident{var}}
\newcommand\aritysymbol        {\ident{arity}}

\newcommand\whilesymbol        {\ident{while}}

\newcommand\nullsymbol         {\ident{null}}
\newcommand\hdsymbol           {\ident{hd}}
\newcommand\tlsymbol           {\ident{tl}}
\newcommand\insymbol           {\ident{in}}
\newcommand\applysymbol        {\ident{app}}
\newcommand\termsymbol         {\ident{term}}
\newcommand\russellsymbol      {\ident{russell}}
\newcommand\sqrtindordsymbol[1]{\ident{sqrtio#1}}
\mathcommand\tightim{\longrightarrow}
\mathcommand\im{\ \tightim\ }
\mathcommand\rs{\:\rulesugar\:\:}
\mathcommand\rulesugar{\longleftarrow}

\mathcommand\doublepp[1]      {\doublesymbol   \beginargs{#1}\allargs}
\mathcommand\aritypp[1]      {\aritysymbol   \beginargs{#1}\allargs}
\mathcommand\lengthpp[1]      {\lengthsymbol   \beginargs{#1}\allargs}
\mathcommand\sizepp[1]      {\sizesymbol   \beginargs{#1}\allargs}
\mathcommand\wellpp[1]      {\wellsymbol   \beginargs{#1}\allargs}
\mathcommand\welltailpp[1]      {\welltailsymbol   \beginargs{#1}\allargs}
\mathcommand\varpp[1]      {\varsymbol   \beginargs{#1}\allargs}
\mathcommand\rempp[2]    {\remsymbol\beginargs{#1}\separgs{#2}\allargs}
\mathcommand\divpp[2]    {\divsymbol\beginargs{#1}\separgs{#2}\allargs}
\mathcommand\divrestpp[2]    {\divrestsymbol\beginargs{#1}\separgs{#2}\allargs}
\mathcommand\diveinspp[2]    {\diveinssymbol\beginargs{#1}\separgs{#2}\allargs}
\mathcommand\divzweipp[3]    {\divzweisymbol\beginargs{#1}\separgs{#2}
\separgs{#3}\allargs}
\mathcommand\diveinstailpp[4]    {\diveinstailsymbol\beginargs{#1}\separgs{#2}
\separgs{#3}\separgs{#4}\allargs}
\mathcommand\divzweitailpp[6]    {\divzweitailsymbol\beginargs{#1}\separgs{#2}
\separgs{#3}\separgs{#4}\separgs{#5}\separgs{#6}\allargs}
\mathcommand\mbppp[2]         {\mbpsymbol   \beginargs{#1}\separgs{#2}\allargs}
\mathcommand\revpp[1]     
{\revsymbol\beginargs{#1}\allargs}
\mathcommand\revppi[2]     
{\revsymbol^{#1}\beginargs{#2}\allargs}
\mathcommand\revtailpp[2]     
{\revtailsymbol\beginargs{#1}\separgs{#2}\allargs}
\mathcommand\revtailppi[3]
{\revtailsymbol^{#1}\beginargs{#2}\separgs{#3}\allargs}
\mathcommand\Permpp[2]     
{\Permsymbol\beginargs{#1}\separgs{#2}\allargs}
\mathcommand\Permppi[3]
{\Permsymbol^{#1}\beginargs{#2}\separgs{#3}\allargs}
\mathcommand\appendpp[2]      
{\appendsymbol \beginargs{#1}\separgs{#2}\allargs}
\mathcommand\appendppi[3]      
{\appendsymbol^{#1}\beginargs{#2}\separgs{#3}\allargs}
\mathcommand\Everypp[2]      
{\Everysymbol \beginargs{#1}\separgs{#2}\allargs}
\mathcommand\RExistspp[1]      
{\RExistssymbol \beginargs{#1}\allargs}
\mathcommand\appendlongpp[2]      
{\appendsymbol\left(\begin{array}{@{}l@{}}{#1}\separgs\\{#2}\end{array}\right)}
\mathcommand\cnspp[2]         {\cnssymbol   \beginargs{#1}\separgs{#2}\allargs}
\mathcommand\cnsppi[3]       {\cnssymbol^{#1}\beginargs{#2}\separgs{#3}\allargs}
\mathcommand\cnsindexpp[3]
{\cnsindexsymbol{#1}\beginargs{#2}\separgs{#3}\allargs}
\mathcommand\dlpp[2]          {\dlsymbol    \beginargs{#1}\separgs{#2}\allargs}
\mathcommand\dloncepp[2]      {\dloncesymbol\beginargs{#1}\separgs{#2}\allargs}
\mathcommand\dlonceppi[3]{\dloncesymbol^{#1}\beginargs{#2}\separgs{#3}\allargs}
\mathcommand\rcpp[2]          {\rcsymbol    \beginargs{#1}\separgs{#2}\allargs}
\mathcommand\brpp[2]          {\brsymbol    \beginargs{#1}\separgs{#2}\allargs}
\mathcommand\orpp[2]          {\orsymbol    \beginargs{#1}\separgs{#2}\allargs}
\mathcommand\andpp[2]         {\andsymbol   \beginargs{#1}\separgs{#2}\allargs}
\mathcommand\shortcnspp[2]    {\csymbol     \beginargs{#1}\separgs{#2}\allargs}
\mathcommand\tightshortcnspp[2]
{\csymbol\beginargs{#1}\tightsepargs{#2}\allargs}
\mathcommand\spp[1]           {\ssymbol     \beginargs{#1}\allargs}
\mathcommand\sppiterated[2]   {\ssymbol^{#1}\beginargs{#2}\allargs}
\mathcommand\sqrtindordpp[3]
                       {\sqrtindordsymbol{#1}\beginargs{#2}\separgs{#3}\allargs}
\mathcommand\ppp[1]           {\psymbol     \beginargs{#1}\allargs}
\mathcommand\pppiterated[2]   {\psymbol^{#1}\beginargs{#2}\allargs}
\mathcommand\zeropp           {\ident 0}
\mathcommand\Julietpp         {\ident{Juliet}}
\mathcommand\Romeopp          {\ident{Romeo}}
\mathcommand\Ipp              {\ident I}
\mathcommand\onepp            {\ident1}
\mathcommand\twopp            {\ident2}
\mathcommand\threepp          {\ident3}
\mathcommand\invertpp[1]      {\invertsymbol\beginargs{#1}\allargs}
\mathcommand\invpp[1]         {\invsymbol\beginargs{#1}\allargs}
\mathcommand\abspp[1]         {\abssymbol\beginargs{#1}\allargs}
\mathcommand\naturalspp[1]    {\naturalssymbol\beginargs{#1}\allargs}
\mathcommand\gensympp[1]      {\gensymsymbol\beginargs{#1}\allargs}
\mathcommand\nilpp            {\ident{nil}}
\mathcommand\falsepp          {\ident{false}}
\mathcommand\truepp           {\ident{true}}
\mathcommand\FALSEpp          {\ident{FALSE}}
\mathcommand\TRUEpp           {\ident{TRUE}}
\mathcommand\UNDEFpp          {\ident{UNDEF}}
\mathcommand\weirdppp         {\ident{weirdp}}
\mathcommand\ambigppp         {\ident{ambigp}}
\mathcommand\zeropredicatepp[1]{\zeropredicatesymbol\beginargs{#1}\allargs}
\mathcommand\cppeins       [1]{\csymbol     \beginargs{#1}\allargs}
\mathcommand\cppzwei       [2]{\csymbol\beginargs{#1}\separgs{#2}\allargs}
\mathcommand\eppeins       [1]{\esymbol     \beginargs{#1}\allargs}
\mathcommand\fppeins       [1]{\fsymbol     \beginargs{#1}\allargs}
\mathcommand\fppeinsindex  [2]{\fsymbol_{#1}\beginargs{#2}\allargs}
\mathcommand\fppeinsiterated[2]{\fsymbol^{#1}\beginargs{#2}\allargs}
\mathcommand\gppeins       [1]{\gsymbol     \beginargs{#1}\allargs}
\mathcommand\gppzwei       [2]{\gsymbol     \beginargs{#1}\separgs{#2}\allargs}
\mathcommand\hppeins       [1]{\hsymbol     \beginargs{#1}\allargs}
\mathcommand\kppeins       [1]{\ksymbol     \beginargs{#1}\allargs}
\mathcommand\appzero          {\ident a}
\mathcommand\bppzero          {\ident b}
\mathcommand\cppzero          {\ident c}
\mathcommand\dppzero          {\ident d}
\mathcommand\eppzero          {\ident e}
\mathcommand\eqindexpp[3]{\eqindexsymbol{#1}\beginargs{#2}\separgs{#3}\allargs}
\mathcommand\ifthenindexpp
[3]{\ifthenindexsymbol{#1}\beginargs{#2}\separgs{#3}\allargs}
\mathcommand\ifthenelseindexpp
[4]{\ifthenelseindexsymbol{#1}\beginargs{#2}\separgs{#3}\separgs{#4}\allargs}
\mathcommand\eqpp[2]{\eqsymbol\beginargs{#1}\separgs{#2}\allargs}
\mathcommand\leqpp[2]{\leqsymbol\beginargs{#1}\separgs{#2}\allargs}
\mathcommand\lespp[2]{\lessymbol\beginargs{#1}\separgs{#2}\allargs}
\mathcommand\lexlespp[2]{\lexlessymbol\beginargs{#1}\separgs{#2}\allargs}
\mathcommand\lexlimlespp[3]
               {\lexlimlessymbol\beginargs{#1}\separgs{#2}\separgs{#3}\allargs}
\mathcommand\lexpp[3]{\lexsymbol\beginargs{#1}\separgs{#2}\separgs{#3}\allargs}
\mathcommand\ackpp[2]{\acksymbol\beginargs{#1}\separgs{#2}\allargs}
\mathcommand\switchpp[1]{\switchsymbol\beginargs{#1}\allargs}
\mathcommand\swatchpp[1]{\swatchsymbol\beginargs{#1}\allargs}
\mathcommand\whilepp[2]{\whilesymbol\beginargs{#1}\separgs{#2}\allargs}
\mathcommand\nullpp[1]{\nullsymbol\beginargs{#1}\allargs}
\mathcommand\nullppiterated[2]{\nullsymbol^{#1}\beginargs{#2}\allargs}
\mathcommand\hdpp[1]{\hdsymbol\beginargs{#1}\allargs}
\mathcommand\hdppiterated[2]{\hdsymbol^{#1}\beginargs{#2}\allargs}
\mathcommand\carpp[1]{\carsymbol\beginargs{#1}\allargs}
\mathcommand\cdrpp[1]{\cdrsymbol\beginargs{#1}\allargs}
\mathcommand\tlpp[1]{\tlsymbol\beginargs{#1}\allargs}
\mathcommand\tlppiterated[2]{\tlsymbol^{#1}\beginargs{#2}\allargs}
\mathcommand\inpp[2]{\insymbol\beginargs{#1}\separgs{#2}\allargs}
\mathcommand\inppiterated[3]{\insymbol^{#1}\beginargs{#2}\separgs{#3}\allargs}
\mathcommand\applypp[2]{\applysymbol\beginargs{#1}\separgs{#2}\allargs}
\mathcommand\applyppiterated
[3]{\applysymbol^{#1}\beginargs{#2}\separgs{#3}\allargs}
\mathcommand\termpp[2]{\termsymbol\beginargs{#1}\separgs{#2}\allargs}
\mathcommand\setpp[1]{\set\beginargs{#1}\allargs}
\mathcommand\russellpp[1]{\russellsymbol\beginargs{#1}\allargs}

\mathcommand\Tpp[6]{\turingmachinesymbol\beginargs{#1}\separgs{#2}\separgs
{#3}\separgs{#4}\separgs{#5}\separgs{#6}\allargs}
\mathcommand\Tppseven[7]{\turingmachinesymbol\beginargs{#1}\separgs{#2}\separgs
{#3}\separgs{#4}\separgs{#5}\separgs{#6}\separgs{#7}\allargs}
\mathcommand\foreverppp[6]{\ident{foreverp}\beginargs{#1}\separgs{#2}\separgs
{#3}\separgs{#4}\separgs{#5}\separgs{#6}\allargs}
\mathcommand\terminatesppp[6]{\terminatespsymbol\beginargs{#1}\separgs
{#2}\separgs{#3}\separgs{#4}\separgs{#5}\separgs{#6}\allargs}
\mathcommand\terminatespppone[1]{\terminatespsymbol \beginargs{#1}\allargs}
\mathcommand\statepp
[3]{\statesymbol\beginargs{#1}\separgs{#2}\separgs{#3}\allargs}
\mathcommand\tightstatepp
[3]{\statesymbol\beginargs{#1}\tightsepargs{#2}\tightsepargs{#3}\allargs}
\mathcommand\cmdpp
[3]{\cmdsymbol  \beginargs{#1}\separgs{#2}\separgs{#3}\allargs}
\mathcommand\tightcmdpp
[3]{\cmdsymbol  \beginargs{#1}\tightsepargs{#2}\tightsepargs{#3}\allargs}
\mathcommand\stoppp           {\ident{stop}}
\mathcommand\leftpp           {\ident{left}}
\mathcommand\rightpp          {\ident{right}}
\mathcommand\nthpp         [2]{\nthsymbol  \beginargs{#1}\separgs{#2}\allargs}
\mathcommand\pppp          [1]{\ppsymbol\beginargs{#1}            \allargs}
\mathcommand\qppp          [2]{\qpsymbol\beginargs{#1}\separgs{#2}\allargs}
\mathcommand\Eppp          [1]{\Epsymbol\beginargs{#1}            \allargs}
\mathcommand\Epppzwei      [2]{\Epsymbol\beginargs{#1}\separgs{#2}\allargs}
\mathcommand\Pppp          [1]{\Ppsymbol\beginargs{#1}            \allargs}
\mathcommand\Ppppeinsindex [2]{\Ppsymbol_{#1}\beginargs{#2}\allargs}
\mathcommand\Ppppdrei      
[3]{\Ppsymbol\beginargs{#1}\separgs{#2}\separgs{#3}\allargs}
\mathcommand\Ppppvier
[4]{\Ppsymbol\beginargs{#1}\separgs{#2}\separgs{#3}\separgs{#4}\allargs}
\mathcommand\Qppp          [2]{\Qpsymbol\beginargs{#1}\separgs{#2}\allargs}
\mathcommand\Qpppeins      [1]{\Qpsymbol\beginargs{#1}\allargs}
\mathcommand\Qpppeinsindex [2]{\Qpsymbol_{#1}\beginargs{#2}\allargs}
\mathcommand\Qpppdrei      
[3]{\Qpsymbol\beginargs{#1}\separgs{#2}\separgs{#3}\allargs}
\mathcommand\Fatherpp      [2]{\Fathersymbol\beginargs{#1}\separgs{#2}\allargs}
\mathcommand\Marriespp     [2]{\Marriessymbol\beginargs{#1}\separgs{#2}\allargs}
\mathcommand\Lovespp       [2]{\Lovessymbol\beginargs{#1}\separgs{#2}\allargs}
\mathcommand\StolenBypp    [2]
{\StolenBysymbol\beginargs{#1}\separgs{#2}\allargs}
\mathcommand\Humanpp       [1]{\Humansymbol\beginargs{#1}\allargs}
\mathcommand\Evenpp        [1]{\Evensymbol\beginargs{#1}\allargs}
\mathcommand\Evenppi       [2]{\Evensymbol^{#1}\beginargs{#2}\allargs}
\mathcommand\Oddpp         [1]{\Oddsymbol\beginargs{#1}\allargs}
\mathcommand\Primepp       [1]{\Primesymbol\beginargs{#1}\allargs}
\mathcommand\EveryPairpp  [2]{\EveryPairsymbol\beginargs{#1}\separgs
{#2}\allargs}
\mathcommand\mindexppeins  [2]{\mindexsymbol{#1}\beginargs{#2}\allargs}
\mathcommand\Givepp        [3]{\Givesymbol
\beginargs{#1}\separgs{#2}\separgs{#3}\allargs}
\mathcommand\mindexppzwei  [3]{\mindexsymbol
{#1}\beginargs{#2}\separgs{#3}\allargs}
\mathcommand\mindexppdrei  [4]{\mindexsymbol
{#1}\beginargs{#2}\separgs{#3}\separgs{#4}\allargs}

\mathcommand\nonnegppp     [1]{\nonnegpsymbol\beginargs{#1}\allargs}

\mathcommand\anonymouscsymbol{c}
\mathcommand\anonymouscindexsymbol[1]{\anonymouscsymbol_{#1}}
\mathcommand\anonymousfsymbol{f}
\mathcommand\anonymouscpp
[2]{\anonymouscsymbol\beginargs{#1}\separgs\ldots\separgs{#2}\allargs}
\mathcommand\anonymouscindexpp
[3]{\anonymouscindexsymbol{#1}\beginargs{#2}\separgs\ldots\separgs{#3}\allargs}
\mathcommand\anonymousfpp
[2]{\anonymousfsymbol\beginargs{#1}\separgs\ldots\separgs{#2}\allargs}
\mathcommand\coerceindexpp[3]{[#3]_{#1}^{#2}}

\mathcommand\Elephantppp    [1]{\Elephantpsymbol\beginargs{#1}\allargs}
\mathcommand\Flowerppp      [1]{\Flowerpsymbol  \beginargs{#1}\allargs}
\mathcommand\Bicycleppp     [1]{\Bicyclepsymbol \beginargs{#1}\allargs}
\mathcommand\Germanppp      [1]{\Germanpsymbol  \beginargs{#1}\allargs}
\mathcommand\Hugeppp        [1]{\Hugepsymbol    \beginargs{#1}\allargs}
\mathcommand\Animalppp      [1]{\Animalpsymbol  \beginargs{#1}\allargs}
\mathcommand\Maleppp        [1]{\Malepsymbol    \beginargs{#1}\allargs}
\mathcommand\Boyppp         [1]{\Boypsymbol     \beginargs{#1}\allargs}
\mathcommand\Girlppp        [1]{\Girlpsymbol    \beginargs{#1}\allargs}
\mathcommand\Femaleppp      [1]{\Femalepsymbol  \beginargs{#1}\allargs}
\mathcommand\Roundppp       [1]{\Roundpsymbol   \beginargs{#1}\allargs}
\mathcommand\Bishoppp       [1]{\Bishopsymbol   \beginargs{#1}\allargs}
\mathcommand\Quadrangularppp[1]{\Quadrangularpsymbol  \beginargs{#1}\allargs}
\mathcommand\Kissedppp[2]{\Kissedpsymbol\beginargs{#1}\separgs{#2}\allargs}
\mathcommand\Metppp[2]   {\Metpsymbol   \beginargs{#1}\separgs{#2}\allargs}

\newcommand\bound     {{\rm bound}}
\newcommand\free      {{\rm free}}

\mathcommand\Vtripleindex[3]{\V\!_{{#1},\,{#2},\,{#3}}}
\mathcommand\Vdoubleindex[2]{\V\!_{{#1},\,{#2}}}
\mathcommand\Vsingleindex[1]{\V\!_{{#1}}}

\mathcommand\Erel[1]{\Gammaoffont\!_{#1}}
\mathcommand\Urel[1]{\Deltaoffont_{#1}}



\mathcommand\theRprimefromstrongtoweak{
  \inparenthesesinlinetight{
     \domres\id{\Vwall\cup\Vsome\setminus\RAN\varsigma}
     \nottight{\nottight\uplus}
     \reverserelation\varsigma
  }
  \nottight{\circ}
  \ranres
    {\transclosureinline R}
    {\Vwall\cup\Vsome\setminus\RAN\varsigma}
  \nottight{\nottight{\nottight{\uplus}}}
  \Vsome\tighttimes\Vsall
}

\mathcommand\deltaminus{\delta^-}
\mathcommand\deltaplus{\delta^+}
\mathcommand\deltaplusplus{\delta^{+^+}}
\mathcommand\deltastar{\delta^*}
\mathcommand\deltastarstar{\delta^{*^*}}

\mathcommand\Vall     {\Vsingleindex\indexdelta         }
\mathcommand\Vwall    {\Vsingleindex\indexdeltaminu     }
\mathcommand\Vsall    {\Vsingleindex\indexdeltaplus     }
\mathcommand\Vgsome   {\Vsingleindex\indexgammaplus     }
\mathcommand\Vsome    {\Vsingleindex\indexgamma         }
\mathcommand\Vfree    {\Vsingleindex\indexfree          }
\mathcommand\Vbound   {\Vsingleindex\indexbound         }
\mathcommand\Vsomesall{\Vsingleindex\indexgammadeltaplus}

\mathapplycommand\VARall      {\VARsingleindex\indexdelta         }
\mathapplycommand\VARwall     {\VARsingleindex\indexdeltaminu     }
\mathapplycommand\VARsall     {\VARsingleindex\indexdeltaplus     }
\mathapplycommand\VARgsome    {\VARsingleindex\indexgammaplus     }
\mathapplycommand\VARsome     {\VARsingleindex\indexgamma         }
\mathapplycommand\VARfree     {\VARsingleindex\indexfree          }
\mathapplycommand\VARbound    {\VARsingleindex\indexbound         }
\mathapplycommand\VARsomesall {\VARsingleindex\indexgammadeltaplus}
\mathcommand\displayVARsall[1]{\VARsingleindex\indexdeltaplus
\!\!\!\:\left(\begin{array}{@{}c@{}}#1\end{array}\right)}

\mathcommand\rigidvari     [2]{#1_{#2}^\indexgammadeltaplus}
\mathcommand\existsvari    [2]{#1_{#2}^\indexgamma    }
\mathcommand\forallvari    [2]{#1_{#2}^\indexdelta    }
\mathcommand\freevari      [2]{#1_{#2}^\indexfree     }
\mathcommand\wforallvari   [2]{#1_{#2}^\indexdeltaminu}
\mathcommand\sforallvari   [2]{#1_{#2}^\indexdeltaplus}
\mathcommand\gexistsvari   [2]{#1_{#2}^\indexgammaplus}
\mathcommand\boundvari     [2]{#1_{#2}}
\mathcommand\vari          [2]{#1_{#2}}
\mathcommand\wforallvarilow[2]{#1_{#2}^
{\raisebox{-.82ex}{\math\indexdeltaminu}}}

\newcommand\indexhelper[1]{{\scriptscriptstyle#1\:\!\!}}
\newcommand\indexdeltaplus
{\indexhelper{\delta^{\raisebox{-.17ex}{\fvesf\hskip-0.14em +}}}}
\newcommand\indexdeltaminu
{\indexhelper{\delta^{\mbox{\fvesf\hskip-0.14em\rule[.2ex]{.7em}{.15ex}}}}}
\newcommand\indexgammaplus
{\indexhelper{\gamma^{\mbox{\fvesf\hskip-0.14em +}}}}
\newcommand\indexgammadeltaplus
{\indexhelper{\gamma\delta^{\raisebox{-.17ex}{\fvesf\hskip-0.14em +}}}}

\newcommand\indexdelta{\indexhelper\delta}
\newcommand\indexgamma{\indexhelper\gamma}
\newcommand\indexfree
{{\scriptscriptstyle\free}}
\newcommand\indexbound
{{\scriptscriptstyle\bound}}

\newcommand\Wellfsymb{\ident{Wellf}}
\mathapplycommand\Wellfpp{\Wellfsymb}

%% file: headersugarterms.tex
\mathcommand\beginargs{(}
\mathcommand\allargs  {)}
\mathcommand\separgs  {,\,}
\mathcommand\tightsepargs{,}

\mathcommand\minusppnoparentheses  [2]{{#1}\,\minussymbol\,{#2}}
\mathcommand\tightminusppnoparentheses  [2]{{#1}\minussymbol{#2}}
\mathcommand\divideppnoparentheses [2]{{#1}\,\dividesymbol\,{#2}}
\mathcommand\plusppnoparentheses   [2]{{#1}\,\plussymbol \,{#2}}
\mathcommand\plusppnoparenthesesi  [3]{{#2}\,\plussymbol^{#1}\,{#3}}
\mathcommand\tightplusppnoparentheses   [2]{{#1}\plussymbol{#2}}
\mathcommand\timesppnoparentheses  [2]{{#1}\,\timessymbol\,{#2}}
\mathcommand\undppnoparentheses    [2]{{#1}\und            {#2}}
\mathcommand\oderppnoparentheses   [2]{{#1}\oder           {#2}}
\mathcommand\impliesppnoparentheses[2]{{#1}\implies        {#2}}
\mathcommand\leqinfixppnoparentheses[2]{{#1}\,\tight\leq\,{#2}}
\mathcommand\geqinfixppnoparentheses[2]{{#1}\,\tight\geq\,{#2}}
\mathcommand\dividepp [2]{(\divideppnoparentheses {#1}{#2})}
\mathcommand\minuspp  [2]{(\minusppnoparentheses  {#1}{#2})}
\mathcommand\pluspp   [2]{(\plusppnoparentheses   {#1}{#2})}
\mathcommand\timespp  [2]{(\timesppnoparentheses  {#1}{#2})}
\mathcommand\undpp    [2]{(\undppnoparentheses    {#1}{#2})}
\mathcommand\oderpp   [2]{(\oderppnoparentheses   {#1}{#2})}
\mathcommand\impliespp[2]{(\impliesppnoparentheses{#1}{#2})}

%% file: quotation.tex
\newcommand\englishtextelevenone
{\englishtexteleventwo{Not only the notorious golden mountain is of gold, 
 but also\\the}.}
\newcommand\englishtexteleventwo[1]
{#1 round quadrangle is just as certainly round as it is quadrangular}

\newcommand\englishtextonehundredandthirteenquotation
{our translation}



\newcommand\englishtextninehundred
{Thus, in mathematics, we have no reasons to assume any meaning of
 `existence' that would be fundamentally different from that of 
 `the validity of axiomatic relations\closesinglequotefullstopnospace}

%% file: namedhelper.tex
\def\citep{\cite}
\def\citet#1{\citeauthor{#1} \shortcite{#1}}
\newcommand\startcite{{\raise.2ex\hbox{[}}}
\newcommand\stopcite {\raise.2ex\hbox{]}}
\newcommand\citehelper[1]{\startcite #1\stopcite}
\newcommand\makeaciteoftwo[2]
{\citehelper{\citeauthor{#1}, \citeyear{#1}; \citeyear{#2}}}

\newcommand\makeaciteofthree[3]
{\citehelper{\citeauthor{#1}, \citeyear{#1}; \citeyear{#2}; \citeyear{#3}}}

\newcommand\makeaciteoffour[4]
{\citehelper{\citeauthor{#1}, \citeyear{#1}; \citeyear{#2}; \citeyear{#3};
\citeyear{#4}}}

%% file: hbdictionary.tex



\newcommand\englishaeuszerlich            
{syntactical}

\newcommand\englishallgemeingueltigkeit 
{\index{validity}validity}

\newcommand\englishAllgemeingueltigkeit 
{\index{validity}Validity}
\newcommand\englishallgemeingueltig     
{\index{validity}valid}


\newcommand\englishalternativedichotomy   
{alternative}
\newcommand\englisheinealternativedichotomy
{an alternative}
\newcommand\englishalternativendichotomies
{alternatives}

\newcommand\englishanalogon               {analog}






\newcommand\englishanzahlenlehre          
{theory of cardinal numbers}



\newcommand\englishauffassenpointofview   {take the point of view}              
\newcommand\englishETWASalsETWASauffassen[2]
{\englishauffassenpointofview\ that #1 is #2}

\newcommand\englishaufloesung             {\index{resolution}resolution}



\newcommand\englishausdruckterm           
{expression}

\newcommand\englishausdruckformel         
{expression}




\newcommand\englishausgezeichnetcanonical
{full}
\newcommand\englishAusgezeichnetcanonical
{Full}

\newcommand\englishAussagenKalkul         {Propositional Calculus}

\newcommand\englishaussagenverbindung     
{propositional \englishverbindung}
\newcommand\englishaussagenverbindungen   
{propositional \englishverbindungen}

\newcommand\englishaussagenverknuepfung   
{propositional \englishverknuepfung}
\newcommand\englishaussagenverknuepfungen 
{propositional \englishverknuepfungen}

\newcommand\englishBereiche               {Domains}

\newcommand\englishbestandteil            
{part}







\newcommand\englishbezeichnung            {designation}
\newcommand\englisheinebezeichnung        {a \englishbezeichnung}


\newcommand\englishbildungsprozesse       
{formation processes}

\newcommand\englishbildungsregeln         
{formation rules}


\newcommand\englishdarstellendekonjunktion
{\index{Konjunktion!darstellende}%
 \index{conjunction, representing}%
 representing conjunction}


\newcommand\englishderjenigewelcher       {that which}











\newcommand\englisheinsetzungsregelindex  
{\index{substitution rule}}

\newcommand\englishentscheidungsproblemnoindex{decision \englishproblem}
\newcommand\englishentscheidungsproblem   
                   {\index{decision problem}\englishentscheidungsproblemnoindex}
\newcommand\englishEntscheidungsProblem   
                              {\index{decision problem}Decision \englishProblem}
\newcommand\englishentscheidungsprobleme  
{\englishentscheidungsproblem s}

\newcommand\englisherfahrungskomplexe     
{experience-complexes}






\newcommand\englishexistentialformeln     
                              {\index{existential formula}existential \formulae}

\newcommand\englishexistenzsatz            
%
{existence sentence}




\newcommand\englishfestumgrenzt
{well-determined and fixed}

\newcommand\englishFinit                  {Fini\-tis\-tic}

\newcommand\englishformelsprache          
{formula language}

\mathcommand\theFormela{(x)\,A(x)\nottight\implies A(a)}

\mathcommand\theFormelinpiteins{a=b\nottight{\nottight{\implies}}\inparentheses
{a=c\nottight{\nottight{\implies}}b=c}}

\mathcommand\theFormelinpitzwei{a=b\nottight{\nottight{\implies}}b=a}

\mathcommand\theFormelinpitdrei{a=b\nottight{\nottight{\implies}}\inparentheses
{b=c\nottight{\nottight{\implies}}a=c}}




\newcommand\englishganzerationalefunktion 
{polynomial function}

\newcommand\englishgegenstand             
{thing}

\newcommand\englishgenetisch              
{genetic}

\newcommand\englishGeschaeftsfuehrendeHerausgeber
{Editors-in-Chief}

\mathcommand\theformelJeins{a=a}

\mathcommand\theformelJzwei{a=b\nottight{\nottight\implies}\inparentheses{
A(a)\nottight\implies A(b)}}

\newcommand\englishgleichheitsbeziehung
{\index{equality!relation}equality relation}





\newcommand\englishhinterglied            
{consequent}

\newcommand\englishindividuensymbol
                                    {\index{individual symbol}individual symbol}
\newcommand\englisheinindividuensymbol    
                                          {an \englishindividuensymbol} 
 
\newcommand\englishindividuensymbole      
                                          {\englishindividuensymbol s}
\newcommand\englishpraedikatenundindividuensymbole
                                       {predicate and \englishindividuensymbole}
\newcommand\englishpraedikatenoderindividuensymbole
                                       {predicate or \englishindividuensymbole}
\newcommand\englishwederpraedikatennochindividuensymbole
                               {neither predicate nor \englishindividuensymbole}





\newcommand\englishkettenschluss          
{\index{chain inference}chain inference}

\newcommand\englishkettenschluesse          
{\index{chain inference}chain inferences}

\newcommand\englishnachpruefen           
{check}




\newcommand\englishnormaldisjunktion
{\index{normal disjunction}normal disjunction}

\newcommand\englishnormierung            
{standardization}



\newcommand\englishPraedikatenKalkul      {Predicate Calculus}

\newcommand\englishEinstelligerpraedikatenkalkulohnepraedikatenkalkul
                                     {\index{predicate calculus!monadic}Monadic}

\newcommand\englishEinstelligerPraedikatenKalkul
{\englishEinstelligerpraedikatenkalkulohnepraedikatenkalkul\
                                                      \englishPraedikatenKalkul}



\newcommand\englishproblem                {problem}
\newcommand\englishProblem                {Problem}







\newcommand\englishsatzverbindung         
{\index{sentential combination}sentential \englishverbindung}
\newcommand\englishsatzverbindungen       
{\index{sentential combination}sentential \englishverbindungen}



\newcommand\englishseinszeichen           
{existential quantifier \englishzeichen}

\newcommand\englishsinnesqualitaeten      
{qualia}

\newcommand\englishsinnlichewahrnehmung   
{sense perception}

\newcommand\englishsprachgebrauch         
{\englishgewoehnlichersprachgebrauch}
\newcommand\englishgewoehnlichersprachgebrauch
{common usage of language}

\newcommand\englishsubstitutionsregel     
{\index{rewrite rules}rewrite rule}


\newcommand\englishtatsaechlichkeit       
{actuality}






\newcommand\englishueberfuehrbar          
{\index{convertibility}convertible}

\newcommand\englishueberfuehrbarkeit      
{\index{convertibility}convertibility}





\newcommand\englishverbindung             {combination}
\newcommand\englishverbindungen           {com\-bina\-tions}




\newcommand\englishverteilungvonwahrheitswerten{distribution of truth values}

\newcommand\englishprepverteilungvonwahrheitswertenauf{on}

\newcommand\englishsichverifizieren    
{prove true}

\newcommand\englishverknuepfung           
{connective}
\newcommand\englishverknuepfungen         {\englishverknuepfung s}
\newcommand\englishverknuepfungprep       
{of}

\newcommand\englishverschaerfen           
{sharpen}

\newcommand\englishschaerfer  
{sharper}

\newcommand\englishverschaerft            
{sharp\-ened}

\newcommand\englishverschaerfung          
{sharpen\-ing}


\newcommand\englishVollstaendigkeit       {\index{completeness}Completeness}

\newcommand\englishvordergliednoindex
{antecedent}






\newcommand\englishwertevorrat
{range of values}

\newcommand\englishwertsystemdervariablen 
{\index{valuation}valuation}




\newcommand\germanwertverteilungdervariablen
{Wertverteilung der Variablen}
\newcommand\englishwertverteilungdervariablen
{\index{\englishverteilungvonwahrheitswerten}%
 \index{\germanwertverteilungdervariablen}%
 \englishverteilungvonwahrheitswerten}
\newcommand\englishprepwertverteilungauf
                                   {\englishprepverteilungvonwahrheitswertenauf}

\newcommand\englishwertverteilungdervariablenexplizit
{\index{\englishverteilungvonwahrheitswerten}%
 \index{\germanwertverteilungdervariablen}%
 \englishverteilungvonwahrheitswerten\
 \englishprepwertverteilungauf\ the variables}

\newcommand\englishWiderspruchsfreiheit   {Consistency}

\newcommand\englishproblemderwiderspruchsfreiheitindex
                                                 {\index{consistency!problem of}}

\newcommand\englishProblemderWiderspruchsfreiheitnoindex
                                        {Problem of \englishWiderspruchsfreiheit}
\newcommand\englishProblemderWiderspruchsfreiheit
{\englishproblemderwiderspruchsfreiheitindex
                                   \englishProblemderWiderspruchsfreiheitnoindex}


\newcommand\englishZahlenTheorienoindex   {Number Theory}  

\newcommand\englishZahlenTheorie    
                              {\index{number theory}\englishZahlenTheorienoindex}

\newcommand\englishzahligidentisch[1]     
{\math{#1}-identical}
     
\newcommand\englishzahligidentischeins[1] 
{\englishzahligidentisch 1}
\newcommand\englishzahligidentischzwei    
{\englishzahligidentisch 2}
\newcommand\englishzahligidentischdrei    
{\englishzahligidentisch 3}
\newcommand\englishzahligidentischvier    
{\englishzahligidentisch 4}

\newcommand\englishzeichen                {symbol}



\newcommand\formulae                      {formulas}


\newcommand\hbsectionhelper[1]{{\large\par\noindent\LINEnomath{{\bf #1}}\par}}
\newcommand\hbsubsectionhelper[1]{{\par\noindent\LINEnomath{{\bf #1}}\par}}
\newcommand\hbsection[2]{\hbsectionhelper{\S\,\,#1.~~~#2}}
\newcommand\hbsubsection[2]{\hbsubsectionhelper{#1.~~~#2}}
\newcommand\thehbsection[2]{\hbsection{#1}{\csname hbsection#1#2\endcsname}}
\newcommand\thehbsubsection
[3]{\hbsubsection{#2}{\csname hbsubsection#1s#2#3\endcsname}}%
\newcommand\sethbsection[4]{%
\expandafter\newcommand\csname hbsection#1#2\endcsname{#3}%
\expandafter\newcommand\csname tochbsection#1#2\endcsname{#4}%
}
\newcommand\sethbsubsection[5]{%
\expandafter\newcommand\csname hbsubsection#1s#2#3\endcsname{#4}%
\expandafter\newcommand\csname tochbsubsection#1s#2#3\endcsname{#5}%
}
\newcommand\tochbsection[3]{\contentsline
 {section}{\numberline{\S\,\,#1.}{\csname tochbsection#1#2\endcsname}}{#3}{}}
\newcommand\tochbsubsection[3] 
{\contentsline{subsection}{\numberline{(#1)}{#2}}{#3}{}}
\newcommand\tochbIIsubsection[4]
{\contentsline
 {subsection}{\numberline{#2.}{\csname tochbsubsection#1s#2#3\endcsname}}{#4}{}}

\sethbsection{1}{I}
{The \label
{problemderwiderspruchsfreiheit1}\englishProblemderWiderspruchsfreiheit\ 
in Axiomatics \nlnomath
~~~as a Logical \englishEntscheidungsProblem}
{The \englishProblemderWiderspruchsfreiheit\ in Axiomatics
as a \\ Logical \englishEntscheidungsProblem}

\sethbsection{2}{I}
{Elementary \englishZahlenTheorie. 
 \ --- \ 
 \englishFinit\
 Inference
 \nlnomath
 ~~~and its Limits}
{Elementary \englishZahlenTheorie. 
 \ --- \ 
 \englishFinit\
 Inference
 and its Limits \ }

\sethbsection{3}{I}
{Formalization of Logical Inference I: 
 \nlnomath
 ~~~The \englishAussagenKalkul\edfootnotemark{45I1}}
{Formalization of Logical Inference I: \ 
 The \englishAussagenKalkul}

\sethbsection{4}{I}
{Formalization of Logical Inference II: 
 \nlnomath
 ~~~The \englishPraedikatenKalkul\edfootnotemark{86I1}}
{Formalization of Logical Inference II: \
 The \englishPraedikatenKalkul}

\sethbsection{5}{I}
{Adding the\edfootnotemark{163I4}  Identity. \  
 \englishVollstaendigkeit\ of the
 \nlnomath
 ~~~\englishEinstelligerPraedikatenKalkul}
{Adding the Identity.\\
 \englishVollstaendigkeit\  of the \englishEinstelligerPraedikatenKalkul}

\sethbsection{6}{I}
{\englishWiderspruchsfreiheit\ 
for 
Infinite \englishBereiche\ of Individuals.
 \nlnomath
 ~~~Beginnings of \englishZahlenTheorie.}
{\englishWiderspruchsfreiheit\
for 
Infinite \englishBereiche\ of Individuals.
\\Beginnings of \englishZahlenTheorie}

\sethbsection{7}{I}
{The Recursive Definitions}
{The Recursive Definitions}

\sethbsection{8}{I}
{The Notion ``\englishderjenigewelcher'' and its Eliminability}
{The Notion ``\englishderjenigewelcher'' and its Eliminability}

\sethbsection{1}{II}
{The Method of Eliminating the Bound Variable
 \nlnomath 
 with the help of
 \hilbert's \math\varepsilon-Symbol.}
{The Method of Eliminating the Bound Variables\\with the help of 
 \hilbert's \math\varepsilon-Symbol}

\sethbsubsection{1}{1}{II}
{The process of the symbolic 
 \label{resolution page}%
 \englishaufloesung\ of 
 \label{existential formula page}%
 \englishexistentialformeln.}
{The process of the symbolic \englishaufloesung\ of \englishexistentialformeln}

\sethbsubsection{1}{2}{II}
{\hilbertsepsilon-symbol and the \math\varepsilon-formula.}
{\hilbertsepsilon-symbol and the \math\varepsilon-formula}

%% file: headerunits.tex
\usepackage{eurosym}
\newcommand\myeuro
{\ifmmode
 \mathchoice{\mbox{\euro}}
            {\mbox{\euro}}
            {\mbox{\scriptsize\euro}}
            {\mbox{\tiny\euro}}
 \else\euro\fi}




\newcommand\km{{\rm\,km}}

\mathcommand\kubikm{{\rm\,m}^3}
\mathcommand\qm{{\rm\,m}^2}
\mathcommand\qmbf{\mathbf{{\rm\bf\,m}^2}}

\mathcommand\gradCelsius{{\rm\,^\circ C}}

\newcommand\kByte{{\rm\,kByte}}

%% file: specialfonts.tex
\let\oefranz\oe

\mathchardef\Gammaoffont="7000
\mathchardef\Gamma="0100
\mathchardef\Deltaoffont="7001
\mathchardef\Delta="0101
\mathchardef\Thetaoffont="7002
\mathchardef\Theta="0102
\mathchardef\Lambdaoffont="7003
\mathchardef\Lambda="0103
\mathchardef\Xioffont="7004
\mathchardef\Xi="0104
\mathchardef\Pioffont="7005
\mathchardef\Pi="0105
\mathchardef\Sigmaoffont="7006
\mathchardef\Sigma="0106
\mathchardef\Upsilonoffont="7007
\mathchardef\Upsilon="0107
\mathchardef\Phioffont="7008
\mathchardef\Phi="0108
\mathchardef\Psioffont="7009
\mathchardef\Psi="0109
\mathchardef\Omegaoffont="700A
\mathchardef\Omega="010A
\mathchardef\itype="017B

\catcode`\@=11

\gdef\allowhyphens{\penalty\@M \hskip\z@skip}

\gdef\set@low@box#1{\setbox\tw@\hbox{,}\setbox\z@\hbox{#1}\dimen\z@\ht\z@
     \advance\dimen\z@ -\ht\tw@
     \setbox\z@\hbox{\lower\dimen\z@ \box\z@}\ht\z@\ht\tw@ \dp\z@\dp\tw@ }
\gdef\set@low@boxsingle#1{\setbox\tw@\hbox{\rm,}\setbox\z@\hbox{#1}\dimen\z@\ht\z@
     \advance\dimen\z@ -\ht\tw@
     \setbox\z@\hbox{\lower\dimen\z@ \box\z@}\ht\z@\ht\tw@ \dp\z@\dp\tw@ }

\gdef\@glqq{%
\ifhmode\edef\@SF{\spacefactor\the\spacefactor}%
\else\let\@SF\empty
\fi
\CheckFamily\font\fraknomath\ifSameFamily ``\relax
\else\CheckFamily\font\swab\ifSameFamily ``\relax
\else\leavevmode\set@low@box{''}\box\z@\kern-.04em\allowhyphens\@SF\relax
\fi\fi}
\gdef\glqq{\protect\@glqq\kern+.07em}
\gdef\@grqq{%
\ifhmode\edef\@SF{\spacefactor\the\spacefactor}%
\else\let\@SF\empty 
\fi 
\CheckFamily\font\fraknomath\ifSameFamily ''\relax
\else\CheckFamily\font\swab\ifSameFamily ''\relax
\else\kern+.07em``\kern.07em\@SF\relax
\fi\fi}
\gdef\grqq{\protect\@grqq}
\gdef\@glq{{\ifhmode \edef\@SF{\spacefactor\the\spacefactor}\else
     \let\@SF\empty \fi \leavevmode
     \set@low@boxsingle{'\/}\box\z@\kern-.04em\allowhyphens\@SF\relax}}
\gdef\glq{\protect\@glq\kern+.07em}
\gdef\@grq{\ifhmode \edef\@SF{\spacefactor\the\spacefactor}\else
     \let\@SF\empty \fi \kern-.0125em`\kern.07em\@SF\relax}
\gdef\grq{\protect\@grq}

\catcode`\@=12

\newcommand\closequotecommanospace{''\nolinebreak\hskip-0.23em,}
\newcommand\closequotecomma      {\closequotecommanospace\         \,}
\newcommand\closequotecommasmallextraspace{\closequotecommanospace\ \,\,}
\newcommand\closequotecommaextraspace{\closequotecommanospace\   \ \,}
\newcommand\closequotefullstop   {\closequotefullstopnospace\      \,}
\newcommand\closequotefullstopextraspace   
                                 {\closequotefullstopnospace\    \ \,}

\newcommand\closequotefullstopnospace
                                 {''\nolinebreak\hskip-0.20em\@.}

\newcommand\closesinglequotecomma{'\nolinebreak\hskip-0.20em,      \,}

\newcommand\closesinglequotefullstopnospace{'\nolinebreak\hskip-0.20em\@.}

\newcommand\commanospace          {,\nolinebreak\hskip-0.23em}
\newcommand\fullstopnospace       {\@.\nolinebreak\hskip-0.23em}

\makeatletter
\if \@ptsize 0
   \newfont{\scriptscriptscriptgoth}{ygoth scaled 760}
   \newfont{\scriptscriptgoth}{ygoth scaled 833}
   \newfont{\scriptgoth}{ygoth scaled 912}
   \newfont{\gothnomath}{ygoth}
   \newfont{\Goth}{ygoth scaled \magstephalf}
   \newfont{\GOth}{ygoth scaled \magstep1}
   \newfont{\GOTh}{ygoth scaled \magstep2}
   \newfont{\GOTH}{ygoth scaled \magstep3}

   \newfont{\scriptscriptscriptswab}{yswab scaled 760}
   \newfont{\scriptscriptswab}{yswab scaled 833}
   \newfont{\scriptswab}{yswab scaled 912}
   \newfont{\swab}{yswab}
   \newfont{\Swab}{yswab scaled \magstephalf}
   \newfont{\SWab}{yswab scaled \magstep1}
   \newfont{\SWAb}{yswab scaled \magstep2}
   \newfont{\SWAB}{yswab scaled \magstep3}

   \newfont{\scriptscriptscriptfrak}{yfrak scaled 760}
   \newfont{\scriptscriptfrak}{yfrak scaled 833}
   \newfont{\scriptfrak}{yfrak scaled 912}
   \newfont{\fraknomath}{yfrak}
   \newfont{\Frak}{yfrak scaled \magstephalf}
   \newfont{\FRak}{yfrak scaled \magstep1}
   \newfont{\FRAk}{yfrak scaled \magstep2}
   \newfont{\FRAK}{yfrak scaled \magstep3}

   \newfont{\init}{yinit}
   \newfont{\Init}{yinit scaled \magstephalf}
   \newfont{\INit}{yinit scaled \magstep1}
   \newfont{\INIt}{yinit scaled \magstep2}
   \newfont{\INIT}{yinit scaled \magstep3}
\fi
\if \@ptsize 1
   \newfont{\scriptscriptscriptgoth}{ygoth scaled 833}
   \newfont{\scriptscriptgoth}{ygoth scaled 912}
   \newfont{\scriptgoth}{ygoth}
   \newfont{\gothnomath}{ygoth scaled \magstephalf}
   \newfont{\Goth}{ygoth scaled \magstep1}
   \newfont{\GOth}{ygoth scaled \magstep2}
   \newfont{\GOTh}{ygoth scaled \magstep3}
   \newfont{\GOTH}{ygoth scaled \magstep4}

   \newfont{\scriptscriptscriptswab}{yswab scaled 833}
   \newfont{\scriptscriptswab}{yswab scaled 912}
   \newfont{\scriptswab}{yswab}
   \newfont{\swab}{yswab scaled \magstephalf}
   \newfont{\Swab}{yswab scaled \magstep1}
   \newfont{\SWab}{yswab scaled \magstep2}
   \newfont{\SWAb}{yswab scaled \magstep3}
   \newfont{\SWAB}{yswab scaled \magstep4}

   \newfont{\scriptscriptscriptfrak}{yfrak scaled 833}
   \newfont{\scriptscriptfrak}{yfrak scaled 912}
   \newfont{\scriptfrak}{yfrak}
   \newfont{\fraknomath}{yfrak scaled \magstephalf}
   \newfont{\Frak}{yfrak scaled \magstep1}
   \newfont{\FRak}{yfrak scaled \magstep2}
   \newfont{\FRAk}{yfrak scaled \magstep3}
   \newfont{\FRAK}{yfrak scaled \magstep4}

   \newfont{\init}{yinit scaled \magstephalf}
   \newfont{\Init}{yinit scaled \magstep1}
   \newfont{\INit}{yinit scaled \magstep2}
   \newfont{\INIt}{yinit scaled \magstep3}
   \newfont{\INIT}{yinit scaled \magstep4}
\fi
\if \@ptsize 2
   \newfont{\scriptscriptscriptgoth}{ygoth scaled 912}
   \newfont{\scriptscriptgoth}{ygoth}
   \newfont{\scriptgoth}{ygoth scaled \magstephalf}
   \newfont{\gothnomath}{ygoth scaled \magstep1}
   \newfont{\Goth}{ygoth scaled \magstep2}
   \newfont{\GOth}{ygoth scaled \magstep3}
   \newfont{\GOTh}{ygoth scaled \magstep4}
   \newfont{\GOTH}{ygoth scaled \magstep5}

   \newfont{\scriptscriptscriptswab}{yswab scaled 912}
   \newfont{\scriptscriptswab}{yswab}
   \newfont{\scriptswab}{yswab scaled \magstephalf}
   \newfont{\swab}{yswab scaled \magstep1}
   \newfont{\Swab}{yswab scaled \magstep2}
   \newfont{\SWab}{yswab scaled \magstep3}
   \newfont{\SWAb}{yswab scaled \magstep4}
   \newfont{\SWAB}{yswab scaled \magstep5}

   \newfont{\scriptscriptscriptfrak}{yfrak scaled 833}
   \newfont{\scriptscriptfrak}{yfrak}
   \newfont{\scriptfrak}{yfrak scaled \magstephalf}
   \newfont{\fraknomath}{yfrak scaled \magstep1}
   \newfont{\Frak}{yfrak scaled \magstep2}
   \newfont{\FRak}{yfrak scaled \magstep3}
   \newfont{\FRAk}{yfrak scaled \magstep4}
   \newfont{\FRAK}{yfrak scaled \magstep5}

   \newfont{\init}{yinit scaled \magstep1}
   \newfont{\Init}{yinit scaled \magstep2}
   \newfont{\INit}{yinit scaled \magstep3}
   \newfont{\INIt}{yinit scaled \magstep4}
   \newfont{\INIT}{yinit scaled \magstep5}
\fi

\makeatother

\newif\ifSameFamily
\def\CheckFamily#1#2{\GetFamilyName{#1}\ArgOne
        \GetFamilyName{#2}\ArgTwo
        \ifx\ArgOne\ArgTwo\SameFamilytrue\else\SameFamilyfalse\fi}
\def\GetFamilyName#1{\edef\Tempa{#1}\def\Tempb{#1}\ifx\Tempa\Tempb
        \edef\Tempa{\fontname#1}\fi
        \edef\Tempa{\Tempa\space}%
        \expandafter\iGetFamilyName\Tempa\\}
\def\iGetFamilyName#1 #2\\#3{\def#3{#1}}
\def\DefFontName#1#2{{\escapechar-1\expandafter\expandafter\expandafter
        \iDefFontName\expandafter{\csname#2\endcsname}%
        \xdef#1{\expandafter\string\Tempa}}}
\def\iDefFontName{\def\Tempa}

\newcommand\unprotectedae
{\CheckFamily\font\fraknomath\ifSameFamily *a\else
 \CheckFamily\font\swab\ifSameFamily\char'212\else\"a\fi\fi}
\newcommand\unprotectedoe
{\CheckFamily\font\fraknomath\ifSameFamily 
*o\else\CheckFamily\font\swab\ifSameFamily\char'232\else\"o\fi\fi}
\newcommand\unprotectedue
{\CheckFamily\font\fraknomath\ifSameFamily 
*u\else\CheckFamily\font\swab\ifSameFamily\char'237\else\"u\fi\fi}

\DefFontName\eccclarge{eccc1200}
\DefFontName\eccc{eccc1000}
\DefFontName\ecccsmall{eccc0900}
\DefFontName\ecccfootnotesize{eccc0800}

\newcommand\unprotectedes
{\CheckFamily\font\fraknomath\ifSameFamily\char'215\else
\CheckFamily\font\swab\ifSameFamily\char'215\else  
s\fi\fi}

\newcommand\unprotectedesi
{\CheckFamily\font\fraknomath\ifSameFamily\char'215\else
\CheckFamily\font\swab\ifSameFamily\char'215\else  
\mbox{s\hskip.04em}\fi\fi
\-\ignorespaces}

\newcommand\unprotectedmyparagraphsymbol
{\CheckFamily\font\fraknomath\ifSameFamily 
\char'244\else\CheckFamily\font\swab\ifSameFamily
\char'244\else\S\fi\fi}
\newcommand\unprotectedzwithdot
{\CheckFamily\font\fraknomath\ifSameFamily 
\accent'137z\else\CheckFamily\font\swab\ifSameFamily z\else\.z\fi\fi}
\newcommand\unprotectedewithtrema
{\-\CheckFamily\font\fraknomath\ifSameFamily 
 "e\else\CheckFamily\font\swab\ifSameFamily "e\else\"e\fi\fi}

\renewcommand\ae{\protect\unprotectedae}
\renewcommand\oe{\protect\unprotectedoe}
\newcommand\ue  {\protect\unprotectedue}

\newcommand\es  {\protect\unprotectedes}

\newcommand\esi {\protect\unprotectedesi}  

\newcommand\fhi {\discretionary{f-} {h}{f\mbox {\hskip.08em}h}}

\newcommand\myparagraphsymbol{\protect\unprotectedmyparagraphsymbol}
\newcommand\zwithdot{\protect\unprotectedzwithdot}
\newcommand\ewithtrema{\protect\unprotectedewithtrema}


%% file: body.tex
\section{A Snapshot of a Decisive Moment in History}%
\label{section Preface}%
\figurewaterfall{t}%
The automation of mathematical theorem proving 
for deductive {\em first-order logic}\/ started in the 1950s, \hskip.1em
and it took about half a century 
to develop software systems that are
sufficiently strong and general to be successfully applied
outside the community of automated theorem proving.\footnote{%
 The currently (\ie\ in 2012) 
 most successful first-order automated theorem prover 
 is \VAMPIRE, \cfnlb\ \eg\ \cite{vampire01}.%
} \hskip.1em 
For more restricted logic languages, \hskip.1em
such as {\em propositional logic}\/ and 
the {\em purely equational fragment}, \hskip.2em
such strong systems were not achieved much earlier.\footnote{%
 \majorheadroom
  A breakthrough toward industrial strength
 in deciding propositional validity
 (\ie\,sentential\,validity) 
 (or\,its\,dual:\,\,propositional\,satisfiability) 
 (which\,are\,decidable,\,\,but\,%
 NP-complete) \hskip.1em
 was the SAT solver \nolinebreak{\sc Chaff},
 \cfnlb\ \eg\ \cite{chaff}. \hskip.3em
 The most successful automated theorem prover 
 for purely 
 equational
 logic is \WALDMEISTER, \cfnlb\,\eg\,\cite{waldmeister},\,\,\cite{HL02}.%
} \hskip.3em
Moreover, 
automation of theorem proving for {\em higher-order logic}\/
has started becoming generally useful only during the last
ten years.\footnote{\label{note higher-order theorem provers}%
 \majorheadroom
 Driving forces in the automation of higher-order 
 theorem proving are the 
 TPTP-competition-winning
 systems \LEOII\ (\cfnlb\ \eg\ \cite{C26})
 and \SATALLAX\ (\cfnlb\ \eg\ \cite{satallax}).%
}%

\begin{figure}[t]%
\noindent\LINEnomath{\includegraphics
[scale=.42,bb=25 4 588 788]{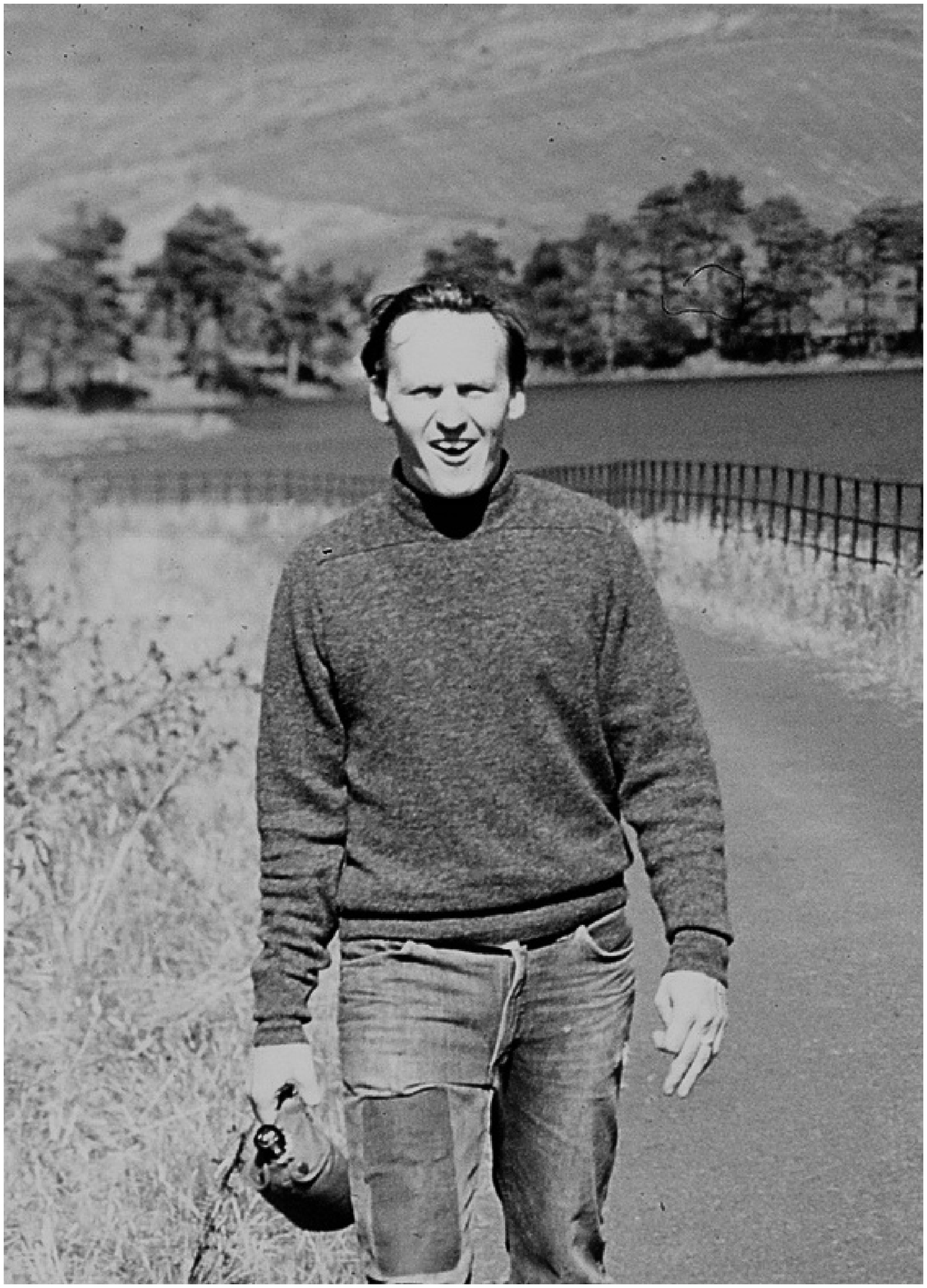}%
\hfill\includegraphics
[scale=.42,bb=64 0 548 792]{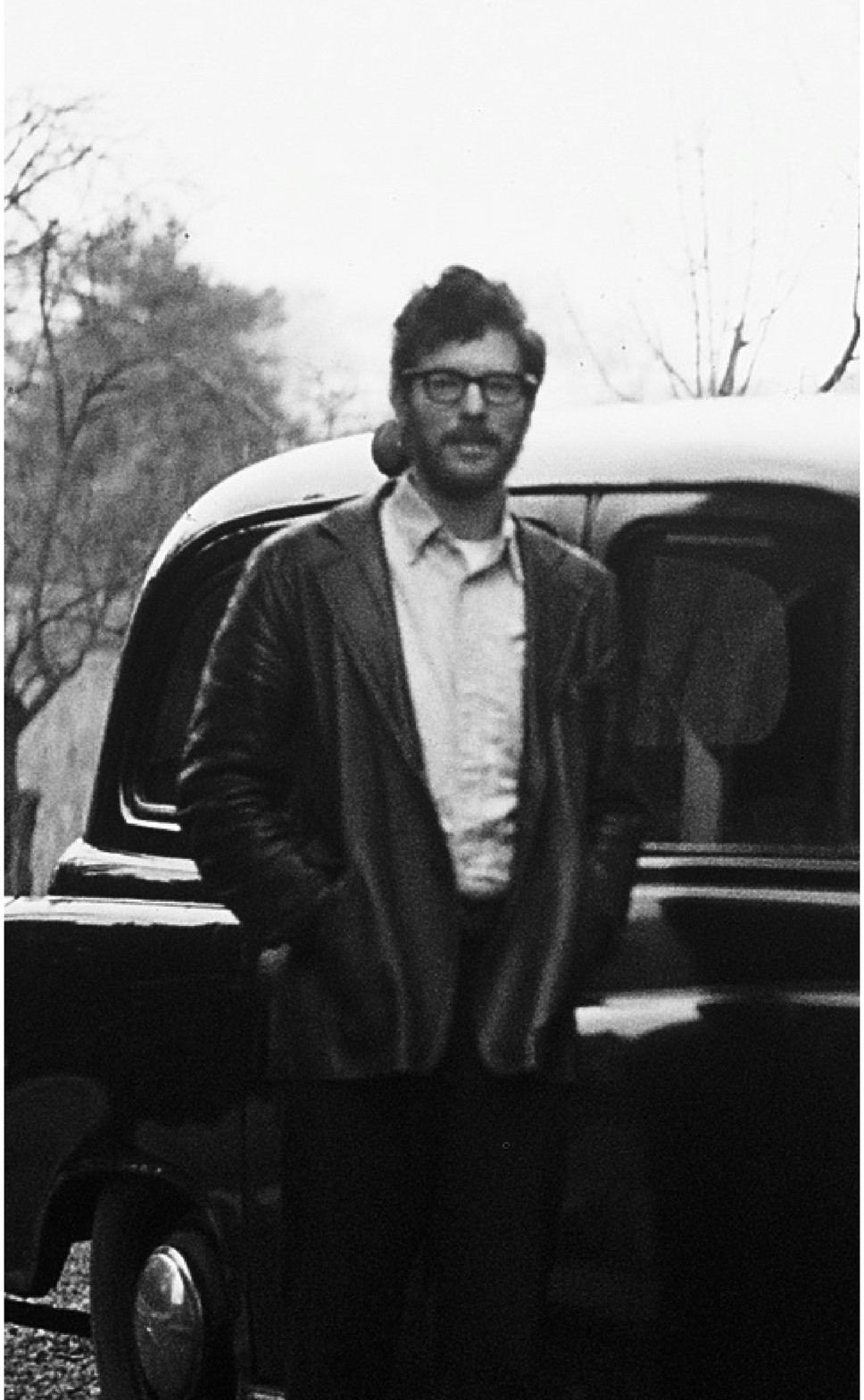}%
}\notop\notop\halftop\halftop\caption
{\boyername\ (1971) (l.)\ and \moorename\ (1972?) (r.)}\yestop\yestop\halftop
\end{figure}%

In this context,
it is 
surprising that 
for the field of quantifier-free first-order \mbox{\em inductive}\/
theorem proving 
based on recursive functions, \hskip.1em
most of the progress 
\mbox{toward}
general usefulness
took place within the 1970s and 
that usefulness was 
clearly 
demonstrated by\,1986.\footnote{%
 \majorheadroom
 See the last paragraph of \sectref{subsection NQTHM}.%
}

\pagebreak

In this article we describe how this 
leap
took place,
and sketch the
further development of automated inductive theorem proving.

The work on this breakthrough in the automation of inductive theorem proving
was started 
in September\,1972, 
by \boyername\ and \moorename, in \EB, Scotland.  
Unlike earlier work on theorem proving,
\boyer\ and \moore\ chose to make induction the focus of 
their 
work.
Most of the crucial steps and their synergetic combination in
the ``waterfall''\footnotemark\
of their
now famous 
theorem provers 
were developed in the span of a single year
and implemented in their ``\PURELISPTP\closequotecomma
presented at \IJCAI\ in \StanfordCA\ in August\,1973,\footnotemark\
and
documented in \moore's \PhDthesis\
\shortcite{moore-1973},
defended in November\,1973.

\addtocounter{footnote}{-1}\footnotetext{%
 See \figuref{figure waterfall} for the \boyermoorewaterfall.
 See \cite{waterfall} for the probably first occurrence of 
 ``waterfall'' as a term in software engineering.
 \boyer\ and \moore, however, 
 were inspired not by this metaphor from 
 software engineering, but again by a real waterfall, 
 as can be clearly seen 
 from \cite[\p\,89]{bm}:\notop\halftop\begin{quote}``%
 A good  metaphor for the organization of these heuristics is
 an initially dry waterfall.  One pours out a clause at the top.
 It trickles down and is split into pieces.  Some pieces evaporate as
 they are proved.  Others are further split up and simplified.  Eventually
 at the bottom a pool of clauses forms whose
 conjunction suffices to prove the original formula.%
 ''\notop\halftop\end{quote}%
}
\addtocounter{footnote}{1}\footnotetext{%
 \majorheadroom
 \Cfnlb\ \cite{boyer-moore-1973}.%
}%
Readers who take a narrow view on the automation of inductive theorem proving
might be surprised that we discuss the waterfall.  
It is impossible, however, 
to build a good inductive
theorem prover without considering how 
to transform the induction conclusion
into the hypothesis
(or, alternatively, how 
to recognize 
that a legitimate 
induction hypothesis 
can dispatch a
subgoal)\@. \hskip.3em 
So we take the expansive view and discuss not just the induction principle and
its heuristic control, 
but 
also
the waterfall architecture that is effectively an integral part of 
\mbox{%
the
success.}

\halftop\halftop\halftop\indent
\boyer\ and \moore\ had met in August\,1971, \hskip.1em
a year before the induction work started, \hskip.15em
when \boyer\ took up the position of a post-doctoral 
research fellow 
at \nolinebreak the \nolinebreak Metamathematics Unit of the 
University of \EB. \hskip.35em
Moore was at
that time starting the second year of his \PhD\ studies in 
``the Unit\closequotefullstopextraspace
Ironically, \hskip.1em
they \nolinebreak 
were both from Texas and they had both come to \EB\ from \theMIT\@. 
\boyer's PhD supervisor, \hskip.1em
\bledsoename,  \hskip.2em
from \unitexasaustin, \hskip.2em
\mbox{spent 1970--71} on sabbatical at \theMIT, \hskip.1em
and \boyer\ accompanied him and
completed his PhD work there. \hskip.1em
\moore\ got his bachelor's degree at \theMIT\
(1966--70) before going to \EB\ for his PhD\@.%

\begin{sloppypar}\par\indent
Being ``warm blooded Texans\closequotecommasmallextraspace 
they shared an office in the 
\mbox{Metamathematics} Unit at 
\linebreak 
9~\HopeParkSquare, 
Meadow Lane. \hskip.2em
The \nth{18}\,century buildings at 
\HopeParkSquare\ were 
\linebreak
the center of \AI\ research in Britain at
a \nolinebreak time when the promises of AI were seemingly just on the
horizon.\footnote{\label{note people and departments}%
 \footnoteonMetamathematicsUnitandMembersofUnivEB} \hskip.2em
In addition to 
mainline work on mechanized reasoning by \hskip.1em
\burstallname, \hskip.1em
\kowalskiname, \hskip.1em
\hayesname, \hskip.1em
\plotkinname, \hskip.1em
\moorename, \hskip.1em
\gordonname, \hskip.1em
\boyername, \hskip.1em
\bundyname, \hskip.1em
and (by\,1973) \hskip.1em
\milnername, \hskip.2em
there was work on new programming paradigms, 
program transformation and synthesis,
natural language,
machine vision,
robotics,
and cognitive modeling. \hskip.2em
\HopeParkSquare\ received a steady 
stream of distinguished visitors from around
the world, \hskip.2em
including \hskip.1em
\robinsonname, \hskip.2em
\mccarthyname, \hskip.2em
\bledsoename, \hskip.2em
\scottname, \hskip.2em
and \hskip.1em
\minskyname. \hskip.3em 
An eclectic series of seminars were on offer
weekly \mbox{to complement} the daily tea times, \hskip.1em
where all researchers gathered
around a table and talked about their \mbox{current problems.}%
\end{sloppypar}

\halftop\halftop\indent
\boyer\ and \moore\ initially worked together on structure sharing in 
\mbox{resolution} theorem proving. 
The inventor of resolution, 
\mbox{\robinsonname~\robinsonlifetime
,} 
created 
and 
awarded them
the\,``1971\,Programming\,Prize''\,on 
\mbox{December\,17,\,1971~---~half} 
{jokingly,} \hskip.1em
half seriously. 
The document, 
handwritten by \robinson, 
\mbox{actually\,says\,in\,part:}\begin{quote}``In 1971,
   the prize is awarded, by unanimous agreement of the Board, 
   to \boyername\ and \moorename\ for their idea, explained in 
   \cite{boyer-moore-1971},
   of representing clauses as their own genesis.
   The Board declared,
   on making the announcement of the award,
   that this idea is `\ldots\ bloody marvelous'.''%
\end{quote}\begin{sloppypar}%
\par\noindent
Their 
structure-sharing
representation of derived clauses in a 
\index{linear resolution}linear resolution 
\mbox{system} is just a stack of resolution steps. \hskip.2em
This suggests the
idea of \mbox{resolution}
being \nolinebreak a \nolinebreak kind \nolinebreak of 
``procedure call.''\footnotemark\
\hskip.6em
Exploiting structure sharing, \hskip.1em
\boyer\ and \moore\ implemented a declarative \LISP-like programming
language called ``\BAROQUE'' \cite{moore-1973}, \hskip.1em
a precursor to \PROLOG.\footnotemark\
\hskip.3em
They then
implemented a \LISP\ 
inter\-preter in \BAROQUE\ and began to use their resolution engine
to prove simple theorems about programs in
\LISP\nolinebreak\hspace*{-.1em}\nolinebreak\@. \hskip.3em  
Resolution
was sufficient to prove such theorems \nolinebreak as 
\hskip.1em ``there 
is 
a list whose length is 3\closequotecommasmallextraspace
whereas
the absence of a rule 
for 
induction
prevented the 
proofs of more interesting theorems like the associativity of 
list
concatenation.%
\end{sloppypar}

\halftop\indent
So, in the summer of 1972, they turned their attention to a theorem prover
designed explicitly to do mathematical induction --- this at a time when
uniform first-order 
proof procedures were all the rage. \hskip.3em 
The fall of 1972
found them taking turns at the blackboard proving theorems about recursive
\LISP\ functions and articulating their reasons for each proof step. \hskip.2em
Only after several months of such \nolinebreak proofs 
did they sit down together to write the
code for the {\PURELISPTP}.

\halftop\indent
Today's
readers might have difficulty imagining the computing infrastructure
in Scotland in the early 1970s.  \boyer\ and \moore\ developed their software
on \eineenglishEBPURELISPMACHINE, \hskip.1em
with \nlbmath{64\kByte} (\nlbmath{128\kByte} in 1972) 
core memory (RAM)\@. \hskip.2em
Paper tape was used for archival storage. \hskip.1em
The machine was physically located
in the \ForrestHillEB\ building of the University of \EB, \hskip.1em
about 1\km\ from \HopeParkSquare. \hskip.25em
A rudimentary time-sharing system allowed several
users at once to run lightweight applications from teletype machines at 
\HopeParkSquare. \hskip.2em
The only high-level programming language supported was {\sc POP--2}, \hskip.2em
a simple stack-based list-processing language with an 
{\sc Algol}-like syntax.\footnotemark
\addtocounter{footnote}{-2}\footnotetext{%
 \Cfnlb\ \cite[Part\,I, \litsectref{6.1}, \PP{68}{69}]{moore-1973}.%
}%
\addtocounter{footnote}{1}\footnotetext{%
 \majorheadroom
 For \BAROQUE\ 
 see \cite[Part\,I, \litsectrefs{6.2}{6.3}, \PP{70}{75}]{moore-1973}.
 \hskip.3em
 For logic programming and \PROLOG\ see
 \cite[Part\,I, \litchapref{6}, \PP{68}{75}]{moore-1973}, \hskip.2em
 \makeaciteoftwo{Kow74}{kowalski-1988}, \hskip.2em 
 and 
 \cite{Prolog}.%
}%
\addtocounter{footnote}{1}\footnotetext{%
 \majorheadroom
 \Cfnlb\ \cite{pop2}.%
}%

\pagebreak

Programs were prepared with a primitive text editor modeled on a paper tape
editor: 
A \nolinebreak disk file could 
be
copied through a one byte buffer to an output file. \hskip.2em
By halting the copying and typing characters into or deleting characters from
the buffer one could edit a file 
---~a process that usually took several passes.
\hskip.3em
Memory limitations of the \mbox\englishEBPURELISPMACHINE\ 
prohibited storing large files in memory
for \nolinebreak editing. \hskip.1em
In \nolinebreak their very early 
collaboration, \hskip.1em
\boyer\ and \moore\ solved this
problem by inventing what has come to be called the 
``piece table\closequotecommasmallextraspace
whereby
an edited document is represented by a linked list of ``pieces'' referring to
the original file which remains on disk. \hskip.3em
Their ``77-editor'' \cite{boyer-moore-davies-1973} 
\hskip.05em (written in
1971 and named for the disk track on which it resided) \hskip.1em
provided an interface like \MIT's Teco, \hskip.1em
but with {\sc POP--2} as the command language.\footnote{%
 The 77-editor was widely used by researchers at \HopeParkSquare\ 
 until the \mbox\englishEBPURELISPMACHINE\ was decommissioned. 
 When \moore\
 went to Xerox PARC 
 in Palo Alto (CA)
 (\Dec\,1973), \hskip.2em 
 the \boyermoore\ representation \cite{moore-1981} was adopted by
 \simonyiname\ \simonyilifetime\ for the Bravo editor on the Alto 
 and subsequently found its way into
 Microsoft Word, \cfnlb\ \cite{simonyi-interview}.%
} \hskip.4em
It was thus with their own editor that \boyer\ and \moore\
wrote the code for the {\PURELISPTP}.

\halftop\halftop\indent
During the day 
\boyer\ and \moore\
worked at \HopeParkSquare, \hskip.1em
with frequent trips by foot
or bicycle through The Meadows to \ForrestHillEB\ to make archival paper tapes
or to pick up line-printer output. \hskip.3em
During the night
---~when they could often have
the \englishEBPURELISPMACHINE\ to themselves~--- 
they often worked at \boyer's home where another
teletype was available.
\notop\halftop\section{Method of Procedure and Presentation}
The excellent survey articles \cite{waltherhandbook} and \cite{bundy-survey}
cover the engineer\-ing and research problems and current 
standards of the field of \mbox{{\em explicit induction}.} \hskip.2em
To \nolinebreak 
cover {\em the \nolinebreak history of the automation of mathematical induction},
we need a wider scope in mathematics and more historical detail.
To keep \englishdiesespapier\ within reasonable limits,
we have to focus more narrowly on those 
developments and systems
which are the respectively first successful and \mbox{historically most 
important ones.}

\halftop\halftop\indent
It is always hard to see the past because we look through the lens of the
present.  Achieving the necessary detachment from the present is especially
hard for the historian of recent history because the ``lens of the present''
is shaped so immediately by the events being studied.

\halftop\halftop\indent
We try to mitigate this problem by avoiding the standpoint
of a disciple of the leading school of explicit induction.
Instead, we put the historic achievements into a broad
mathematical context and a space of time from the ancient Greeks to
a possible future, based on 
a most general approach to {\em recursive definition}\/
(\cfnlb\ \sectref{section Recursive Definitions}), \hskip.3em
and on 
{\em\descenteinfinienoindex}\/ 
as a general, implementation-neutral approach to mathematical induction
(\cfnlb\ \sectref{subsection DescenteInfinie}). \hskip.2em
Then we \nolinebreak can see the 
achievements in the field
with the surprise 
they historically deserve
---
after all, 
until 1973 mathematical induction was considered too creative an activity to be automated.

\pagebreak

As a historiographical text, \hskip.1em 
\englishdiesespapier\ 
should be accessible to an audience that goes
beyond the technical experts and programmers of the day, \hskip.1em 
should use common mathematical language and representation, \hskip.1em 
focus on the global and eternal ideas and their developments, \hskip.1em 
and 
paradigmatically 
display the {\em historically most significant achievements}.%

\halftop\halftop\indent
Because these achievements 
in the automation of inductive theorem proving 
manifest themselves 
for the first time 
mainly in the line of 
the \boyermooretheoremprovers, 
we cannot avoid the 
confrontation of the reader with some more ephemeral forms of representation
found in these software systems. \hskip.2em
In particular,
we cannot avoid some small expressions 
in the list programming language 
\nolinebreak\LISP\nolinebreak\hspace*{-.1em}\nolinebreak,\footnote{%
 \Cf\ \cite{LISP}. \hskip.2em
 Note that we use the historically correct capitalized 
 ``\LISP'' for general reference,
 but not for more recent, special dialects such as \COMMONLISP.%
} \hskip.1em
simply because the \boyermooretheoremprovers\ we discuss in 
\englishdiesespapier,
namely the \PURELISPTP, \THM, \NQTHM, and \ACLTWO, \hskip.1em
all have {\em logics}\/ based on a subset of \LISP\@. \hskip.2em

\halftop\halftop\indent
Note that we do not necessarily refer to the {\em implementation language}\/ 
of these  
software systems,
but to the {\em logic language}\/ used both for representation of formulas
and for communication with the user.

\halftop\halftop\indent
For the first system in this line of development, 
\boyer\ and \moore\ had a free choice, \hskip.1em
but 
wrote:
\notop\halftop
\begin{quote}
``We use a subset of \LISP\ as our language
because recursive list processing functions are easy to
write in \LISP\ and because theorems can be naturally
stated in \LISP; \hskip.2em
furthermore, \LISP\ has a simple syntax
and is universal in Artificial Intelligence. We
employ a \LISP\ interpreter to `run' our theorems and a
heuristic which produces induction formulas from information
about how the interpreter fails.  We combine
with the induction heuristic a set of simple rewrite
rules of \LISP\ and a heuristic for generalizing the
theorem being proved.''\footnotemark
\end{quote}

\noindent
\footnotetext{%
 \majorheadroom
 \Cf\ \cite[\p\,486, left column]{boyer-moore-1973}.
}%
Note that the choice of \LISP\ was influenced by the \role\ of
the \LISP\ interpreter in \mbox{induction}\@.  
\LISP\ was important for another reason:  
\boyer\ and \moore\ were building a {\em computational-logic}\/
theorem prover:
\notop\halftop
\begin{quote}
``The structure of the program is remarkably simple by artificial intelligence
standards.  
This is primarily because the control structure is embedded in the syntax of the theorem.
This means that the system does not contain two languages, 
the `object language\closesinglequotecomma 
\LISP, and the `meta-language\closesinglequotecomma
predicate calculus.  They are identified.  
This mix of computation and deduction was largely inspired by the view that
the two processes are actually identical. \hskip.2em
Bob \kowalskiindex\kowalski, \hayesname, 
and the nature of \LISP\ deserve the credit for this
unified view.''\footnote{%
 \majorheadroom
 \Cfnlb\ \cite[p.\,207\f]{moore-1973}.%
}%
\end{quote}
This view was prevalent in the Metamathematics Unit by 1972\@. \hskip.2em  
Indeed, 
``the Unit''
was by then officially renamed the Department of Computational Logic.%
\arXivfootnotemarkref{note people and departments}

\pagebreak

\halftop\indent
In general, 
inductive theorem proving with recursively defined functions
requires a logic 
in which\notop\begin{quote}
a {\em method of symbolic evaluation}\/ can be obtained from
an interpretation procedure 
by generalizing the ground terms of computation 
to terms with free variables
that are implicitly universally quantified.\notop\halftop\end{quote}
So
candidates 
to be considered
today (besides a subset of \LISP\ or of \math\lambda-calculus)
are the typed functional programming languages 
\ml\ and \HASKELL\commanospace\footnote{%
 \Cfnlb\ \cite{haskell} for \HASKELL, \cite{Pau96} for \ml, 
 which started as the \underline meta-\underline language for implementations of
 \LCF\ (the {\em Logic of Computable Functions}
 with a single undefined element \nlbmath\bot,
 invented by \citet{scott-LCF})
 with structural induction over \nlbmath\bot,
 \nlbmath\zeropp, and \nlbmaths\ssymbol,
 but without original contributions to the automation of induction,
 \cfnlb\ \cite[\p\,8]{milner-1972},\,\cite{fromLCFtoHOL}.%
} \hskip.05em
which, however, were not available in\,\,1972\@. \hskip.4em
\LISP\ and \ml\ are to be preferred to \HASKELL\ as the logic 
of an inductive theorem prover
because of their innermost evaluation strategy,
which gives preference to the 
constructor terms that represent the constructor-based data types, \hskip.1em
which again establish the most interesting domains 
in hard- and software verification
and 
the major elements
of mathematical 
induction.
\par\indent
Yet another candidate today would be the 
rewrite systems of \cite{wgjsc} and \makeaciteoftwo{wirth-master}{wirth-jsc}
\hskip.1em
with 
\index{constructor variables}%
{\em constructor variables}\/\footnote{%
 \majorheadroom
 See \sectref{subsection Constructor Variables} of \englishdiesespapier.%
}
and {\em\pnc\ equations}, \hskip.1em
designed and developed for the specification,
interpretation, 
and symbolic evaluation of recursive functions in the
context of inductive theorem proving in the domain of constructor-based
data types. \hskip.2em
\mbox{Neither} this tailor-made theory,
nor even the general theory of rewrite systems in which its
development is rooted,\footnote{%
 \majorheadroom
 See \cite{term-rewriting-one} for
 the theory in which the rewrite systems of 
 \cite{wgjsc}, \makeaciteoftwo{wirth-master}{wirth-jsc}
 are rooted. 
 One may try to argue that the paper that launched the whole field
 of rewrite systems,
 \cite{KB70}, \hskip.1em
 was already out in\,\,1972,
 but the relevant parts of rewrite theory
 for unconditional equations were 
 developed only in the late 1970s and the 1980s.
 Especially relevant in the given context are \cite{huet} and \cite{toyama}.
 \hskip.2em
 The rewrite theory of {\em\pnc}\/ equations,
 however,
 started to become an intensive area of research only
 with the burst of creativity at 
 \thefirstCTRSeightyseven; \hskip.2em
 \cfnlb\ \cite{firstCTRSeightyseven}.%
}
were available in\,\,1972\@. \hskip.3em
And still today, 
the applicative subset of \COMMONLISP\ that provides
the logic language for \ACLTWO\ \hskip.1em
(= \nlbmath{\inpit{\rm ACL}^2} 
 = \underline A \nolinebreak\underline Computational \underline Logic for 
 \underline Applicative 
 \underline{\sc C}{\sc ommon} \underline{\sc L}{\sc isp}) \hskip.2em
is again 
to be 
preferred to these \pnc\ rewrite systems
for reasons of efficiency:
The applications of \ACLTWO\ in hardware verification and testing 
require a performance that is still at the very limits of
today's computing technology.
This challenging efficiency demand requires, among other aspects,
that the logic of the theorem prover is so close to its 
own programming language that 
---~after certain side conditions have been checked
~---
the theorem prover can defer the interpretation of ground
terms to the analogous interpretation in its own programming language.
\par\indent
For 
most 
of our illustrative 
examples in \englishdiesespapier, however,
we will use the higher flexibility and conceptual adequacy
of \pnc\ rewrite systems. \hskip.1em
They are so close to standard logic that we can 
dispense their semantics to the reader's intuition,\footnotemark\
\hskip.1em
and they can immediately serve as an intuitively clear replacement 
of the {\em\boyermooremachines}.\footnotemark

\pagebreak

\addtocounter{footnote}{-1}%
\footnotetext{%
 The readers interested into the precise details are referred to 
 \cite{wirth-jsc}.%
}\addtocounter{footnote}{1}\footnotetext{%
 \majorheadroom
 \Cf\ \cite[\p\,165\f]{bm}.%
}%
Moreover, the typed (many-sorted) approach of the 
\pnc\ equations 
allows the presentation of formulas in a form that is 
much easier to grasp for human readers than the 
corresponding sugar-free \LISP\ notation with its 
overhead of 
explicit type restrictions.

Another reason for avoiding \LISP\ notation is that
we \nolinebreak want to make it most obvious that the achievements of the
\boyermooretheoremprovers\ are not limited to their \LISP\ logic.

For the same reason, \hskip.1em
we also prefer examples from 
arithmetic to examples from list theory,
which might be considered to be especially supported by the 
\LISP\ logic\@. \hskip.1em
The reader can find the famous examples from list theory 
in almost any other publication on the subject.\footnote{%
 \majorheadroom
 \Cf\ \eg\ 
 \cite{moore-1973},
 \makeaciteofthree{bm}{boyermoore}{boyermooresecondedition},
 \cite{waltherhandbook},
 \cite{bundy-survey},
 \makeaciteoftwo{ACLTWO-CASESTUDIES}{ACLTWO}.%
}

In general,
we tend to present the challenges and their historical solutions 
with the help of small intuitive examples
and refer the readers interested in the very details of
the implementations of the theorem provers 
to the published and easily accessible documents
on which our description is mostly based.

Nevertheless, \hskip.1em
small \LISP\ expression cannot completely be avoided because 
we \nolinebreak have 
to \nolinebreak describe the crucial parts of the historically most significant
implementations and ought to \nolinebreak show 
some of the  advantages of \LISP's
untypedness.\footnote{%
 \majorheadroom
 See the advantages of the untyped, type-restriction-free
 declaration of the 
 \index{shells}%
 shell {\tt CONS} in \nlbsectref{subsection theorem prover}.%
}
The readers,
however, do not have to know more about \LISP\ than the following:
A \LISP\ term is either a variable symbol, \hskip.1em
or
a function call of the form 
\mbox{\tt(\math f \math{t_1} \math\cdots\ \math{t_n})},
\hskip.3em
where \math f is a function symbol, \maths{t_1}, \ldots, \math{t_n}
are \LISP\ terms, and \math n is one of the natural numbers,
which we assume to include \nlbmaths 0.

\section{Organization of \englishdiesesPapier}
\englishDiesespapier\ is further organized as follows.

\sectrefs{section What is mathematical induction?}
{section Recursive Definitions} offer a self-contained reference
for the readers who are not 
familiar 
with the field 
of mathematical induction and its automation. \hskip.2em
In \nlbsectref{section What is mathematical induction?}
we introduce the essentials of mathematical induction. \hskip.1em
In \sectref{section Recursive Definitions} we have to become
more formal regarding recursive function definitions,
their consistency, termination, and 
induction templates and schemes. \hskip.2em
The main part is \nlbsectref{section Explicit Induction}, \hskip.1em
where we present the historically most important
systems in automated induction,
and discuss the details
of software systems for explicit induction,
with a focus on the\,1970s. 
After describing the application context in \nlbsectref
{subsection The Application Context of Automated Explicit Induction}, \hskip.1em
we 
present
the following \boyermooretheoremprovers: \hskip.2em
the \PURELISPTP\ (\sectref{The Pure Lisp Theorem Prover}),
\THM\ (\sectref{subsection theorem prover}),
\NQTHM\ (\sectref{subsection NQTHM}),
and \ACLTWO\nolinebreak\ (\sectref{subsection ACLTWO}). \hskip.3em
The historically most important
remaining explicit-induction systems are sketched in 
\sectref{subsection Further Most Noteworthy Explicit-Induction Systems}. 
\hskip.2em
Alternative approaches to the automation of induction
that do not follow the paradigm of explicit induction
are discussed in 
\nlbsectref{section Alternative Approaches to the Automation of Induction}.
After summarizing the lessons learned in 
\sectref{section Lessons Learned}, \hskip.1em
we \nolinebreak conclude with \sectref{section Conclusion}.
\vfill\pagebreak
\section{Mathematical Induction}\label{section What is mathematical induction?}%
In this section,
we 
introduce mathematical induction 
and clarify the difference between
{\it\descenteinfinienoindex}\/ and
{\em\noetherian}, {\em structural}, and {\em explicit induction}.
\par\indent
According to \aristotleindex\aristotle,
{\em induction}\/ means to go from the special to the general,
and to realize the {\em general}\/ \hskip.1em
from the memorized perception of particular cases.
\mbox{Induction} plays a major \role\ in the generation of conjectures
in mathematics and the natural \mbox{sciences}.
Modern scientists design experiments
to \nolinebreak falsify a conjectured law of nature,
and they accept the law as a scientific fact
only after many trials have all failed to falsify \nolinebreak it.
In \nolinebreak the tradition of \euclidname,
mathematicians accept a 
mathematical conjecture
as a theorem
only after a rigorous proof 
has been provided.
According to \kantindex\kant,
induction is {\em synthetic}\/ in the sense that it properly extends
what we think to \nolinebreak know --- 
in \nolinebreak opposition to {\em deduction},
which is {\em analytic}\/ in the sense that 
it cannot provide us with any information 
not implicitly contained in the initial judgments,
though we \nolinebreak
can hardly be aware of all deducible \mbox{consequences.}
\par\indent
Surprisingly, 
in this well-established and time-honored terminology,
{\em mathematical induction}\/ is not induction,
but a special form of deduction for which
\mbox{---~}in \nolinebreak the \nth{19}\,century~--- 
the term ``induction'' was introduced and became standard 
in German and English mathematics.\footnote{%
 First in German (\cfnlb\ \noteref{note fries}), \hskip.2em
 soon later in English
 (\cfnlb\ \cite{cajori-1918}).%
}

In spite of this misnomer,
for the sake of \nolinebreak brevity,
the term ``induction'' will always refer to mathematical induction
in \nolinebreak what follows.
\par\indent
Although it received its current name only in the \nth{19}\,century,
mathematical 
\mbox{induction} has been 
a standard method of every working mathematician
at all \nolinebreak times.
It has been conjectured\footnote{%
 \majorheadroom
 It is conjectured in \cite{hippa} that \hippasos\ has proved 
that there is no pair of natural numbers that can describe the ratio of
the lengths of the sides of a pentagram and its enclosing pentagon.
Note that this ratio, seen as an irrational number, 
is equal to the golden number, which, however,
was conceptualized in
entirely different terms in ancient Greek mathematics.%
}
that 
\hippasosname\ 
\hippasoslifetime\ 
applied a form of mathematical induction,
later named 
{\em\descenteinfinie\ \mbox{(ou ind\'efinie)}}\/ \hskip.2em
by \fermat. \hskip.2em
We \nolinebreak find another form of induction, 
nowadays called \structuralinductionindex\mbox{\em structural induction},
in a text of \platoname\ \platolifetime.\footnotemark
\footnotetext{%
 \majorheadroom
 \Cfnlb\ \cite{plato-induction}.%
}%

In \nolinebreak \euclidindex\euclid's famous 
``Elements''\,\shortcite{elements}, \hskip.1em
we \nolinebreak find several applications of {\it\descenteinfinie}\/
and in a way also of \structuralinductionindex
structural induction.\footnote{%
 \majorheadroom
 An example for {\it\descenteinfinie}\/ is 
 Proposition\,31 of \Vol\,VII of the Elements.
 Moreover, 
 the \nolinebreak proof in the Elements of \litpropref{8} of \Vol\,IX
 seems to be sound according to mathematical standards; \hskip.2em
 and so we can see it only as a proof by 
 \structuralinductionindex structural induction
 in a very poor linguistic and logical form. \hskip.2em
 This is in accordance with \cite{freudenthal}, 
 but not with \cite{unguru-one} and \cite{plato-induction}. \hskip.4em
 See \nolinebreak\cite{Greeks-induction} and  
 \cite[\litsectref{2.4}]{fermatsproof} \hskip.1em 
 for further discussion.%
}   
\structuralinductionindex%
Structural induction was known to
the Muslim mathematicians around the year 1000, 
and occurs in a Hebrew book of 
\gersonname\ (Orange and \Avignon) \gersonlifetime.\footnote{%
 \majorheadroom
 \Cf\ \cite{Ravinovitch-Gershon}. \hskip.3em
 Also summarized in \cite{katz-history}.%
} 
Furthermore, 
\structuralinductionindex%
structural induction was used by
\maurolycusname\ (\maurolycuslifeplace) \maurolycuslifetime,\footnotemark\
and by \pascalname\ \pascallifetime.\footnotemark\
After an absence of more than one millennium
(besides copying ancient proofs),
{\em\descenteinfinie}\/ was reinvented by 
\mbox{\fermatname\,\fermatlifetime.}\footnotemark\,\footnotemark

\subsection{\WellFoundedness\ and Termination}\label
{subsection Well-Foundedness and Termination}%
\halftop
\noindent 
\addtocounter{footnote}{-3}%
\footnotetext{%
 \Cf\ \cite{maurolycus}.%
}%
\addtocounter{footnote}{1}%
\footnotetext{%
 \majorheadroom
 \Cf\ \cite[\p\,103]{pascal}.%
}%
\addtocounter{footnote}{1}%
\footnotetext{%
 \majorheadroom
 There is no consensus on \fermatindex\fermat's year of birth.
 Candidates are 
 1601%
 ,
 1607 (\cite{fermat-birth-date}),
 and 1608. \hskip.3em
 Thus,
 we write ``\fermatindex\fermatbirthyear\closequotecomma
 following \cite
 {catherine-goldstein-fermat-priceton-companion-math-2008}.%
}%
\addtocounter{footnote}{1}%
\footnotetext{
 \majorheadroom
 The best-documented example 
 of \fermatindex\fermat's applications of
 {\em\descenteinfinie}\/ 
 is the proof of the theorem:
 {\em The area of a rectangular triangle with positive integer side lengths 
 is not the square of an integer}\/; \hskip.4em
 \cf\ \eg\ \cite{fermatsproof}.%
}%
A relation \nlbmath < is \index{well-foundedness}{\em\wellfounded}\/
if, \hskip.1em
for each proposition \nlbmath{\app Q w} that is not constantly false, \hskip.1em
there is a \math<-minimal \nlbmath m among the objects for which
\math Q holds, \hskip.3em
\ie\ there is an \nlbmath m with \nlbmaths{\app Q m}, \hskip.2em 
for which there is no \nolinebreak\mbox{\math{u<m}}
with \app Q u. 

\halftop\indent
Writing ``\Wellfpp <'' for 
``\math<~is~\wellfounded\closequotecommasmallextraspace
we can formalize this {\em definition}\/
as follows:\smallfootroom
\par\halftop\halftop\noindent\math{\begin{array}{@{}l@{~~~~~~~~}r@{}l@{}}
 \inpit{\Wellfpp{\tight<}}
&\forall Q\stopq
&\,\inparentheses{\headroom\footroom
     \exists w\stopq\app Q w
   {\nottight{\nottight{\nottight\implies}}}
     \exists m\stopq\inparenthesestight{\app Q m\und
     \neg\exists u\tight<m\stopq\app Q u}}
\\\end{array}}
\par\halftop\halftop\indent\mediumheadroom
Let \transclosureinline < 
denote the transitive closure of \nlbmaths <, \hskip.3em
and \refltransclosureinline < the reflexive closure of \transclosureinline <.

\halftop\indent
\math < is an (irreflexive) {\em ordering}\/
if it is an irreflexive and transitive relation. 

\begin{sloppypar}\halftop\indent
There is not much difference between a \wellfounded\ {\em relation}\/ and a
\wellfounded\ \mbox{{\em ordering\/}:\,}\footnote{%
 \majorheadroom
 \Cf\ \litlemmref{2.1} of \cite[\litsectref{2.1.1}]{wirthcardinal}.%
}\par\end{sloppypar}

\halftop
\begin{lemma}\label{lemma 1} \ \ \math < is \wellfounded\ 
\ifandonlyif\ \hskip.3em \transclosureinline <
 is a \wellfounded\ ordering.\end{lemma}

\halftop\halftop\noindent
Closely related to the \wellfoundedness\ of a relation \nlbmath< \hskip.2em
is the \index{termination}termination of its 
{\em reverse relation}\/ \hskip.1em
written as \hskip.15em\nolinebreak\nlbmath{\reverserelation{\tight <}} \hskip.1em
or \hskip.1em\maths>, \hskip.3em
and defined as
\nlbmaths{\setwith{\pair u v}{\pair v u\tightin\tight<}}. \,\,\,

\halftop\indent
A relation \hskip.1em\nolinebreak\nlbmath{\tight>} \hskip.2em
is {\em terminating}\/ \hskip.2em if 
it has no non-terminating sequences, \hskip.2em
\ie\ if \nolinebreak there is no infinite sequence of the form \hskip.2em
\math{x_0} \tight> \math{x_1} \tight> \math{x_2} \tight> \math{x_3} \ldots.

\halftop\indent
If \nolinebreak\hskip.1em\nlbmath > \hskip.1em
has a non-terminating sequence,
then this sequence, taken as a set, is a witness for
the non-\wellfoundedness\ of \nlbmaths{\tight<}. \hskip.5em
The converse implication, however, is a weak form of the
\axiomofchoice;\footnote{%
 \majorheadroom
 See \cite[\litsectref{2.1.2}, \p\,18]{wirthcardinal} for the 
 equivalence to the \principleofdependentchoice,
 found 
 in \cite[\p 19]{axiomofchoice}, \hskip.1em
 analyzed 
 in
 \mbox{\cite[\p\,30, Form\,43]{weakaxiomofchoice}.}%
} \hskip.2em
indeed, 
it
allows 
us to pick a non-terminating sequence for \nlbmath> \hskip.2em 
from the
set witnessing the non-\wellfoundedness\ of \nlbmath <. \hskip.4em

\halftop\indent
So 
\wellfoundedness\ is slightly stronger than termination of the
reverse relation, \hskip.1em
and the difference is relevant here
because we cannot take the \axiomofchoice\ for granted in a discussion of
the 
foundations of induction, \hskip.1em
as will be explained \nolinebreak
in \nlbsectref{subsection Beyond Noetherian induction}.
\vfill\pagebreak
\subsection{The \theoremofnoetherianinduction}\label
{subsection noetherian induction}
\halftop\noindent
In its modern standard meaning,
the method of mathematical induction 
is easily seen to be a form of deduction,
simply because it can be formalized as
the application of the {\em\theoremofnoetherianinduction}\/:
\begin{quote}
A proposition \nlbmath{\app P w} 
can be shown to hold (for all \nlbmath w) \hskip.1em
by {\em\noetherian\ induction}\/ over 
a \wellfounded\ relation \nlbmath< \hskip.1em
as follows: \hskip.3em
{\em Show \hskip.1em
(for \nolinebreak every \nlbmath v\nolinebreak\hspace*{-.08em}\nolinebreak) 
\hskip.1em
that \app P v follows from the assumption that \app P u holds for all\/ 
\nlbmaths{u<v}.}\halftop\halftop
\end{quote}\begin{sloppypar}\noindent
Again writing \hskip.2em``\Wellfpp <'' \hskip.2em
for \hskip.2em``\math< is \wellfounded\closequotecommaextraspace
we can formalize the {\em\theoremofnoetherianinduction}\/
as follows:\footnote{%
 When we write an implication \nlbmath{A\hskip.1em\tightimplies B} \hskip.1em
 in the reverse form of \nlbmaths{B\tightantiimplies A}, \hskip.2em
 we do this to indicate that a proof attempt
 will typically start from \nlbmath B \hskip.1em
 and try to reduce it to \nlbmaths A.%
}%
\end{sloppypar}
\par\halftop\halftop\halftop\noindent\math{\begin{array}{@{}l@{~~~~~~~~}r@{}l@{}}
 \inpit{\ident{N}}
&\forall P\stopq
&\inparentheses{\headroom
     \forall w\stopq\app P w
   {\nottight{\nottight{\nottight\antiimplies}}}
     \exists\tight<\stopq\inparenthesesoplist{
      \forall v.\inparentheses{\!\!
           \app P  v
         {\nottight\antiimplies}
           \theinnermostpartofW\!\!}\!\!
    \oplistund
       \Wellfpp <}\!\!\!}\!
\\\end{array}}
\par\halftop\halftop\halftop\halftop\noindent
The today commonly used term ``\noetherian\ induction'' 
is a tribute to the famous female German 
mathematician
\noethername\ \noetherlifetime. \hskip.3em
It occurs as the ``Generalized principle of induction (\noetherian\ induction)''
in \mbox{\cite[\p\,20]{cohn-1965}.} \hskip.3em
Moreover,
it occurs as Proposition\,7 (``Principle of \noetherian\ Induction'') \hskip.1em
in \cite[Chapter\,III, \litsectref {6.5}, \p\,190]{bourbaki-english} \hskip.2em
---~a translation of the French original in its second edition 
\cite[\litsectref{6.5}]{bourbaki-set-theory-chapter-3-2nd-edn}, \hskip.2em
where it occurs as Proposition\,7
(``principe de r\'ecurrence n\oefranz th\'erienne'')\fullstopnospace\footnote{%
 \majorheadroom
 The peculiar French spelling 
 ``n\oefranz th\'erienne''
 imitates the
 German pronunciation of ``\noether\closequotecommasmallextraspace
 where the ``oe'' is to be pronounced 
 neither as a long~``o'' 
 (the default, as in ``Itzehoe''), \hskip.1em
 nor as two separate vowels as indicated by the diaeresis in 
 ``o\ewithtrema\closequotecommasmallextraspace
 but as an umlaut,
 typically written in German as the ligature ``\oe\closequotefullstopextraspace
 Neither \emmy\ nor her father \noetherfathername\ \noetherfatherlifetime\
 (mathematics professor \aswell) 
 \hskip.1em
 used this ligature, \hskip.1em 
 found however
 in some of their official \mbox{German documents.}%
} \hskip.3em
We \nolinebreak do \nolinebreak not know whether 
``\noetherian'' was used as a name of an induction principle 
before\,\,1965;\footnote{%
 \majorheadroom
 \label{note complete induction}%
 In 1967, ``\noetherian\ Induction'' was not generally 
 used as a name for the \theoremofnoetherianinduction\ yet:
 For instance,
 this theorem 
 ---~instantiated with the ordering of the natural numbers~---
 is called the {\em principle of complete induction}\/
 in \cite[\p\,205]{schoenfield}, \hskip.1em
 but more often called \index{induction!course-of-values}%
 {\em course-of-values induction},
 \cfnlb\ \eg\ \url
 {http://en.wikipedia.org/wiki/Mathematical_induction#Complete_induction}.
 ``Complete induction\closequotecomma
 however,
 is a most confusing name hardly used in English.
 Indeed,
 \index{induction!complete}``complete induction''
 is the literal translation of 
 the German technical term
 ``{\germanfont vollst\ae ndige Induction}\closequotecomma
 which traditionally means \structuralinductionindex%
 \mbox{structural induction}
 (\cfnlb\ \noteref{note fries}) 
 ---~and these two kinds of mathematical induction are different from 
 each \nolinebreak other.%
} \hskip.3em
in particular, \hskip.1em 
it \nolinebreak does not occur 
in the first French edition \cite{bourbaki-set-theory-chapter-3-1st-edn}
of \nolinebreak\cite{bourbaki-set-theory-chapter-3-2nd-edn}.\footnote{%
 \majorheadroom
 Indeed, 
 the main text of \nolinebreak\hskip.1em\nolinebreak\litsectref{6.5} 
 in the \nth 1\,edition
 \cite{bourbaki-set-theory-chapter-3-1st-edn} \hskip.1em
 ends (on \litspageref{98}) \hskip.1em
 three lines before the text of Proposition\,7 begins
 \hskip.1em
 in the \nth 2\,edition 
 \nolinebreak\cite{bourbaki-set-theory-chapter-3-2nd-edn}
 (on \litspageref{76} of \nolinebreak\hskip.1em\nolinebreak\litsectref{6.5}).%
}%
\vfill\pagebreak
\subsection{An Induction Principle 
Stronger than 
\noetherian\ Induction?}\label
{subsection Beyond Noetherian induction}%
\halftop\noindent
Let us try to find a weaker replacement for the precondition of
\mbox\wellfoundedness\ in \noetherian\ induction, \hskip.1em
in the sense that we try to replace ``\Wellfpp <'' in 
the \theoremofnoetherianinduction\ \inpit{\ident N} in 
\nlbsectref{subsection noetherian induction} 
with some weaker property, \hskip.1em
which we \nolinebreak will designate with
``\mbox{\math{\ident{Weak}(\tight <,P)}}'' \hskip.4em
(such \nolinebreak 
 that \hskip.1em\math{\forall P\stopq\ident{Weak}(\tight <,P)
 \nottight\antiimplies\Wellfpp <}).
\hskip.6em
This would result in the formula 
\tinyfootroom\par\halftop\halftop\noindent
\math{\begin{array}{@{}l@{~~~~~~~~}r@{}l@{}}
 \inpit{\ident{N}'}
&\forall P\stopq
&\inparentheses{\headroom
     \forall w\stopq\app P w
   {\nottight{\nottight{\nottight\antiimplies}}}
     \exists\tight<\stopq\inparenthesesoplist{
      \forall v.\inparentheses{\!\!
           \app P  v
         {\nottight\antiimplies}
           \theinnermostpartofW\!\!}\!\!
    \oplistund
       \ident{Weak}(\tight <,P)}\!\!\!}.
\\\end{array}}
\par\halftop\halftop\noindent
If we assume \nlbmaths{\inpit{\ident N'}},
however, 
we get the converse
\hskip.3em\maths{
\forall P\stopq\ident{Weak}(\tight <,P)
\nottight\implies
\Wellfpp <
}.\mbox{\,}\footnotemark\ \hskip.4em
This means that a proper weakening is possible only \wrt\
{\em certain}\/ \nlbmaths P, \hskip.2em
and the \theoremofnoetherianinduction\ {\em is the strongest among those
induction principles of the form \nlbmath{\inpit{\ident N'}}
where \nlbmath{\ident{Weak}(\tight <,P)} does not depend on \nlbmaths P.}

\math C is a {\em\math <-chain}\/ if \transclosureinline < 
 is a total ordering on \nlbmaths C. \hskip.3em
Let us write ``\math{u\tight<C}\nolinebreak\hskip.02em\nolinebreak'' 
for \maths{\forall c\tightin C.\,u\tight<c},
\hskip.2em 
and ``\nolinebreak\hskip.01em\nlbmath
{\forall u\tight<C\stopq F}\nolinebreak\hskip.02em\nolinebreak'' 
as usual for \maths{\forall u.\inpit{u\tight<C\implies F}}. \hskip.5em
In \cite{geseraxiomofchoice}, \hskip.1em
we find applications of 
an induction principle that roughly
has the form \nlbmath{\inpit{\ident N'}} \hskip.05em
where \math{\ident{Weak}(\tight <,P)} 
\nolinebreak\hskip.1em\nolinebreak is:\mediumfootroom\notop\halftop\begin{quote}
For every non-empty \math <-chain \nlbmath C \hskip.1em
\opt{without a \math <-minimal element}%
:
\\\mediumheadroom\LINEmaths{
\exists v\tightin C\stopq\app P v\nottight{\nottight\antiimplies}
\forall u\tight<C\stopq\app P u}.\mediumfootroom\notop\halftop\end{quote}
The resulting induction principle can be given an elegant form:
If \nolinebreak 
we drop the part of \hskip.05em\math{\ident{Weak}(\tight <,P)} \hskip.1em 
given in optional brackets \opt{\ldots\hskip-.16em}, \hskip.1em
then we can drop the conjunction in \nlbmath{\inpit{\ident N'}}
together with its first element,
because \math{\{v\}} \mbox{is a non-empty \nlbmath <-chain.}
\par\footnotetext{%
 {\em Proof.} \hskip.5em\hfill
 Let \ranres{\tight <}A denote the range restriction of 
 \nlbmath < to\,\nlbmath A\hskip.3em\hfill
 (\ie\ \maths{u\tight{\ranres{\tight <}A}v}{}
  \ifandonlyif\ \maths{u\,\tight<\,v\tightin A}). \ \ \hfill
 \\
 Let us take \math{P(w)} to be \Wellfpp{\ranres{\tight <}{\app A w}} for 
 \maths{\app A w:=\setwith{w'}{w'\refltransclosureinline <\,w}}. \ 
 Then the reverse implication 
 \\[-.3ex]%
 follows from \inpit{\ident N'}
 because 
 \maths{\app P v\antiimplies\theinnermostpartofW}{}
 holds for any \nlbmaths v,\footnotemark\ 
 and \maths{\forall w\stopq\app P w}{} implies \nlbmaths{\Wellfpp <}.%
}\footnotetext{%
 \majorheadroom
 {\em Proof.} \hskip.5em
 To show \maths{\app P v}, \hskip.2em
 it suffices to find, \hskip.15em 
 for an arbitrary, not constantly false proposition \nlbmaths Q, \hskip.1em
 an \math m with \nlbmaths{\app Q m}, \hskip.1em 
 for which, \hskip.1em
 in case of \maths{m\tightin\app A v}, \hskip.2em
 there is no \math{m'\tight<\,m} with \maths{\app Q{m'}}.
 \par
 If we have \app Q m for some \math m with \math{m\tightnotin\app A v},
 \hskip.2em
 then we are done. 
 \par
 If we have \app Q{u'} for some \nlbmath{u<v} and some 
 \math{u'\in\app A u}, \hskip.2em
 then, \hskip.1em
 for \math{\app{Q'}{u''}} being the conjunction of \math{\app Q{u''}}
 and \nlbmath{u''\tightin\app A u}, \hskip.2em
 there\,is 
 (because of the assumed \math{\app P u}) \hskip.1em
 an \math m with \nlbmaths{\app{Q'}m}, \hskip.2em
 for which there is no \math{m'\tight<\,m} with \nlbmaths{\app{Q'}{m'}}.
 \hskip.3em
 Then we have \maths{\app Q m}. \hskip.3em
 If there were an \math{m'\tight<\,m} with
 \nlbmaths{\app{Q}{m'}}, \hskip.2em
 then we would have \nlbmaths{\app{Q'}{m'}}. \hskip.3em
 Thus, 
 there cannot be such an \nlbmaths{m'}, \hskip.2em
 and so \math m satisfies our requirements.
 \par
 Otherwise, \hskip.1em
 if none of these two cases is given, \hskip.1em
 \math Q can only hold for \nlbmaths v. \hskip.3em
 As \math Q is not constantly false,
 we get \app Q v and then \math{v\tight\nless\,v}
 (because otherwise the second case is given for \math{u:=v} and \math{u':=v}). 
 \hskip.3em
 Then \math{m:=v} satisfies our requirements.%
}%
Then the following equivalent 
is obtained by switching 
from proposition \nlbmath P \hskip.1em
to its class of counterexamples \math Q: \hskip.4em
``If, \hskip.2em
  for every non-empty \math <-chain \nlbmaths{C\subseteq Q}, \hskip.4em
  there is a \math{u\in Q} \hskip.2em
  with \nolinebreak\hskip.1em\nlbmaths{u\tight<C}, \hskip.4em
  then \hskip.1em\maths{Q\tightequal\emptyset}.'' \hskip.4em
Under the assumption that \math Q is a set, 
this is an equivalent of the \axiomofchoice\
(\cf\ \cite{geseraxiomofchoice}, \cite{axiomofchoice}).

This means that the axiomatic status of induction principles
ranges from 
the \theoremofnoetherianinduction\
up to the \axiomofchoice. \hskip.3em
If we took the \axiomofchoice\ for granted, \hskip.1em
this difference in status between a theorem and an axiom would collapse
and our discussion of the axiomatic status of mathematical induction
would 
degenerate%
. \hskip.3em
So the care with which we distinguished 
termination of the reverse relation from \wellfoundedness\ in 
\nlbsectref{subsection Well-Foundedness and Termination} \hskip.1em
is justified.

\vfill
\subsection{The Natural Numbers}\label
{subsection Mathematical Induction and the Natural Numbers}
\halftop
\noindent
The field of application of mathematical induction 
most familiar in mathematics is the domain of the
natural numbers \mbox{\maths 0, \maths 1, \maths 2, \ldots}. \hskip.3em
Let us formalize the natural numbers 
with the help of two 
\index{constructor function symbols}constructor function symbols, 
namely one for the constant zero and one for 
the direct successor of a natural number:
\par\halftop\noindent\LINEmaths{\begin{array}{r@{\,\,}c@{\,\,\,}l}\zeropp
 &:
 &\nat
\\\ssymbol
 &:
 &\FUNSET\nat\nat
\\\end{array}}{}
\par\noindent\smallheadroom 
Moreover, let us assume in \englishdiesespapier\
that the variables \maths x, \nlbmath y \hskip.1em
always range over the natural numbers, and that free variables in 
formulas are implicitly universally quantified 
(as is standard in mathematics),
\hskip.1em
such that, for example, a formula with the free variable \nlbmath x
can be seen as having the implicit outermost quantifier
\nolinebreak\hskip.3em\nlbmath
{\forall\prettytighthastype{\boundvari x{}}\nat\stopq}
\\\indent
After the definition \nlbmath{\inpit{\Wellfpp<}} and the theorem
\nlbmath{\inpit{\ident N}}, \hskip.2em
let us now consider some standard {\em axioms}\/ \hskip.03em
for specifying the
natural numbers, \hskip.1em
namely that a natural number is either zero or 
a direct successor of another natural number \nlbmaths{\inpit{\nat 1}}, 
\hskip.1em
that zero is not a successor \nlbmaths{\inpit{\nat 2}}, \hskip.1em
that the successor function is injective \nlbmaths{\inpit{\nat 3}}, \hskip.1em
and that the so-called {\em\axiomofstructuralinduction\ over\/ \nlbmath\zeropp\
and\/ \nlbmath\ssymbol}\/\hskip.45em 
holds; \hskip.3em
formally:\smallfootroom
\\\math{\begin{array}{@{}l@{~~~~~~~}l@{}}
 \inpit{\nat 1}
 &
     \boundvari x{}\tightequal\zeropp
   \nottight{\nottight\oder}
     \exists
\boundvari y{}
\stopq
     \inparentheses{
     \boundvari x{}\tightequal\spp{\boundvari y{}}}
 \footroom\\\inpit{\nat 2}
 &
 \spp x\tightnotequal\zeropp
 \footroom\\\inpit{\nat 3}
 &
  \spp x\tightequal\spp y
  \nottight{\nottight\implies}x\tightequal y
  \footroom\\
  \inpit{\ident S}
 &\forall P\stopq
  \inparentheses{\headroom\footroom
  \forall
\boundvari x{}
\stopq
   \app P  x
   {\nottight{\nottight{\nottight\antiimplies}}}
   \app P \zeropp\nottight\und
   \forall
\boundvari y{}
\stopq
   \inparentheses{\app P {\spp y}\antiimplies\app P  y}}
\\\end{array}}
\par\noindent
\dedekindname\ \dedekindlifetime\
proved the \axiomofstructuralinduction\ \nlbmath{\inpit{\ident S}} \hskip.05em
for his model of the natural numbers in \cite{dedekind-1888},
where he states that the proof method resulting from the application
of this axiom is known under the name
``{\germanfont vollst\ae ndige Induction}\closequotefullstopnospace
\footnote{\label{note fries}%
 In \nolinebreak the tradition
 of \aristotelianlogic, 
 the technical term ``{\germanfontfootnote vollst\ae ndige Induction}''
 (in Latin: ``inductio completa\closequotecomma 
  \cfnlb\ \eg\ \cite[Part\,I, \litsectref{478}, \p\,369]{wolff-rationalis-1740})
 denotes a complete case analysis, 
 \cfnlb\ \eg\ \cite
 [Dianoiologie, \litsectref{287}; Alethiologie, \litsectref{190}]{lambert-1764}.
  \hskip.2em
 Its misuse as \englisheinebezeichnung\ of 
 \structuralinductionindex%
 structural induction 
 originates
 in \cite[\p\,46\f]{fries-vollstaendige-induction}, \hskip.2em 
 and was perpetuated by 
 \dedekindindex\dedekind.
 Its literal translation 
 \index{induction!complete}``complete induction''
 is misleading,
 \cfnlb\ \noteref{note complete induction}. \hskip.3em
 By the 1920s,
 ``{\germanfontfootnote vollst\ae ndige Induction}''
 had become a very vague notion that 
 is best translated as ``mathematical induction\closequotecomma
 as done in \cite[\p 130]{heijenoort-source-book} \hskip.1em
 and as it is standard today, 
 \cf\ \eg\ \cite[\litnoteref{23.4}]%
 {grundlagen-german-english-edition-volume-one-one}.%
}
\par
Now we can go on by defining 
---~in two equivalent\footnote{%
 \majorheadroom
 For the equivalence transformation between 
 \index{constructor style}%
 constructor and 
 \index{destructor style}%
 destructor style
 see \examref{example and back} in 
 \nlbsectref{subsubsection Destructor Elimination in}.%
}  ways~---
the destructor function
\FUNDEF\psymbol\nat\nat, \hskip.3em
returning the \underline predecessor of a positive natural number: 
\par\halftop\noindent\math{\begin{array}{@{}l@{~~~~~~~}l@{\ =\ }l}
  (\psymbol 1)
 &\ppp{\spp x}
 &x
\\\end{array}}%
\par\halftop\noindent\math{\begin{array}{@{}l@{~~~~~~~}l@{\ =\ }l}
  (\psymbol 1')
 &\ppp{x'}
 &x\nottight\antiimplies x'\tightequal\spp x
\\\end{array}}
\par\halftop\noindent
The definition via \inpit{\psymbol 1} is in 
\index{constructor style}{\em constructor style}, \hskip.1em
where constructor terms may occur 
on the left-hand side of the \pnc\ equation
as arguments of the function being defined. \hskip.3em
The alternative definition via \inpit{\psymbol 1'} is in 
\index{destructor style}{\em destructor style}, \hskip.1em
where only variables may occur as arguments on the left-hand side.%

\pagebreak

For both definition styles, the term on the left-hand side must
be \index{linear terms}{\em linear}\/ 
(\ie\ all its variable occurrences must be distinct variables) 
\hskip.1em
and have the function symbol to be defined as the top symbol.

Let us define some recursive functions over the natural numbers, \hskip.1em
such as 
addition and multiplication
\hskip.1em\maths{\FUNDEF{\plussymbol,\,\timessymbol}{\nat,\,\nat}\nat}, 
\hskip.25em
the irreflexive ordering of the natural numbers
\hskip.1em\math{\FUNDEF\lessymbol{\nat,\,\nat}\bool} \hskip.1em
(see \sectref{subsubsection Boolean Values}
 for the data type \bool\ of \myboolean\ values), \hskip.2em
and the \ackermannfunction\
\hskip.1em\math{\FUNDEF\acksymbol{\nat,\,\nat}\nat\,}:\,\footnote{%
 \petername\ \peterlifetime\
 (a woman in the fertile community 
 of \Budapest\ mathematicians 
 and, like most of them, 
 of Jewish parentage) \hskip.1em
 published a simplified version \shortcite{peter-1951} of the 
 first recursive, but not primitive recursive function
 developed by \ackermannname\ \ackermannlifetime\ \cite{ackermann-1928a}. 
 \hskip.3em
 It is actually \peterrosza's version that is simply called 
 ``the \ackermannfunction'' today.%
}%
\par\halftop\noindent\math{\begin{array}
  {@{}l@{~~~~~~~}r@{\ =\ }l@{~~~~~~~}|@{~~~~~~~}l@{~~~~~~~}r@{\ =\ }l}
  \inpit{\plussymbol 1}
 &\plusppnoparentheses\zeropp y
 &y
 &\inpit{\timessymbol 1}
 &\timesppnoparentheses\zeropp y
 &\zeropp
  \smallfootroom
\\\inpit{\plussymbol 2}
 &\plusppnoparentheses{\spp x}y
 &\spp{\plusppnoparentheses x y}
 &\inpit{\timessymbol 2}
 &\timesppnoparentheses{\spp x}y
 &\plusppnoparentheses y{\timespp x y}
\\\end{array}}
\par\halftop\noindent\math{\begin{array}{@{}l@{~~~~~~~}l@{\ =\ }l}
  (\lessymbol 1)
 &\lespp x\zeropp
 &\falsepp
  \smallfootroom
\\(\lessymbol 2)
 &\lespp\zeropp{\spp y}
 &\truepp
  \smallfootroom
\\\inpit{\lessymbol 3}
 &\lespp{\spp x}{\spp y}
 &\lespp x y
\\\end{array}}
\par\halftop\noindent\math{\begin{array}{@{}l@{~~~~~~~}l@{\ =\ }l}
  \inpit{\acksymbol 1}
 &\ackpp\zeropp y
 &\spp y 
  \smallfootroom
\\\inpit{\acksymbol 2}
 &\ackpp{\spp x}\zeropp
 &\ackpp x{\spp\zeropp}
  \smallfootroom
\\\inpit{\acksymbol 3}
 &\ackpp{\spp x}{\spp y}
 &\ackpp x{\ackpp{\spp x}y}
\\\end{array}}
\par\halftop\halftop\noindent
The relation from a natural number to its direct successor can be
formalized by the binary relation \bigmaths{
\lambda
\boundvari x{},\boundvari y{}
\stopq\inpit{
\spp{\boundvari x{}}\tightequal\boundvari y{}}}. \hskip.4em
Then 
\bigmaths{\Wellfpp{
\lambda
\boundvari x{},\boundvari y{}
\stopq\inpit{
\spp{\boundvari x{}}\tightequal\boundvari y{}}}}{} 
states the \wellfoundedness\ of this relation,
which means according to \lemmref{lemma 1} that its
transitive closure 
---~\ie\ the irreflexive ordering of the natural numbers~---
is a \wellfounded\ ordering; \hskip.2em
so, 
in particular, 
we have
\Wellfpp{
\lambda
\boundvari x{},\boundvari y{}
\stopq\inpit{
\lespp x y\tightequal\truepp}}.
\par\halftop\halftop\halftop\noindent
Now the natural numbers can be specified up to isomorphism
either by\footnote{%
 \majorheadroom
 \Cfnlb\ \cite[\litsectref{1.1.2}]{wirthcardinal}.%
}\begin{itemize}\item
\math{(\nat 2)}, \nlbmath{(\nat 3)}, and \nlbmath{(\ident S)} \hskip.4em
\getittotheright{---~~following \peanoname\ \peanolifetime,}
\noitem\end{itemize}\mediumfootroom\headroom 
or \nolinebreak else by\noitem\begin{itemize}\item
\math{\inpit{\nat 1}}
and 
\math{\Wellfpp{\lambda
\boundvari x{},\boundvari y{}
\stopq
\inpit{\spp x\tightequal y}}}
\getittotheright{---~~following \pieriname\ \pierilifetime.\footnotemark}%
\footnotetext{\label{note pieri standard}%
 \majorheadroom
 \citet{pieri} stated these axioms informally and showed their
 equivalence to the version of the \peanoaxiom s 
 \cite{peanonovamethodo}
 given in \cite{padoa-1913}. \hskip.2em
 For a discussion and an 
 English translation see \cite{smith-pieri}. \hskip.3em
 \citet{pieri} has also a version where, 
 instead of the symbol \nlbmath\zeropp,
 there is only the statement that there is a natural number,
 and where \inpit{\nat 1} is replaced with the weaker statement 
 that there is at most one \ssymbol-minimal element:
 \\\LINEmaths{\neg\exists\boundvari y 0\stopq\inpit{
     \boundvari x 0\tightequal\spp{\boundvari y 0}}
     \und
     \neg\exists\boundvari y 1\stopq\inpit{
     \boundvari x 1\tightequal\spp{\boundvari y 1}}
     \nottight\implies\boundvari x 0\tightequal\boundvari x 1}.
 \\
 That non-standard natural numbers cannot exist 
 in \pieri's specification is easily shown as follows: \hskip.2em
 For every natural number \nlbmath x we can form the 
 set of all elements that can be reached from \nlbmath x
 by the reverse of the successor relation; \hskip.3em
 by \wellfoundedness\ of \nlbmaths\ssymbol, 
 this set contains the unique \ssymbol-minimal element 
 (\zeropp); 
 \hskip.5em 
 thus, \hskip.2em
 we \nolinebreak have 
 \hskip.1em\mbox{\maths{x\tightequal\sppiterated n\zeropp}{}} \hskip.2em
 for some standard meta-level natural number \nlbmath n.%
}
\end{itemize}

\vfill\pagebreak\par\indent
Immediate consequences of the axiom \inpit{\nat 1} \hskip.1em 
and the definition \inpit{\psymbol 1} \hskip.1em are the 
lemma~\inpit{\ssymbol 1}
and its flattened\footnote{%
 {\em Flattening}\/ is a logical equivalence transformation that replaces 
 a subterm (here: \ppp{x'}) with a fresh variable (here: \nlbmath x) 
 and adds a condition that equates the variable with the subterm.%
} 
version \inpit{\ssymbol 1'}:
\par\halftop\noindent\math{\begin{array}{@{}l@{~~~~~~~}r@{\ =\ }l}
  \inpit{\ssymbol 1}
 &\spp{\ppp{x'}}
 &x'
  \nottight{\nottight\antiimplies}x'\tightnotequal\zeropp
  \majorfootroom
\\\inpit{\ssymbol 1'}
 &\spp x
 &x'
  \nottight{\nottight\antiimplies}x'\tightnotequal\zeropp\nottight\und
  x\tightequal\ppp{x'}
\\\end{array}}
\par\halftop\halftop\halftop\noindent
Moreover,
on the basis of the given axioms we can most easily show
\halftop\par\noindent\math
{\begin{array}{@{}l@{~~~~~~~}l@{\ =\ }l}
  \inpit{\lessymbol 4}
 &\lespp x{\spp{x}}
 &\truepp
  \majorfootroom
\\\inpit{\lessymbol 5}
 &\lespp x{\spp{\plusppnoparentheses x y}}
 &\truepp
\\\end{array}}
\par\halftop\noindent
by \structuralinductionindex%
{\em structural induction on}\/ \nlbmaths x, \hskip.4em
\ie\ by taking the predicate variable \nlbmath P in the
\axiomofstructuralinduction\ \nlbmath{\inpit{\ident S}} \hskip.05em
to be 
\bigmaths{\lambda x\stopq\inpit{
\lespp x{\spp x}\tightequal\truepp}}{}
in case of \inpit{\lessymbol 4}, \hskip.2em
and \bigmaths{\lambda x\stopq\forall y\stopq\inpit{
\lespp x{\spp{\plusppnoparentheses x y}}\tightequal\truepp}}{}
in case of \inpit{\lessymbol 5}. \hskip.3em

\halftop\halftop\halftop\noindent
Furthermore
---~to see the necessity of doing induction on 
    several variables in parallel~---
we \nolinebreak will present\footnote{%
 \majorheadroom 
 We will prove \inpit{\lessymbol 7} twice: 
 once in \examref{example first proof of (less7)} in
 \sectref{section example first proof of (less7)}, \hskip.2em
 and again in \examref{example second proof of (less7)} in
 \sectref{section example second proof of (less7)}.%
}
the more complicated proof of the
{
 strengthened transitivity of the irreflexive 
 ordering of the natural numbers}, \hskip.2em
\ie\ of
\par\halftop\noindent\math{\begin{array}{@{}l@{~~~~~~~}l}
  \inpit{\lessymbol 7}
 &\lespp{\spp{x}}z\tightequal\truepp
  \nottight{\nottight\antiimplies}
  {\lespp x y\tightequal\truepp
    \nottight\und\lespp y z\tightequal\truepp}
\\\end{array}}
\par\halftop\halftop\halftop\noindent
We \nolinebreak will also prove 
the commutativity lemma \nlbmath{\inpit{\plussymbol 3}}\footnote{%
 \majorheadroom
 We will prove \inpit{\plussymbol 3} twice: 
 once in \examref{example first proof of (+3)} in
 \sectref{section example first proof of (+3)}, \hskip.2em
 and again in \examref{example second proof of (+3)} in
 \sectref{section example second proof of (+3)}.%
} \hskip.1em 
and the simple 
lemma \nlbmath{\inpit{\acksymbol 4}} \hskip.1em
about the \ackermannfunction:\smallfootroom\footnote{%
 \majorheadroom
 We will prove \inpit{\acksymbol 4} \hskip.1em
 in \examref{example proof of (ack4)} in
 \sectref{section example proof of (ack4)}.%
} 
\par\noindent\math{\begin{array}{@{}l@{~~~~~~~}l@{\ =\ }l}
  (\plussymbol 3)
 &\plusppnoparentheses x y
 &\plusppnoparentheses y x,
  \mediumfootroom
\\\end{array}}
\par\noindent\math{\begin{array}{@{}l@{~~~~~~~}l@{\ =\ }l}
  (\acksymbol 4)
 &\lespp y{\ackpp x y}
 &\truepp
\\\end{array}}%

\subsection{Standard Data Types}\label{subsection Standard Data Types}
\halftop\noindent
As we are interested in the verification of hardware and software, \hskip.1em
more important for \nolinebreak us than natural numbers
are the standard data types of higher-level programming languages,
such as lists, arrays, and records.

\halftop\halftop\halftop\noindent
To clarify the inductive character of data types defined by constructors,
and to show the additional complications arising from constructors 
with no or more than one argument,
let us present the data types
\nlbmath{\bool} (of \myboolean\ values)
and \nlbmath{\app\lists\nat} (of lists over natural numbers), \hskip.1em
which we \nolinebreak also need for our further examples.

\pagebreak

\subsubsection{\myBoolean\ Values}\label
{subsubsection Boolean Values}%
A special case is the data type \bool\ of the \myboolean\ values
given by the two constructors \hastype{\truepp,\falsepp}\bool\ \hskip.1em
without any arguments,
for which we get only the following two axioms by analogy
to the axioms for the natural numbers.
We \nolinebreak globally declare the variable \nlbmaths{\hastype b\bool};
\hskip.4em 
so \nlbmath b \nolinebreak will always range
over the \myboolean\ values.
\par\halftop\noindent\math{\begin{array}{@{}l@{~~~~~~~}l@{}}
  \inpit{\bool 1}
 &\boundvari b{}\tightequal\truepp\nottight{\nottight\oder} 
  \boundvari b{}\tightequal\falsepp
  \majorfootroom
\\\inpit{\bool 2}
 &\truepp\tightnotequal\falsepp
\\\end{array}}
\par\halftop\noindent
Note that the analogy of the axioms of \myboolean\ values to the axioms of the
natural numbers 
(\cfnlb\ \sectref{subsection Mathematical Induction and the Natural Numbers}) 
\hskip.1em 
is not perfect:  An axiom \inpit{\bool3} analogous to \inpit{\nat 3}
 cannot exist because there are no constructors for \bool\ 
 that take arguments.
 Moreover,  
 an axiom analogous to \nlbmath{\inpit{\ident S}} is superfluous
 because it is implied by \inpit{\bool 1}.%
\par\halftop\indent
Furthermore, let us define the \myboolean\ function \nlbmaths{
\FUNDEF\andsymbol{\bool,\bool}\bool}\,:
\par\noindent\math{\begin{array}{@{}l@{~~~~~}l@{\ =\ }l}
  \inpit{\andsymbol 1}
 &\andpp\falsepp b
 &\falsepp
  \smallfootroom
\\\inpit{\andsymbol 2}
 &\andpp b\falsepp
 &\falsepp
  \smallfootroom
\\\inpit{\andsymbol 3}
 &\andpp\truepp\truepp
 &\truepp
\\\end{array}}%
\subsubsection{Lists over Natural Numbers}
Let us now formalize the data type of the (finite) lists over natural numbers
with the help of the following two constructors: \hskip.2em
the constant symbol 
\\\noindent\mediumheadroom\LINEmaths{\hastype\nilpp{\app\lists\nat}}{}
\\\noindent for the empty list, \hskip.2em
and the function symbol 
\\\noindent\mbox{~~~~~~~~~~~~~~~~~~~~}\LINEmaths{\FUNDEF\cnssymbol
{\nat\comma\app\lists\nat}{\app\lists\nat}},
\\\noindent 
which takes a natural number and a list of natural numbers,
and returns the list where the number has been added 
to the input list as a new first element.
We \nolinebreak
globally declare 
the variables \nolinebreak\hskip.15em
\nlbmaths{\hastype{k,l}{\app\lists\nat}}.

\halftop\indent
By analogy to natural numbers,
the axioms of this data type are the following:\tinyfootroom
\par\halftop\noindent\math{\begin{array}{@{}l@{~~~~~}l@{}}
 \inpit{\app\lists\nat 1}
 &
     \boundvari l{}\tightequal\nilpp
   {\nottight\oder}
     \exists
\boundvari y{}
,
\boundvari k{}
\stopq
     \inparentheses{
     \boundvari l{}\tightequal\cnspp{\boundvari y{}}{\boundvari k{}}}
 \majorfootroom\\\headroom\inpit{\app\lists\nat 2}
 &
 \cnspp x l\tightnotequal\nilpp
 \majorfootroom\\\headroom\inpit{\app\lists\nat 3_1}
 &
  \cnspp x l\tightequal\cnspp y k
  \nottight{\nottight\implies}x\tightequal y
\\\inpit{\app\lists\nat 3_2}
 &
  \cnspp x l\tightequal\cnspp y k
  \nottight{\nottight\implies}l\tightequal k
  \majorfootroom\\\majorheadroom
  \inpit{\app\lists\nat\ident S}
 &\forall P\stopq
  \inparentheses
  {\!\!\forall
\boundvari l{}
\stopq\app P l\ 
  {\nottight\antiimplies}\,\,
    \inparentheses
      {
        \!\!
        \app P\nilpp
      \nottight\und
        \forall
\boundvari x{}
,
\boundvari k{}
\stopq
        \inparentheses{\!\!
            \app P{\cnspp x k}\antiimplies\app P k
        \!\!}
        \!\!}
        \!\!}
\\\end{array}}
\par\halftop\halftop\indent
Moreover, let us define the recursive functions 
\FUNDEF{\lengthsymbol,\sizesymbol}{\app\lists\nat}\nat, 
returning the length and the size of a list: 
\par\noindent\math{\begin{array}{@{}l@{~~~~~}l@{\ =\ }l}
  \inpit{\lengthsymbol1}
 &\lengthpp\nilpp
 &\zeropp
\\\inpit{\lengthsymbol2}
 &\lengthpp{\cnspp{\boundvari x{}}{\boundvari l{}}}
 &\spp{\lengthpp{\boundvari l{}}}
\\\end{array}}
\par\noindent\math{\begin{array}{@{}l@{~~~~~}l@{\ =\ }l}
  \inpit{
  \sizesymbol
  1}
 &\sizepp\nilpp
 &\zeropp
\\\inpit{
  \sizesymbol
  2}
 &\sizepp{\cnspp{\boundvari x{}}{\boundvari l{}}}
 &\spp{\plusppnoparentheses{\boundvari x{}}{\sizepp{\boundvari l{}}}}
\\\end{array}}\pagebreak
\par\halftop\indent
Note that the analogy of the axioms of lists to the 
axioms of the natural numbers is again not perfect:\begin{enumerate}\noitem\item
There is an additional axiom \nlbmaths{\inpit{\app\lists\nat 3_1}},
which has no \englishanalogon\ among the 
axioms of the natural numbers.
\noitem\item
Neither of the axioms
\nlbmath{\inpit{\app\lists\nat 3_1}}
and
\nlbmath{\inpit{\app\lists\nat 3_2}}
is implied by the axiom \math{\inpit{\app\lists\nat 1}}
together with the axiom
\\\mediumheadroom\LINEmaths{
\Wellfpp{\lambda
\boundvari l{},\boundvari k{}
\stopq
\exists
\boundvari x{}
\stopq
\inpit{\cnspp x l\tightequal k}}},
\\\mediumheadroom which is the \englishanalogon\ 
to \pieriindex\pieri's second axiom for the natural numbers.\footnotemark
\noitem\item
The latter axiom
is weaker than each of the two axioms
\\\mediumheadroom\LINEmaths{
\Wellfpp{\lambda
\boundvari l{},\boundvari k{}
\stopq
\inpit{
\lespp{\lengthpp l}
{\lengthpp k}
\tightequal\truepp
}}},
\\\mediumheadroom\LINEmaths{
\Wellfpp{\lambda
\boundvari l{},\boundvari k{}
\stopq
\inpit{
\lespp{\sizepp l}
{\sizepp k}
\tightequal\truepp
}}},
\\\mediumheadroom which state the \wellfoundedness\ of 
bigger\footnotemark\
relations. \hskip.2em
In spite of their relative strength,
the \wellfoundedness\ of these relations
is already implied by the \wellfoundedness\ that \pieriindex\pieri\ used
for his specification of the natural numbers.%
\end{enumerate}%
\addtocounter{footnote}{-1}\footnotetext{%
 See \sectref{subsection Mathematical Induction and the Natural Numbers}
 for \pieriindex\pieri's specification of the natural numbers.
 The axioms
 \nlbmath{\inpit{\app\lists\nat 3_1}}
 and \math{\inpit{\app\lists\nat 3_2}}
 are not implied because all axioms besides
 \nlbmath{\inpit{\app\lists\nat 3_1}}
 or
 \nlbmath{\inpit{\app\lists\nat 3_2}}
 are satisfied in the structure 
 where both 
 natural numbers and lists are
 isomorphic to the standard model of the natural numbers,
 and where lists differ only 
 in their sizes.
}\addtocounter{footnote}{1}\footnotetext{%
 \majorheadroom
 Indeed, 
 in case of \bigmaths{\cnspp x l=k},
 we have \bigmaths{\lespp{\lengthpp l}{\lengthpp k}=}{}\\\bigmaths{
 =\lespp{\lengthpp l}{\lengthpp{\cnspp x l}}
 =\lespp{\lengthpp l}{\spp{\lengthpp l}}
 =\truepp}{} because of \inpit{\lessymbol 4},
 and we also have
 \bigmaths{\lespp{\sizepp l}{\sizepp k}
 =\lespp{\sizepp l}{\sizepp{\cnspp x l}}
 =}{}\\\bigmaths{\lespp{\sizepp l}{\spp{\plusppnoparentheses x{\sizepp l}}}
 =\truepp}{}
 because of \inpit{\plussymbol 3} and \inpit{\lessymbol 5}.%
}%
\par\noindent
Therefore, 
the lists of natural numbers can be specified up to isomorphism
by a specification of the natural numbers up to isomorphism 
(see \sectref{subsection Mathematical Induction and the Natural Numbers}), 
{\hskip.2em}plus the axioms \inpit{\app\lists\nat 3_1} and
\inpit{\app\lists\nat 3_2}, 
{\hskip.2em}plus one of the following sets of axioms:\begin{itemize}\noitem\item
\inpit{\app\lists\nat 2}, \
\inpit{\app\lists\nat\ident S}
\getittotheright{---~~in the style of \peanoindex\peano,}\item
\inpit{\app\lists\nat 1}, \
\bigmaths{\Wellfpp{\lambda
\boundvari l{},\boundvari k{}
\stopq
\exists
\boundvari x{}
\stopq
\inpit{\cnspp x l\tightequal k}}}{}
\getittotheright{---~~in the style of \pieriindex\pieri,\footnotemark}%
\footnotetext{%
 \majorheadroom
 This option is essentially the choice of the 
 \index{shells}%
 \index{shell principle}%
 ``shell principle''
 of \cite[\p37\ff]{bm}: \
 The one but last axiom of \lititemref{\inpit 1} of the 
 \index{shell principle}%
 shell principle
 means \inpit{\app\lists\nat 2} in our formalization,
 and guarantees that \lititemref{\inpit 6} 
 implies
 \Wellfpp{\lambda\boundvari l{},\boundvari k{}\stopq\exists\boundvari x{}\stopq
 \inpit{\cnspp x l\tightequal k}}.%
}\item
\inpit{\app\lists\nat 1}, \
\inpit{\lengthsymbol1\mbox{--}2}
\getittotheright{---~~refining the style of \pieriindex\pieri.\footnotemark}%
\footnotetext{%
 \majorheadroom
 Although \inpit{\app\lists\nat 2} follows from 
 \inpit{\lengthsymbol1\mbox{--}2} \
 and \inpit{\nat 2}, \
 it \nolinebreak should be included in this standard specification
 because of its frequent applications.%
}\end{itemize}
Today it is standard to \nolinebreak avoid higher-order axioms in the way
exemplified in the last of these three items,\footnote{%
 \majorheadroom
 For this avoidance, 
 however,
 we have to admit the additional function \lengthsymbol\@.
 The same can be achieved with \sizesymbol\ instead of 
 \lengthsymbol, which is only possible, however, 
 for lists over element types that have a mapping into the natural numbers.%
}
and to get along with one second-order axiom for the natural numbers,
or even with the first-order instances of that axiom.%
\pagebreak
\par\halftop\halftop\indent
Moreover, 
as some of the most natural functions on lists,
let us define 
the destructors \FUNDEF\carsymbol{\app\lists\nat}\nat\ 
and \FUNDEF\cdrsymbol{\app\lists\nat}{\app\lists\nat}, \hskip.2em
both in 
\index{constructor style}%
constructor and 
\index{destructor style}%
destructor style. \hskip.2em
Furthermore, let us define 
the recursive 
member predicate \nlbmaths{\FUNDEF\mbpsymbol{\nat\comma\app\lists\nat}\bool},
\hskip.3em and 
\maths{\FUNDEF\dloncesymbol{\app\lists\nat}{\app\lists\nat}},
a recursive 
function
that deletes the first occurrence of a natural number 
in a list:
\par\halftop\noindent\math{\begin{array}{@{}l@{~~~~~}r@{\ =\ }l@{}l}
  \inpit{\carsymbol 1}
 &\carpp{\cnspp x l}
 &x
  \footroom
\\\inpit{\cdrsymbol 1}
 &\cdrpp{\cnspp x l}
 &l
  \footroom
\\\inpit{\carsymbol 1'}
 &\carpp{l'}
 &x
 &\nottight\antiimplies l'\tightequal\cnspp x l
  \footroom
\\\inpit{\cdrsymbol 1'}
 &\cdrpp{l'}
 &l
 &\nottight\antiimplies l'\tightequal\cnspp x l
\\\end{array}}
\par\noindent\math{\begin{array}{@{}l@{~~~~~}l@{\ =\ }l@{}l}
  \inpit{\mbpsymbol 1}
 &\mbppp x\nilpp
 &\falsepp
\\\inpit{\mbpsymbol 2}
 &\mbppp x{\cnspp y l}
 &\truepp
 &\nottight\antiimplies x\tightequal y
\\\inpit{\mbpsymbol 3}
 &\mbppp x{\cnspp y l}
 &\mbppp x l
 &\nottight\antiimplies x\tightnotequal y
\\\end{array}}
\par\noindent\math{\begin{array}{@{}l@{~~~~~}l@{\ =\ }l@{}l}
  \inpit{\dloncesymbol 1}
 &\dloncepp x{\cnspp y l}
 &l
 &\nottight\antiimplies x\tightequal y
\\\inpit{\dloncesymbol 2}
 &\dloncepp x{\cnspp y l}
 &\cnspp y{\dloncepp x l}
 &\nottight\antiimplies x\tightnotequal y
\\\end{array}}
\par\halftop\noindent

\par\indent
Immediate consequences of the axiom \inpit{\app\lists\nat 1} 
and the definitions \inpit{\carsymbol 1} and \inpit{\cdrsymbol 1} 
are the lemma \inpit{\cnssymbol 1}
and its flattened version \inpit{\cnssymbol 1'}:
\par\halftop\noindent\math{\begin{array}{@{}l@{~~~~~~~}r@{\ =\ }l}
  \inpit{\cnssymbol 1}
 &\cnspp{\carpp{l'}}{\cdrpp{l'}}
 &l'
  \nottight\antiimplies l'\tightnotequal\nilpp
  \footroom
\\\inpit{\cnssymbol 1'}
 &\cnspp x l
 &l'
  \nottight\antiimplies l'\tightnotequal\nilpp
  \nottight\und x\tightequal\carpp{l'}
  \nottight\und l\tightequal\cdrpp{l'}
  \\\end{array}}
\par\halftop\halftop\indent
Furthermore, let us define the \myboolean\ function 
\maths{
\FUNDEF{\lexlessymbol}{\app\lists\nat,\app\lists\nat}\bool}, \hskip.2em
which lexicographically
compares lists according to the ordering of the natural numbers,
and 
\maths{
\FUNDEF\lexlimlessymbol{\app\lists\nat,\app\lists\nat,\nat}\bool}, \hskip.2em
which further restricts the length of the first argument 
to be less than the number given as third argument:
\par\noindent\math{\begin{array}{@{}l@{~~~~~}l@{\ =\ }l@{}l}
  \inpit{
  \lexlessymbol
  1}
 &\lexlespp l\nilpp
 &\falsepp
\\\inpit{
   \lexlessymbol
   2}
 &\lexlespp\nilpp{\cnspp y k}
 &\truepp
\\\inpit{
  \lexlessymbol
  3}
 &\lexlespp{\cnspp x l}{\cnspp y k}
 &\lexlespp l k
 &\nottight\antiimplies x\tightequal y
\\\inpit{
  \lexlessymbol
  4}
 &\lexlespp{\cnspp x l}{\cnspp y k}
 &\lespp x y
 &\nottight\antiimplies x\tightnotequal y
\\\end{array}}
\par\noindent\math{\begin{array}{@{}l@{~~~~~}l@{\ =\ }l@{}l}
  \inpit{
  \lexlimlessymbol
  1}
 &\lexlimlespp l k x
 &\andpp{\lexlespp l k}{\lespp{\lengthpp l}x}
\\\end{array}}
\par\halftop\indent Such 
\index{lexicographic combination}lexicographic combinations play 
an important \role\ in \wellfoundedness\ arguments of induction proofs,
because they combine given \wellfounded\ orderings into new 
\wellfounded\ orderings, \hskip.1em
provided there is an upper bound
for the length of the list:\footnote{%
 The length limit is required because otherwise we 
 have the following counterexample to termination:
 \maths
 {\inpit{\spp\zeropp}\comma\inpit{\zeropp,\spp\zeropp}\comma
  \inpit{\zeropp,\zeropp,\spp\zeropp}\comma
  \inpit{\zeropp,\zeropp,\zeropp,\spp\zeropp}\comma\ldots}. \hskip.4em
 Note that the need to compare lists of different lengths 
 typically arises in mutual induction proofs where the
 induction hypotheses have a different number of free
 variables at \index{measured positions}measured positions. \hskip.3em
 See \cite[\litsectref{3.2.2}]{wirthcardinal} for a nice example.%
}%
\par\noindent\headroom\math{\begin{array}{@{}l@{~~~~~}l}
  \inpit{
  \lexlimlessymbol
  2}
 &\Wellfpp{\lambda l,k\stopq\inpit{\lexlimlespp l k x=\truepp}}
\\\end{array}}
\par\halftop\halftop\halftop\halftop\halftop\noindent
Finally note that analogous axioms can be used to specify any other data type
generated by constructors,
such as pairs of natural numbers or binary trees over such pairs.%
\pagebreak
\subsection{The  Standard High-Level Method of Mathematical Induction}\label
{subsection The  Standard High-Level Method of Mathematical Induction}

In general,
the intuitive and procedural aspects of a mathematical proof method 
are not completely captured by its 
logic formalization.
\hskip.1em
For actually finding and automating proofs by induction,
we also need effective heuristics.

In the everyday mathematical practice of an advanced \mbox{theoretical journal,} 
\hskip.2em
the common inductive arguments 
are hardly ever carried out explicitly.
\mbox{Instead,} 
\mbox{the proof} 
reads something like
\structuralinductionindex%
``by structural induction on \math n, \qedabbrev''\ \hskip.1em or 
\hskip.05em
\mbox{``by (\noetherian) induction on \pair x y} \mbox{over \math <, 
\qedabbrev\closequotecommasmallextraspace}
expecting that the mathematically 
\mbox{educated} reader could easily expand the 
proof if in \nolinebreak doubt. \hskip.2em
In \nolinebreak contrast, 
difficult inductive arguments, sometimes covering several pages,\footnote{%
 Such difficult inductive arguments
 are the proofs of \hilbert's
 {\em first \mbox{\math\varepsilon-theorem}} 
 \cite{grundlagen-second-edition-volume-two}, \hskip.2em 
 \gentzen's {\em Hauptsatz} \nolinebreak\cite{gentzen}, 
 and 
 \index{confluence}%
 confluence theorems such as the ones in \nolinebreak\cite
 {gwrta}, \cite{wirth-jsc}.%
}
require considerable ingenuity and have to be carried out
in the journal explicitly.

In case of a proof on natural numbers,
the experienced mathematician might engineer his proof roughly according to
the following pattern:
\begin{quote}%
He starts with the conjecture and simplifies it by case analysis,
\mbox{typically} based on the axiom \nlbmath{\inpit{\nat 1}}. \hskip.1em
When he realizes that the current goal is
similar to an instance of the conjecture, 
he applies the instantiated conjecture just like a lemma, 
but keeps in mind that he has actually applied an induction hypothesis. 
\hskip.2em
Finally, using the free variables of the conjecture, 
he constructs some ordering
whose \wellfoundedness\ follows from
the axiom \nlbmath{\Wellfpp{\lambda
{\boundvari x{},\boundvari y{}}
\stopq
\inpit{\spp x\tightequal y}}}
\hskip.2em
and in which all instances of the conjecture 
applied as induction hypotheses 
are smaller than the original conjecture.
\end{quote}
\noindent\label{section items}%
The hard tasks of a proof by mathematical induction are thus:
\begin{description}\item[(Induction-Hypotheses Task) ]\mbox{}\\to find the
 numerous induction hypotheses\commanospace\footnote{%
 \majorheadroom
 As, \eg, 
 in the proof of \gentzen's Hauptsatz on Cut-elimination.} and\noitem
\item[(Induction-Ordering Task) ]\mbox{}\\to construct 
 an {\em induction ordering}\/ for the proof, \ie\ 
 a \wellfounded\ ordering that satisfies
 the ordering constraints of all
 these induction hypotheses in parallel.\footnotemark
\end{description}%
\footnotetext{%
 \majorheadroom
 For instance, this was the 
 hard
 part in the elimination of the \math\varepsilon-formulas
 in the proof of the \nth 1\,\math\varepsilon-theorem
 in \cite{grundlagen-second-edition-volume-two}, \ 
 and in the proof of 
 the \index{consistency}consistency of arithmetic by the \nlbmath\varepsilon-substitution
 method in 
 \nlbcite{ackermann-consistency-of-arithmetic}.%
}%
\begin{sloppypar}
\par\noindent
The above induction method 
can be formalized as
an application of the \theoremofnoetherianinduction.
For non-trivial proofs, 
mathematicians indeed prefer the axioms of \pieriindex\pieri's specification
in combination with the \theoremofnoetherianinduction\ 
\nlbmath{\inpit{\ident{N}}} \hskip.1em
to \peanoindex\peano's alternative with the \axiomofstructuralinduction\ 
\nlbmaths{\inpit{\ident S}}, \hskip.2em
because the instances for \math P and \nlbmath< in \nlbmath{\inpit{\ident{N}}}
are often easier to find than the instances for \math P in 
\nlbmath{\inpit{\ident S}} are.%
\vfill\pagebreak
\end{sloppypar}
\subsection{{\DescenteInfinie}}\label
{subsection DescenteInfinie}
The soundness of the induction method of \sectref
{subsection The Standard High-Level Method of Mathematical Induction}
is most easily seen when the argument is structured as a proof by contradiction,
assuming a counterexample. \hskip.1em
For \fermatindex\fermat's historic reinvention of the method,
it \nolinebreak is thus just natural that he developed 
the method 
in terms of assumed counterexamples.\footnotemark\ \hskip.2em
Here is \fermatindex\fermat's Method of {\em\DescenteInfinie}\/ 
in modern language, very roughly speaking:%
\footnotetext{%
 \Cfnlb\ \cite{fermat-oeuvres}, \cite{fermat-career}, 
 \cite{From-Fermat-to-Gauss}, \cite{fermatsproof}.%
}%
\begin{quote}\sloppy
A proposition \nlbmath{\app P w}
can be proved by {\em\descenteinfinie}\/ as follows:
{\em Show that for each 
assumed counterexample\/ \nlbmath v of\/ \nlbmath{P}
there is a smaller counter\-example\/ \nlbmath u of\/ \nlbmath{P}
\wrt\ a \wellfounded\ relation\/ \nlbmath<, \hskip.3em
which does not depend on the counterexamples.}
\end{quote}
If this method is executed successfully,
we have proved \nlbmath{\forall w\stopq\app P w}
because no counterexample can be a \math<-minimal one, \hskip.1em
and so the \wellfoundedness\ of \nlbmath< implies that
there are no counterexamples at all.

It was very hard for \fermatindex\fermat\
to obtain a positive version of his counterexample method.\footnote{%
 \majorheadroom
 \fermatindex\fermat\ reported in his letter for \huygensname\
 \huygenslifetime\ 
 that he had had problems
 applying the Method of {\em\DescenteInfinie}\/ 
 to positive mathematical statements. \hskip.2em
 See \cite[\p\,11]{fermatsproof} and the references there, 
 in particular \cite[\Vol\,II, \p\,432]{fermat-oeuvres}. \par
 Moreover,
 a natural-language presentation via {\em\descenteinfinie}\/
 (such as \fermatindex\fermat's representation in Latin) \hskip.1em
 is often simpler than a presentation 
 via the \theoremofnoetherianinduction,
 because it is easier to speak of one counterexample \nlbmath v and to find
 one smaller counterexample \nlbmath u, 
 than to manage the dependences of universally quantified variables.%
} \hskip.2em
Nowadays every logician immediately realizes that a formalization of 
the method of {\em\descenteinfinie}\/ is obtained from the 
\theoremofnoetherianinduction\ \nlbmath{\inpit{\ident N}}
(\cfnlb\ 
 \sectref{subsection noetherian induction})
\hskip.1em
simply by replacing 
\\[-.5ex]\noindent\LINEmaths{\app P  v
         {\nottight\antiimplies}
           \theinnermostpartofW}{}
\\with its contrapositive
\\[-.5ex]
\mbox{~\,}\LINEmaths{\neg\app P  v
         {\nottight\implies}\exists u\tight<v\stopq\neg\app P  u}.
\begin{sloppypar}%
\par\halftop\indent 
For the history of the {\em automation}\/ of induction,
however,
that difference between an implication and its contrapositive 
is not crucial. \hskip.3em
Indeed, 
for this endeavor, 
the relevant mathematical logic was formalized during the 
\nth{19}~and the \nth{20}~centuries and we may confine ourselves 
to \mbox{\em classical} \mbox{(\ie\ two-valued)} logics. \hskip.2em
What actually matters here is the heuristic task of finding proofs. \hskip.2em
Therefore
---~overlooking that difference~---
we will take {\it\descenteinfinie}\/ 
in the remainder of \englishdiesespapier\footnotemark\ \hskip.1em
simply as a synonym for the modern
standard high-level method of mathematical induction described 
\nolinebreak in
\nlbsectref{subsection The Standard High-Level Method of Mathematical Induction}.%
\footnotetext{%
 \majorheadroom
 In general, in the tradition of \cite{wirthcardinal}, \hskip.1em
 {\it\descenteinfinie}\/ is 
 nowadays 
 taken 
 as a synonym for the
 standard high-level method of mathematical induction as 
 described in \nlbsectref
 {subsection The Standard High-Level Method of Mathematical Induction}.
 \hskip.3em
 This way of using the term {\it``\descenteinfinie''}\/ \hskip.2em 
 is found in
 \makeaciteoftwo{DBLP:conf/lics/BrotherstonS07}{brotherston-cut-elimination},
 \cite{voicu-li-di},
 \makeaciteoffour
 {zombie}
 {swp200601}
 {SR--2011--01}
 {wirth-manifesto-ljigpl}. \par
 If,
 however,
 the historical perspective before the \nth{19}\,century is taken,
 then 
 this identification is not appropriate because a more fine-grained 
 differentiation is required, 
 such as found in \cite{From-Fermat-to-Gauss},
 \cite{fermatsproof}.%
}%
\end{sloppypar}

\halftop\halftop\indent
Let us now prove the lemmas \inpit{\plussymbol 3} and \inpit{\lessymbol 7} of
\sectref{subsection Mathematical Induction and the Natural Numbers} \hskip.1em
(in the axiomatic context of 
 \sectref{subsection Mathematical Induction and the Natural Numbers}) \hskip.1em
by {\em\descenteinfinie}, \hskip.2em
seen as the standard high-level method of 
mathematical induction described in 
\sectref{subsection The Standard High-Level Method of Mathematical Induction}.%
\pagebreak
\halftop\halftop
\label{section example first proof of (+3)}%
\begin{example}[Proof of \inpit{\plussymbol 3} by {\em\descenteinfinie}\/]
\label{example first proof of (+3)}%
\par
\noindent
By application of
the \theoremofnoetherianinduction\ \nlbmath{\inpit{\ident N}}
(\cfnlb\ \sectref{subsection noetherian induction}) \hskip.2em
with \nlbmath P set to
\maths{\lambda x,y\stopq\inpit{\plusppnoparentheses x y
=\plusppnoparentheses y x}}, \hskip.3em
and the variables \nlbmaths v, \nlbmath u
renamed to \pair x y, \pair{x''}{y''}, \hskip.2em
respectively, \hskip.1em
the conjectured lemma \nlbmath{\inpit{\plussymbol 3}} \hskip.1em
reduces to 
\par
\noindent\LINEmaths\bigexampleformulaone.
\par
\noindent
Let us focus on the sub-formula 
\bigmaths{\plusppnoparentheses x y
=\plusppnoparentheses y x}.
Based on axiom \nlbmath{\inpit{\nat 1}} 
we \nolinebreak can reduce this task to the 
two cases \bigmaths{x\tightequal\zeropp}{} and \bigmathnlb
{x\tightequal\spp{x'}}{}
with the two goals
\par
\noindent\LINEmaths{\begin{array}{l@{~~~~~~~~~~~~~~~~~~~~~~~~~~~~}r}
  \plusppnoparentheses\zeropp y=\plusppnoparentheses y\zeropp;
 &\plusppnoparentheses{\spp{x'}}y=\plusppnoparentheses y{\spp{x'}};
\\\end{array}}{}
\par
\noindent
respectively. \hskip.2em
They simplify by \inpit{\plussymbol 1} and \inpit{\plussymbol 2} to
\par
\noindent\LINEmaths{
\begin{array}{l@{~~~~~~~~~~~~~~~~~~~~~~~~~~~~~~~~~}r}
  y=\plusppnoparentheses y\zeropp;
 &\spp{\plusppnoparentheses{x'}y}=\plusppnoparentheses y{\spp{x'}};
\\\end{array}}{}
\par
\noindent 
respectively. \hskip.2em
Based on axiom \nlbmath{\inpit{\nat 1}} we can reduce each of these goals to the 
two cases \bigmaths{y\tightequal\zeropp}{} and \bigmathnlb
{y\tightequal\spp{y'}},
which leaves us with the four open goals
\par
\noindent\LINEmaths{
\begin{array}{l@{~~~~~~~~~~~~~~~~}r}
  \zeropp=\plusppnoparentheses\zeropp\zeropp;
 &\spp{\plusppnoparentheses{x'}\zeropp}=\plusppnoparentheses\zeropp{\spp{x'}};
  \mediumfootroom
\\\spp{y'}=\plusppnoparentheses{\spp{y'}}\zeropp;
 &\spp{\plusppnoparentheses{x'}{\spp{y'}}}
  =\plusppnoparentheses{\spp{y'}}{\spp{x'}}.
\\\end{array}}{}
\par
\noindent 
They simplify by \inpit{\plussymbol 1} and \inpit{\plussymbol 2} to
\par
\noindent\LINEmaths{
\begin{array}{l@{~~~~~~~~~~~~~~~~}r}
  \zeropp=\zeropp;
 &\spp{\plusppnoparentheses{x'}\zeropp}=\spp{x'};
  \mediumfootroom
\\\spp{y'}=\spp{\plusppnoparentheses{y'}\zeropp};
 &\spp{\plusppnoparentheses{x'}{\spp{y'}}}
  =\spp{\plusppnoparentheses{y'}{\spp{x'}}};
\\\end{array}}{}
\par
\noindent 
respectively. \hskip.2em
Now
we 
instantiate the induction hypothesis
that is available in the context\footnote{%
 On how this availability can be understood formally, 
 see \cite{sergetableau}.%
} 
given by our above formula
in four different forms, namely we 
instantiate 
\pair{x''}{y''} with
\pair{x'}{\zeropp}, \hskip.1em
\pair{\zeropp}{y'}, \hskip.1em
\pair{x'}{\spp{y'}}, \hskip.1em
and
\pair{\spp{x'}}{y'}, \hskip.1em
respectively.
Rewriting with these instances, 
the four goals become:
\par\noindent\LINEmaths{
\begin{array}{l@{~~~~~~~~~~~~~~~~}r}
  \zeropp=\zeropp;
 &\spp{\plusppnoparentheses\zeropp{x'}}=\spp{x'};
  \mediumfootroom
\\\spp{y'}=\spp{\plusppnoparentheses\zeropp{y'}};
 &\spp{\plusppnoparentheses{\spp{y'}}{x'}}
  =\spp{\plusppnoparentheses{\spp{x'}}{y'}};
\\\end{array}}{}
\par
\noindent 
which simplify by \inpit{\plussymbol 1} and \inpit{\plussymbol 2} to
\par
\noindent\LINEmaths{
\begin{array}{l@{~~~~~~~~~~~~~~~~~~~~~}r}
  \zeropp=\zeropp;
 &\spp{x'}=\spp{x'};
  \mediumfootroom
\\\spp{y'}=\spp{y'};
 &\spp{\spp{\plusppnoparentheses{y'}{x'}}}
  =\spp{\spp{\plusppnoparentheses{x'}{y'}}}.
\\\end{array}}{}
\par
\noindent 
Now the first three goals follow directly from the 
reflexivity of equality,
whereas the last goal also needs 
an application of our induction hypothesis:
This time we \nolinebreak have to instantiate
\pair{x''}{y''} with 
\pair{x'}{y'}.
\par
Finally, we instantiate our induction ordering \nlbmath <
to the lexicographic combination of length less than\,\,3 of 
the ordering of the natural numbers.
If \nolinebreak we \nolinebreak read our pairs as two-element lists,
\ie\ \pair{x''}{y''} as \cnspp{x''}{\cnspp{y''}\nilpp},
then we can set \nlbmath < to
\par\noindent\LINEmaths{\lambda l,k\stopq\inpit{
\lexlimlespp l k{\spp{\spp{\spp\zeropp}}}=\truepp}}, 
\par\noindent which is \wellfounded\ according to 
\inpit{
\lexlimlessymbol
2} 
(\cfnlb\ \sectref{subsection Standard Data Types}). \hskip.2em
Then it is trivial to show that
\pair{\spp{x'}}{\spp{y'}} is greater than each of \hskip.2em  
\pair{x'}{\zeropp},  \hskip.2em 
\pair{\zeropp}{y'},  \hskip.2em 
\pair{x'}{\spp{y'}}, \hskip.2em 
\pair{\spp{x'}}{y'}, \hskip.2em 
\pair{x'}{y'}. 

This completes the proof of our conjecture by {\it\descenteinfinie}.
\getittotheright\qed\end{example}\pagebreak
\label{section example first proof of (less7)}%
\begin{example}[Proof of \inpit{\lessymbol 7} by {\em\descenteinfinie}]
\label{example first proof of (less7)}%
\sloppy
\\[+1.3ex]
\noindent
In the previous proof in \examref{example first proof of (+3)}
we 
made 
the application of the 
\theoremofnoetherianinduction\ most explicit,
and so its presentation was rather formal \wrt\ the underlying logic.
Contrary to this,
let \nolinebreak us now proceed more 
in the vernacular of a working mathematician.
Moreover, instead of \bigmaths{p\tightequal\truepp}, 
let us just write \nlbmaths p. \hskip.3em

To prove the strengthened transitivity of \lessymbol\ as expressed 
in lemma \inpit{\lessymbol 7} in the axiomatic context of 
\sectref{subsection Mathematical Induction and the Natural Numbers}, \hskip.2em
we 
have to show
\par\halftop\noindent\mbox{~~~~~~~~~~~~~~~~~~~~~~~~~~~~~~~~~~~}\maths{
  \lespp{\spp{x}}z
  \nottight\antiimplies
  {\lespp x y
   \und\hskip.05em
   \lespp y z}}.
\par\halftop\noindent 
Let us reduce the last literal.
To this end,
we apply the axiom \nlbmath{\inpit{\nat 1}} 
once to \math y and once to \maths z. \hskip.3em
Then, after reduction with \inpit{\lessymbol 1}, 
the two base cases have an atom \nlbmath{\falsepp} in their conditions,
abbreviating \bigmaths{\falsepp\tightequal\truepp},
which is false according to \inpit{\bool 2}, 
and so the base cases are true 
(\hskip-.05em{\it ex falso quodlibet}\/\nolinebreak\hskip.05em\nolinebreak). 
\hskip.3em
The remaining case, 
where we have both \bigmathnlb{y\tightequal\spp{y'}}{} and
\bigmathnlb{z\tightequal\spp{z'}},
reduces with \inpit{\lessymbol 3} to 
\par\halftop\noindent\mbox{~~~~~~~~~~~~~~~~~~~~~~~~~~~~~~~~~~~~~~}\math{
  \lespp{x}{z'}\nottight\antiimplies
  {\lespp x{\spp{y'}}
  \tightund\hskip.05em
  \lespp{y'}{z'}
}}
\par\halftop\noindent
If we apply the induction hypothesis instantiated via 
\bigmaths{\{
 y\tight\mapsto y'
\comma
 z\tight\mapsto z'
\}}{}
to match the last 
literal,
then we obtain the two goals
\par\halftop\noindent
\mbox{~~~~~~~~~~~~~~~~~~~~~~~~~~~~~~~~~~~~~~}\maths{
  \lespp{x}{z'}
  \nottight\antiimplies
   \lespp x{\spp{y'}}
   \tightund\hskip.05em
   \lespp{y'}{z'}
   \tightund\hskip.05em
   \lespp{\spp{x}}{z'}
  }{}
\par\noindent\mbox{~~~}\math{
  {
  \lespp x{y'}
  \tightoder\hskip.05em
  \lespp{\spp{x}}{z'}
  \tightoder\hskip.05em
  \lespp{x}{z'}
  }
  \nottight\antiimplies
  {\lespp x{\spp{y'}}
  \tightund\hskip.05em
  \lespp{y'}{z'}}}
\par\halftop\noindent
By elimination of irrelevant literals,
the first goal can be reduced to the valid conjecture 
\bigmaths{\lespp{x}{z'}\nottight\antiimplies\lespp{\spp{x}}{z'}},
but we cannot obtain a lemma 
simpler than our initial conjecture \inpit{\lessymbol 7}
by generalization and elimination of irrelevant literals
from the second goal.
This means that the application of the given instantiation of the 
induction hypothesis is useless.

Thus, \hskip.1em
instead of induction-hypothesis application, \hskip.1em
we had better apply the 
axiom \nlbmath{\inpit{\nat 1}}
also to \maths x, \hskip.3em
obtaining
the cases \bigmaths{x\tightequal\zeropp}{}
and \bigmaths{x\tightequal\spp{x'}}{} with the two goals
\ ---~~after reduction with \inpit{\lessymbol 2} and \inpit{\lessymbol 3}~~---
\par\noindent\mbox{~~~~~~~~~~~~~~~~~~~~~~~~~~~~~~~~~~~~~~}\math{
  \lespp{\zeropp}{z'}
  \nottight\antiimplies
  \lespp{y'}{z'}
}
\par\noindent\mbox{~~~~~~~~~~~~~~~~~~~~~~~~~~~~~~\,~~}\maths{
  \lespp{\spp{x'}}{z'}
  \nottight{\nottight\antiimplies}
  {\lespp{x'}{y'}\und\lespp{y'}{z'}}},
\par\halftop\noindent respectively. \hskip.3em
The first is trivial by \inpit{\lessymbol 1}, \hskip.1em
\inpit{\lessymbol 2} \hskip.1em
after another application of the axiom \math{\inpit{\nat 1}} to 
\nlbmaths{z'}. \hskip.6em
The second is just an instance of the induction hypothesis
via 
\bigmaths{\{
 x\tight\mapsto x'
\comma
 y\tight\mapsto y'
\comma
 z\tight\mapsto z'
\}}. \hskip.1em
As the induction ordering we can 
select
any of the variables 
of the original conjecture \wrt\ the irreflexive ordering on the natural
numbers or \wrt\ the successor relation.

This completes the proof of the conjecture by {\it\descenteinfinie}.

Note that we also have made clear that the given proof can only be successful
with an induction hypotheses where all variables are instantiated with
predecessors.
It is actually possible to show that this simple example
{\it---~ceteris paribus~---}
requires an induction hypothesis resulting from an instance 
\maths{\{
 x\tight\mapsto x''
,\,\,
 y\tight\mapsto y''
,\,\,
 z\tight\mapsto z''
\}}{}
where, \hskip.2em
for some 
meta-level natural number \nlbmaths n, \hskip.4em
we have 
\ \bigmaths{ 
 x\tightequal\sppiterated{n+1}{x''}\nottight\und
 y\tightequal\sppiterated{n+1}{y''}\nottight\und
 z\tightequal\sppiterated{n+1}{z''}}.\getittotheright\qed\pagebreak\end{example}
\subsection{Explicit Induction}\label{subsection Explicit Induction}%
\subsubsection{From the \theoremofnoetherianinduction\ to Explicit Induction}
\index{induction!explicit|(}%
To admit the realization of 
the standard high-level method of mathematical induction 
as described in \sectref
{subsection The  Standard High-Level Method of Mathematical Induction}, 
\hskip.2em
a proof calculus should have an explicit concept of an induction hypothesis.
Moreover, 
it would have to cope in some form with the second-order variables 
\nlbmath P and \nlbmath < \hskip.2em
in the 
\theoremofnoetherianinduction\ \nlbmath{\inpit{\ident N}}
(\cfnlb\ \sectref{subsection noetherian induction}), \hskip.2em
and with the second-order variable \nlbmath Q in the 
definition of \wellfoundedness\ \nlbmath{\inpit{\Wellfpp{\tight<}}}
(\cfnlb\ \sectref{subsection Well-Foundedness and Termination}). \hskip.2em
\\\indent
Such an implementation needs special care
regarding the calculus and its heuristics. 
\mbox{For example,}
the theorem provers for higher-order logic 
with the strongest automation today%
\arXivfootnotemarkref{note higher-order theorem provers}
are yet not able
to prove standard inductive theorems by just adding 
the \theoremofnoetherianinduction,
which immediately effects an explosion of the search space. \hskip.3em
It \nolinebreak is a main obstacle to practical usefulness of 
higher-order theorem provers that they are still
poor in the automation of \mbox{induction.}
\\\indent
Therefore, it is probable that 
---~on the basis of
    the logic calculi and the computer technology of the 1970s~---
\boyer\ and \moore\ 
would also have failed to implement induction via 
these human-oriented and higher-order features.
Instead,
they confined the concept of an induction
hypothesis to the internals of single reductive
inference steps 
---~namely the applications of the so-called {\em\inductionrule}~--- 
and restricted all other
inference steps to quantifier-free first-order deductive reasoning.  
These decisions were crucial to their success.
\\\indent
Described in terms of the \theoremofnoetherianinduction,
this {\em\inductionrule}\/ immediately instantiates the higher-order variables
\nlbmath P and \nlbmath < \hskip.2em
with \firstorder\ predicates. \hskip.1em
This is rather straightforward for the predicate variable \nlbmaths P,
\hskip.2em
which simply becomes the (properly simplified and generalized) \hskip.1em
quantifier-free
\firstorder\ conjecture that is to be proved
by induction,
and 
the tuple of the free \firstorder\ variables of this conjecture
takes the place of the single argument of 
\nlbmaths P; \hskip.1em
\mbox{\cfnlb\,\examref{example second proof of (+3)}\,below.}
\\\indent
The instantiation of the 
higher-order variable 
\nlbmath < \hskip.1em
is \nolinebreak more difficult: \hskip.2em
Instead of a simple instantiation,
the whole context of its two occurrences is transformed. \hskip.3em
For the first occurrence, namely the one in 
the sub-formula \bigmaths{\theinnermostpartofW},
the whole sub-formula is replaced with a conjunction of instances of 
\nlbmaths{\app P u}, \hskip.2em
for which \math u is known to be smaller than \nlbmath v in some 
lexicographic combination of given orderings 
that are already known to be \wellfounded. \hskip.2em
As a consequence, the second occurrence of \nlbmaths <, \hskip.3em
\ie\ the one in \Wellfpp <, \hskip.2em
simplifies to true, \hskip.1em
and so we \nolinebreak can drop the conjunction that contains it.
\\\indent
At a first glance,
it seems highly unlikely that 
there could be 
any framework of proof-search heuristics
in which such an \inductionrule\ could succeed in implementing
all applications of the \theoremofnoetherianinduction, \hskip.1em
simply because this rule
has to solve the two hard tasks of an induction proof, \hskip.1em
namely the Induction-Hypotheses Task and the Induction-Ordering Task
(\cfnlb\ 
 \sectref{subsection The Standard High-Level Method of Mathematical Induction}),
\hskip.2em
right at the beginning of the proof attempt, \hskip.1em
before the proof has been sufficiently developed
to exhibit its structural difficulties.
\\\indent
Most surprisingly, but as a matter of fact,
the \inductionrule\ has proved to be most successful 
in realizing all applications of the \theoremofnoetherianinduction\
required within the proof-search heuristics of the \boyermoorewaterfall\
(\cfnlb\ \figuref{figure waterfall}). \hskip.3em
Essential for this success
is the relatively weak quantifier-free \firstorder\ logic:
\begin{itemize}\noitem\item No new symbols have to be introduced during the 
proof, such as the ones of quantifier elimination.
Therefore,
the required instances of the induction hypothesis can already be denoted
when the \inductionrule\ is applied.\footnotemark\pagebreak\par
\footnotetext{%
 \Cfnlb\ \noteref{note Newman}.%
}\noitem\item
A general peculiarity of induction,\footnote{%
 See \lititemref 2 of \nlbsectref
 {subsection Proof-Theoretical Peculiarities of Mathematical Induction}.%
} \hskip.1em
namely
that the 
formulation of 
lemmas often requires the definition 
of new recursive functions, \hskip.1em
is aggravated by the weakness of the logic; \hskip.2em
and the user 
is actually required to provide
further guidance for the \inductionrule\ 
via these new function definitions.\footnote{%
 \Cfnlb\ \sectref{section Conclusion}.%
}\noitem
\end{itemize}
Moreover, this success crucially depends on the possibility to 
generate additional lemmas
that are proved by subsequent inductions, \hskip.1em
which is best shown by an example. \
\label{section example second proof of (+3)}%
\begin{example}
[Proof of \inpit{\plussymbol 3} by \index{induction!explicit}Explicit Induction]
\label{example second proof of (+3)}%
\\\noindent
Let us prove \inpit{\plussymbol 3}
in the context of 
\sectref{subsection Mathematical Induction and the Natural Numbers}, \hskip.2em
just as we have done already in \examref{example first proof of (+3)}
(\cfnlb\ \sectref{section example first proof of (+3)}), \hskip.2em
but now with the \inductionrule\ as the only way to 
apply the \theoremofnoetherianinduction.
\\\indent
As the conjecture is already properly simplified and concise,
we instantiate \nlbmath{\app P w} 
in the \theoremofnoetherianinduction\
again to the whole conjecture
and reduce this conjecture by application of 
the \theoremofnoetherianinduction\
again to 
\par\noindent\LINEmaths\bigexampleformulaone.
\par\noindent Based, \hskip.1em
roughly speaking, \hskip.1em
on a termination analysis for the function \nlbmath\plussymbol, \hskip.3em
the \mbox{heuristic} of the 
\inductionrule\ of \index{induction!explicit}explicit induction 
suggests to 
instantiate \nlbmath < \hskip.1em
to 
\bigmaths{\lambda\pair{x''}{y''},\pair x y\stopq\inpit
{\spp{x''}\tightequal x}}.%
\\\noindent
As this relation is known to be \wellfounded, \hskip.1em
the \inductionrule\ reduces the task 
based on axiom \nlbmath{\inpit{\nat 1}} \hskip.1em
to two goals, 
namely the base case 
\par\noindent\LINEmaths{\plusppnoparentheses\zeropp y
=\plusppnoparentheses y\zeropp};\mbox{~~~~~~~~~}
\\[-.5ex]\noindent and the step case
\\[-.5ex]\noindent\mbox{~~~~~~~~~~~~~~~~~~~}\LINEmaths
{\inpit{\plusppnoparentheses{\spp{x'}}y
=\plusppnoparentheses y{\spp{x'}}}
\nottight\antiimplies\inpit{\plusppnoparentheses{x'}{y}
=\plusppnoparentheses{y}{x'}}}.
\par\indent
This completes the application of the \inductionrule.
Thus, 
instances of the induction hypothesis 
can no longer be applied in the further proof
(except the ones that have already been added explicitly 
 as conditions of step cases by the \inductionrule).
\\\indent
The \inductionrules\ of the \boyermooretheoremprovers\
are not able to find the many instances 
we applied in the proof of \examref{example first proof of (+3)}\@. \hskip.1em
This is different for a theoretically more powerful
\inductionrule\ suggested 
by \walthername\ \waltherlifetime, \hskip.2em 
which actually finds the proof of 
\examref{example first proof of (+3)}.\footnote{%
 See \cite[\p\,99\f]{waltherIJCAI93}. 
 On \litspageref{100},
 the most interesting 
 step case computed by \walther's \inductionrule\ is
 (rewritten to constructor-style):
 \par\halftop\noindent\LINEmaths{\plusppnoparentheses{\spp x}{\spp y}
 =\plusppnoparentheses{\spp y}{\spp x}
 \nottight\antiimplies\inparentheses{\plusppnoparentheses x{\spp y}
 =\plusppnoparentheses{\spp y}x
 \nottight\und\forall z\stopq
 \inpit{\plusppnoparentheses z y=\plusppnoparentheses y z}}}.
 \par\halftop\noindent In practice, however, \walther's \inductionrule\ 
 has turned out to be 
 overall
 less successful when applied 
 within a heuristic framework similar to the \boyermoorewaterfall\
 (\cfnlb\ \figuref{figure waterfall}).%
}
\hskip.1em
In general, 
however,
for harder conjectures,
a simulation of {\it\descenteinfinie}\/ 
by the \inductionrule\ of \index{induction!explicit}explicit induction 
would require an arbitrary look-ahead
into the proofs,
depending on the size of the structure of these proofs; \hskip.2em
thus, 
because the \inductionrule\ is understood to have a limited look-ahead
into the proofs,
such a simulation would not fall under the paradigm of \index{induction!explicit}explicit
induction 
any more.
Indeed, 
the look-ahead of \inductionrules\ into the proofs is
typically not more than a single unfolding of a single
occurrence of a recursive function symbol,
for each such occurrence in the conjecture.
\\\indent
Note that the two above goals of the base and the step case 
can also be obtained by reducing 
the input conjecture 
with an instance of axiom~\inpit{\ident S} 
(\cfnlb\ 
\nlbsectref{subsection Mathematical Induction and the Natural Numbers}), 
\hskip.3em
\pagebreak
\ie\ with 
the \axiomofstructuralinduction\ over \nlbmath\zeropp\ and \nlbmaths\ssymbol.
\hskip.4em
Nevertheless, 
the \inductionrule\ of the \boyermooretheoremprovers\ is, 
in general, 
able to produce much more complicated 
base and step cases than
those that can be obtained by \mbox{reduction} with
the axiom \nlbmath{\inpit{\ident S}}.
\\\indent
Now the first goal is simplified again to 
\bigmaths{y=\plusppnoparentheses y\zeropp},
and then another application of the \inductionrule\ results in two goals 
that can be proved without further induction.
\\\indent
The second goal is simplified to 
\par\noindent\mbox{~~~~~~~~~~~~~~~~~~~~~~~~~~~~}\LINEmaths{
\inpit{\spp{\plusppnoparentheses{x'}y}
=\plusppnoparentheses y{\spp{x'}}}
\nottight\antiimplies\inpit{\plusppnoparentheses{x'}{y}
=\plusppnoparentheses{y}{x'}}}.
\par\noindent Now we use the condition from {\em left}\/ to right 
for rewriting only the {\em left}\/-hand side of the conclusion
and then we throw away the condition completely,
with the intention to obtain a stronger induction hypothesis 
in a subsequent induction proof.
This is the famous 
\index{cross-fertilization}{\em``cross-fertilization''}\/ of the 
\boyermoorewaterfall\ (\cfnlb\ \figuref{figure waterfall}). \hskip.4em
By \nolinebreak this, 
the simplified second goal reduces to 
\\[-1.4ex]\noindent\LINEmaths{
\spp{\plusppnoparentheses y{x'}}=\plusppnoparentheses y{\spp{x'}}}.
\par\noindent Now the \inductionrule\ triggers a
\structuralinductionindex%
structural induction on \nlbmaths y, \hskip.1em
which is successful without further induction.
\\\indent
All in all,
although the \inductionrule\ of the \boyermooretheoremprovers\ does not find 
the more complicated induction hypotheses of the 
{\it\descenteinfinie}\/ proof of 
\examref{example first proof of (+3)}
in \sectref{section example first proof of (+3)}, \hskip.2em
it is well able prove our original conjecture
with the help of the additional lemmas
\bigmathnlb{y=\plusppnoparentheses y\zeropp}{}
and
\bigmathnlb{\spp{\plusppnoparentheses y{x'}}
=\plusppnoparentheses y{\spp{x'}}}.
\\\indent
It is crucial here
that the heuristics of the \boyermoorewaterfall\ 
discover these lemmas automatically,
and that this is also typically the case in general.
\\\indent
From a logical viewpoint, 
these lemmas are redundant because they
follow from the original conjecture and the definition of 
\nlbmaths\plussymbol. \hskip.5em
From a heuristic viewpoint,
\mbox{however,}
they are more useful than the original conjecture,
because 
---~oriented for rewriting from right to left~---
their application tends to terminate 
in the context of the overall simplification by symbolic evaluation, \hskip.1em
which constitutes the first stage of
the waterfall.\getittotheright\qed\end{example}

\halftop\noindent Although 
the two proofs of the very simple conjecture 
\nlbmath{\inpit{\plussymbol 3}}
given in 
\examrefs{example first proof of (+3)}{example second proof of (+3)} 
can only give a very rough idea on the advantage of
{\it\descenteinfinie} for hard induction proofs,\footnote{\label{note Newman}%
 For some of the advantages of {\it\descenteinfinie},
 see \examref{example merging} 
 in \sectref{section example merging}, 
 and especially the more difficult, complete formal proof of 
 \newmanname's famous lemma in \cite[\litsectref{3.4}]{wirthcardinal},
 \hskip.2em 
 where the reverse of a \wellfounded\ relation is shown to be confluent 
 in case of local \index{confluence}confluence 
 {\em---~by induction \wrt\ this \wellfounded\ relation itself}. 
 The \inductionrule\ of \index{induction!explicit}explicit induction 
 cannot be applied here 
 because an eager induction hypothesis generation is not possible:
 The required instances of the induction hypothesis
 contain \math\delta-variables that can only be generated later 
 during the proof by quantifier elimination.\par
 Though \index{confluence}confluence is the \churchrosserproperty,
 the \newmanlemma\ has nothing to do with
 the \churchrossertheorem\ 
 stating the \index{confluence}confluence of the rewrite relation of 
 \mbox{\math{\alpha\beta}-reduction}
 in untyped \math\lambda-calculus,
 which has actually been
 verified with a \boyermooretheoremprover\ in the first half
 of the 1980s by \citet{church-rosser-bm} 
 (see the last paragraph of \sectref{subsection NQTHM} and \noteref
 {note shankar}) \hskip.1em
 following the short \tait/\loef\ proof found 
 \eg\ in \citep[\p\,59\ff]{lambda-calculus-final}. \hskip.3em
 Unlike the \newmanlemma,
 \shankar's proof 
 proceeds by 
 \structuralinductionindex%
 structural induction on the \math\lambda-terms,
 not by \noetherian\ induction \wrt\ the reverse of the 
 rewrite relation; \hskip.1em
 indeed, untyped \math\lambda-calculus \mbox{is not terminating.}%
}
\hskip.2em
these two proofs
nicely demonstrate how the \inductionrule\ 
of \index{induction!explicit}explicit induction
manages to prove simple theorems very efficiently
and with additional benefits for the further 
performance of the simplification procedure.

Moreover, 
for proving very hard theorems
for which the overall waterfall heuristic fails, \hskip.1em
the user can state 
hints and additional lemmas with additional notions
in any \boyermooretheoremprover,
except\,the\,\PURELISPTP.
\index{induction!explicit|)}
\subsubsection{Theoretical Viewpoint on Explicit Induction}\label
{subsubsection Theoretical Viewpoint}%
From a theoretical viewpoint,
we have to be aware of the possibility that 
the \mbox{intended} models of
specifications in \index{induction!explicit}{\em explicit-induction systems}\/
may also include non-standard models.

For the natural numbers, \hskip.1em 
for instance, \hskip.1em
there may be \mbox{\ZZ-chains}
in addition to the natural 
numbers \nlbmaths\N, \hskip.1em
whereas the higher-order specifications of \peanoindex\peano\ and 
\pieriindex\pieri\ specify exactly the natural numbers \nlbmaths\N\
up to isomorphism.\footnote{%
 Contrary to the \ZZ-chains
 (which are structures similar to the integers \ZZ,
  injectively generated from an arbitrary element 
  via \ssymbol\ and its inverse,
  where every 
  element is greater than
  every standard natural number),
 ``\ssymbol-circles'' cannot exist because
 it is possible to show by 
 \structuralinductionindex%
 structural induction on \nlbmath x \hskip.1em
 the two lemmas \bigmaths{\lespp x x\tightequal\falsepp}{}
 and \bigmaths{\lespp x{\sppiterated{n+1}x}\tightequal\truepp}{}
 for each standard meta-level natural number \nlbmaths n.%
} \hskip.3em
This is indeed the case for the case of the
\boyermooretheoremprovers\ as explained in \noteref{note kaufmann non-standard}.
\hskip.3em
These \ZZ-chains 
cannot be excluded 
because 
the inference rules realize only 
first-order deductive reasoning, \hskip.1em
except for the \inductionrule\ to which 
all applications of 
the \theoremofnoetherianinduction\ 
are confined 
and 
which does not use any higher-order properties, \hskip.1em
but only \wellfounded\ orderings that are
defined in the first-order logic of the 
\index{induction!explicit}explicit-induction system. \hskip.3em
\subsubsection{Practical Viewpoint on Explicit Induction}\label
{subsubsection Practical Viewpoint}%
Note
that the application of the \inductionrule\ 
of \index{induction!explicit}explicit induction is not implemented via a reference to
the \theoremofnoetherianinduction, \hskip.2em
but directly handles the following practical tasks
and their heuristic decisions.

In general, \hskip.1em
the {\em induction stage}\/ of the \boyermoorewaterfall\
(\cfnlb\ \figuref{figure waterfall}) \hskip.1em
applies the \inductionrule\ once to its input formula,
which results in a conjunction ---~or conjunctive set~---
of base and 
step cases to which the input conjecture reduces, \hskip.1em
\ie\ whose validity implies the validity of the input conjecture. \hskip.1em

Therefore%
, a working mathematician 
would expect that 
the \inductionrule\ of \index{induction!explicit}explicit induction 
solves 
the following two tasks:
\begin{enumerate}\noitem
\item Choose some of the variables in the conjecture as 
\index{induction variables}{\em induction variables}, \hskip.2em
and 
split the conjecture into several base and step cases,
based on the induction variables' demand on which 
governing conditions and 
\index{constructor substitutions}%
constructor substitutions\footnotemark\
have to be added to be able to unfold 
---~without further case analysis~---
some of the recursive function calls
that contain the 
\index{induction variables}induction variables as direct arguments.
\item
Eagerly generate the induction hypotheses 
for the step cases.\noitem
\end{enumerate}
\noindent\mediumheadroom%
\footnotetext{%
 This adding of 
 \index{constructor substitutions}%
 constructor substitutions 
 refers to the application of axioms like \inpit{\nat 1}
 (\cfnlb\ \sectref
 {subsection Mathematical Induction and the Natural Numbers}), \hskip.2em
 and is required whenever 
 \index{constructor style}%
 constructor style either is found
 in the recursive function definitions or is to be used for the step cases.
 In the \PURELISPTP,
 only the latter is the case.
 In \nolinebreak\THM,
 none is the case.%
}%
The actual realization of these tasks in the \inductionrule,
however, 
is quite different from these expectations:
Except the very early days of
\index{induction!explicit}explicit induction 
in the \PURELISPTP\ (\cfnlb\ \examref{example induction rule}), \hskip.2em
\index{induction variables}induction variables play only a very minor \role\ 
toward the end of the procedure
(in the deletion of flawed induction schemes, 
\cfnlb\ \sectref{subsubsection Proof-Time Recursion Analysis in}). \hskip.4em 
The focus, however, is on complete step cases including eagerly generated
induction hypotheses, 
and the complementing bases case are generated 
only \mbox{%
at
the very end.\footnotemark}\footnotetext{%
 See, \eg, \examref{example induction schemes constructor style} of
 \sectref{section example induction schemes constructor style}.%
}%
\vfill\pagebreak
\subsection{Generalization}\label
{subsection Generalization}%
\index{generalization|(}%
Contrary to merely deductive, analytic theorem proving,
an input conjecture for a proof by induction is not only 
a task (as induction conclusion) but also a tool
(as induction hypothesis) in the proof attempt.
Therefore, 
a stronger conjecture is often easier to prove
because it supplies us with a stronger induction hypothesis
during the proof attempt.

Such a step from a weaker to a stronger input conjecture is 
called {\em generalization}. \hskip.2em

Generalization is to be handled with great care
because it is a sound, but {\em unsafe}\/ reduction step
in the sense that it may reduce 
valid goal to an invalid goal, 
causing the proof attempt to fail; \hskip.2em
such a reduction is called {\em over-generalization}.

Generalization 
of input conjectures directly supplied by humans 
is rarely helpful
because stating sufficiently general theorems
is part of the standard mathematical training in induction. \hskip.2em
As we have seen in 
\examref{example second proof of (+3)} of
\nlbsectref{section example second proof of (+3)}, \hskip.2em
however,
\index{induction!explicit}explicit induction often has to start another induction during the 
proof,
and then
the secondary, machine-generated input conjecture
often requires generalization.

The two most simple syntactical generalizations are 
the replacement of terms with fresh universal variables
and 
the removal of irrelevant side conditions. \hskip.3em

In \nolinebreak the vernacular of \boyermooretheoremprovers,
the first is simply called ``generalization'' and the 
second is called 
\index{elimination of irrelevance}%
``elimination of irrelevance\closequotefullstopextraspace
They are dealt with in two consecutive stages 
of these names in the \boyermoorewaterfall,
which come right before the induction stage.

The removal of irrelevant side conditions
is intuitively clear.
For formulas in clausal form,
it simply means to remove irrelevant literals.
More interesting are the heuristics of its realization, 
which we discuss in \nlbsectref{subsubsection Elimination of Irrelevance in}.

The less clear process of
generalization typically proceeds by the replacement of all occurrences of a 
non-variable\footnote{%
 Besides the replacement of 
 \hskip.1em (typically all) \hskip.1em 
 the occurrences of a non-variable term, \hskip.1em
 there is also the possibility of replacing some 
 ---~{\em but not all}~---
 occurrences of a variable with a fresh variable. \hskip.2em
 This is a very delicate process,
 but heuristics for it were discussed very early,
 namely in \cite[\litsectref{3.3}]{aubin-1976}.%
}
term with a fresh variable.

This is especially promising for a subsequent induction
if the same non-variable term 
has multiple
occurrences in the conjecture,
and becomes even more promising if these occurrences are found
on both sides of the same positive 
equation or in literals of different polarity,
say in a conclusion and a condition of an implication.

To avoid {\em over-generalization},
subterms are to be preferred to their super-terms,\footnote{%
 \majorheadroom
 This results in a weaker conjecture
 and the stronger one remains available by a further generalization.%
}
and one should never generalize 
a term
of any of the following forms:
a constructor term, 
a top level term,
a term with a logical operator 
(such as implication or equality)
as top symbol,
a \nolinebreak direct argument of a logical operator,
or the first argument of a 
conditional ({\tt IF}). \hskip.3em
Indeed, for any of these forms, 
the information loss by generalization 
is typically so high that the generalization results in an invalid conjecture.

How powerful generalization can be is best seen by the multitude of 
its successful automatic applications, which often surprise humans.
Here is \nolinebreak one \nolinebreak of \nolinebreak these:%
\pagebreak
\yestop
\label{section example proof of (ack4)}\begin{example}%
[Proof of \inpit{\acksymbol 4} by Explicit Induction and Generalization]\label
{example proof of (ack4)}%
\par\noindent
Let us prove \inpit{\acksymbol 4}
in the context of 
\sectref{subsection Mathematical Induction and the Natural Numbers}
by \index{induction!explicit}explicit induction.
It is obvious that such a proof has to follow the definition of
\nlbmath\acksymbol\ in the three cases 
\inpit{\acksymbol 1},
\inpit{\acksymbol 2},
\inpit{\acksymbol 3},
using the termination ordering of \nlbmaths\acksymbol,
which is just the lexicographic combination of its arguments. \hskip.2em
So 
the \inductionrule\ of \index{induction!explicit}explicit induction
reduces the input formula \nlbmath{\inpit{\acksymbol 4}}
to the following goals:\footnote{%
 See \examref{example induction schemes constructor style}
 of \sectref{section example induction schemes constructor style} \hskip.1em
 on how these step cases are actually found in \index{induction!explicit}explicit induction.%
}
\par\indent\math{\lespp{y}{\ackpp\zeropp y}=\truepp;}
\par\indent\math{\lespp\zeropp{\ackpp{\spp{x'}}\zeropp}\tightequal\truepp
\nottight\antiimplies
\lespp{\spp\zeropp}{\ackpp{x'}{\spp\zeropp}}\tightequal\truepp;
}
\par\indent\math
{\lespp{\spp{y'}}{\ackpp{\spp{x'}}{\spp{y'}}}\tightequal\truepp}
\getittotheright{\maths{\nottight\antiimplies
\inparenthesesoplist{
\lespp
{y'}
{\ackpp{\spp{x'}}{y'}}
\tightequal\truepp
\oplistund
\lespp{\ackpp{\spp{x'}}{y'}}
{\ackpp{x'}{\ackpp{\spp{x'}}{y'}}}
\tightequal\truepp
}}.}
\par\noindent
After simplifying with \inpit{\acksymbol 1},
\inpit{\acksymbol 2},
\inpit{\acksymbol 3},
respectively,
we obtain:
\par\indent\math{\lespp{y}{\spp y}=\truepp;}
\par\indent\math
{\lespp\zeropp{\ackpp{x'}{\spp\zeropp}}\tightequal\truepp
\nottight\antiimplies
\lespp{\spp\zeropp}{\ackpp{x'}{\spp\zeropp}}\tightequal\truepp;
}
\par\indent\math{
\lespp{\spp{y'}}{\ackpp{x'}{\ackpp{\spp{x'}}{y'}}}\tightequal\truepp}
\getittotheright{\maths{\nottight\antiimplies
\inparenthesesoplist{
\lespp
{y'}
{\ackpp{\spp{x'}}{y'}}
\tightequal\truepp
\oplistund
\lespp{\ackpp{\spp{x'}}{y'}}
{\ackpp{x'}{\ackpp{\spp{x'}}{y'}}}
\tightequal\truepp
}}.}
\par\noindent
Now the base case is simply an instance of our lemma \nlbmaths
{\inpit{\lessymbol 4}}. \hskip.3em
Let us simplify the two step cases by 
introducing variables for their common subterms:
\par\indent\maths
{\lespp\zeropp z\tightequal\truepp
\nottight\antiimplies\inparentheses{
\lespp{\spp\zeropp}z\tightequal\truepp
\nottight\und
z\tightequal\ackpp{x'}{\spp\zeropp}
}};
\par\indent\maths{
\lespp{\spp{y'}}{z_2}\tightequal\truepp
\nottight\antiimplies
\inparenthesesoplist{
\lespp
{y'}
{z_1}
\tightequal\truepp
\nottight\und
\lespp{z_1}{z_2}\tightequal\truepp
\oplistund z_1\tightequal\ackpp{\spp{x'}}{y'}
\mbox{\,}\nottight\und z_2\tightequal\ackpp{x'}{z_1}
}}.
\par\noindent
Now the first follows from applying \nlbmath{\inpit{\nat 1}} 
to \nlbmaths z. \hskip.3em
Before we can prove the second by another induction, however,
we have to generalize it to the lemma \inpit{\lessymbol 7}
of \sectref{subsection Mathematical Induction and the Natural Numbers}
by deleting the last two literals from the condition.
\getittotheright\qed\end{example}
\par
\halftop\halftop\halftop
\noindent
In combination with \index{induction!explicit}explicit induction,
generalization becomes especially powerful in the 
invention of new lemmas of general interest,
because the step cases of \index{induction!explicit}explicit induction tend to have common occurrences
of the same term in their conclusion and their condition. \hskip.2em
Indeed, 
the lemma \nlbmath{\inpit{\lessymbol 7}}, \hskip.1em
which we have just discovered in \examref{example proof of (ack4)}, \hskip.1em
is one of the most useful lemmas in the theory of natural numbers.

It should be noted that all \boyermooretheoremprovers\ except
the \PURELISPTP\ are
able to do this whole proof
completely automatically
and invent the lemma \nlbmath{\inpit{\lessymbol 7}}
by generalization of the second step case; \hskip.2em
and they do this 
even when they work with an arithmetic theory that was redefined,
so that no decision procedures or other special knowledge on the natural
numbers can be used by the system.
Moreover, 
as shown in \litsectref{3.3} of \cite{wirthcardinal}, \hskip.1em
in a slightly richer logic,
these heuristics can actually synthesize 
the lower bound in the first argument of \nlbmath\lessymbol\
from the weaker input conjecture \bigmathnlb
{\exists z\stopq\inpit{\lespp z{\ackpp x y}\tightequal\truepp}}, \hskip.07em
simply because \nlbmath\lessymbol\ does not contribute to
the choice of the base and step cases.%
\index{generalization|)}%
\pagebreak
\subsection{Proof-Theoretical Peculiarities of Mathematical Induction}\label
{subsection Proof-Theoretical Peculiarities of Mathematical Induction}%

\halftop\noindent
The following 
two proof-theoretical peculiarities of 
induction compared to \firstorder\ deduction may be considered 
noteworthy:\footnote{\label{note no difference in practice}%
 Note, however,
 that these peculiarities of induction 
 do not make a difference to \firstorder\ deductive theorem 
 proving {\em in practice}. 
 See \noterefs{note practice one}{note practice two}.%
}\begin{enumerate}
\item
A \nolinebreak calculus for arithmetic cannot be complete,
simply because the theory of the arithmetic of natural numbers 
is not enumerable.\footnote{\label{note practice one}%
 \majorheadroom
 This theoretical result is given by \goedelsfirstincompletenesstheorem\ 
 \shortcite{goedel}. \hskip.3em
 In practice, however, it does not matter whether our proof attempt fails because
 our theorem 
 will not be enumerated ever, 
 or will not be enumerated before doomsday.%
}%
\item
According to \nolinebreak\gentzen's Hauptsatz,\footnote{%
 \majorheadroom
 \Cfnlb\ \cite{gentzen}.%
}
\hskip.2em
a proof of a \firstorder\ theorem
can always be restricted to the ``sub''-formulas of this theorem. \hskip.2em
In contrast to lemma application in a deductive proof tree, 
however,
the application of induction hypotheses and lemmas 
inside an inductive reasoning cycle cannot generally
be eliminated in the sense that the
``sub''-formula property could be obtained.\footnote{%
 \majorheadroom
 \Cfnlb\ \cite{induction-no-cut}.%
}
As a consequence, 
in \firstorder\ inductive theorem proving, 
``creativity'' cannot be restricted to 
finding just the proper instances, 
but may require the invention of new lemmas and notions.\footnote{%
 \majorheadroom
 \label{note practice two}%
 In practice, 
 however, 
 proof search for harder theorems often requires 
 the introduction of lemmas, functions, and relations,
 and it is only a matter of degree 
 whether we have to do this for principled reasons (as in induction) 
 or for tractability (as required in \firstorder\ deductive theorem proving,
 \cfnlb\ \cite{baazleitschcolllog}).%
}%
\end{enumerate}%
\halftop
\subsection{Conclusion}
In this section, \hskip.1em
after briefly presenting the induction method in
its rich historical \mbox{context,}
we \nolinebreak have offered a formalization and a first
practical description
.
Moreover, 
we have explained why we can take \fermatindex\fermat's term 
{\it``\descenteinfinie''}\/ \hskip.2em
in our modern context as a synonym for the standard high-level method of
mathematical induction. \hskip.3em
Finally, 
we \nolinebreak have introduced 
\index{induction!explicit}explicit induction and generalization.

\noetherian\ 
induction requires domains for its \wellfounded\ orderings; \hskip.3em
and these domains
are typically built-up by constructors.
Therefore, the discussion of the method of induction 
required the introduction of some paradigmatic data types, 
such as natural numbers and lists.

To express the relevant notions 
on 
these data types,
we need {\em recursion}, \hskip.1em
a method of definition,
which we have often used in this section intuitively.
We \nolinebreak did not 
discuss its formal admissibility requirements yet.
We will do so in \nlbsectref{section Recursive Definitions},
\hskip.2em
with \nolinebreak a \nolinebreak focus on modes of recursion that admit an
effective \index{consistency}consistency test,
\mbox{including} termination aspects such as induction templates and schemes.
\vfill\pagebreak
\section
{Recursion, Termination, and Induction}\label
{section Recursive Definitions}
\subsection{Recursion and the Rewrite Relation on Ground Terms}
\index{recursion}%
{\em Recursion}\/ is a form of programming or 
definition where a 
newly defined notion may even occur in its {\em definientia}. \hskip.3em
Contrary to {\em \index{induction!explicit}explicit}\/ definitions, \hskip.1em
where we can always get rid of the new notions by reduction \hskip.2em
(\ie\ by rewriting the {\em definienda}
({\em left-hand sides}\/ of \nolinebreak the defining equations) \hskip.1em
to the {\em definientia}\/ ({\em right-hand sides}\/)), \hskip.3em
reduction with {\em recursive}\/ definitions may run forever.
\par
We have already seen 
some recursive function definitions 
in \sectrefs{subsection Mathematical Induction and the Natural Numbers}
{subsection Standard Data Types}, \hskip.2em
such as the ones of \plussymbol,
\lessymbol, \lengthsymbol, and \sizesymbol, \hskip.1em
where these function symbols
occurred in some of the right-hand sides of the equations of their own
definitions; \hskip.2em
for instance, the function symbol \nlbmath\plussymbol\ 
occurs in the right-hand side of \inpit{\plussymbol 2} in
\sectref{subsection Mathematical Induction and the Natural Numbers}.
\par
The steps of rewriting with recursive definitions can be formalized as
a binary relation on terms, namely as the
\index{rewrite relation}{\em rewrite relation}\/ that results from reading the
defining equations as reduction rules, \hskip.1em
in the sense that they allow us 
to replace occurrences
of left-hand sides of instantiated equations with their respective
right-hand sides, 
provided that their conditions are fulfilled.\footnote{%
 For the technical meaning of {\em fulfilledness}\/
 in the recursive definition of the rewrite relation see
 \cite{wirth-jsc}, \hskip.1em
 where it is also explained why 
 the rewrite relation respects 
 the straightforward purely logical,
 model-theoretic semantics of \pnc\ equation equations,
 provided that the given admissibility conditions are satisfied
 (as is the case for all our examples).%
}%
\\\indent
A 
\index{ground terms}%
{\em ground}\/ term is a term without variables. 
We can restrict our considerations here to rewrite relations 
{\em on ground terms}.%
\subsection{Confluence}\label
{subsection Confluence}%
\index{confluence|(}%
The restriction that is to be required for every
recursive function definition is the 
{\em confluence}\/\footnote{%
 \majorheadroom
 A relation \redsimple\ is {\em confluent}\/ 
 (or has the ``\churchrosserproperty'') \hskip.3em
 if two sequences of steps with \nlbmaths\redsimple,
 starting from the same element,
 can always be joined by an arbitrary number of further steps on each side; 
 \hskip.3em formally:
 \bigmaths{\antitrans\circ\trans
 {\nottight{\nottight\subseteq}}
 \refltrans\circ\antirefltrans}. \ \hskip.3em
 Here 
 \math\circ\ denotes the concatenation of binary relations; \hskip.3em
 for the further notation see 
 \sectref{subsection Well-Foundedness and Termination}.%
} 
of 
this rewrite relation on ground terms.
\par
The \index{confluence}confluence restriction
guarantees that no distinct objects of the data types
can be equated by the recursive function definitions.\footnote{%
 \majorheadroom
 As constructor terms are irreducible \wrt\ this rewrite relation, \hskip.2em
 if the application of a defined function symbol rewrites to two
 constructor terms, 
 they
 must be identical in case of 
 \index{confluence}confluence.%
} 

This is essential for \index{consistency}consistency
if we assume 
axioms such as \inpit{\nat 2\mbox{--}3} (\cfnlb\ 
\sectref{subsection Mathematical Induction and the Natural Numbers}) \hskip.1em
or \inpit{\app\lists\nat 2\mbox{--}3} (\cfnlb\ 
\sectref{subsection Standard Data Types}).
\par
Indeed, 
without \index{confluence}confluence,
a definition of a recursive function 
could destroy the data type in the sense that the specification has no 
model anymore; \hskip.2em
for example, 
if we added \bigmaths{\ppp x=\zeropp}{} as a further defining equation to 
\nlbmaths{\inpit{\psymbol1}}, \hskip.3em
then we would get \bigmaths{\spp{\zeropp}=\ppp{\spp{\spp\zeropp}}=\zeropp},
in contradiction to the axiom \nlbmath{\inpit{\nat 2}} of 
\nlbsectref{subsection Mathematical Induction and the Natural Numbers}. 
\pagebreak
\par\indent
For the recursive function definitions admissible in the 
\boyermooretheoremprovers, 
\index{confluence}confluence results from the restrictions
that there is only one (unconditional) defining equation for each new
function symbol,\footnote{%
 \Cf\ \lititemref{\inpit a} of the 
 ``definition principle'' of \cite[\p\,44\f]{bm}. \hskip.1em
 \index{confluence}Confluence is also discussed under the label ``uniqueness''
 on \litspageref{87\ff}\ of \cite{moore-1973}.%
}
and that all variables occurring on the right-hand side of the 
definition also occur on the left-hand side of the 
defining equation.\footnote{%
 \majorheadroom
 \Cf\ \lititemref{\inpit c} of the 
 ``definition principle'' of \cite[\p\,44\f]{bm}.
}%
\par
These two restrictions are an immediate consequence of the 
general definition style of the list-programming language 
\LISP\nolinebreak\hspace*{-.1em}\nolinebreak\@. \hskip.4em
More precisely,
recursive functions are to be defined 
in all \boyermooretheoremprovers\
in the more restrictive style of {\em applicative}\/ 
\LISP\nolinebreak\hspace*{-.1em}\nolinebreak\@.\footnotemark
\label{section example PLUS}%
\begin{example}[A Recursive Function Definition in Applicative \LISP]\label
{example PLUS}%
\mbox{}\\%
Instead 
of our two equations 
\inpit{\plussymbol 1}, \inpit{\plussymbol 2}
for \plussymbol, \hskip.1em
we find the following single equation 
on \litspageref{53} 
of 
the standard reference for the 
\boyermoore\ heuristics \cite{bm}:
\\\noindent\LINEnomath
{\tt\begin{tabular}{l}(PLUS~X~Y)~=~(IF~(ZEROP~X)
\\\mbox{}~~~~~~~~~~~~~~~~~(FIX Y)
\\\mbox{}~~~~~~~~~~~~~~~~~(ADD1 (PLUS (SUB1 X) Y)))
\\\end {tabular}}
\par\halftop\noindent
Note that 
\hskip.1em \mbox{\tt(IF \math x \math y \math z)} \hskip.2em
is nothing but
 the conditional \hskip.1em
 ``{\tt IF~\math z}~then~{\math y}~else~{\math z}\closequotecommaextraspace
 that {\tt ZEROP} is a \myboolean\ function
 checking for being zero, \hskip.1em
 that {\tt (FIX Y)} returns {\tt Y} \mbox{if {\tt Y} is a} natural number,
 \hskip.1em
 and that {\tt ADD1} is the successor function \nlbmaths\ssymbol.
\\\indent
The primary difference to
\inpit{\plussymbol 1}, \inpit{\plussymbol 2} \hskip.1em
is that {\tt PLUS} is defined in 
\index{constructor style}%
{\em destructor style}\/ instead of 
the
\index{destructor style}%
{\em constructor style}\/ of our equations \inpit{\plussymbol1}, 
\inpit{\plussymbol2} in 
\sectref{subsection Mathematical Induction and the Natural Numbers}. \hskip.3em
As
a 
\index{constructor style}%
constructor-style definition can always be transformed into an equivalent 
\index{destructor style}%
destructor-style definition,
\hskip.1em
let us 
do so for
our definition of \plussymbol\ 
via \inpit{\plussymbol 1}, \nlbmaths{\inpit{\plussymbol 2}}. 
\\\indent
In place of the untyped destructor {\tt SUB1}, \hskip.1em 
let us use the typed destructor \nlbmath\psymbol\ defined 
by either by \inpit{\psymbol 1} or by \inpit{\psymbol 1'}
of 
\sectref{subsection Mathematical Induction and the Natural Numbers}, \hskip.2em
which ---~just as {\tt SUB1}~---
returns the predecessor of a positive natural number. \hskip.2em
Now our 
\index{destructor style}%
destructor-style definition of \nlbmaths\plussymbol\ 
consists of the following
two \pnc\ equations:
\par\noindent\math{\begin{array}{@{}l@{~~~~~~~}l@{\ =\ }l@{\ }l}
  \inpit{\plussymbol 1'}
 &\plusppnoparentheses x y
 &y
 &\antiimplies\ x\tightequal\zeropp
  \mediumfootroom
\\\inpit{\plussymbol 2'}
 &\plusppnoparentheses x y
 &\spp{\plusppnoparentheses{\ppp x}{y}}
 &\antiimplies\ x\tightnotequal\zeropp
\\\end{array}}
\par\noindent\smallheadroom
If we compare this definition of \nlbmath\plussymbol\ to the 
one via the equations \inpit{\plussymbol 1}, \inpit{\plussymbol2}, \hskip.2em
then we \nolinebreak find that 
the constructors \zeropp\ and \ssymbol\ have been removed from 
the left-hand sides of the defining equations; \hskip.2em
they are replaced with the destructor \nlbmath\psymbol\ on the right-hand side
and with some conditions.
\\\indent
Now 
it is easy
to see that \inpit{\plussymbol 1'}, \inpit{\plussymbol 2'}
represent the above definition of {\tt PLUS} in \pnc\ equations,
provided that we ignore that \boyermooretheoremprovers\ have 
no types and no typed variables.
\getittotheright\qed\end{example}
\footnotetext{\label{note applicative}%
 \majorheadroom
 See \cite{LISP} for the definition of \LISP\@.
 The ```applicative'' subset of \LISP\ lacks side effects 
 via global variables and
 the imperative commands of \LISP,
 such as variants of {\tt PROG}, {\tt SET}, {\tt GO}, and {\tt RETURN},
 as \nolinebreak well as all functions or special forms 
 that depend on the concrete allocation on the system heap,
 such \nolinebreak as {\tt EQ}, {\tt RPLACA}, and {\tt RPLACD},
 which can be used in \LISP\ to realize circular structures or to
 save space on the system heap.%
}%
\vfill\pagebreak
\halftop\noindent
If we considered the recursive equation \nlbmath{\inpit{\plussymbol2}} \hskip.1em
together with the alternative recursive equation 
\nlbmaths{\inpit{\plussymbol2'}}, \hskip.3em
then we could rewrite \plusppnoparentheses{\spp x}y \hskip.2em
on \nolinebreak the one hand with 
\nlbmath{\inpit{\plussymbol2}} \hskip.1em
into 
\hskip.1em\spp{\plusppnoparentheses x y}, \hskip.2em
and, on the other hand, with \nlbmath{\inpit{\plussymbol2'}} \hskip.1em
into \hskip.1em\spp{\plusppnoparentheses{\ppp{\spp x}}y}. \hskip.3em
This does not seem to be problematic, 
because the latter result
can be rewritten to the former one by \nlbmaths{\inpit{\psymbol1}}.

In \nolinebreak general, 
however, 
\index{confluence}confluence is undecidable and criteria
sufficient for \index{confluence}confluence
are extremely hard to develop. \hskip.2em
The only 
known 
decidable criterion that is sufficient for \index{confluence}confluence 
of conditional equations and applies to all our example specifications,
but does not require termination, 
is found in \cite{wirth-jsc}\fullstopnospace\footnotemark\ \hskip.3em
\footnotetext{%
 The effective \index{confluence}confluence test of 
 \cite{wirth-jsc} requires
 {\em binding-triviality}\/ or {\em -complementary}\/ of every critical peak, 
 and {\em effective weak-quasi-normality},
 \ie\ that each equation in the condition must be restricted to 
 \index{constructor variables}%
 constructor variables (\cfnlb\,\sectref{subsection Constructor Variables}),
 or that one of its top terms either is a constructor term or occurs
 as the argument of a definedness literal in the same condition.%
}%
It \nolinebreak can be more easily tested than the admissibility 
conditions 
of the \boyermooretheoremprovers\ 
and avoids divergence even in case of non-termination; \hskip.2em
the proof that it indeed guarantees \index{confluence}confluence is \mbox{very involved}.%
\index{confluence|)}%
\vfill
\subsection{Termination and Reducibility}\label
{subsection Termination and Reducibility}%
\halftop\noindent
There are two restrictions
that are additionally required for any function definition in 
the \boyermooretheoremprovers, 
namely
\index{termination}%
{\em termination}\/ of the rewrite relation and
\index{reducibility}%
{\em reducibility}\/ of all ground terms
that contain a defined function symbol 
\wrt\ the rewrite relation.
\par
The requirement of termination should be intuitively clear; \hskip.2em
we will further discuss it in \nlbsectref{subsection Termination}.
\par
Let us now discuss the requirement of reducibility.

First of all, 
note that it is not only so that we can check the soundness of 
\inpit{\plussymbol 1'} \mbox{and \inpit{\plussymbol 2'}}
independently from each other,
we can even omit one of the equations, resulting in a 
partial definition of the function \nlbmaths\plussymbol. \hskip.3em
Indeed, for the function \nlbmath\psymbol\ we did not specify any value
for \ppp\zeropp; \hskip.3em
so \ppp\zeropp\ is not reducible in the rewrite relation
that results from reading the specifying equations as reduction rules.
\par
A function defined in a \boyermooretheoremprover, 
however,
must always be specified completely,
in the sense that every application of such a function
to (constructor) ground terms must be reducible. \hskip.2em
This reducibility immediately results from the 
\LISP\ definition style, 
which requires all arguments of the function symbol
on the  left-hand side of its defining equation
to be distinct variables.\footnote{%
 \majorheadroom
 \Cf\ \lititemref{\inpit b} of the 
 ``definition principle'' of \cite[\p\,44\f]{bm}.%
}%
\vfill\vfill\pagebreak
\subsection{Constructor Variables}\label
{subsection Constructor Variables}%
\halftop
\noindent
\index{constructor variables}%
These restrictions of
reducibility and termination of the rewrite relation  
are not essential; \hskip.2em 
neither for the semantics of 
recursive function definitions with data types given by constructors,\footnote{%
 \Cf\ \cite{wgcade}.%
} 
nor for \index{confluence}confluence and \index{consistency}consistency.\footnote{%
 \majorheadroom
 \Cf\ \cite{wirth-jsc}.%
}%
\par\indent
Note that these two 
restrictions imply that 
only {\em total recursive}\/ functions\footnote{%
 \majorheadroom
 You may follow the explicit reference to \cite{schoenfield}
 as the basis for the logic of the \PURELISPTP\ 
 on \litspageref{93} of \cite{moore-1973}.%
}
are admissible in the \boyermooretheoremprovers.
\par\indent
As a termination 
restriction
is not in the spirit of the \LISP\ logic of the \boyermooretheoremprovers,
we have to ask 
why \boyer\ and \moore\ brought up 
this additional restriction.%

When
both 
reducibility and termination are 
given,
then 
---~similar to the classical case of explicitly defined notions~---
we can get rid of all recursively defined function symbols by rewriting,
but in general only 
for 
{\em ground}\/
terms.%
\par\indent
A better potential answer is found on \litspageref{87\ff}\ of 
 \cite{moore-1973},
 where \index{confluence}confluence of the rewrite relation is discussed
 and a reference to \russellsparadox\ serves as an argument
 that \index{confluence}confluence alone would not be sufficient for \index{consistency}consistency. \hskip.3em
 The \nolinebreak argumentation is essentially the following: \hskip.3em
 First, \hskip.1em
 a \myboolean\ function \nlbmath\russellsymbol\ 
 is recursively defined by
 \par\halftop\noindent\math{\begin{array}{@{}l@{~~~~~}l@{\ =\ }l@{}l}
  \inpit{\russellsymbol 1}
 &\russellpp b
 &\falsepp
 &\nottight\antiimplies\russellpp b\tightequal\truepp
  \mediumfootroom
\\\inpit{\russellsymbol 2}
 &\russellpp b
 &\truepp
 &\nottight\antiimplies\russellpp b\tightequal\falsepp
\\\end{array}}
\par\halftop\indent
Then it is claimed that this function definition would 
result in an inconsistent specification on the basis of 
the axioms \inpit{\bool 1\mbox{--}2} of \sectref{subsection Standard Data Types}.
\par\indent
This \index{consistency}inconsistency, however, arises \onlyif\ 
the variable \nlbmath b of the axiom \inpit{\bool 1}
can be instantiated with the term \nlbmaths{\russellpp b}, \hskip.2em
which is actually not our intention and which 
we do not have to permit: \hskip.2em
If all variables we have introduced so far are
\index{constructor variables}%
\mbox{\em constructor variables}\/\nolinebreak\hskip.1em\nolinebreak\footnote{%
 \majorheadroom
 \index{constructor variables}%
 Such {\em constructor variables}\/ were formally introduced 
 for the first time in 
 \cite{wgkp}
 and became an essential part of the 
 frameworks found in 
 \makeaciteoftwo{wgjsc}{wgcade},
 \makeaciteoftwo{kwspec}{kwspec2},
 \makeaciteoftwo{wirthdiss}{wirth-jsc}
 \cite{kuehlerdiss},
 \cite{quodlibet-cade},
 and \makeaciteofthree{samoacalculemus}{samoa-phd}{jancl}.%
}
in \nolinebreak the \nolinebreak sense that they can only be instantiated with 
terms formed from 
\index{constructor function symbols}%
constructor function symbols (\incl\ constructor constants) 
\hskip.1em and 
\index{constructor variables}%
constructor variables, \hskip.1em
then irreducible terms such as \russellpp b
can denote {\em junk objects}\/
different from \truepp\ and \falsepp,
and no \index{consistency}inconsistency arises.\footnote{%
 \majorheadroom
 For the appropriate semantics see 
 \cite{wgcade},
 \cite{kwspec2}.%
}%
\par\indent
Note that these 
\index{constructor variables}%
constructor variables are implicitly part
of the \LISP\ semantics with its innermost evaluation strategy.
For instance, 
in \examref{example PLUS} of \nlbsectref{section example PLUS}, \hskip.2em
neither the \LISP\ definition of PLUS 
nor its representation via the \pnc\ equations
\inpit{\plussymbol 1'}, \inpit{\plussymbol 2'}
is intended to be applied to a non-constructor term
in the sense that {\tt X} or \math x should be instantiated
to a term that is a function call of a (partially) defined function
symbol that may denote a junk object.
\par\indent
Moreover, 
there is evidence that \moore\ considered the variables 
already in\,1973 as 
\index{constructor variables}%
constructor variables: \hskip.2em
On \litspageref{87} in \cite{moore-1973},
we \nolinebreak find formulas on defined\-ness and \index{confluence}confluence, 
which make sense only for 
\index{constructor variables}%
constructor variables; \hskip.2em
the one on defined\-ness of the \myboolean\ 
function {\tt(AND X Y)} reads\footnote
{\label{note on IF and COND}%
 In the logic of the \PURELISPTP,
 the special form {\tt IF} is actually 
 called ``{\tt COND}\closequotefullstopextraspace
 This is most confusing because 
 {\tt COND} is a standard special form in \LISP,
 different from \nolinebreak {\tt IF}\@. \hskip.2em
 Therefore, we will ignore this peculiarity and tacitly write ``{\tt IF}''
 here and in what follows for every ``{\tt COND}'' of the \PURELISPTP.}
\par\halftop\noindent\LINEmaths{\exists\mbox{\tt Z (IF X (IF Y T NIL) NIL) = Z}},
\par\halftop\noindent
which is trivial for a general variable {\tt Z} and makes sense \onlyif\,
{\tt Z} is taken to be a 
\index{constructor variables}%
constructor variable.
\par\indent
Finally, 
the way termination is established 
via induction templates in 
\boyermooretheoremprovers\ 
and 
as we will describe it in \nlbsectref{subsection Termination}, \hskip.1em
is sound for the rewrite relation of the defining equations
\onlyif\ we consider the variables of these equations to be constructor
variables (or if we restrict the termination result to an innermost
rewriting strategy and require that all function definitions are total).%
\yestop
\subsection{Termination and General Induction Templates}\label
{subsection Termination}%
\halftop\noindent
\index{termination|(}%
\index{induction templates|(}%
In addition to the restricted style of recursive definition 
that is found 
in \LISP\ and that guarantees reducibility of terms
with defined function symbols and \index{confluence}confluence as described 
in \sectrefs{subsection Termination and Reducibility}
{subsection Constructor Variables}, \hskip.2em
the theorem provers for \index{induction!explicit}explicit induction
require termination of 
the rewrite relation
that results from reading the specifying equations as reduction rules.
More precisely, in all \boyermooretheoremprovers\ except the 
\PURELISPTP,\footnote{%
 \majorheadroom
 Note that termination is not proved in the \PURELISPTP; \hskip.1em
 instead, the \nolinebreak soundness of the induction proofs comes with the 
 {\it proviso}\/ that the rewrite relation of all defined
 function symbols terminate.%
} \hskip.1em
{\em before}\/ a new function symbol \nlbmath{f_k}
is admitted to the specification, \hskip.1em
a ``valid induction template''
---~which immediately implies termination~---
has to be constructed
from the defining equation of \nlbmaths{f_k}.\footnote{%
 \majorheadroom
 See also \lititemref{\inpit d} of the 
 ``definition principle'' of \cite[\p\,44\f]{bm} \hskip.1em
 for a formulation that avoids the technical term 
 ``induction template\closequotefullstopnospace
}%

Induction templates were first used in \THM\ and 
received their name when they were first described in \cite{bm}.

Every time a new recursive function \nlbmath{f_k} is defined, \hskip.1em
a system for \index{induction!explicit}explicit induction 
immediately tries to construct {\em valid induction templates}\/; \hskip.2em
if it does not find any, \hskip.1em
then the new function symbol is rejected \wrt\ the given definition; \hskip.2em
otherwise the system 
links
the function name 
with its definition and 
its valid induction templates.

The induction templates 
serve actually
two purposes: 
as witnesses for termination and
as the basic tools of the \inductionrule\ of 
\index{induction!explicit}explicit induction 
for generating the step cases.

\pagebreak

For a finite number of mutually 
recursive functions \nlbmath{f_k} with arity \nlbmath{n_k}
\nlbmaths{\inpit{k\tightin K}}, \hskip.1em
an \nolinebreak induction template in the most general form 
consists of the following:\begin{enumerate}
\item
A 
\index{relational descriptions}{\em relational description}\/\footnote{%
 The name \index{relational descriptions}``relational description'' comes from 
 \makeaciteoftwo{waltherLPAR92}{waltherIJCAI93}.%
} 
of the changes in the argument pattern
of these recursive functions
as found in their recursive defining equations:
\\
For each \nlbmath{k\in K} and for each \pnc\ equation with 
a left-hand side of the form
\maths{f_k(t_1,\ldots,t_{n_k})}, \hskip.3em
we take the set \nlbmath R of recursive function calls of the 
\nlbmath{f_{k'}} \inpit{k'\tightin K} \hskip.1em
occurring in the right-hand side or the condition,
and some case condition \nlbmaths C, \hskip.1em
which must be a subset of the conjunctive condition literals of the defining
equation. 
Typically, \hskip.1em
\math C is empty 
(\ie\ always true) 
in 
the 
case of constructor-style definitions, \hskip.1em
and just sufficient to guarantee proper destructor applications
in 
the 
case of 
\index{destructor style}%
destructor-style definitions.
\\
Together they form the triple \trip{f_k(t_1,\ldots,t_{n_k})} R C, \hskip.2em
and a set containing such a triple for each such defining equation 
forms the \index{relational descriptions}relational description.
\\
For our definition of \plussymbol\  
via \inpit{\plussymbol1}, \inpit{\plussymbol2} \hskip.1em
in \sectref{subsection Mathematical Induction and the Natural Numbers}, 
\hskip.2em 
there is only one recursive equation and only one relevant 
\index{relational descriptions}relational description, 
namely the following one with an empty case condition: 
\par\noindent\LINEmaths{\displayset{\displaytrip
{\plusppnoparentheses{\spp x}y}
{\{\plusppnoparentheses x y\}}
{\emptyset}}}.%
\par\noindent 
Also for our definition of \plussymbol\  
with \inpit{\plussymbol1'}, \inpit{\plussymbol2'} \hskip.1em
in \examref{example PLUS}, \hskip.2em 
there is only one recursive equation and only one relevant 
\index{relational descriptions}relational description, namely 
\par\noindent\LINEmaths{\displayset{\displaytrip
{\plusppnoparentheses x y}
{\{\plusppnoparentheses{\ppp x}y\}}
{\{x\tightnotequal\zeropp\}}}}.\item
For each \nlbmath{k\in K}, \hskip.2em
a variable-free 
weight term \nlbmath{w_{f_k}} 
in which the position numbers 
\par\noindent\LINEmaths{\inpit 1,\ldots,\inpit{n_k}}{}
\par\noindent are used in place of variables. \hskip.3em
The position numbers actually occurring in the term are called
the \index{measured positions}{\em measured positions}.
\\\headroom
For our two \index{relational descriptions}relational descriptions, only the weight term \inpit 1
(consisting just of a position number) \hskip.1em
makes sense as \nlbmaths{w_+}, \hskip.3em
resulting in the set of \index{measured positions}measured positions \nlbmaths{\{1\}}. \hskip.4em
Indeed, \plussymbol\ terminates in both definitions
because the argument in the first position gets smaller.\item
A binary predicate \nlbmath < \hskip.1em 
that is known to represent a \wellfounded\ relation.
\\\headroom
For our two \index{relational descriptions}relational descriptions, the predicate
\bigmaths{\lambda x,y\stopq\inpit{\lespp x y\tightequal\truepp}},
is appropriate.\end{enumerate}
\noindent
Now, an induction template is {\em valid}\/
if for each element of the \index{relational descriptions}relational description 
as given above, and for each \nlbmaths{f_{k'}(t'_1,\ldots,t'_{n_{k'}})\in R},
\hskip.2em the following conjecture is valid:%
\par\halftop\noindent\LINEmaths{
w_{f_{k'}}\{
\inpit{1}     \tight\mapsto t'_1,\ldots,
\inpit{n_{k'}}\tight\mapsto t'_{n_{k'}}\}
\nottight{\nottight<}
w_{f_{k}}\{
\inpit{1}     \tight\mapsto t_1,\ldots,
\inpit{n_{k}}\tight\mapsto  t_{n_{k}}\}
\nottight{\nottight{\nottight\antiimplies}}\bigwedge C}.
\par\halftop\noindent
For our two \index{relational descriptions}relational descriptions, \hskip.1em
this amounts to showing 
\bigmaths{
\lespp x{\spp x}\tightequal\truepp
}{}
and \bigmaths{
\lespp{\ppp x}x\tightequal\truepp\nottight\antiimplies x\tightnotequal\zeropp
}, 
respectively; \hskip.3em
so their templates are both valid by 
lemma \math{\inpit{\lessymbol 4}}
and axioms \nlbmath{\inpit{\nat 1\mbox{--}2}} and \nlbmaths{\inpit{\psymbol 1}}.

\pagebreak

\halftop\begin{example}%
[Two Induction Templates with different Measured Positions]%
\label{example induction template two measured subsets constructor style}%
\mbox{}\\
For the ordering predicate \lessymbol\ 
as defined by \inpit{\lessymbol 1\mbox{--}3} of 
\sectref{subsection Mathematical Induction and the Natural Numbers}, \hskip.2em
we get two appropriate induction templates with the 
sets of \index{measured positions}measured positions \math{\{1\}} and \nlbmaths{\{2\}}, \hskip.2em
respectively, \hskip.1em
both with the \index{relational descriptions}relational description 
\\\LINEmaths{
\displayset{\displaytrip{\lespp{\spp x}{\spp y}}
{\{\lespp x y\}}\emptyset}},
\\and both with the 
\wellfounded\ ordering \nlbmaths{\lambda x,y\stopq\inpit{\lespp x y
\tightequal\truepp}}. \hskip.5em
The first template has the weight term \nlbmath{\inpit 1} \hskip.1em
and
the second one has the weight term \nlbmath{\inpit 2}. \hskip.3em
The validity of both templates is given by lemma \inpit{\lessymbol 4}
of \sectref{subsection Mathematical Induction and the Natural Numbers}.
\getittotheright\qed\notop\halftop\footroom\end{example}
\begin{example}[One Induction Template with Two Measured Positions]%
\label{example non-singleton measured subset constructor style}%
\mbox{}\\
For the \ackermannfunction\ \acksymbol\ 
as defined by \inpit{\acksymbol 1\mbox{--}3} of
\sectref{subsection Mathematical Induction and the Natural Numbers}, \hskip.2em
we \nolinebreak get only one appropriate induction template.
The set of its \index{measured positions}measured positions is \nlbmaths{\{1,2\}}, \
because of the weight function \cnspp{\inpit 1}{\cnspp{\inpit 2}\nilpp},
which we will abbreviate in the following with 
\maths{[\inpit 1, \inpit 2]}. 
\hskip.5em
The \wellfounded\ relation is the lexicographic ordering
\bigmaths{\lambda l,k\stopq\inpit{\lexlimlespp l k{\spp{\spp{\spp\zeropp}}}
\tightequal\truepp}}. 
The \index{relational descriptions}relational description has two elements: \hskip.3em
For the equation \inpit{\acksymbol 2} \hskip.1em we get
\\\LINEmaths{\displaytrip
{\ackpp{\spp x}\zeropp}{\{\ackpp x{\spp\zeropp}\}}\emptyset},
\\and for the equation \inpit{\acksymbol 3} \hskip.1em we get
\\\phantom\qed\LINEmaths{\displaytrip
{\ackpp{\spp x}{\spp y}}
{\{\ackpp{\spp x}{y}\comma\ackpp{x}{\ackpp{\spp x}{y}}\}}
\emptyset}.
\\The validity of the template is expressed in the three equations
\\\LINEmaths{\begin{array}[b]{l@{}l@{\,}l@{\,}l@{}l}\lexlimlessymbol(
 &[x,\spp\zeropp],
 &[\spp x,\zeropp],
 &\spp{\spp{\spp\zeropp}}
 &)\nottight{\nottight{\nottight{\nottight{\nottight{\nottight=}}}}}\truepp;
\\\lexlimlessymbol(
 &[\spp x,y],
 &[\spp x,\spp y],
 &\spp{\spp{\spp\zeropp}}
 &)\nottight{\nottight{\nottight{\nottight{\nottight{\nottight=}}}}}\truepp;
\\\lexlimlessymbol(
 &[x,\ackpp{\spp x}{y}],
 &[\spp x,\spp y],
 &\spp{\spp{\spp\zeropp}}
 &)\nottight{\nottight{\nottight{\nottight{\nottight{\nottight=}}}}}\truepp;
\\\end{array}}{}
\\which follow deductively from \inpit{\lessymbol 4}, \inpit{\lexlimlessymbol 1},
\inpit{\lexlessymbol 2\mbox{--}4}, \inpit{\lengthsymbol 1\mbox{--}2}.%
\getittotheright\qed\end{example}
\noindent For 
induction templates of 
\index{destructor style}%
destructor-style definitions see \examrefs
{example induction template two measured subsets}
{example non-singleton measured subset} in \nlbsectref
{subsubsection Induction Templates}.

\subsection{Termination of the Rewrite Relation on Ground Terms}
\noindent
Let us prove that the existence of a valid induction template 
for a new set of 
recursive functions \nlbmath{f_k} \inpit{k\tightin K} \hskip.1em
actually implies termination
of the rewrite relation 
after addition of 
the new \pnc\ equations for the \nlbmaths{f_k}, \hskip.2em
assuming any arbitrary model \nlbmath{\mathcal M}
of all (\pnc) equations
with free constructors 
to be given.\footnote{%
 A {\em model with free constructors}\/ is a model where
 two constructor ground terms are equal in \nlbmath{\mathcal M}
 \onlyif\ they are syntactically equal. \hskip.3em
 Because the confluence result of \cite{wirth-jsc} \hskip.1em
 applies in our case without requiring termination, \hskip.1em
 there is always an initial model with free constructors
 according to \litcororef{7.17} of \cite{wirthdiss}, \hskip.2em
 namely the factor algebra of the ground term algebra modulo
 the equivalence closure of the rewrite relation.%
}%
\\\indent
For \areductioadabsurdum, \hskip.1em
suppose that there is an infinite sequence of
rewrite steps on ground terms. \hskip.2em
Consider each term in this sequence to be replaced with the multiset
that contains, \hskip.1em
for each occurrence of a function call \nlbmath{f_k(t_1,\ldots,t_{n_k})} 
with \nlbmaths{k\tightin K}, \hskip.2em
the value of its weight term 
\nlbmath{w_{f_{k}}\{
\inpit{1}    \tight\mapsto t_1,\ldots,
\inpit{n_{k}}\tight\mapsto  t_{n_{k}}\}} in \nlbmaths{\mathcal M}.
\\\indent
Then the rewrite steps with instances of the {\em old}\/ equations
of previous function definitions 
(of symbols not among the \nlbmath{f_k}) \hskip.1em
can change the multiset only by deleting some elements
for the following two reasons:
Instances that do not contain any new function 
symbol have no effect on the values in \nlbmaths{\mathcal M}, \hskip.1em
because \math{\mathcal M} is a model of the old equations. \hskip.2em
There are no other instances
because 
the new function symbols do not occur in the old equations,
and because
we consider all our variables to be 
\index{constructor variables}%
constructor variables
as explained in \nlbsectref{subsection Constructor Variables}.\footnotemark
\pagebreak
\\\indent\footnotetext{\label{note on general variables}%
 Among the old equations here,
 we may even admit projective equations with {\em general}\/ 
 variables, \hskip.1em
 such as for destructors and the conditional function
 \bigmaths{\FUNDEF{\ifthenelseindexsymbol\nat}{\bool,\nat,\nat}\nat}:
 \\\linemaths{\begin{array}{l@{\,}l@{\,}l}\ppp{\spp X}
  &=
  &X
 \\
 \\\end{array}\hfill
 \begin{array}{|l@{\,}l@{\,}l}\carpp{\cnspp X L}
  &=
  &X
 \\\cdrpp{\cnspp X L}
  &=
  &L
 \\\end{array}\hfill
 \begin{array}{|l@{\,}l@{\,}l}\ifthenelseindexpp\nat\truepp  X Y
  &=
  &X
 \\\ifthenelseindexpp\nat\falsepp X Y
  &=
  &Y
 \\\end{array}}{} 
 for general
 variables 
 \hastype{X,Y}\nat, \hskip.2em
 \hastype L{\app\lists\nat}, \hskip.3em
 ranging over general terms (instead of constructor terms only).
}%
Moreover, 
a rewrite step with a {\em new}\/ equation 
reduces only a single innermost occurrence 
of a new function symbol,
because only a single new function symbol occurs on the left-hand side 
of the equation and because we consider all our variables
to be constructor variables.
The other occurrences in the multiset are not affected because
\math{\mathcal M} \nolinebreak is a model of the new equations.
Thus, such a rewrite step
reduces the multiset 
in a \wellfounded\ relation,
namely the multiset extension of the \wellfounded\ relation of the template
in the assumed model \nlbmaths{\mathcal M}. \hskip.3em
Indeed,
this follows from the fulfilledness of 
the conditions of the equation and the validity of the template. 
\\\indent
Thus, 
in each rewrite step,
the multiset gets smaller in a \wellfounded\ ordering 
or does not change. \hskip.2em
Moreover, \hskip.1em
if we assume 
that rewriting with the old equations 
terminates, \hskip.1em
then the new equations must be applied infinitely often in this sequence, 
\hskip.1em
and so the multiset gets smaller in infinitely many steps, \hskip.1em
which is impossible in a \wellfounded\ ordering.%
\index{termination|)}%
\subsection{Applicable Induction Templates for Explicit Induction}
\index{induction!explicit}%
We restrict the discussion in this section to 
recursive functions that are not mutually recursive,
partly for simplicity and 
partly because 
induction templates are
hardly helpful for finding proofs
involving non-trivially mutually recursive functions.\footnote{%
 See, however, \cite{mutualexplicitinduction} for 
 \index{induction!explicit} explicit-induction heuristics
 applicable to simple forms
 of mutual recursion.%
}
\\\indent
\mbox{Moreover,}
\mbox{in principle,}
users can always encode mutually recursive functions \math{f_k(\ldots)} 
\hskip.1em
by means of a single recursive function \nlbmaths{f(k,\ldots)}. \hskip.4em
Via such an encoding, humans tend to provide 
additional heuristic information 
relevant for induction templates, 
namely by the way they 
standardize the argument list \wrt\ length and position 
(\cfnlb\ the ``changeable positions'' below).
\\\indent
Thus, \hskip.1em
all the \math{f_k} with arity \nlbmath{n_k}
of \nlbsectref{subsection Termination}
simplify to one symbol \nlbmath f with arity \nlbmaths n.
Moreover, under this restriction it is easy to
partition the \index{measured positions}measured positions of a template 
into 
\index{changeable positions}``changeable'' and 
``unchangeable'' ones.\footnote{%
 This partition into \index{changeable positions}changeable and \index{changeable positions}unchangeable positions (actually: variables)
 originates in \cite[\p\,185\f]{bm}.%
}
\\\indent
\index{changeable positions}{\em Changeable}\/ are those 
\index{measured positions}measured positions \nlbmath i of the template 
which sometimes change in the recursion, 
\ie\ for which there is a triple \trip{f(t_1,\ldots,t_n)} R C \hskip.1em
in the \index{relational descriptions}relational description of the template,
and an \nlbmath{f(t'_1,\ldots,t'_n)\in R} \hskip.1em
such that \maths{t'_i\tightnotequal t_i}. \hskip.5em
The remaining \index{measured positions}measured positions of the template
are called {\em unchangeable}. \hskip.2em
\index{changeable positions}Unchangeable positions 
typically result from the inclusion of 
a global variable into the argument list of a function
(to observe an applicative programming style).
\\\indent
To improve the applicability
of the induction hypotheses of the 
step cases \mbox{produced} by the \inductionrule, \hskip.1em
these induction hypotheses should mirror the
recursive calls of the unfolding of 
the definition of a function \nlbmath{f} occurring in the \inductionrule's 
input formula, say
\\[-.5ex]\noindent\LINEmaths{A[f(t''_1,\ldots,t''_{n})]}.

\par\halftop\noindent 
An induction template is {\em applicable}\/ to the indicated occurrence of
its function symbol \nlbmath f \hskip.1em
if the terms \nlbmath{t''_i} at 
the \index{changeable positions}changeable positions \nlbmath i of the template
are {\em distinct variables}\/ 
and none of these variables occurs in the terms \nlbmath{t''_{i'}} \hskip.1em
that fill the \index{changeable positions}unchangeable positions \nlbmath{i'} of the template.\footnotemark\
For templates of constructor-style equations we additionally have to 
require here that the first element \nlbmath{f(t_1,\ldots,t_n)} of each triple
of the \index{relational descriptions}relational description of the template matches
\math{\inpit{f(t''_1,\ldots,t''_{n})}\xi} 
for some \index{constructor substitutions}{\em constructor substitution}\/
\nlbmath\xi\ that may replace the variables of \nlbmath{f(t''_1,\ldots,t''_{n})}
\hskip.1em with constructor terms,
\ie\ terms consisting of constructor symbols and variables,
such that \bigmaths{t''_i\xi\tightequal t''_i}{}
for each \index{changeable positions}unchangeable position \nlbmath i of the template.%
\halftop
\label{section example Applicable Induction Templates constructor style}%
\begin{example}[Applicable Induction Templates]\label
{example Applicable Induction Templates constructor style}%
\\Let us consider the conjecture \inpit{\acksymbol 4} from
\sectref{subsection Mathematical Induction and the Natural Numbers}.
From the three induction templates of \examrefs
{example induction template two measured subsets constructor style}
{example non-singleton measured subset constructor style},
only the one of 
\examref{example non-singleton measured subset constructor style}
is applicable.
The two of 
\examref{example induction template two measured subsets constructor style}
are not applicable because \lespp{\spp x}{\spp y} cannot be matched to 
\math{\inpit{\lespp y{\ackpp x y}}\xi} \hskip.1em
for any 
\index{constructor substitutions}%
constructor substitution \nlbmaths\xi.
\getittotheright\qed%
\index{induction templates|)}%
\end{example}

\subsection{Induction Schemes}\label
{subsection Induction Schemes}
\noindent
\index{induction schemes|(}%
Let us recall that for every recursive call 
\nlbmath{f(t'_{j',1},\ldots,t'_{j',n})} \hskip.2em
in a \pnc\ equation with left-hand side \nlbmaths{f(t_1,\ldots,t_{n})}, 
\hskip.4em
the \index{relational descriptions}relational description of an \index{induction templates}induction template for \nlbmath{f}
contains a triple
\par\noindent\LINEmaths{\displaytrip
{f(t_1,\ldots,t_{n})}
{\setwith{f(t'_{j,1},\ldots,t'_{j,n})}
{j\tightin J}}C},
\par\noindent such that \math{j'\tightin J} \hskip.2em
(by definition of an \index{induction templates}induction template).

Let us assume 
that the \index{induction templates}induction template is valid and applicable to the 
occurrence indicated in the formula
\math{A[f(t''_1,\ldots,t''_{n})]} \hskip.1em
given as input 
to the \inductionrule\ of \index{induction!explicit}explicit induction. \hskip.1em
Let \math{\sigma} be the substitution
whose domain are the variables
of \math{f(t_1,\ldots,t_{n})} \hskip.1em and 
which matches the first element \nlbmath{f(t_1,\ldots,t_{n})} \hskip.1em
of the triple to
\math{\inpit{f(t''_1,\ldots,t''_{n})}\xi} \hskip.1em
for some 
\index{constructor substitutions}%
constructor substitution \nlbmath\xi\
whose domain are the variables of 
\nlbmaths{f(t''_1,\ldots,t''_{n})}, \hskip.2em
such that \bigmathnlb{t''_i\xi\tightequal t''_i}{}
for each \index{changeable positions}unchangeable position \nlbmath i of the template. \hskip.1em
Then we have \nlbmath{t_i\sigma=t''_i\xi} \hskip.2em 
for \maths{i\in\{1,\ldots,n\}}. \hskip.2em

Now, for the \wellfoundedness\ of the generic step-case formula
\par\halftop\noindent\LINEmaths{
\inparentheses{\mediumheadroom\mediumfootroom
\inparenthesestight{A[f(t''_1,\ldots,t''_n)]}\xi
\nottight{\nottight\antiimplies}
\bigwedge_{j\in J}\inparenthesestight{A[f(t''_1,\ldots,t''_n)]}\mu_j}
\nottight{\nottight\antiimplies}\bigwedge C\sigma}{}
\par\halftop\noindent to be implied by the validity of the \index{induction templates}induction template,
it suffices
to take substitutions \math{\mu_j}
whose domain \nlbmath{\DOM{\mu_j}} \hskip.1em
is the set of variables of \nlbmaths{f(t''_1,\ldots,t''_n)},
\hskip.3em
such that the constraint 
\hskip.2em\maths{t''_i\mu_j\tightequal t'_{j,i}\sigma}{} \hskip.2em
is satisfied for each \index{measured positions}measured position \nlbmath i of the template
and for each \nlbmath{j\in J} \hskip.2em
(because of 
 \nolinebreak\hskip.1em\nlbmath{t''_i\xi\tightequal t_i\sigma}).

If \nolinebreak \math i is an \index{changeable positions}unchangeable position of the template, \hskip.1em
then we have \math{t_i\tightequal t'_{j,i}} 
and \mbox{\maths{t''_i\xi\tightequal t''_i}.} \hskip.5em
Therefore, we can satisfy the constraint by requiring \nlbmath{\mu_j}
to be the identity on the variables of \nlbmath{t''_i}, \hskip.2em
simply because then we have
\bigmaths{t''_i\mu_j\tightequal t''_i\tightequal t''_i
\xi\tightequal t_i\sigma\tightequal 
t'_{j,i}\sigma}.

If \math i is a \index{changeable positions}changeable position, 
then we know by the applicability of the template
that \math{t''_i} is a variable not occurring
in another \index{changeable positions}changeable or \index{changeable positions}unchangeable position in 
\nlbmaths{f(t''_1,\ldots,t''_n)}, \hskip.2em
and we can satisfy the constraint simply by defining 
\maths{t''_i\mu_j:=t'_{j,i}\sigma}.

\pagebreak
\footnotetext{%
 This definition of applicability originates in \cite[\p\,185\f]{bm}.%
}%
On the remaining variables of \nlbmaths{f(t''_1,\ldots,t''_n)}, \hskip.2em
we define \math{\mu_j} in a way that we get \maths
{t''_i\mu_j\tightequal t'_{j,i}\sigma}{}
for as many \index{measured positions}unmeasured positions \nlbmath i as 
possible,
and otherwise as the identity.
This is not required for \wellfoundedness, 
but it improves the 
likeliness of applicability of the induction hypothesis
\nlbmath{\inpit{A[f(t''_1,\ldots,t''_n)]}\mu_j} \hskip.2em
after unfolding \nlbmaths{f(t''_1,\ldots,t''_n)\xi}{} \hskip.1em
in \nlbmaths{\inpit{A[f(t''_1,\ldots,t''_n)]}\xi}. \hskip.5em
Note that such an eager instantiation is required in \index{induction!explicit}explicit induction
unless the logic admits one of the following:
existential quantification, existential
variables,\footnotemark\
\index{induction!lazy}%
lazy induction-hypothesis generation.
\par\halftop\halftop\halftop\noindent
An {\em induction scheme}\/ for the given input formula
consists of the following items:\begin{enumerate}\item
The \index{position sets}{\em position set}\/ contains the position of 
\nlbmaths{f(t''_1,\ldots,t''_n)}{} \hskip.1em
in \nlbmaths{A[f(t''_1,\ldots,t''_n)]}. \hskip.4em
Merging of induction schemes may lead to non-singleton position sets later.%
\noitem\item\sloppy
The set of the 
\index{induction variables}\mbox{\em induction variables}, \hskip.2em
which are defined as the variables 
at the \index{changeable positions}changeable positions of the \index{induction templates}induction template
in \nlbmaths{f(t''_1,\ldots,t''_n)}.%
\noitem\item
To obtain a \index{step-case descriptions}{\em step-case description}\/
for all step cases 
by means of the generic step-case formula 
displayed above, \hskip.1em
each triple in the \index{relational descriptions}relational description of the considered form
is replaced with the new triple 
\\\LINEmaths{\displaytrip
\xi{\setwith{\!\mu_j\!}{\!j\tightin J\!}}{C\sigma}}.
\\To make as many induction hypotheses available as possible
in each case,
we \nolinebreak assume that \index{step-case descriptions}step-case descriptions are 
implicitly 
kept normalized
by the following associative commutative operation:
If two triples are identical in their first elements and in
their last elements, we replace them with the single triple 
that has the same first and last elements
and the union of the middle elements
as new middle element. 
\noitem\item
We also add the 
\index{hitting ratio}{\em hitting ratio}\/\footnotemark\
of all 
substitutions \nlbmath{\mu_j} with 
\nlbmaths{j\in J}{} \hskip.2em given by
\\\noindent\LINEmaths{\displaystyle\frac{\CARD{\setwith
{\pair j i\in J\tighttimes\{1,\ldots,n\}}{t''_i\mu_j\tightequal t'_{j,i}\sigma}}}
{\CARD{J\tighttimes\{1,\ldots,n\}}}},
\\\noindent\mediumheadroom where \math J \hskip.1em
actually 
has to be the disjoint sum over all the \nlbmath J occurring as index sets of
second elements of triples like the one displayed above.\end{enumerate}%
Note that the resulting \mbox{step-case} description is a set
describing all step cases of an induction scheme; \hskip.3em
these step cases are guaranteed to be \wellfounded,\footnotemark\
\hskip.2em 
but 
---~for providing a sound induction formula~---
they still have to be complemented by base cases, \hskip.1em
which may be analogously described by triples 
\nlbmaths{\trip{\xi}\emptyset{C}}, \hskip.2em
such that all substitutions in the first elements of the triples
together describe a 
distinction of cases that is complete for constructor terms and, \hskip.1em
for each of these substitutions, \hskip.1em
its case conditions describe a complete distinction of cases again.%

\addtocounter{footnote}{-2}\footnotetext{%
 \majorheadroom
 Existential variables are called ``free variables'' in modern tableau systems
 (see the \nth 2\,\rev\,\edn\,\cite{fitting}, 
  but not its \nth 1\,\edn\,\cite{Fitting90}) \hskip.1em
 and occur with extended functionality under different names in 
 the inference systems of 
 \makeaciteofthree
 {wirthcardinal}%
 {wirth-jsc-non-permut}%
 {SR--2011--01}%
 .%
}\addtocounter{footnote}{1}\footnotetext{%
 \majorheadroom
 We newly introduce this name here in the hope that it helps the readers
 to remember that this ratio measures how well
 the induction hypotheses hit the recursive calls.%
}\addtocounter{footnote}{1}\footnotetext{%
 \majorheadroom
 \Wellfoundedness\ is indeed guaranteed according to the above discussion.
 As a consequence, the induction scheme does not need 
 the weight term and the \wellfounded\ relation of 
 the \index{induction templates}induction template anymore.%
}%
\pagebreak 
\halftop\halftop\label{section example induction schemes constructor style}%
\begin{example}[Induction Scheme]\label
{example induction schemes constructor style}%
\mbox{}\par\noindent
The template for \acksymbol\ of \examref
{example non-singleton measured subset constructor style}
is the only one that is applicable to \inpit{\acksymbol 4}
according to \examref{example Applicable Induction Templates constructor style}.
\hskip.3em
It yields the following induction scheme.
\par
The \index{position sets}{\em position set}\/ is \nlbmath{\{1.1.2\}}. \hskip.5em
It describes the occurrence of \nlbmath\acksymbol\ in 
the second subterm of 
the left-hand side of 
the first literal of 
the formula \inpit{\acksymbol 4} \hskip.1em
as input
to the \inductionrule\ of \index{induction!explicit}explicit induction:
\\[+.5ex]\LINEmaths
{\inpit{\acksymbol 4}\,/\,1.1.2\nottight{\nottight{\nottight=}}
\ackpp x y}.
\par
The set of \index{induction variables}{\em induction variables}\/ 
is \nlbmaths{\{x,y\}}, \hskip.2em
because both 
\mbox{positions} of the \index{induction templates}induction template are \index{changeable positions}changeable. 

The \index{relational descriptions}relational description of the \index{induction templates}induction template
is replaced with the \index{step-case descriptions}{\em step-case description}
\\[+.5ex]\LINEmaths{\displayset{
\displaytrip{\xi_1}{\{\mu_{1,1}\}}\emptyset\comma\ \
\displaytrip{\xi_2}{\{\mu_{2,1},\mu_{2,2}\}}\emptyset}}. 
\\[+.7ex]that is given as follows.
\par
The first triple of the \index{relational descriptions}relational description, 
namely
\\[+.7ex]\noindent\LINEmaths{\displaytrip
{\ackpp{\spp x}\zeropp}{\{\ackpp x{\spp\zeropp}\}}\emptyset}{}
\\[+.5ex](obtained from the equation \inpit{\acksymbol 2}) \hskip.2em 
is replaced with 
\par\noindent\LINEmaths{\displaytrip{\xi_1}{\{\mu_{1,1}\}}\emptyset},
\par\noindent where \hskip.1em
\mbox{\math{\xi_1=\{x\tight\mapsto\spp{x'}\comma y\tight\mapsto\zeropp\}}}
\hskip.2em
and \nolinebreak\hskip.1em
\nlbmaths{\mu_{1,1}=\{x\tight\mapsto x'\comma y\tight\mapsto\spp\zeropp\}}. 
\hskip.4em
This can be seen as follows.
The substitution called \math\sigma\ in the above discussion
\ ---~~which has to match the first element of the triple
 to \math{\inpit{\inpit{\acksymbol 4}/1.1.2}\xi_1}~~--- \
has to satisfy 
\bigmaths
{\inpit{\ackpp{\spp x}\zeropp}\sigma=\inpit{\ackpp x y}\xi_1}.
Taking \math{\xi_1} as the minimal 
\index{constructor substitutions}%
constructor substitution 
given above, \hskip.1em
this determines \nlbmaths{\sigma=\{x\tight\mapsto x'\}}. \hskip.3em
Moreover, \hskip.1em
as both positions of the template are \index{changeable positions}changeable, \hskip.2em
\math{\mu_{1,1}} has to match \math{\inpit{\acksymbol 4}/1.1.2}
\hskip.2em
to the \nlbmath\sigma-instance of the single element of the
second element of the triple,
which determines \math{\mu_{1,1}} as given.
\par
The second triple of the \index{relational descriptions}relational description,
namely
\\[+.7ex]\phantom\qed\LINEmaths{\displaytrip
{\ackpp{\spp x}{\spp y}}
{\{\ackpp{\spp x}{y}\comma\ackpp{x}{\ackpp{\spp x}{y}}\}}
\emptyset}{}
\\[+.5ex](obtained from the equation \inpit{\acksymbol 3}) \hskip.2em 
is replaced with \nolinebreak\hskip.2em
\nlbmaths{\displaytrip{\xi_2}{\{\mu_{2,1},\mu_{2,2}\}}\emptyset},
\hskip.4em
where \nolinebreak\hskip.5em
\nlbmaths{\xi_2=\{x\tight\mapsto\spp{x'}\comma y\tight\mapsto\spp{y'}\}},
\hskip.8em 
\maths{\mu_{2,1}=\{x\tight\mapsto\spp{x'}\comma y\tight\mapsto y'\}}, \hskip.8em 
and \hskip.1em
\mbox{\maths{\mu_{2,2}=\{x\tight\mapsto x'\comma y\tight\mapsto
\ackpp{\spp{x'}}{y'}\}}.}
\hskip.5em
This can be seen as follows.
The substitution called \math\sigma\ in the above discussion
has to satisfy
\bigmaths
{\inpit{\ackpp{\spp x}{\spp y}}\sigma=\inpit{\ackpp x y}\xi_2}.
Taking \math{\xi_2} as the minimal 
\index{constructor substitutions}%
constructor substitution 
given above, \hskip.1em
this determines \nlbmaths
{\sigma=\{x\tight\mapsto x'\comma y\tight\mapsto y'\}}. \hskip.4em
Moreover, 
we get the constraints
\bigmaths{\inpit{\ackpp x y}\mu_{2,1}=
\inpit{\ackpp{\spp x}{y}}\sigma}{} and \bigmaths{
\inpit{\ackpp x y}\mu_{2,2}=\inpit{\ackpp x{\ackpp{\spp x}{y}}}\sigma}, 
which determine \math{\mu_{2,1}} and \math{\mu_{2,2}} as given above.
\par
The \index{hitting ratio}hitting ratio for the three constraints on the two arguments 
of \math{\inpit{\acksymbol 4}/1.1.2} \hskip.1em
is \hskip.2em\nlbmaths{{6\over 6}=1}. \hskip.4em
This is optimal: the induction hypotheses
are 100\% identical to the expected recursive calls.
\par
To achieve completeness of the substitutions \nlbmath{\xi_k} 
for constructor terms we have to add the base case
\trip{\xi_0}\emptyset\emptyset\ \hskip.1em
with \nlbmath{\xi_0=\{x\tight\mapsto\zeropp\comma y\tight\mapsto y\}} \hskip.1em
to the \index{step-case descriptions}step-case description.
\par
The three new triples now describe exactly the three formulas 
displayed at the beginning of 
\examref{example proof of (ack4)} in
\sectref{section example proof of (ack4)}.
\getittotheright\qed
\index{induction schemes|)}%
\end{example}
\halftop\section{Automated Explicit Induction}\label
{section Explicit Induction}%
\halftop\halftop\subsection
{The Application Context of Automated Explicit Induction}\label
{subsection The Application Context of Automated Explicit Induction}%
\halftop\halftop\noindent\index{induction!explicit|(}%
Since the development of programmable computing machinery 
in the middle of the \nth{20}\,cen\-tu\-ry, \hskip.2em
a major problem of hard- and software has been and still is the 
uncertainty 
that
they actually always do what they should do.

\halftop\indent
It is almost never the case  
that the product of the possible initial states, input threads, 
and schedulings of a computing system is a 
small 
number.
Otherwise,
however,
even the most carefully chosen test series cannot cover
the often very huge or even infinite number of possible cases; \hskip.2em
and then, no matter how many bugs have been found by testing,
there can never be certainty that none remain.
\par\halftop\halftop\halftop\noindent
Therefore,
the only viable solution to this problem seems 
to be:
\begin{quote}
Specify the intended functionality in a language of formal logic,
and then supply a formal mechanically checked proof that the program actually 
satisfies the specification!\end{quote}
Such an approach also requires formalizing the platforms on which the
system is implemented.  This may include the hardware, operating system,
programming language, sensory input, \etc\ \hskip.3em
 One may additionally formalize and prove that the underlying platforms are
 implemented correctly and this may ultimately involve proving, for example,
 that a network of logical gates and wires implements a given abstract
 machine.  
 Eventually, however, 
 one must make an engineering judgment that
 certain physical objects (\eg\ printed circuit boards, gold plated pins,
 power supplies, \etc)\ reliably behave as specified.
 To be complete,
 such an approach would also require a verification 
 that the verification system is sound and correctly implemented.%
 \footnote{%
 See, for example, \cite{davis-2009}.%
 }%
\par\halftop\indent
A crucial problem, however,
is the cost ---~in time and money~--- 
of doing 
the many 
proofs required, given
the huge amounts of application 
hard- and software in our modern 
economies.
Thus, 
we can expect formal verification only in 
areas where the managers 
expect that mere testing does not suffice,
that the costs of the verification process are lower than
the costs of bugs in the hard- or software,
and that the competitive situation 
admits the verification investment. \hskip.2em
Good candidates are the areas of
central processing units (CPUs) 
in standard processors and of security protocols.
\par\halftop\indent
To reduce the costs of verification,
we can hope to automate it with automated theorem-proving systems. \hskip.2em
This automation has to include mathematical induction
because induction is essential for the verification of the properties of 
most data types used in digital design
(such as natural numbers, arrays, lists, and trees), \hskip.1em
for the repetition in processing (such as loops), \hskip.1em
and for parameterized systems (such as a generic \math n-bit adder).
\\\mbox{}
\pagebreak

Decision methods
(many of them exploiting finiteness, \eg\ the use of 32-bit data paths) 
\hskip.1em
allow automatic verification of some modules,
but 
---~barring a completely un\-expected breakthrough in the future~---
the verification of a new hard- or software system will always require 
human users who help the theorem-proving systems to explore and develop
the notions and theories that properly match the new 
system.
\par\halftop\indent
Already today,
however, 
\ACLTWO\ often achieves complete automation in 
verifying minor modifications of previously verified modules --- an
activity called {\em{proof maintenance}} which is increasingly important 
in the microprocessor-design industry.
\subsection{The\/ \PURELISPTP}\label
{The Pure Lisp Theorem Prover}%
\halftop\halftop\noindent
Our overall task is to answer
---~from a historical perspective~---
the question:
\begin{quote}How could \boyername\ and \moorename\ 
---~starting virtually from zero\footnote{%
 \label{note burstall quotation}%
 No heuristics at all were explicitly described, 
 for instance,
 in \burstall's 1968 work on program verification 
 by induction over recursive functions
 in \cite{burstall-1969}, 
 where 
 the proofs were not even formal,
 and an implementation seemed to be more or less utopian:\notop\halftop\halftop
 \begin{quote}``%
 The proofs presented will be mathematically rigorous but not formalised
 to the point where each inference is presented as a mechanical application
 of elementary rules of symbol manipulation.
 This is deliberate since I feel that our first aim should be to 
 devise methods of proof which will prove the validity of non-trivial programs
 in a natural and intelligible manner.
 Obviously we will wish at some stage to formalise the reasoning to a point
 where it can be performed by a computer to give a mechanised debugging service.%
 '' \getittotheright{\cite[\p\,41]{burstall-1969}}\end{quote}
 As far as we are aware, 
 besides interactively invoked induction in 
 resolution theorem proving 
 (\eg\ by starting a resolution proof for 
  the two clauses resulting from \skolemization\ 
  of \math{\inpit{\Pppp\zeropp\und\neg\Pppp x}
  \implies\exists y\stopq\inpit{\Pppp y\und\neg\Pppp{\spp y}}}
 \cite{darlington-1968}), \hskip.3em
 the only implementation of an automatically invoked
 mathematical-induction heuristic prior to\,1972 \hskip.1em
 is in a set-theory prover by
 \citet{Ble71},
 which uses 
 \structuralinductionindex%
 structural induction over \nlbmath\zeropp\ and \nlbmath\ssymbol\ 
 (\cfnlb\ \sectref{subsection Mathematical Induction and the Natural Numbers}) 
 \hskip.1em
 on a randomly picked, universally quantified variable of type \nlbmaths\nat.%
}
in the summer of 1972~---
actually invent their long-lived solutions
to the hard heuristic problems in the automation of induction
and implement them in the sophisticated theorem prover 
\THM\ as described in \cite{bm}?
\end{quote}
As already described in \sectref{section Preface}, \hskip.1em
the breakthrough in the heuristics for automated inductive theorem proving
was achieved with
the 
\mbox{``\PURELISPTP\closequotecomma}
devel\-oped and implemented by 
\boyer\ and \moore.
It was presented by \moore\ 
at the third \IJCAI\  \cite{boyer-moore-1973}, \hskip.1em
which took 
place in \StanfordCA\ in August\,1973, \hskip.1em  
and it \nolinebreak is best documented in Part\,II of \moore's \PhDthesis\
\shortcite{moore-1973}, \hskip.1em
defended in November\,1973.
\par\halftop\indent
The \PURELISPTP\ was given no name in the before-mentioned publications.
The only occurrence of the name in publication seems to be in
\cite[\p\,1]{moore-1975}, \hskip.2em
where it is actually called
``the \boyermoore\ \PURELISPTP\closequotefullstopnospace
\pagebreak\par\halftop\halftop\noindent
To make a long story short, 
the fundamental insights were\begin{itemize}\noitem\item 
to exploit the duality of recursion and induction 
to formulate explicit induction hypotheses,\noitem\item
to abandon ``random'' search and focus on simplifying the goal
by rewriting and normalization techniques to lead to opportunities to use the
induction hypotheses, and\noitem\item
to support generalization to prepare subgoals
for subsequent inductions.\end{itemize}
\indent
Thus, 
it is not enough for us to focus here just on the
induction heuristics {\it per \nolinebreak se}, but it is necessary 
to place them in the context of
the development of 
the \boyermoorewaterfall\ (\cfnlb\ \figuref{figure waterfall}).%

To understand the achievements a bit better,
let us now discuss the material of Part\,II of
\moore's \PhDthesis\ in some detail,
because it provides some explanation 
of 
how \boyer\ and \moore\
could be so surprisingly successful.
Especially helpful for understanding the process of creation
are those procedures of the \PURELISPTP\
that are provisional \wrt\ their refinement in
later \boyermooretheoremprovers.
Indeed, 
these provisional procedures help to decompose
the 
leap from nothing to \THM,
which was achieved by 
two men 
in less than eight years of work.

As \bledsoename\ \bledsoelifetime\ was \boyer's \PhD\ advisor,
it is no surprise that the \PURELISPTP\ shares many design features with
\bledsoe's provers.
 In \cite[\p 172]{moore-1973} we read on the
 \PURELISPTP:\notop\halftop\begin{quote}``%
 The design of the program,
 especially the straightforward approach of `hitting' the theorem over and over
 again with rewrite rules until it can no longer be changed,
 is largely due to the influence of \bledsoename.%
 ''\notop\halftop\end{quote}
\begin{sloppypar}\noindent
\boyer\ and \moore\ report\footnote{%
 \Cfnlb\ \cite{boyer-moore-2012}.%
}
that 
in late\,\,1972 and early\,\,1973 \hskip.1em
they were doing proofs about list data structures
on the blackboard and verbalizing to each other the heuristics
behind their choices on how to proceed with the proof.
This means that,
although \index{induction!explicit}explicit induction is not the approach
humans would choose for non-trivial induction tasks, \hskip.1em
the heuristics of the \PURELISPTP\
are learned from human heuristics after all.

Note that \boyer's and \moore's method 
of learning computer heuristics from 
their own human behavior in mathematical logic
was a step of two young men against the spirit of the time:
the use of vast amounts of computational power to {\em search}\/ an even more
enormous space of possibilities. \hskip.1em 
\boyer's and \moore's goal,
however,  
was in a sense more modest:\notop\halftop\begin{quote}%
``The program was designed to behave properly on simple functions.
The overriding consideration was that it should be automatically able to prove theorems about simple \LISP\
functions in the straightforward way we prove them.''
\getittotheright{\cite[\p\,205]{moore-1973}}\notop\halftop\end{quote}
\pagebreak 

\indent
It may be that the orientation toward 
human-like or ``intelligible''
methods and heuristics in the automation of theorem proving 
had also some tradition
in \EB\ at the time,\footnote{%
 \Cf\ \eg\ the quotation from \cite{burstall-1969} 
 in \noteref{note burstall quotation}.%
} \hskip.1em 
but, \hskip.1em 
also in this aspect, \hskip.1em 
the major influence on \boyer\ and \moore\ is again
\bledsoename.\footnote{%
 \majorheadroom
 \Cfnlb\ \eg\ \cite{BBH72}.%
}

The source code of the \PURELISPTP\ 
was written in the programming language {\sc POP--2}\@.\footnote{%
 \majorheadroom
 \Cfnlb\ \cite{pop2}.%
} \hskip.1em
\boyer\ and \moore\ were the only programmers involved in the implementation.
The average time in the central processing unit (CPU) 
of the \englishEBPURELISPMACHINE\
for the proof of a theorem is reported to 
be about ten seconds.\footnotemark\ \hskip.3em
\footnotetext{%
 \majorheadroom
 Here is the actual wording of the timing result found on 
 \litspageref{171\f}\,\,of \cite{moore-1973}:
 \begin{quote}
 ``Despite theses inefficiencies, 
 the `typical' theorem proved requires only 8\,\,to\,10 seconds of CPU time.
 For comparison purposes, 
 it should be noted that the time for {\tt CONS} in 4130 {\sc POP--2}
 is 400\,\,microseconds,
 and {\tt CAR} and {\tt CDR} are about 50\,\,microseconds each.
 The hardest theorems solved, 
 such as those involving {\tt SORT}, 
 require 40 to 50\,\,seconds each.''%
 \end{quote}%
}%
This was considered fast at the time, compared to the search-dominated proofs by
resolution systems. \hskip.1em  
\moore\ explains the speed:%
\begin{quote}
``Finally, it should be pointed out that the program uses no search.  At no time
does it `undo' a decision or back up.  
This is both the primary reason it is
a fast theorem prover, and strong evidence that its methods allow the theorem
to be proved in the way a programmer might `observe' it.  
The program is
designed to make the right guess the first time, and then pursue one goal with
power and perseverance.''
\getittotheright{\cite[\p\,208]{moore-1973}}
\end{quote}%
One remarkable omission in the \PURELISPTP\ is lemma application.
As \nolinebreak a \nolinebreak consequence,
the success of proving a set of theorems cannot depend on the order
of \nolinebreak their presentation to the theorem prover. \hskip.2em
Indeed, 
just as the resolution theorem provers of the time,
the \PURELISPTP\ starts every proof right from scratch
and does not improve its behavior 
with the help of previously proved lemmas. \hskip.2em
This was a design decision; \hskip.1em
one of the reasons was:
\begin{quote}
``Finally, one of the primary aims of this project has been to demonstrate
clearly that it is possible to prove program properties entirely automatically.
A total ban on all built-in information about user defined function%
s
thus removes any taint of user supplied information.''
\getittotheright{\cite[\p\,203]{moore-1973}}
\end{quote}%
Moreover, all induction orderings in the \PURELISPTP\ 
are
recombinations of constructor relations,
such that all inductions it can do are 
\structuralinductionindex%
structural inductions
over combinations of constructors.
As a consequence,
contrary to later \boyermooretheoremprovers,
the \wellfoundedness\ of the induction orderings 
does not depend on the termination 
of the recursive function definitions.\footnotemark
\pagebreak
\end{sloppypar}

\par\footnotetext{%
 Note that the \wellfoundedness\ of the constructor relations
 depends on distinctness of the constructor ground terms in the models,
 but this does not really depend on the termination of the recursive
 functions because 
 (as discussed in \sectref{subsection Confluence}) \hskip.1em
 \index{confluence}confluence is sufficient here.%
}%
Nevertheless,
the soundness of the \PURELISPTP\ depends
on the termination of the recursive function definitions,
but only in one aspect: \hskip.3em
It simplifies and evaluates expressions 
under the assumption of termination. \hskip.4em
For instance, \hskip.2em
both
\ \mbox{\tt (IF\hskip.05em\footnotemark\ \math a \math d \math d\hskip.07em)}%
\footnotetext{%
 \majorheadroom
 \Cf\ \noteref{note on IF and COND}.%
} \ 
and 
\ \mbox{\tt (CDR (CONS \math a \math d\hskip.05em))} \ 
simplify to \nlbmaths d, \hskip.2em
no matter whether \mbox{\math a terminates;}
\hskip.2em
and it is admitted to rewrite with a recursive function definition 
even if an argument of the function call does not terminate.
\hskip.2em
Note that such a lazy form of evaluation is sound \wrt\ the given logic 
\onlyif\ each eager call terminates and returns a constructor ground term,
simply
because all functions are meant to be defined in terms of 
\index{constructor variables}%
constructor variables
(\cfnlb\ \sectref{subsection Constructor Variables}).\footnote{%
 \majorheadroom
 There is a work-around 
 for projective functions 
 as indicated in 
 \noteref{note on general variables} and in \nlbcite{wirth-jsc}.%
} 

The termination of the recursively defined functions,
however, is not checked by the\/ \PURELISPTP,
but comes as a {\it proviso}\/ for its soundness.

The logic of the \PURELISPTP\ is an applicative\footnotemark\
subset of the logic of \LISP\@. \hskip.3em
The only {\em destructors}\/ \hskip.1em in this logic 
are {\tt CAR} and {\tt CDR}\@. \hskip.3em
They are overspecified on the 
only {\em constructors}\/ \hskip.1em {\tt NIL} and {\tt CONS} \hskip.1em
by the following equations:
\\[+.9ex]\noindent\LINEnomath{\tt\begin{tabular}[t]
{l@{\ }l c l@{\mbox{~~~~}}|@{\mbox{~~~~}}l@{\ }l c l}(CAR 
 &(CONS \math a \math d\hskip.05em)) 
 &= 
 &\math a
 &(CAR 
 &NIL) 
 &=
 &NIL
\\(CDR 
 &(CONS \math a \math d\hskip.05em)) 
 &= 
 &\math d
 &(CDR 
 &NIL) 
 &=
 &NIL
\\\end{tabular}}
\\[+.9ex]\footnotetext{%
 \majorheadroom
 \Cf\ \noteref{note applicative}.%
}
As standard in \LISP,
every term of the form
\mbox{\tt (CONS \math a \math d\hskip.07em)} \hskip.2em
is taken to be \truepp\
in the logic of the \PURELISPTP\
if it occurs at an argument position with \myboolean\ intention.
The actual truth values (to be returned by \myboolean\ functions) \hskip.1em
are {\tt NIL} (representing \falsepp) \hskip.1em
and \nolinebreak{\tt T}, \hskip.1em
which is an abbreviation for 
\mbox{\tt (CONS NIL NIL)} and represents \nlbmaths\truepp.\footnote{%
 \majorheadroom
 \Cf\ \nth 2\,paragraph of \litspageref{86} of \cite{moore-1973}.%
} 
\hskip.2em
Unlike conventional \LISP s (both then and now), the natural numbers
are represented by lists of {\tt NIL}s to keep the logic simple; \hskip.2em
the natural number \nlbmath\zeropp\ is represented by 
{\tt NIL} and the successor function \nlbmath{\spp{d\hskip.05em}} is
represented by \mbox{\tt (CONS NIL \math d\hskip.07em)}.\footnote{%
 \majorheadroom
 \Cf\ \nth 2\,paragraph of \litspageref{87} of \cite{moore-1973}.%
}%

\par\halftop\halftop\halftop\halftop\noindent
Let us now discuss the behavior of the \PURELISPTP\ 
by describing 
the instances of the stages of the \boyermoorewaterfall\
(\cfnlb\ \figuref{figure waterfall}) \hskip.1em
as \nolinebreak they are described in \moore's \PhDthesis.

\vfill\pagebreak

\subsubsection{Simplification in the\/ \PURELISPTP}\label
{subsubsection Simplification in the pure}%

\begin{sloppypar}%
\index{simplification|(}%
\halftop\halftop\halftop\noindent
The first stage of the \boyermoorewaterfall\ 
---~
``simplification'' in \figuref{figure waterfall}~---
is \nolinebreak called  
\index{normalation}``normalation''\footnote{%
 During the oral defense of the dissertation, 
 \moore's committee abhorred the
 non-word and instructed him to choose a word.  Some copies of the
 dissertation call the process ``simplification.'' }
in the \PURELISPTP\@. 
\mbox{It applies} the following simplification procedures 
to \LISP\ expressions
until the result does not change %
any more:
``evaluation\closequotecomma
``normalization\closequotecomma
and ``reduction\closequotefullstop
\end{sloppypar}
\par\halftop\halftop\halftop\halftop\halftop\indent
{\em``Normalization''}\/ tries 
to
find sufficient conditions 
for a given expression to have 
the soft type ``\myboolean'' and to normalize logical expressions. \hskip.2em
Contrary to clausal logic over equational atoms,
\LISP\ admits {\tt EQUAL} and 
{\tt IF} to appear
not only at the top level, but in nested terms.
To free later tests and heuristics from checking for their triggers
in every equivalent form, such a normalization \wrt\ 
propositional logic and equality is part of 
most theorem provers today.

\par\halftop\halftop\halftop\halftop\halftop\indent
{\em``Reduction''}\/ \hskip.1em
is a form of what today is called {\em contextual rewriting}. \hskip.2em
It is based on the fact that ---~in the logic of the \PURELISPTP~--- 
in the conditional expression
\par\noindent\LINEnomath{\mbox{\tt (IF \math c \math p \math n)}}
\par\noindent\mediumheadroom we can simplify occurrences 
of \nlbmath c in \nlbmath p
to \hskip.1em \mbox{\tt(CONS (CAR \math c) (CDR \math c))}, \hskip.3em
and in \nlbmath n to \nolinebreak{\tt NIL}\@. 
The replacement with
\hskip.1em \mbox{\tt(CONS (CAR \math c) (CDR \math c))} \hskip.2em
is executed only at positions with \myboolean\ intention
and can be improved in the following two 
special cases:\begin{enumerate}\item
If we know that \math c \hskip.1em
is of soft type ``\myboolean\closequotecommasmallextraspace
then we rewrite all occurrences of \nlbmath c in \nlbmath p
actually to \nolinebreak{\tt T}\@.\item
If \math c is of the
form \mbox{\tt(EQUAL \math l \math r)}, 
then we can rewrite occurrences of \math l in \math p to \math r
(or vice versa). \hskip.2em
Note that we have to treat the variables in \math l and \math r
as constants in this rewriting.
The \PURELISPTP\ rewrites 
in this case
\onlyif\ either
\nolinebreak\hskip.15em\nlbmath l or \nlbmath r \hskip.1em
is a ground term;\footnote{%
 \majorheadroom
 Actually, this ground term (\ie\ a term without variables) 
 here is always a {\em constructor}\/ 
 ground term 
 (\ie\ a term built-up exclusively from 
  \index{constructor function symbols}%
  constructor function symbols) 
 because the previously applied ``evaluation'' procedure
 has reduced any ground term to a constructor ground term,
 provided that the termination {\it proviso}\/ is satisfied.%
} \hskip.2em
then the other cannot be a ground term
because the equation would otherwise have been simplified to {\tt T}
or {\tt NIL} in the previously applied 
``evaluation\closequotefullstopextraspace
So replacing the latter term with the ground term
everywhere in \nlbmath p must terminate,
and this is all the contextual rewriting with equalities 
that the \PURELISPTP\ does in ``reduction\closequotefullstopnospace\footnote{%
 \majorheadroom
 Note,
 however, 
 that further contextual rewriting with equalities is applied
 in a later stage of the \boyermoorewaterfall,
 named \index{cross-fertilization}{\em cross-fertilization}.%
}\end{enumerate}
\vfill\pagebreak\par\indent
{\em``Evaluation''}\/ is a procedure that evaluates expressions partly by
simplification within the elementary logic as given by \myboolean\ operations
and the equality predicate. Moreover, 
``evaluation'' executes some rewrite steps with the 
equations defining the recursive functions.
Thus,
``evaluation'' can roughly be seen as normalization
with the rewrite relation resulting from the elementary logic
and from the recursive function definitions.
The rewrite relation is applied according to the innermost left-to-right
rewriting strategy, which is standard in \LISP\@.%

By ``evaluation\closequotecomma
ground terms are completely evaluated to their normal forms.
Terms containing (implicitly universally quantified) variables,
however, have to be handled in addition.
Surprisingly,
the considered rewrite relation is not necessarily
terminating on non-ground terms, although the \LISP\ evaluation of ground
terms terminates because of the assumed termination of
recursive function definitions (\cfnlb\ \sectref{subsection Termination}).
\hskip.2em
The reason for this  non-termination is 
the following: \hskip.2em
Because of the \LISP\ definition style via {\em un}\/conditional equations,
the positive/negative conditions are 
actually part of the {\em right-hand sides}\/ 
of the defining equations,
such that the rewrite step can 
be executed even if the conditions evaluate neither to 
false nor to true. \hskip.2em
For instance,
in \examref{example PLUS} of \nlbsectref{section example PLUS}, \hskip.1em
a rewrite step with the definition of {\tt PLUS}
can always be executed, 
whereas a rewrite step with \inpit{\plussymbol 1'} or \inpit{\plussymbol 2'} 
\hskip.1em
requires \bigmathnlb{x\tightequal\zeropp}{} to be definitely true or
definitely false. \hskip.1em
This means that non-termination may result from the rewriting of cases
that do not occur in the evaluation of any ground instance.\footnotemark
\footnotetext{%
 It becomes clear 
 in the second paragraph on \litspageref{118} of \cite{moore-1973} \hskip.1em
 that the code of both the positive and the negative case of a conditional
 will be evaluated,
 unless one of them can be 
 canceled by the complete evaluation of the governing condition
 to true or false.
 Note that the evaluation of both cases is necessary indeed
 and cannot be avoided in practice.
 \par
 Moreover, 
 note that a stronger termination requirement that guarantees
 termination independent of the governing condition is not 
 feasible for recursive function definitions in practice.
 \par
 Later \boyermooretheoremprovers\ also use lemmas for rewriting during 
 symbolic evaluation, which is another source of possible non-termination.%
}%

As the final aim of the stages of the \boyermoorewaterfall\
is a formula that provides 
concise and sufficiently 
strong induction hypotheses in the last of these stages, \hskip.1em
symbolic evaluation must be prevented from unfolding 
function definitions unless the context admits us to expect
an effect of simplification.\footnote{%
 \majorheadroom
 In \QUODLIBET\ 
 this is achieved by {\em contextual rewriting}\/
 where evaluation stops when the governing conditions cannot be
 established from the context. \Cfnlb\ \makeaciteoftwo{samoa-phd}{jancl}.%
}

Because the main function of ``evaluation'' 
---~only to be found in this first one of the 
\boyermooretheoremprovers
~---
is to collect data on which base and step cases
should be chosen later by the \inductionrule, \hskip.1em
the \PURELISPTP\ applies a unique
procedure to stop the unfolding of recursive function definitions: \hskip.2em

A rewrite step with 
an equation defining a recursive function \nlbmath f
is canceled if there is a {\tt CAR} or a {\tt CDR}
in an argument to an occurrence of \nlbmath f in the right-hand side
of the defining equation that
is encountered during the control flow of 
``evaluation\closequotecommasmallextraspace
and if this {\tt CAR} or {\tt CDR} is not removed by the ``evaluation'' of the 
arguments of this occurrence of \nlbmath f
under the current environment updated by 
matching the left-hand side of the equation to the redex.
\hskip.2em
For instance,
``evaluation'' of \mbox{\tt(PLUS (CONS NIL X) Y)} returns
\mbox{\tt(CONS NIL (PLUS X Y))}; \hskip.3em
whereas 
``evaluation'' of \mbox{\tt(PLUS X Y)} 
returns \mbox{\tt(PLUS X Y)} 
and informs the \inductionrule\ that only \mbox{\tt(CDR X)} occurred
in the recursive call during the trial to rewrite with the 
definition of {\tt PLUS}\@. \hskip.2em
 In \nolinebreak general, 
 such occurrences
 indicate which induction hypotheses should be generated by the 
 \inductionrule.\footnotemark
\footnotetext{\label{note no destructor elimination}%
 Actually,
 ``evaluation'' also informs which 
 occurrences of {\tt CAR} or {\tt CDR}
 besides the arguments of recursive occurrences of {\tt PLUS} 
 were permanently introduced during that trial to rewrite. \hskip.2em
 Such occurrences trigger an additional case analysis to be 
 generated by the \inductionrule, mostly as a compensation for
 the omission of the stage of 
 \index{destructor elimination}%
 ``destructor elimination'' in the \PURELISPTP.%
}%
\,\footnotemark\par\begin{sloppypar}\footnotetext{%
 \majorheadroom
 The mechanism for partially enforcing termination of ``evaluation''
 according to this procedure
 is vaguely described in the last paragraph on \litspageref{118} of 
 \moore's \PhDthesis. 
 As this kind of ``evaluation'' is only an intermediate solution
 on the way to more refined control information for the
 \inductionrule\ in later \boyermooretheoremprovers,
 the rough information given here may suffice.%
}%
``Evaluation''
provides a 
crucial 
link between symbolic evaluation and 
the \mbox{induction} rule of \index{induction!explicit}explicit induction.
The question 
``Which case distinction on which variables
  should be used for the induction proof and how should
  the step cases look?''\ \hskip.1em 
is \nolinebreak reduced to the quite different question 
``Where do destructors like {\tt CAR} and 
  {\tt CDR}
  heap up during symbolic evaluation?\closequotefullstopextraspace
This reduction helps to understand by which intermediate steps 
it was possible to develop the most surprising, sophisticated 
\index{recursion analysis}%
recursion analysis of later \boyermooretheoremprovers.%
\index{simplification|)}%
\end{sloppypar}
\subsubsection{Destructor Elimination in the\/ \PURELISPTP}\label
{subsubsection Destructor Elimination in the PURE}
There is no such stage in the \PURELISPTP.\footnote{%
 \majorheadroom
 See, 
 however, 
 \noteref{note no destructor elimination}
 and the discussion of the \PURELISPTP\ in 
 \nlbsectref{subsubsection Destructor Elimination in}.%
}
\subsubsection{(Cross-) Fertilization in the\/ \PURELISPTP}
\label{fertilization in pure lisp theorem prover}%

\index{cross-fertilization|(}%
{\em Fertilization} is just contextual rewriting with an equality,
described before for the ``reduction'' that is part of the
simplification of the \PURELISPTP\ 
(\cfnlb\ \sectref{subsubsection Simplification in the pure}), \hskip.2em
but now with an equation between {\em two non-ground}\/ terms.

The most important case of fertilization is called 
{\em``cross-fertilization''}. \hskip.4em
It occurs very often in step cases of induction proofs of equational theorems,
and we have seen it already in \examref{example second proof of (+3)} 
of \sectref{section example second proof of (+3)}. \hskip.3em

Neither \boyer\ nor \moore\ ever explicitly explained why 
cross-fertilization is ``cross\closequotecommaextraspace
but in
\cite[\p\,142]{moore-1973} \hskip.1em
we read:\notop\halftop
\begin{quote}
``When two equalities are involved and the fertilization was right-side''
[of the induction hypothesis put]
``into left-side''
[of the induction conclusion,]
``or left-side into right-side, 
it is called `cross-fertilization'.''%
\notop\halftop
\end{quote}
``Cross-fertilization'' is actually a term from 
genetics referring to the alignment of 
haploid genetic code from male and female
to a diploid code in the egg cell.
This image may help to recall that 
only that side (\ie\ left- or right-hand side of the equation) \hskip.1em
of the induction conclusion
which was activated by a successful simplification
is further rewritten during cross-fertilization,
namely
{\em everywhere where the same side of the induction hypothesis
 occurs as a redex}\/
---~just like two haploid chromosomes have to start at the same (activated)
sides for successful recombination.
In \cite[\p\,139]{moore-1973}
we find the reason for this:
cross-fertilization 
frequently produces a new goal that is easy to prove because its 
uniform ``genre'' in the sense 
that its subterms uniformly come from just one side of the 
\mbox{original equality.}

Furthermore
---~for getting a sufficiently powerful new induction hypothesis
    in a follow-up induction~---
it is crucial to delete the equation used for rewriting 
(\ie\nolinebreak\ the old induction hypothesis), \hskip.2em
which can be remembered by the fact that 
\mbox{---~in the image~---}
only one (diploid) genetic code remains.

The only noteworthy difference 
between cross-fertilization 
in the \PURELISPTP\
and later \boyermooretheoremprovers\ is that the generalization
that consists in the deletion of the used-up equations 
is done in a hal\fhi earted way:
the resulting formula is equipped with a link to 
the deleted equation.%
\index{cross-fertilization|)}%
\subsubsection{Generalization in the\/ \PURELISPTP}\label
{subsubsection Generalization pure lisp theorem prover}%
\index{generalization|(}%
Generalization in the \PURELISPTP\ 
works as described in 
\sectref{subsection Generalization}. \hskip.3em
The only difference to our presentation there is the following:
Instead of just replacing 
all occurrences of a non-variable subterm \nlbmath t with a new variable
\nlbmath z,
the definition of the top function symbol of \nlbmath t 
is \nolinebreak used to generate the definition of a new predicate \nlbmath p,
such that \app p t holds.
Then the generalization of \nolinebreak\hskip.2em\nlbmath{T[t]} \hskip.1em
becomes
\bigmathnlb{T[z]\antiimplies\app p z}{}
instead of just \nolinebreak\hskip.1em\nlbmaths{T[z]}. \hskip.3em
The version of this automated function synthesis 
actually implemented in the \PURELISPTP\ 
is just able to generate simple type properties,
such as being a number or being a \myboolean\ \nolinebreak value.\footnote
{\label{note failure of synthesis}%
 See \litsectref{3.7} of \cite{moore-1973}. \hskip.3em
 As explained on \litspageref{156\f}\ of \cite{moore-1973}, \hskip.1em
 \boyer\ and \moore\ failed with the trial to improve the implemented version
 of the function synthesis,
 so that it could generate a predicate on a list being ordered
 from a simple sorting-function.%
}

Note that generalization is essential for the \PURELISPTP\
because it does not use lemmas,
and so it cannot build up a more and
more complex theory successively.
It is clear that this limits the complexity of the theorems 
it can prove, 
because a proof can only be successful if the implemented 
non-backtracking heuristics
work out all the way from the theorem down to the most elementary theory.%
\index{generalization|)}%
\subsubsection{Elimination of Irrelevance in the\/ \PURELISPTP}
\notop\halftop\noindent
There is no such stage in the \PURELISPTP.
\subsubsection{Induction in the\/ \PURELISPTP}\label
{subsubsection Induction of pure lisp theorem prover}%
\notop\halftop\noindent
This stage of the \PURELISPTP\ applies the \inductionrule\ of 
\index{induction!explicit}explicit induction as described 
in \sectref{subsection Explicit Induction}.
Induction is tried only after the goal formula has been maximally simplified and
generalized by repeated trips through the waterfall. 
The induction heuristic takes a formula as input 
and returns a conjunction of base and step cases
to which the input formula reduces. \hskip.3em
Contrary to later \boyermooretheoremprovers\ that gather the relevant 
information via induction \index{induction schemes}schemes gleaned by preprocessing recursive definitions,\footnote{%
 \majorheadroom
 \Cfnlb\ \sectref{section example induction schemes constructor style}.%
} \hskip.1em
the \inductionrule\ of the \PURELISPTP\ is based solely on the
information provided by ``evaluation'' 
as described \nolinebreak 
in \nlbsectref{subsubsection Simplification in the pure}.\par

\pagebreak

Instead of trying to describe the general procedure, 
let us just put the \inductionrule\ of the \PURELISPTP\ to test
with two paradigmatic examples.
In these examples we ignore the here irrelevant fact that the \PURELISPTP\ 
actually uses a list representation for the natural numbers. \hskip.2em
The only effect of this is that the 
destructor \nlbmath\psymbol\ takes over the \role\
of the destructor \nolinebreak{\tt CDR}\@.

\halftop
\label{section example induction rule}\begin{example}[Induction Rule in the 
Explicit Induction Proof of \nlbmath{\inpit{\acksymbol 4}}]%
\label{example induction rule}%
\index{induction!explicit}%
\mbox{}\par
\noindent Let us see how the \inductionrule\ of the \PURELISPTP\
proceeds \wrt\ the proof of \nlbmath{\inpit{\acksymbol 4}} \hskip.05em
that we have seen in \examref{example proof of (ack4)} of 
\nlbsectref{section example proof of (ack4)}.
The substitutions \nlbmaths{\xi_1}, \nlbmath{\xi_2} \hskip.1em
computed as instances for the induction conclusion in 
\examref{example induction schemes constructor style} of
\nlbsectref{section example induction schemes constructor style}
suggest an overall case analysis with a base case given by
\math{\{x\mapsto\zeropp\}}, \hskip.3em and two step cases given by
\math{\xi_1=\{x\mapsto\spp{x'}\comma y\mapsto\zeropp\}} \hskip.2em
and \maths{\xi_2=\{x\mapsto\spp{x'}\comma y\mapsto\spp{y'}\}}. \hskip.5em
The \PURELISPTP\ \nolinebreak requires the axioms \inpit{\acksymbol 1},
\inpit{\acksymbol 2},
\inpit{\acksymbol 3} 
to be in 
\index{destructor style}%
destructor instead of 
\index{constructor style}%
constructor style:
\par\halftop\noindent\math{\begin{array}{@{}l@{~~~~~~~}l@{\ =\ }l@{}l@{}}
  \inpit{\acksymbol 1'}
 &\ackpp x y
 &\spp y 
 &\nottight\antiimplies x\tightequal\zeropp
\\\inpit{\acksymbol 2'}
 &\ackpp{x}y
 &\ackpp{\ppp x}{\spp\zeropp}
 &\nottight\antiimplies x\tightnotequal\zeropp
  \nottight\und y\tightequal\zeropp
\\\inpit{\acksymbol 3'}
 &\ackpp{x}{y}
 &\ackpp{\ppp x}{\ackpp{x}{\ppp y}}
 &\nottight\antiimplies x\tightnotequal\zeropp
  \nottight\und y\tightnotequal\zeropp
\\\end{array}}
\par\halftop\noindent
``Evaluation'' does not rewrite the input conjecture with this definition,
but writes a ``fault description'' for the permanent
occurrences of \nlbmath\psymbol\ as arguments of the three occurrences of
\nlbmath\acksymbol\ on the right-hand sides,
essentially consisting of the following three ``pockets'':
\inpit{\ppp x}, 
\inpit{\ppp x, \ppp y}, and
\inpit{\ppp y}, respectively.
Similarly, the pockets gained from the fault descriptions of rewriting
the input conjecture with the definition of \nlbmath\lessymbol\
essentially consists of the pocket \inpit{\ppp y,\ppp{\ackpp x y}}. \hskip.3em
Similar to the non-applicability of the \index{induction templates}induction template
for \lessymbol\ in 
\examref{example Applicable Induction Templates constructor style} of
\nlbsectref{section example Applicable Induction Templates constructor style},
\hskip.2em
this fault description does not suggest any induction because
one of the arguments of \nlbmath\psymbol\ in one of the pockets
is not a variable. \hskip.2em
As this is not the case for the previous fault description,
it suggests the set of all arguments of \nlbmath\psymbol\ 
in all pockets as \index{induction variables}induction variables. \hskip.2em
As this is the only suggestion,
no merging of suggested inductions is required here.

So the \PURELISPTP\ picks the right set of 
\index{induction variables}induction variables.
Nevertheless, it fails to generate appropriate base and step cases,
because the overall case analysis results in two base cases given by 
\nlbmath{\{x\mapsto\zeropp\}} \hskip.2em
and
\nlbmaths{\{y\mapsto\zeropp\}}, \hskip.2em
and a step case given by 
\nlbmaths{\{x\mapsto\spp{x'}\comma y\mapsto\spp{y'}\}}.\footnote{%
 We can see this from 
 a similar case on \litspageref{164} 
 and from the explicit description on the bottom of \litspageref{166}
 in \cite{moore-1973}.%
} \hskip.3em
This turns the first step case 
of the proof of \examref{example proof of (ack4)}
into a base case. \hskip.1em
The \PURELISPTP\ finally fails
(contrary to all other \boyermooretheoremprovers, see
\examrefss{example proof of (ack4)}
{example induction schemes constructor style}
{example induction schemes}) 
with the step case it actually generates:\mediumfootroom
\par\halftop\indent\maths
{\lespp{\spp{y'}}{\ackpp{\spp{x'}}{\spp{y'}}}\tightequal\truepp
\nottight\antiimplies
\lespp
{y'}
{\ackpp{x'}{y'}}
\tightequal\truepp}.
\par\halftop\noindent\mediumheadroom This step case has only one
hypothesis, which is neither of the two we need.
\getittotheright\qed\end{example}\vfill\pagebreak
\label{section example merging}%
\label{section example second proof of (less7)}%
\begin{example}[Proof of \inpit{\lessymbol 7} by Explicit Induction with Merging]\sloppy\par\noindent
\label{example second proof of (less7)}%
\label{example merging}%
\index{induction!explicit}%
Let us write \nlbmath{T(x,y,z)} for \nlbmath{\inpit{\lessymbol 7}} of 
\sectref{subsection Mathematical Induction and the Natural Numbers}. \hskip.3em
From the proof of \inpit{\lessymbol 7} in 
\examref{example first proof of (less7)} of 
\sectref{section example first proof of (less7)}
we can learn the following: \hskip.3em
The proof becomes simpler when we take 
\maths{
T(\zeropp,\spp{y'},\spp{z'})}{} \hskip.1em
as base case
(besides say \nlbmath{T(x,y,\zeropp)} \hskip.1em
 and 
 \nlbmath{T(x,\zeropp,\spp{z'})}), \hskip.3em
instead of any of \hskip.2em\maths{
T(\zeropp,y,\spp{z'})\comma
T(\zeropp,\spp{y'},z)\comma
T(\zeropp,y,z)}. \hskip.4em
The crucial lesson from \examref{example first proof of (less7)},
however,
is that the step case of \index{induction!explicit}explicit induction has to be 
\par\halftop\noindent\LINEmaths{
T(\spp{x'},\spp{y'},\spp{z'})\antiimplies T(x',y',z')}.
\par\halftop\noindent
Note that the \boyermoore\ heuristics for using the 
\inductionrule\ of \index{induction!explicit}explicit induction 
look only one rewrite step ahead,
separately for each occurrence of 
a recursive function in the conjecture.

This means that there is no way for their heuristic
to apply case distinctions
on variables step by step, 
most interesting first,
until finally we end up with an instance of the induction hypothesis
as in \examref{example first proof of (less7)}. \hskip.2em

Nevertheless,
even the \PURELISPTP\ 
manages the pretty hard task of 
suggesting exactly the right step case. \hskip.3em
It requires 
all axioms to be in 
\index{destructor style}%
destructor style,
so instead of 
\inpit{\lessymbol 1},
\inpit{\lessymbol 2},
\inpit{\lessymbol 3},
\hskip.2em
we have to take:
\par\halftop\noindent\math{\begin{array}{@{}l@{~~~~~~~}l@{\ =\ }l@{}l@{}}
  {(\lessymbol 1')}
 &\lespp x y
 &\falsepp
 &\nottight\antiimplies y\tightequal\zeropp
\\{(\lessymbol 2')}
 &\lespp x y
 &\truepp
 &\nottight\antiimplies y\tightnotequal\zeropp
  \nottight\und x\tightequal\zeropp
\\{(\lessymbol 3')}
 &\lespp x y
 &\lespp{\ppp x}{\ppp y}
 &\nottight\antiimplies y\tightnotequal\zeropp
  \nottight\und x\tightnotequal\zeropp
\\\end{array}}
\par\halftop\noindent\smallheadroom
``Evaluation'' does not rewrite any of the occurrences of 
\nlbmath\lessymbol\ in the input conjecture with this definition,
but writes one ``fault description'' for each of these occurrences about 
the permanent occurrences of \nlbmath\psymbol\ as argument 
of the one occurrence of \nlbmath\lessymbol\ on the right-hand sides,
resulting in one ``pocket'' in each fault description,
which essentially consist of 
\hskip.2em
\inpit{\inpit{\ppp z}}, \hskip.3em
\inpit{\inpit{\ppp x, \ppp y}}, \hskip.2em
and \hskip.2em
\inpit{\inpit{\ppp y, \ppp z}}, \hskip.3em
\mbox{respectively}. \hskip.4em
The \nolinebreak\PURELISPTP\ merges these three fault descriptions to the
single one \inpit{\inpit{\ppp x, \ppp y, \ppp z}}, and
so suggests the proper step case indeed,
although it suggests the base case \nlbmath{T(\zeropp,y,z)}
instead of \nlbmaths{T(\zeropp,\spp{y'},\spp{z'})}, \hskip.1em
which requires some extra work, 
but does not result in
a failure.\getittotheright\qed\end{example}
\subsubsection{Conclusion on the\/ \PURELISPTP}\label
{subsubsection Conclusion on the pure lisp theorem prover}%
The \PURELISPTP\ 
establishes the historic breakthrough regarding the 
heuristic automation of inductive theorem proving 
in theories specified by recursive function definitions.

Moreover,
it is the first implementation of a prover for \index{induction!explicit}explicit induction
going beyond most simple 
\structuralinductionindex%
structural inductions 
over \nlbmath\ssymbol\ and \nlbmaths\zeropp.

Furthermore,
the \PURELISPTP\ 
 has most of the stages of the \boyermoorewaterfall\
(\cfnlb\ \figuref{figure waterfall}), \hskip.2em
and these stages occur in the final order and 
with the final overall behavior of throwing the formulas back to the 
center pool after a stage was successful in changing them.

\pagebreak

As we have seen in \examref{example induction rule} of 
\nlbsectref{subsubsection Induction of pure lisp theorem prover}, \hskip.2em
the main weakness of the \PURELISPTP\ is the realization of 
its \inductionrule,
which ignores most of the structure of the recursive calls
in the right-hand sides of recursive function definitions.\footnote{%
 There are indications that the \inductionrule\ 
 of the \PURELISPTP\ had to be implemented in a hurry.
 For instance, on top of \litspageref{168} of \cite{moore-1973}, \hskip.1em
 we read on the \PURELISPTP: 
 ``The case for \math n term induction is much more complicated,
 and is not handled in its full generality by the program.''%
}
In the \PURELISPTP,
all information on this structure
that is
taken into account by the \inductionrule\ 
comes from the fault descriptions of previous applications of 
``evaluation\closequotecomma
which
store only a small part of the 
information that is actually required for finding
the proper instances for 
the eager instantiation of 
induction hypotheses required in \index{induction!explicit}explicit induction.

As a consequence, 
all induction hypotheses and conclusions 
of 
the \PURELISPTP\ 
are instantiations of 
the input formula with mere constructor terms.
Nevertheless,
the \PURELISPTP\ can generate multiple hypotheses for astonishingly
complicated step cases, 
which go far beyond the simple ones typical for 
\structuralinductionindex%
structural induction
over \nlbmath\ssymbol\ and \nlbmaths\zeropp.

Although
the induction stage of the \PURELISPTP\ 
is pretty underdeveloped
compared to the sophisticated 
\index{recursion analysis}%
{\em recursion analysis}\/ 
of the later \boyermooretheoremprovers, \hskip.1em
it somehow contains all essential later ideas in a rudimentary form,
such as 
\index{recursion analysis}%
recursion analysis and the merging of step cases. \hskip.2em
As \nolinebreak 
we have seen in \examref{example second proof of (less7)}, \hskip.1em
the simple merging procedure of the \PURELISPTP\ 
is surprisingly successful.

The \PURELISPTP\ cannot succeed,
however, 
in the rare cases
where a step case has to follow a destructor different
from {\tt CAR} and \nolinebreak{\tt CDR} 
\hskip.1em(such as \dloncesymbol\ in
\nlbsectref{subsection Standard Data Types}), \hskip.2em
or in the more general case that   
the arguments of the recursive calls contain
recursively defined functions at the \index{measured positions}measured positions 
(such \nolinebreak as 
 the \ackermannfunction\ in \examref{example induction rule}).

The weaknesses and provisional procedures of the \PURELISPTP\ 
we \nolinebreak have documented, help to decompose the 
leap
from
nothing to \THM, and
so fulfill our historiographic intention expressed at the
beginning of \nlbsectref{The Pure Lisp Theorem Prover}.

Especially the crucial link between symbolic evaluation and the induction 
rule of \index{induction!explicit}explicit induction described at the end of 
\nlbsectref{subsubsection Simplification in the pure}
may be 
crucial for the success of the entire development of 
\index{recursion analysis}%
recursion analysis 
and \index{induction!explicit}explicit induction.

\vfill\pagebreak
\subsection{\THM}\label
{subsection theorem prover}%
\THMindexstart%
``\THM'' 
is the name used in this article for a release of 
the prover described in \cite{bm}.
Note that the clearness, precision, and detail of the 
natural-language descriptions of heuristics 
in \cite{bm} is 
probably unique.%
\footnotemark\
\hskip.3em
To the best of our knowledge,
there is no similarly broad treatment of heuristics in theorem 
proving, 
at \nolinebreak least not 
in subsequent publications about \boyermooretheoremprovers.
\par\footnotetext{%
 In \cite[\p\,xi]{boyermoore} and \cite[\p\,xv]{boyermooresecondedition}
 we can read about the book \cite{bm}:\begin{quote}``%
 The main purpose of the book was to describe in
 detail how the theorem prover worked, its organization, proof techniques,
 heuristics, etc. One measure of the success of the book is 
 that we know of three
 independent successful efforts to construct the theorem prover from the book.%
 ''\end{quote}%
}%
Except for \ACLTWO, \hskip.15em
\boyer\ and \moore\ never gave names to their theorem pro\-vers.\footnote{%
 The only further exception seems to be \cite[\p\hskip.05em 1]{moore-1975},
 where the \PURELISPTP\ is called 
 ``the \boyermoore\ Pure \LISP\ Theorem Prover\closequotecomma
 because \moore\ wanted to stress that,
 though \boyer\ appears in the references of \cite{moore-1975}
 only in \cite{boyer-moore-1975},
 \boyer\ has had an equal share in contributing to the \PURELISPTP\ 
 right from the start.%
} \hskip.1em
The names \nolinebreak``\THM'' (for ``theorem prover''), \hskip.3em
``\QTHM'' (``quantified \THM''), \hskip.3em
and ``\NQTHM'' (``new quantified \THM'') \hskip.1em
were actually the directory names under which the 
different versions of their theorem provers 
were developed and maintained.\footnote{%
 \majorheadroom
 \Cfnlb\ \cite{boyer-2012}.%
} \hskip.2em
\QTHM\ was never released and its development was 
discontinued soon after the ``quantification'' in \NQTHM\
had turned out to be superior; \hskip.3em
so the name ``\QTHM'' was never used in public. \hskip.1em
Until today,
it seems that ``\THM'' appeared 
in publication only as a mode in \NQTHM,\footnote{%
 \majorheadroom
 For the occurrences of ``{\tt THM}'' 
 in publications,
 and for the exact differences between the
 {\tt THM} and {\tt NQTHM} modes and logics,
 see \litspagefromtoref{256}{257} and 308 
 in \cite{boyermoore}, 
 as well as \litspagefromtoref{303}{305}, 326, 357, and 386 
 in the second edition \cite{boyermooresecondedition}.%
} 
which simulates the release previous to the release of \NQTHM\
(\ie\ before ``quantification'' was introduced) \hskip.1em
with a logic that is a further development of the one described in \cite{bm}.
\hskip.4em
It was \kaufmannname\ \kaufmannlifetime\ \hskip.1em
who started calling the prover ``\NQTHM\closequotecommasmallextraspace
in the second half of the 1980s.\footnote{%
 \majorheadroom
 \Cfnlb\ \cite{boyer-2012}.%
} \hskip.3em
The name ``\NQTHM'' appeared for the first time in publication 
in \cite{boyermoore}, \hskip.2em
namely as the name of a mode in \NQTHM.

In this section we describe the enormous heuristic improvements
documented in \cite{bm} \hskip.1em as compared to \cite{moore-1973} 
(\cfnlb\ \sectref{The Pure Lisp Theorem Prover}). \hskip.3em
In case of the minor differences of the logic described in \cite{bm} \hskip.1em
and of the later released version that is simulated by the {\tt THM}
mode in \NQTHM\ as documented in 
\makeaciteoftwo{boyermoore}{boyermooresecondedition}, \hskip.3em
we try to follow the later descriptions,
partly because of their elegance,
partly because \NQTHM\ is still an available program. \hskip.3em
For this reason, \hskip.1em
we have entitled this section ``\THM''
instead of ``The standard reference on the \boyermoore\ heuristics
\cite{bm}\closequotefullstopnospace

From 1973 to 1981 \boyer\ and \moore\ were 
researchers at Xerox Palo Alto Research Center
(\moore\ only) \hskip.1em
and ---~just a few miles away~---
at SRI International in Menlo Park (CA)
\@. \hskip.3em
From 1981 they were both professors at
\unitexasaustin\ or
scientists at Computational Logic \Inc\ in Austin (TX)\@. \hskip.2em
So they could easily meet and work together. \hskip.3em
And ---~just like the \PURELISPTP~---
the provers \THM\ and \NQTHM\ were again developed and implemented 
exclusively by \boyer\ and \moore.\footnotemark
\footnotetext{%
 In both  
 \cite[\p\,xv]{boyermoore} and \cite[\p\,xix]{boyermooresecondedition} we read:%
 \notop\halftop\begin{quote}``%
 Notwithstanding the contributions of all our friends and supporters, we would
 like to make clear that ours is a very large and complicated system that was
 written entirely by the two of us. Not a single line of \LISP\ in our system was
 written by a third party. Consequently, every bug in it is ours alone. Soundness
 is the most important property of a theorem prover, and we urge any user who
 finds such a bug to report it to us at once.''\end{quote}%
}

In the six years separating \THM\ from the \PURELISPTP,
\boyer\ and \moore\ extended
the system in four important ways that especially affect inductive theorem proving.
The first major extension is
the provision for an arbitrary number of inductive data types, 
where the \PURELISPTP\
supported only {\tt CONS}\@. \hskip.2em
The second is the formal provision of a definition principle with its explicit
termination analysis based on well-founded relations which we discussed
in \nlbsectref{subsection Termination}.
The third major extension is the expansion of the proof techniques
used by the waterfall, notably including the use of previously proved theorems, most often as rewrite
rules via what would come to be called ``contextual rewriting\closequotecomma 
and by which the
\THM\ user can ``guide'' the prover by posing lemmas that the system cannot discover on its own.
The fourth major extension is the synthesis of 
induction 
\index{induction schemes}schemes 
from definition-time termination analysis and
the application and manipulation of those \index{induction schemes}schemes at proof-time to create ``appropriate'' inductions for a given formula, in place of
the \PURELISPTP's less structured reliance on symbolic evaluation. 
We discuss \THM's inductive
data types, waterfall, and induction schemes below.

By means of the new 
\index{shell principle}%
{\em shell principle},\/\footnote{%
 \majorheadroom
 \Cfnlb\ \cite[\p\,37\ff]{bm}.%
} \hskip.1em
it is now possible to define new data types by 
describing the 
\index{shells}%
{\em shell}, \hskip.1em
a constructor with at least one argument, 
each of whose arguments may have a simple type restriction,
and the optional {\em base object}, \hskip.2em
a nullary constructor.\footnote{%
 \majorheadroom 
 Note that this restriction to at most two constructors,
 including exactly one with arguments, 
 is pretty uncomfortable. 
 For instance, it neither admits simple enumeration types
 (such as the \myboolean\ values), \hskip.1em
 nor 
 disjoint unions (\eg, as part of the popular record types with variants, say of 
 \cite{wirth-pascal}).
 Moreover, 
 mutually recursive data types are not possible,
 such as and-or-trees, 
 where each element is a list of or-and-trees,
 and vice versa, 
 as given by the following four constructors:
 \par\noindent\LINEmaths{\begin{array}[t]{c@{\ \ \ \ \ \ \ \ }c}
   \hastype{\mbox{\ident{empty\mbox-or\mbox-tree}}}{\ortreesort};
  &\FUNDEF{\orsymbol}{\andtreesort\comma\ortreesort~}{~\ortreesort};
 \\\hastype{\mbox{\ident{empty\mbox-and\mbox-tree}}}{\andtreesort};
  &\FUNDEF{\andsymbol}{\ortreesort\comma\andtreesort~}{~\andtreesort}.
 \\\end{array}}{}%
} \hskip.2em
Each argument of the 
\index{shells}%
shell can be accessed\footnote{\label{note jargon one}%
 \majorheadroom
 Actually,
 in the jargon of \makeaciteofthree{bm}{boyermoore}{boyermooresecondedition},
 \hskip.1em
 the destructors are called 
 \index{accessor functions}{\em accessor functions}, \hskip.1em
 and the type predicates are called 
 \index{recognizer functions}{\em recognizer functions}.%
} 
by its destructor,
for which a name and a default value 
(for \nolinebreak the sake of totality) \hskip.1em
have to be given in addition. \hskip.2em
The user also has to supply a name for the
predicate that 
that recognizes\arXivfootnotemarkref{note jargon one} 
the objects
of the new data type
(as the logic remains untyped).

{\tt NIL} lost its elementary status and is now 
an element of the 
\index{shells}%
shell {\tt PACK} of symbols.\footnotemark\ \hskip.4em
{\tt T} \nolinebreak and \nolinebreak{\tt F} 
now abbreviate the nullary function calls
{\tt (TRUE)} and {\tt (FALSE)}, respectively,
which are the only \myboolean\ values.
Any argument with \myboolean\ intention besides {\tt F} 
is taken to be {\tt T}
(including {\tt NIL}).
\pagebreak\par\footnotetext{\label{note PACK}%
 There are the following two different declarations for the 
 \index{shells}%
 shell {\tt PACK}:
 \hskip.2em
 In \cite{bm}, \hskip.1em
 the shell {\tt CONS} is defined after the shell {\tt PACK}
 because {\tt NIL} is the default value for the destructors {\tt CAR} and
 {\tt CDR}; \hskip.2em
 moreover, 
 {\tt NIL} is an abbreviation for {\tt (NIL)}, \hskip.2em
 which is the base object of the shell {\tt PACK}\@.\par
 In \makeaciteoftwo{boyermoore}{boyermooresecondedition}, \hskip.1em
 however,
 the shell {\tt PACK} is defined after the shell {\tt CONS}, \hskip.1em
 we \nolinebreak have 
 \ \maths{\mbox{\tt (CAR NIL)} = \mbox{\tt 0}}, \
 the shell {\tt PACK} has no base object, \hskip.1em
 and 
 {\tt NIL} just abbreviates 
 \\\LINEnomath
 {\mbox{\tt (PACK (CONS 78 (CONS 73 (CONS 76 0))))}.}
 \\
 When we discuss the logic of \cite{bm}, \hskip.1em
 we tacitly use the shells {\tt CONS} and {\tt PACK}
 as described in 
 \makeaciteoftwo{boyermoore}{boyermooresecondedition}.%
}
Instead of discussing the 
\index{shell principle}%
shell principle in detail with all its
intricacies resulting from the untyped framework, \hskip.1em
we just present the first two 
\index{shells}%
shells:\begin{enumerate}\item\sloppy 
The shell \ \mbox{\tt (ADD1 X1)} \
of the {\em natural numbers}, \hskip.2em with\noitem\begin{itemize}\item 
type restriction \ 
 \mbox{\tt(NUMBERP X1)},\noitem\item
 base object \ {\tt(ZERO)}, \ abbreviated by 
 \nolinebreak\hskip.2em{\tt0},\noitem\item
 destructor\arXivfootnotemarkref{note jargon one} 
 \ {\tt SUB1} \ with default value \ {\tt0}, \ 
 and\noitem\item
 type predicate\arXivfootnotemarkref{note jargon one} \
 {\tt NUMBERP}.\noitem\end{itemize}
\noitem\item 
The 
\index{shells}%
shell \ \mbox{\tt(CONS X1 X2)} \ of \hskip.15em{\em pairs},\/ \hskip.2em 
with\begin{itemize}\notop\item
destructors \ \begin{tabular}[t]{@{}l@{}}
  {\tt CAR} \ with default value  \ {\tt0}, \ 
\\{\tt CDR} \ with default value  \ {\tt0}, \ and
\\\end{tabular}\noitem\item
type predicate \ {\tt LISTP}.\noitem\end{itemize}\end{enumerate}
According to the 
\index{shell principle}%
shell principle,
these two 
\index{shells}%
shell declarations add 
axioms to the theory, \hskip.1em
which are equivalent to the following ones
:\mediumfootroom
\par\vfill\noindent\LINEnomath
{\small\begin{tabular}{@{\,}l@{\,}||l|l@{}}\#
 & Axioms Generated by Shell {\tt ADD1}
 & Axioms Generated by Shell {\tt CONS}
\\\hline\hline 0.1
 &{\tt(NUMBERP X)\tightequal T\nottight\oder(NUMBERP X)\tightequal F}
 &{\tt(LISTP   X)\tightequal T\nottight\oder(LISTP   X)\tightequal F}
\\\hline 0.2
 &{\tt(NUMBERP  (ADD1 X1))\tightequal T}
 &{\tt(LISTP (CONS X1 X2))\tightequal T}
\\0.3
 &{\tt(NUMBERP 0)\tightequal T}
 &
\\\hline 0.4
 &{\tt(NUMBERP T)\tightequal F}
 &{\tt(LISTP   T)\tightequal F}
\\0.5
 &{\tt(NUMBERP F)\tightequal F}
 &{\tt(LISTP   F)\tightequal F}
\\0.6
 &
 &{\tt(LISTP X)\tightequal F\nottight\oder(NUMBERP X)\tightequal F}
\\\hline 1
 &{\tt(ADD1 (SUB1        X))\tightequal X}
 &{\tt(CONS (CAR X) (CDR X))\tightequal X}
\\
 &\mbox{}\hfill{\tt\nottight\antiimplies X\tightnotequal 0\und 
  (NUMBERP X)\tightequal T}
 &\mbox{}\hfill{\tt\nottight\antiimplies(LISTP   X)\tightequal T}
\\\hline 2
 &{\tt(ADD1 X1)\tightnotequal 0}
\\\hline 3
 &{\tt(SUB1 (ADD1 X1))\tightequal X1}
 &{\tt(CAR (CONS X1 X2))\tightequal X1}
\\
 &\mbox{}\hfill{\tt\nottight\antiimplies(NUMBERP X1)\tightequal T}
 &{\tt(CDR (CONS X1 X2))\tightequal X2}
\\\hline 4
 &{\tt(SUB1 0)\tightequal 0}
\\\hline 5.1
 &{\tt(SUB1 X)\tightequal 0\nottight\antiimplies(NUMBERP X)\tightequal F}
 &{\tt(CAR X)\tightequal 0\nottight\antiimplies (LISTP X)\tightequal F}
\\
 &
 &{\tt(CDR X)\tightequal 0\nottight\antiimplies (LISTP X)\tightequal F}
\\\hline 5.2
 &{\tt(SUB1 (ADD1 X1))\tightequal 0}
\\
 &\mbox{}\hfill{\tt\nottight\antiimplies(NUMBERP X1)\tightequal F}
\\\hline\hline L1\,\footnotemark\headroom
 &{\tt(ADD1 X)\tightequal(ADD1 0)}
\\[-.5ex]
 &\mbox{}\hfill{\tt\nottight\antiimplies (NUMBERP X)\tightequal F}
\\[+.5ex]
  L2\,\footnotemark
 &{\tt(NUMBERP (SUB1 X))\tightequal T}
  \footroom
\\\end{tabular}}%
\vfill\pagebreak
Note that the two occurrences of ``\mbox{\tt(NUMBERP X1)}''
in Axioms 3 and 5.2
are \mbox{exactly}
the ones that
result from the type restriction of {\tt ADD1}. \hskip.2em
Moreover, \hskip.1em
the occurrence of ``\mbox{\tt(NUMBERP X)}'' in Axiom\,0.6
is allocated at the right-hand side \mbox{because the} shell~{\tt ADD1} is 
declared {\em before}\/ the shell~{\tt CONS}\@. \hskip.3em
\addtocounter{footnote}{-1}\footnotetext{%
 Proof of Lemma\,L1 from 0.2, 1--2, 5.2: \hskip.2em
 Under the assumption of \hskip.1em \mbox{\tt(NUMBERP X)\tightequal F},  
 \hskip.2em
 we show \ \mbox
   {\tt(ADD1 X)\tightequal(ADD1 (SUB1 (ADD1 X)))\tightequal(ADD1 0)}. \ \
 The first step is a backward application of the conditional 
 equation\,\,1 via \math{\{\mbox{\tt X}\mapsto\mbox{\tt(ADD1 X)}\}}, \hskip.2em
 where the  condition is fulfilled because of\,\,2 and\,\,0.2\@. \
 The second step is an application of\,\,5.2, \hskip.2em
 where the condition is fulfilled by assumption.%
}\addtocounter{footnote}{1}\footnotetext{%
 \majorheadroom
 Proof of Lemma\,\,L2 from 0.1--0.3, 1--4, 5.1--5.2 \hskip.1em
 by \reductioadabsurdum: \hskip.3em
 \\
 For a counterexample {\tt X}, \hskip.3em
 we get \mbox{\tt(SUB1 X)\tightnotequal 0} \ by\,\,0.3, \hskip.4em
 as well as \mbox{\tt(NUMBERP (SUB1 X))\tightequal F} \ by\,\,0.1\@. \hskip.4em
 From the first we get 
 \ \mbox{\tt X\tightnotequal 0} \ by\,\,4, \hskip.3em
 and 
 \ \mbox{\tt(NUMBERP X)\tightequal T} \ by\,\,5.1 and\,\,0.1\@. \hskip.3em
 Now we get the contradiction
 \mbox{\tt (SUB1 X)\tightequal 
 (SUB1 (ADD1 (SUB1 X)))\tightequal
 (SUB1 (ADD1 0))\tightequal 0}; \
 the first step is a backward application of the conditional 
 equation 1, \hskip.3em
 the second of L1, \hskip.3em
 and the last of\,\,3 (using\,\,0.3)\@.%
}%

Let us discuss the axioms generated by declaration of
the 
\index{shells}%
shell~{\tt ADD1}\@. \hskip.3em
Roughly speaking, \hskip.1em
Axioms\,\,0.1--0.3 \hskip.1em are return-type declarations, \hskip.2em
Axioms\,\,0.4--0.6 \hskip.1em are about disjointness of types, \hskip.2em
Axiom\,1 and Lemma\,\,L2 imply the axiom \nlbmath{\inpit{\nat 1}}
from \sectref{subsection Mathematical Induction and the Natural Numbers},
\hskip.2em
Axioms 2 and 3 imply axioms \nlbmath{\inpit{\nat 2}} 
and \nlbmaths{\inpit{\nat 3}}, 
respectively. 
Axioms 4 and \mbox{5.1--5.2} \hskip.1em 
overspecify {\tt SUB1}. \hskip.3em
Note that Lemma\,L1 is equivalent to 5.2 under
0.2--0.3 and \nolinebreak 1--3\@.%

Analogous to Lemma\,L1, \hskip.1em
every 
\index{shells}%
shell forces each argument not satisfying its type restriction
into behaving like the default object of the argument's destructor.

By contrast,
the arguments of the 
\index{shells}%
shell~{\tt CONS} \hskip.1em
(just as every shell argument without type restriction) \hskip.1em
are not forced like this, \hskip.2em
and so 
---~a clear advantage of the untyped framework~---
even objects of later defined 
\index{shells}%
shells 
(such as {\tt PACK}) \hskip.1em
can be properly paired by
the shell {\tt CONS}\@. \hskip.3em
\mbox{For instance}, \hskip.2em
although 
{\tt NIL} belongs to the 
\index{shells}%
shell {\tt PACK} defined after the shell {\tt CONS}
\hskip.1em (and so 
\hskip.2em\maths
{\mbox{\tt(CDR NIL)}=\mbox{\tt0}}),\arXivfootnotemarkref{note PACK} \hskip.5em
we have \hskip.2em\maths{\mbox{\tt(CAR (CONS NIL NIL))}=\mbox{\tt NIL}}{} \
by Axiom\,3.

Nevertheless, the 
\index{shell principle}%
shell principle also allows us
to declare a 
\index{shells}%
shell 
\\\LINEnomath
{\mbox{\tt(CONSNAT X1 X2)}}
\\of the {\em lists of natural numbers only}
---~similar to the ones of \sectref{subsection Standard Data Types}~---
say,
with a type predicate {\tt LISTNATP}, \hskip.1em
type restrictions \mbox{\tt(NUMBERP X1)}, \mbox{\tt(LISTNATP X2)}, \hskip.1em
base object {\tt (NILNAT)}, \hskip.1em
and
destructors {\tt CARNAT}, {\tt CDRNAT} \hskip.1em
with default values {\tt 0}, {\tt (NILNAT)}, \hskip.2em respectively%
. 

Let us now come to the admissible definitions 
of new functions in \THM\@. \hskip.3em
In \nlbsectref{section Recursive Definitions} \hskip.1em
we \nolinebreak have already discussed 
the {\em definition principle}\/\footnote{%
 \majorheadroom
 \Cfnlb\ \cite[\p\,44\f]{bm}.%
} \hskip.2em
of \THM\ in detail. \hskip.2em
The definition of recursive functions 
has not changed compared
to the \PURELISPTP\
besides that a 
function definition is admissible now only after a termination proof,
which proceeds as explained in \sectref{subsection Termination}.
To this end, \hskip.1em
\THM\nolinebreak\ 
can apply
its additional axiom of the \wellfoundedness\ of 
the irreflexive ordering \nolinebreak{\tt LESSP} \hskip.1em
on the natural numbers,\footnotemark\
and the theorem of the \wellfoundedness\ 
of the lexicographic combination of two \wellfounded\ 
orderings.

\halftop\halftop\noindent
Just as in \sectref{The Pure Lisp Theorem Prover}, \hskip.1em
we
will now 
again follow the \boyermoorewaterfall\ 
(\cfnlb\ \figuref{figure waterfall}) \hskip.1em
and sketch how the stages of the waterfall are realized in \THM\
in comparison to the \PURELISPTP\@.

\subsubsection{Simplification in\/ \THM}\label
{subsubsection simplification in}%
\index{simplification|(}%
We discussed simplification in the \PURELISPTP\
in \sectref{subsubsection Simplification in the pure}.
Simplification in \THM\ is covered in \litchapfromtoref{VI}{IX}
of \cite{bm}, \hskip.1em
and the reader interested in the details is strongly encouraged
to read these 
descriptions of heuristic procedures
for simplification.\footnotetext{\label{note kaufmann non-standard}%
 See \litspageref{52\f}\ of \cite{bm} for the
 informal statement of 
 this axiom on \wellfoundedness\ of {\tt LESSP}.
 \par
 Because \THM\ is able to prove \mbox{\tt(LESSP X (ADD1 X))}, \hskip.2em
 \wellfoundedness\ of {\tt LESSP} would imply
 ---~together with Axiom\,1 and Lemma\,L2~---
 that
 \THM\ admits only the standard model of the natural numbers,
 as explained in \noteref{note pieri standard}.
 \par
 \kaufmannname, however, 
 was so kind and made clear in a private \eMAIL\ communication
 that non-standard models are not excluded, because the statement 
 ``We assume {\tt LESSP} to be a \wellfounded\ relation.'' of 
 \cite[\p\,53]{bm} \hskip.2em
 is actually to be read as the 
 \wellfoundedness\ of the formal definition of 
 \sectref{subsection Well-Foundedness and Termination} \hskip.1em
 with the {\em additional assumption}\/ that the predicate \nlbmath Q
 must be definable in \nolinebreak\THM.\par
 Note that in \pieriindex\pieri's 
 argument on the exclusion of non-standard models
 (as described in \noteref{note pieri standard}), \hskip.1em
 it is not possible to replace the reflexive and transitive closure 
 of the successor relation \nlbmath\ssymbol\ \hskip.1em
 with the \THM-definable predicate
 \bigmathnlb{\displaysetwith{\tt Y}
 {\mbox{\tt(NUMBERP Y)\tightequal T}
  \nottight{\nottight\und}\inpit{\mbox{\tt(LESSP Y X)\tightequal T}
  \nottight\oder\mbox{\tt Y}\tightequal\mbox{\tt X}}}}, 
 because 
 (by the \THM-\englishanalogon\ of 
  axiom \inpit{\lessymbol 2'} of \examref{example second proof of (less7)}
  in \nlbsectref{section example second proof of (less7)}) \hskip.2em
 this predicate will contain {\tt0} as a minimal element
   even for a non-standard natural number \nolinebreak{\tt X}; \hskip.3em
 thus, \hskip.1em
 in non-standard models,
 {\tt LESSP} is a {\em proper}\/ super-relation of the
 reflexive and transitive closure of \nlbmaths\ssymbol.%
}

To compensate for the extra complication of the untyped approach
in \THM, which has a much higher number of 
interesting soft types than the \PURELISPTP,
soft-typing rules are computed for each new function symbol based 
on types that are disjunctions (actually: bit-vectors) \hskip.1em
of the following disjoint types: one for {\tt T}, \hskip.1em
one for \nolinebreak{\tt F}, \hskip.1em
one for each 
\index{shells}%
shell, 
and one for objects not belonging to any of these.\footnote{%
 \majorheadroom
 See \litchapref{VI} in \cite{bm}.%
}
These soft-typing rules are pervasively applied in all stages of the
theorem prover,
which we cannot discuss here in detail. \hskip.3em
Some of these rules can be expressed in the \LISP\ logic language as a theorem
and presented in this form to the human users.
Let us see two examples on this.

\begin{example} \getittotheright
{\em (continuing \examref{example PLUS} of\/ \sectref{section example PLUS})}
\\As \THM\ knows \mbox{\tt(NUMBERP (FIX X))} and \mbox{\tt(NUMBERP (ADD1 X))},
\hskip.3em
it produces the theorem \mbox{\tt(NUMBERP (PLUS X Y))} \hskip.1em
immediately after 
the termination proof for the definition of {\tt PLUS} in \examref{example PLUS}.
\hskip.3em Note that this would neither hold in case of non-termination 
of {\tt PLUS}, \hskip.2em
nor if there were a simple {\tt Y} instead of \mbox{\tt(FIX Y)}
in the definition of {\tt PLUS}\@. \hskip.3em
In the latter case,
\THM\ would only register that the return-type of {\tt PLUS} is among
{\tt NUMBERP} and the types of its second argument \nolinebreak{\tt Y}\@.
\getittotheright\qed\end{example}%
\begin{example}
As \THM\ knows that the type of {\tt APPEND} is among 
{\tt LISTP} and the type of its second argument, \hskip.2em
it produces the theorem \mbox{\tt(LISTP (FLATTEN X))}
immediately after the termination proof for the following definition:
\par\indent{\tt
(FLATTEN~X)~=~(IF~(LISTP~X)
\\\indent\mbox{}~~~~~~~~~~~~~~~~~~(APPEND (FLATTEN (CAR X)) (FLATTEN (CDR X)))
\\\indent\mbox{}~~~~~~~~~~~~~~~~~ (CONS X NIL))}
\getittotheright\qed\end{example}%
\vfill\pagebreak\par\indent
The standard representation of a propositional expression 
has improved from the multifarious 
\LISP\ representation of the \mbox\PURELISPTP\ toward
today's standard of clausal representation.
A {\em clause}\/ is a disjunctive list of literals.
{\em Literals}, however,
deviating from the standard 
of 
being optionally negated atoms,
are just \LISP\ terms here, 
because every \LISP\ function 
can \nolinebreak be \nolinebreak seen \nolinebreak as 
\nolinebreak a \nolinebreak\mbox{predicate.}

This means that the ``water'' of the waterfall now consists of clauses,
and the conjunction of all clauses in the waterfall represents the proof task.

Based on this clausal representation, we find a full-fledged description
of {\em contextual rewriting}\/
in \litchapref{IX} of \cite{bm}, \hskip.1em 
and its applications in \litchapfromtoref{VII}{IX}\@.
This description comes some years before the term ``contextual rewriting''
became popular in automated theorem proving,
and the term does not appear in \cite{bm}. \hskip.3em
It is probably the first description of contextual rewriting in
the history of logic, 
unless one counts the rudimentary contextual rewriting in 
the ``reduction'' of the \PURELISPTP\ as such.\footnote{%
 \Cf\ \sectref{subsubsection Simplification in the pure}.%
}

As indicated before, 
the essential idea of contextual rewriting
is the \mbox{following}: 
While focusing on one literal of a clause for simplification,
we can assume all other literals
---~the {\em context}~--- to be false,
simply because the literal in focus is irrelevant otherwise.
Especially useful are literals that are negated equations,
because they can be used as a ground term-rewrite system.
A \nolinebreak non-equational literal \nlbmath{t} can always be taken to be the 
negated equation 
\nolinebreak\hskip.2em\nlbmaths{\inpit{t\tightnotequal{\tt F}}}. \hskip.3em
The free universal variables of a clause have to be treated as constants during
contextual rewriting.\footnote{%
 This has the advantage that we could take any \wellfounded\ 
 ordering that is total on ground terms and run the terminating ground version
 of a \KNUTHBENDIX\ completion procedure \cite{KB70} 
 for all literals in a clause representation
 that have the form \nlbmath{l_i\tightnotequal r_i}, \hskip.2em
 and replace the literals of this form with the resulting 
 confluent and terminating rewrite system and normalize
 the other literals of the clause with it.
 Note that this transforms a clause into a logically equivalent one.
 None of the \boyermooretheoremprovers\ does this, however.%
}

To bring contextual rewriting to full power,
all occurrences of the function symbol {\tt IF} in the
literals of a clause are expelled from the literals as follows.
If the condition of an \mbox{{\tt IF}-expression} can be simplified
to be definitely false {\tt F} or definitely true
\mbox{(\ie\ non-{\tt F}}, \eg\ if {\tt F} is not set in the
 bit-vector as a potential type), \hskip.2em
then the \mbox{{\tt IF}-expression} is replaced with its respective case.
Otherwise, after the \mbox{{\tt IF}-expression} could not be removed by
those rewrite rules for \nolinebreak{\tt IF}
whose soundness depends on termination,\footnotemark\
it is moved to the top position (outside-in),
by replacing each case with itself in the {\tt IF}'s context,
such that the literal \nlbmath
{C[\mbox{\tt(IF \math{t_0} \math{t_1} \math{t_2})}]} 
is intermediately replaced with
\mbox{\tt(IF \math{t_0} \math{C[t_1]} \math{C[t_2]})}, \hskip.2em
and then this literal splits its clause in two: \hskip.2em
one with the two literals \mbox{\tt(NOT \math{t_0})} and \nlbmath{C[t_1]} 
in place of the old one, \hskip.1em
and one with \math{t_0} and \nlbmath{C[t_2]} instead.\footnotetext{%
 These rewrite rules 
 whose soundness depends on termination are 
 \mbox{\tt(IF X Y Y) = Y};
 \mbox{\tt(IF X X F) = X}; \hskip.1em and
 for \myboolean\ {\tt X}:
 \mbox{\tt(IF X T F) = X}; \hskip.1em
 tested for applicability in the given order.%
} 

\THM\ eagerly removes variables in solved form:
If the variable {\tt X} does not occur in the term \nlbmaths t, \hskip.2em
but the literal 
\nolinebreak\hskip.1em\nlbmath{\inpit{{\tt X}\tightnotequal t}} \hskip.2em
occurs in a clause,
then we can remove that literal after rewriting all occurrences of 
{\tt X} in the clause to \nlbmaths t. \hskip.2em
This removal is a logical equivalence transformation,
because the single remaining occurrence of \hskip.1em {\tt X} 
is implicit\-ly universally quantified and so 
\inpit{{\tt X}\tightnotequal t} \hskip.2em
must be false because 
it \nolinebreak \mbox{implies} \nolinebreak\inpit{t\tightnotequal t}.
Alternatively, the removal can be seen as a resolution step
with the axiom of reflexivity.

\halftop\halftop\indent
It now remains to describe the rewriting with 
function definitions and with lemmas 
tagged 
for rewriting,
where the context of the clause is involved again.

\indent
Non-recursive function definitions are always unfolded by \THM\@. \hskip.1em

\indent
Recursive function definitions are treated in a way very similar 
to that of the \PURELISPTP\@. \hskip.1em
The criteria on the unfolding of a function call of a
recursively defined function \nlbmath f
still depend solely on the terms introduced as arguments 
in the recursive calls of \nlbmath f in the body of \nlbmaths f, \hskip.1em
which are accessed during the simplification of the body.
But now, instead of rejecting the unfolding 
in case of
the presence of new destructor
terms in the simplified recursive calls, rejections are based on whether the
simplified recursive calls contain subterms not occurring elsewhere in the
clause.  That is, an unfolding is approved if all subterms of the simplified
recursive calls already occur in the clause. \hskip.15em
This 
basic {\em occurrence heuristic}\/
is one of the keys to \THM's
success at induction. \hskip.2em
As we 
will 
see, instead of the \PURELISPTP's phrasing of inductive
arguments with ``constructors in the conclusion\closequotecomma
such as \bigmathnlb{P(\spp x))\antiimplies P(x)}, 
\THM\ uses ``destructors in the hypothesis\closequotecomma
such as 
\bigmathnlb{\inpit{P(x)\antiimplies P(\ppp x)}\antiimplies x\tightnotequal 0}. 
\hskip.1em
Thanks to the occurrence heuristic, 
the very presence of a well-chosen induction hypothesis gives the
rewriter ``permission'' to unfold certain recursive functions in the
induction conclusion (which is possible because all function definitions
are in 
\index{destructor style}%
destructor style).

\indent
There are also two less important criteria 
which individually suffice to unblock the unfolding of 
recursive function definitions:\begin{enumerate}\noitem\item
An increase of the number of arguments
of the function to be unfolded
that are constructor ground terms.\noitem\item
A decrease of the number of function symbols 
in the arguments of the function to be unfolded
at the \index{measured positions}measured positions of an \index{induction templates}induction template for that function.
\\
So the clause \\[-1.3ex]\LINEmaths{C[\lespp x{\spp y}]}{}
\par\noindent will be expanded by \inpit{\lessymbol 2'}, \inpit{\lessymbol 3'},
and \nlbmath{\inpit{\psymbol 1}}
into the clauses
\par\noindent\LINEmaths{x\tightnotequal\zeropp\comma C[\truepp]
}{}\mbox{~~~~~~~~~~~~~~~~~~~~\,}
\\and
\\[-1.3ex]\LINEmaths{
x\tightequal\zeropp\comma C[\lespp{\ppp x}y]}{}\mbox{~~~~~~~~\,}
\par\noindent---~even if \ppp x is a newly occurring subterm!~---
because the second argument position of \lessymbol\ 
is such a set of \index{measured positions}measured positions according to 
\examref{example induction template two measured subsets}
of \sectref{section example induction template two measured subsets}.%
\footnote{%
 See \litspageref{118\f}\ of \cite{bm} 
 for the details of the criteria for unblocking the unfolding of 
 function definitions.%
}%
\end{enumerate}

\begin{sloppypar}\halftop\halftop\halftop\noindent
\THM\ is able to exploit previously proved lemmas.  When the user submits a
theorem for proof, 
the user tags it with tokens indicating how it is to be
used in the future {\em{if it is proved}}.  
\THM\ supports four 
non-exclusive 
tags and they
indicate that the lemma is to be used as a rewrite rule, as a rule to
eliminate destructors, 
as a rule to restrict generalizations, or as a
rule to suggest inductions. \hskip.2em 
The paradigm of tagging theorems for use by
certain proof techniques focus the user on developing general ``tactics''
(within a limited framework of \mbox{very abstract control),} \hskip.1em 
while \nolinebreak allowing the user to think mainly
about relevant mathematical truths. \hskip.2em 
This paradigm has been a hallmark of all
\boyermooretheoremprovers\ since \THM\ 
and partially accounts for their reputation
of being ``automatic\closequotefullstopnospace
\par\indent
Rewriting with lemmas that have been proved and then 
tagged for rewriting
\mbox{---~so-called} {\em rewrite lemmas}~---
differs from rewriting with recursive function definitions
mainly in one aspect: There is no need to block them
because the user has tagged them explicitly for rewriting,
and because rewrite lemmas have the form of conditional equations
instead of unconditional ones.
Simplification with lemmas tagged for rewriting and the 
heuristics behind the process are nicely described in \cite{jancl},
where a rewrite lemma is not just tagged for rewriting,
but where the user can also mark the condition literals 
on how they should be dealt with. \hskip.2em
In \THM\ there is no lazy rewriting with rewrite lemmas,
\ie\ no case splits are introduced to be able to apply the lemma.\footnote{%
 \kaufmannname\ and \moorename\ added
 support for ``forcing'' and ``case split'' annotations 
 to \ACLTWO\ in the mid-1990s.%
}
This means that all conditions of the rewrite lemma have to be 
shown to be fulfilled in the current context.
In partial compensation there is a process of backward chaining,
\ie\ the conditions can be shown to be fulfilled by the application
of further conditional rewrite lemmas.
The termination of this backward chaining is achieved
by avoiding the generation of conditions into which the 
previous conditions can be homeomorphically embedded.\footnote{%
 See \litspageref{109\ff}\ of \cite{bm} for the details.%
}
In addition, rewrite lemmas can introduce {\tt IF}-expressions, splitting the rewritten clause
into cases.
There are provisions to instantiate extra variables of conditions
eagerly, 
which is necessary because there are no existential variables.\footnote{%
 See \litspageref{111\f}\ of \cite{bm} for the details.%
}

Some collections of rewrite lemmas can cause 
\THM's rewriter not to terminate.\footnote{%
 Non-termination of rewriting 
 caused the \boyermooretheoremprovers\ 
 to run forever or exhaust the \LISP\ stack or heap ---
 except \ACLTWO, which maintains its own
 user-adjustable stack size
 and gives a coherent error on
 stack overflow without crashing the \LISP\ system. 
 \NQTHM\ introduced special tools to track down the rewriting process
 via the rewrite call stack
 (namely
 {\tt BREAK-REWRITE}, after setting {\tt (MAINTAIN-REWRITE-PATH T)}) \hskip.1em
 and to count the applications of a rewrite rule (namely
 {\tt ACCUMULATED-PERSISTENCE}), \hskip.2em
 so the problematic rules can easily be detected and the user can
 disable them. \hskip.3em
 See \litsectref{12} of 
 \makeaciteoftwo{boyermoore}{boyermooresecondedition} \hskip.1em
 for the details.%
}
\hskip.2em
For \nolinebreak permutative rules like commutativity,
however,
termination is assured by simple term ordering heuristics.\footnotemark
\footnotetext{%
 See \litspageref{104\f}\ of \cite{bm} for the details.%
}%
\index{simplification|)}%
\end{sloppypar}
\subsubsection{Destructor Elimination in\/ \THM}\label
{subsubsection Destructor Elimination in}%
\index{destructor elimination|(}%
We have already seen constructors such as \ssymbol\ (in \THM: {\tt ADD1}) and
\cnssymbol\ ({\tt CONS}) with the destructors 
\psymbol\ ({\tt SUB1}) and \carsymbol\ ({\tt CAR}), \cdrsymbol\ ({\tt CDR}),
respectively.%
\begin{example}[From Constructor to Destructor Style and back]%
\label{example and back}%
\index{destructor style}%
\mbox{}\par\noindent
We have presented several function definitions both in constructor
and in 
\index{destructor style}%
destructor style. 
Let us do careful and generalizable 
equivalence transformations (reverse step justified in parentheses)
starting with
the constructor-style rule \inpit{\acksymbol 3} 
of \sectref{subsection Mathematical Induction and the Natural Numbers}:%
\par
\indent\maths{
\ackpp{\spp x}{\spp y}\tightequal\ackpp x{\ackpp{\spp x}y}}.\nopagebreak
\par
\noindent Introduce (delete) the solved variables 
\math{x'} and \math{y'} for the
constructor terms
\spp x and \spp y occurring on the left-hand side, respectively,
and add (delete) two further conditions by applying the definition
\inpit{\psymbol 1'} 
(\cfnlb\ \sectref{subsection Mathematical Induction and the Natural Numbers})
\hskip.1em
twice.%
\mathcommand\thestandardcondition[2]{\nottight\antiimplies\inparenthesesoplist{
x'\tightequal\spp x\nottight\und
\ppp{x'}\tightequal x#1
\oplistund 
y'\tightequal\spp y\nottight\und
\ppp{y'}\tightequal y#2}}%
\par
\indent\maths{\ackpp{\spp x}{\spp y}\tightequal\ackpp x{\ackpp{\spp x}y}
\thestandardcondition{}{}}.%
\\[+.5ex]\noindent 
Normalize the conclusion with leftmost equations of the condition
  from right to left (left to right).
\\[-.9ex]\indent\maths{
\ackpp{x'}{y'}\tightequal\ackpp x{\ackpp{x'}y}\thestandardcondition{}{}}.
\\[+.5ex]\noindent 
Normalize the conclusion with rightmost equations of the condition
    from right to left (left to right).
\\[-.5ex]\indent\maths{\ackpp{x'}{y'}\tightequal
\ackpp{\ppp{x'}}{\ackpp{x'}{\ppp{y'}}}\thestandardcondition{}{}}.
\par
\noindent
Add (Delete) two conditions by applying axiom \inpit{\nat 2} twice.
\par
\indent\maths{
\ackpp{x'}{y'}\tightequal\ackpp{\ppp{x'}}{\ackpp{x'}{\ppp{y'}}}
\thestandardcondition
{\nottight\und x'\tightnotequal\zeropp}
{\nottight\und y'\tightnotequal\zeropp}}.
\par
\noindent 
Delete (Introduce) the leftmost equations of the condition 
by applying lemma \inpit{\ssymbol 1'} 
(\cfnlb\ \sectref{subsection Mathematical Induction and the Natural Numbers}) 
\hskip.1em
twice,
and delete (introduce) the  
solved variables \math x and \math y for the destructor terms
\ppp{x'} and \ppp{y'} occurring in the left-hand side of the equation
in the conclusion,
respectively.
\par
\indent\maths{\ackpp{x'}{y'}\tightequal\ackpp{\ppp{x'}}{\ackpp{x'}{\ppp{y'}}}
\nottight\antiimplies x'\tightnotequal\zeropp\und y'\tightnotequal\zeropp}.
\par
\noindent Up to renaming of the variables,
this is the 
\index{destructor style}%
destructor-style rule \inpit{\acksymbol 3'} of 
\examref{example induction rule} 
(\cfnlb\ \sectref{section example induction rule}).
\getittotheright\qed\end{example}
\halftop\halftop\halftop\noindent
Our data types are defined inductively over 
constructors.\footnote{%
 Here the term ``inductive'' means the following:
 We start with the empty set and take the smallest
 fixpoint under application of the constructors,
 which contains only finite structures, such as natural numbers and lists.
 Co-inductively over the destructors we would obtain different data types,
 because we \nolinebreak start with 
 the universal class and obtain the greatest fixed point
 under inverse application of the destructors,
 which typically contains infinite structures. \hskip.2em
 For instance, for the unrestricted destructors \carsymbol, \cdrsymbol\
 of the list of natural numbers \app\lists\nat\ of 
 \sectref{subsection Standard Data Types}, \hskip.1em 
 we co-inductively 
 obtain the data type of infinite streams of natural numbers.%
} \hskip.2em
Therefore constructors play the main \role\ in our semantics,
and practice shows that step cases of simple induction proofs work out 
much better with constructors than with the respective destructors,
which are secondary (\ie\ defined) operators in our semantics and 
have a more complicated case analysis in applications.%

\indent
For this reason
---~contrary to the \PURELISPTP~--- 
\THM\ applies destructor elimination to the clauses in the waterfall, \hskip.1em
but not (as in \examref{example and back}) \hskip.1em
to the defining equations. \hskip.2em
This application of \index{destructor elimination}destructor elimination 
  has actually 
two further positive effects%
:\begin{enumerate}\noitem\item
It tends to standardize the representation of a clause
in the sense that the numbers of occurrences of identical subterms 
tend to be increased.\noitem\item
\index{destructor elimination}%
Destructor elimination also brings the subterm property
in line with the sub-structure property; \hskip.2em
\eg, {\tt Y} is both a sub-structure of \mbox{\tt(CONS X Y)}
and a subterm of it, 
whereas \mbox{\tt(CDR Z)} is a sub-structure of \nolinebreak{\tt Z} 
in case of \mbox{\tt(LISTP Z)},
but not a subterm of \nolinebreak{\tt Z}\@.
\noitem\end{enumerate}
Both effects improve the chances that the clause 
passes the follow-up stages of 
\index{cross-fertilization}cross-fertilization and generalization 
with good success.\footnote{%
 See \litspageref{114\ff}\ of \cite{bm} for a nice example
 for the advantage of 
 \index{destructor elimination}%
 destructor elimination for \index{cross-fertilization}cross-fertilization.%
}%
\pagebreak

As noted earlier,
the \PURELISPTP\ does induction using step cases with constructors,
such as \bigmathnlb{\app P{\spp x}\antiimplies \app P x}, 
whereas
\THM\ does induction using step cases with destructors,
such as 
\\[-1.3ex]\mbox{~~~}\LINEmaths{
\inparentheses{\app P x\nottight\antiimplies\app P{\ppp x}}
\nottight\antiimplies x\tightnotequal\zeropp}.
\par\noindent
So 
\index{destructor elimination}%
destructor elimination was not so urgent in 
the \PURELISPTP,
simply because there were fewer destructors around. \hskip.2em
Indeed, 
the stage 
\index{destructor elimination}%
``destructor elimination''
does not exist in the \PURELISPTP\@.

\THM\ does not do induction with constructors because
there are generalized destructors that do not have a 
straightforward constructor (see below), 
and because the \inductionrule\ of 
\index{induction!explicit}explicit induction has to fix
in advance whether the step cases are 
\index{destructor style}%
destructor or 
\index{constructor style}%
constructor style.
So with 
\index{destructor style}%
destructor style in all step cases and in all function definitions,
\index{induction!explicit}explicit induction and recursion 
in \THM\ choose the style that is always applicable.
\index{destructor elimination}%
Destructor elimination then confers the advantages of constructor-style 
proofs when possible.
\begin{example}[A Generalized Destructor Without Constructor]%
\label{example d l once}%
\mbox{}\\
A generalized destructor that does not have a straightforward constructor is the
function \dloncesymbol\ defined in \sectref{subsection Standard Data Types}.
To verify the correctness of a deletion-sort algorithm based on 
\dloncesymbol, a useful step case for an induction proof
is of the form\footnotemark
\\\noindent\LINEmaths{\inparentheses{\app P l\nottight\antiimplies
\app P{\dloncepp{\app{\ident{max}}l}l}}\nottight\antiimplies 
l\tightnotequal\nilpp}.
\\\noindent A constructor version of this 
induction \index{induction schemes}scheme would need something like
an insertion function with an additional free variable 
indicating the position of insertion
---~a complication that further removes the proof
obligations from the algorithm being verified.\getittotheright\qed\end{example}

\halftop\halftop\noindent
Proper destructor functions take only one argument.
The generalized destructor \dloncesymbol\ 
we \nolinebreak have seen in \examref{example d l once}
has actually two arguments; \hskip.2em
the second one is the {\em proper destructor argument}
and the first is a {\em parameter}. \hskip.3em
After the 
\index{destructor elimination}%
elimination of a set of destructors,
the terms at the parameter positions of the destructors 
are typically still present,
whereas all the terms at the positions of the proper destructor arguments 
are removed.%
\par
\begin{example}[Division with Remainder as a pair of Generalized Destructors]%
\label{example quotient and remainder}%
\mbox{}\\
In case of \maths{y\tightnotequal\zeropp}, \hskip.1em
we can construct each natural number \math x in the form of 
\plusppnoparentheses{\timespp q y}r
with \lespp r y\tightequal\truepp.
The related generalized destructors are 
the quotient \divpp x y of \math x by \maths y, \hskip.2em
and its remainder \rempp x y. \hskip.3em
Note that in both functions,
the first argument is the proper destructor argument and
the second the parameter,
which must not be \nlbmaths\zeropp. \hskip.3em
The \role\ that the definition \inpit{\psymbol 1'}
and the lemma \nlbmath{\inpit{\ssymbol 1'}} of 
\nlbsectref{subsection Mathematical Induction and the Natural Numbers} \hskip.1em
play in \examref{example and back}
(and which the definitions \inpit{\carsymbol 1'}, \inpit{\cdrsymbol 1'}
and the lemma \inpit{\cnssymbol 1'} of 
\nlbsectref{subsection Standard Data Types} \hskip.1em
play in the equivalence transformations between 
\index{constructor style}%
constructor and 
\index{destructor style}%
destructor style for lists) \hskip.15em
is here taken by the following lemmas on the generalized destructors 
\divsymbol\ and \remsymbol\
and on the generalized \mbox{constructor
\math{\lambda q,r\stopq\pluspp{\timespp q y}r}:}
\par\noindent\math{\begin{array}{@{}l@{~~~~~~~}l@{\ =\ }l@{}l@{}}
  \inpit{\divsymbol 1'}
 &\divpp x y
 &q
 &\nottight\antiimplies y\tightnotequal\zeropp
  \nottight\und\plusppnoparentheses{\timespp q y}r\tightequal x
  \nottight\und\lespp r y\tightequal\truepp
\\\inpit{\remsymbol 1'}
 &\rempp x y
 &r 
 &\nottight\antiimplies y\tightnotequal\zeropp
  \nottight\und\plusppnoparentheses{\timespp q y}r\tightequal x
  \nottight\und\lespp r y\tightequal\truepp
\\\inpit{\plussymbol 9'}
 &\plusppnoparentheses{\timespp q y}r
 &x
 &\nottight\antiimplies y\tightnotequal\zeropp
  \nottight\und
    q\tightequal\divpp x y
  \nottight\und
    r\tightequal\rempp x y
\\\end{array}}
\par\noindent
If we have a clause with the literal \maths{y\tightequal\zeropp}, \hskip.1em
in which the destructor 
terms \divpp x y or \rempp x y occur,
we can ---~just as in the 
of \examref{example and back}
(reverse direction)~---
introduce the new literals
\math{\divpp x y\tightnotequal q} and \math{\rempp x y\tightnotequal r}
for fresh \maths q, \maths r,
and 
apply lemma \inpit{\plussymbol 9'} to introduce 
the literal \maths{x\tightnotequal\plusppnoparentheses{\timespp q y}r}.
Then we can normalize with the first two literals, 
and afterwards with the third.
Then all occurrences of \divpp x y, \rempp x y, 
and \math x are gone.\footnotemark\
\getittotheright\qed\pagebreak\end{example}
\halftop\halftop\noindent
\addtocounter{footnote}{-1}%
\footnotetext{%
 See \litspageref{143\f}\ of \cite{bm}.%
}%
\addtocounter{footnote}{1}%
\footnotetext{%
 For a nice, but non-trivial example 
 on why proofs tend to work out much easier after this transformation,
 see \litspageref{135\ff} of \cite{bm}.%
}%
To enable the form of 
\index{destructor elimination}%
elimination of generalized destructors
described in \examref{example quotient and remainder},
\THM\ allows the user to tag lemmas of the form
\inpit{\ssymbol 1'},
\inpit{\cnssymbol 1'}, or
\inpit{\plussymbol 9'} as {\em elimination lemmas}\/ to perform
\index{destructor elimination}%
destructor elimination.
In clause representation, this form is in general the following:
The first literal of the clause is of the form \inpit{t^c\tightequal x},
where \math{x} is a variable which does not occur in the 
(generalized)
constructor 
term \nlbmaths{t^{\rm c}}. \hskip.3em
Moreover, 
\math{t^{\rm c}} \nolinebreak contains some distinct variables 
\math{y_0,\ldots,y_n},
which occur only on the left-hand sides of the first literal 
and of 
the last \nlbmath{n\tight+1}~literals of the clause, which 
are of the form
\bigmaths{
\inpit{y_0\tightnotequal t^{\rm d}_0}\comma\ldots\comma
\inpit{y_n\tightnotequal t^{\rm d}_n}},
for distinct (generalized) destructor terms 
\mbox{\maths{t^{\rm d}_0},\ldots,\nlbmaths{t^{\rm d}_n}.}\footnotemark

The idea of application for 
\index{destructor elimination}%
destructor elimination in a given clause
is, of course, the following:
If,
for an instance of the elimination lemma,
the literals not mentioned above 
(\ie\ in the middle of the clause, 
such as \maths{y\tightnotequal\zeropp}{} \hskip.1em in \inpit{+9'})
\hskip.1em
occur in 
the given clause,
and if 
\maths{t^{\rm d}_0}, \ldots, \nlbmath{t^{\rm d}_n} \hskip.3em
occur in the given clause as subterms,
then rewrite all their occurrences 
with 
\inpit{y_0\tightnotequal t^{\rm d}_0}\comma\ldots\comma
\inpit{y_n\tightnotequal t^{\rm d}_n} \hskip.35em
from right to left
and then use the first literal of the elimination lemma from right to left
for further normalization.\footnotemark
\par
\addtocounter{footnote}{-1}\footnotetext{%
 \THM\ adds one more restriction here,
 namely that 
 the generalized destructor terms have to consist of 
 a function symbol applied to a list containing exactly the
 variables of the clause,
 besides \nlbmaths{y_0,\ldots,y_n}.\par
 Moreover, 
 note that 
 \THM\ actually does not use our flattened form of the
 elimination lemmas, 
 but the one that results from replacing each \math{y_i} in the clause with 
 \nlbmaths{t^{\rm d}_i}, and then removing the literal 
 \inpit{y_i\tightnotequal t^{\rm d}_i}. 
 Thus, 
 \THM\ would accept 
 only the non-flattened versions of our elimination lemmas, 
 such as \inpit{\ssymbol 1} instead of \inpit{\ssymbol 1'}
 (\cfnlb\,\sectref{subsection Mathematical Induction and the Natural Numbers}),
 \hskip.1em
 and such as \inpit{\cnssymbol 1} instead of 
 \inpit{\cnssymbol 1'} (\cfnlb\,\sectref{subsection Standard Data Types}).%
}\addtocounter{footnote}{1}\footnotetext{%
 If we add the last literals of the elimination lemma to the given clause,
 use them for contextual rewriting,
 and remove them only if this can be achieved safely
 via application of the definitions of the destructors
 (as we could do in all our examples), \hskip.1em
 then the 
 \index{destructor elimination}%
 elimination of destructors is an equivalence transformation. 
 Destructor elimination in \THM, however, may (over-) generalize the conjecture,
 because 
 these last literals are not present in the non-flattened 
 elimination lemma of \THM\
 and its variables \nlbmath{y_i} are actually introduced in \THM\
 by generalization.
 Thus, instead of trying to delete the last literals of our deletion 
 lemmas safely, 
 \THM\ never adds them.%
}
After a clause enters the 
\index{destructor elimination}%
destructor-elimination stage of \THM, \hskip.1em
its most simple 
(actually: the one defined first) \hskip.1em
destructor that can be eliminated
\hskip.1em
is eliminated, \hskip.1em
and 
\index{destructor elimination}%
destructor elimination is continued
until all destructor terms introduced by 
\index{destructor elimination}%
destructor elimination 
are eliminated if possible.
Then, before further destructors are eliminated,
the resulting clause is returned to the center pool of the waterfall.
So the clause will enter the simplification stage
where the (generalized) constructor introduced by destructor elimination
may be replaced with a (generalized) destructor. \hskip.2em
Then the resulting clauses re-enter the 
\index{destructor elimination}%
destructor-elimination stage,
which may result in infinite looping. \hskip.2em

For example, 
\index{destructor elimination}%
destructor elimination turns the clause
\\[+.5ex]
\mbox{~~~~}\LINEmaths{
x'\tightequal\zeropp\comma C[\lespp{\ppp{x'}}{x'}]\comma C'[\ppp{x'},x']}{}
\\[+.5ex]by the elimination lemma \nlbmath{\inpit{\ssymbol 1}} into the clause
\\[+.5ex]\mbox{}\LINEmaths{
\spp x\tightequal\zeropp\comma C[\lespp x{\spp x}]\comma\ \ C'[x,\spp x]}.
\\[+.5ex]
Then, \hskip.2em
in the simplification stage of the waterfall, \hskip.3em
\lespp x{\spp x} \hskip.2em
is unfolded, \hskip.2em
resulting in the clause
\\[-2.1ex]\mbox{~~~}\LINEmaths{
x\tightequal\zeropp\comma C[\lespp{\ppp x}x]\comma\ \ C'[x,\spp x]}{}
\\[+.9ex]and another one.\footnotemark
\footnotetext{%
 The latter step is given in more detail in the context of 
 the second of the two less important criteria of
 \sectref{subsubsection simplification in} for unblocking the unfolding 
 of \nlbmaths{\lespp x{\spp y}}.%
}%
\pagebreak

Looping could result from 
\index{destructor elimination}%
eliminating the destructor introduced by simplification
(such as it is actually the case for 
 our destructor \nlbmath\psymbol\ in the last clause). \hskip.2em
To \nolinebreak avoid looping, \hskip.1em
before returning a clause to the center pool of the waterfall,
the variables 
introduced by 
\index{destructor elimination}%
\mbox{destructor} elimination 
(such as our variable \nlbmath x) \hskip.1em
are marked.
(Generalized) destructor terms containing 
marked variables
are blocked for further 
\index{destructor elimination}%
destructor elimination.
This marking is removed only when the clause reaches 
the induction stage of the waterfall.\footnote{%
 See \litspageref{139} of \cite{bm}. 
 In general, for more sophisticated details of 
 \index{destructor elimination}%
 destructor elimination in \THM,
 we have to refer the reader to 
 \litchapref{X} of \mbox{\cite{bm}.}%
}%
\index{destructor elimination|)}%
\subsubsection{(Cross-) Fertilization in\/ \THM}
\index{cross-fertilization|(}%
This stage has already been described in 
\sectref{fertilization in pure lisp theorem prover}. \hskip.2em
There is no noticeable difference 
between the \PURELISPTP\ and \THM\ here,
besides some heuristic fine tuning.\footnote{%
 See \litspageref{149} of \cite{bm}.%
}%
\index{cross-fertilization|)}%
\subsubsection{Generalization in\/ \THM}
\index{generalization|(}%
\THM\ adds only one new rule 
to the universally applicable heuristic rules 
for generalization on a term \nlbmath t
mentioned in 
\sectref{subsection Generalization}:\notop\halftop\begin{quote}
``Never generalize on a destructor term \nlbmath{t\,}!'' \hskip.2em
\notop\halftop\end{quote}
This new rule makes sense
in particular after the preceding stage of 
\index{destructor elimination}%
destructor elimination
in the sense that destructors that outlast their elimination 
probably carry some relevant information.
Another reason for not generalizing on destructor terms is that the
clause will enter the center pool in case another generalization is
possible, and then the 
\index{destructor elimination}%
destructor elimination might eliminate the 
destructor term more carefully than generalization would do.\footnote{%
 See \litspageref{156\f}\ of \cite{bm}.%
}

The main improvement of generalization in \THM\
over the \PURELISPTP, however, is the following: \hskip.3em
Suppose again that the term \nlbmath t is to be replaced 
at all its occurrences in the clause \nlbmath{T[t]} with the
fresh variable \nlbmaths z. \hskip.4em
\mbox{Recall} 
that the \PURELISPTP\ restricts the fresh variable 
with a predicate synthesized from
the definition of the top function symbol of the replaced term.  
\THM\ instead restricts
the new variable in two ways.  
Both ways add additional literals to the clause before the
term is replaced by the fresh variable:
\begin{enumerate}\noitem\item
Assuming all literals of the clause \nlbmath{T[t]} to be false 
(\ie\ of type \nolinebreak{\tt F}),
the bit-vector describing the soft type of \math t is computed and
if only one bit is set
(say \nolinebreak the bit expressing {\tt NUMBERP}), \hskip.1em
then, 
for the respective type predicate,
a \nolinebreak new literal is added to the clause
(such as \hskip.1em\mbox{\tt(NOT (NUMBERP \math t))}).\item
The user can tag certain lemmas as 
{\em generalization lemmas}\/; \hskip.3em
such as 
\\\LINEnomath{\mbox{\tt(SORTEDP (SORT X))}}
\\for a sorting function {\tt SORT}; \hskip.3em
and if \mbox{\tt(SORT X)} matches \nlbmath t,
\hskip.2em
the respective \mbox{instance} of \mbox{\tt(NOT (SORTEDP (SORT X)))}
is added to \nolinebreak\hskip.1em\nlbmaths{T[t]}.\footnote{%
 \Cf\ \noteref{note failure of synthesis}.%
} \hskip.5em
In general, 
for the addition of such a literal
\mbox{\tt(NOT \math{t'})}, \hskip.2em
a proper subterm \nlbmath{t'} of a generalization lemma 
must match \nlbmaths t.\footnotemark
\index{generalization|)}%
\end{enumerate}\pagebreak
\footnotetext{%
 Moreover, the literal is actually added to the generalized clause
 only if the top function symbol of \math t
 does no longer occur in the literal 
 after replacing \nlbmath t with \nlbmaths x. \hskip.3em
 This means that, for \nolinebreak a \nolinebreak generalization lemma 
 \mbox{\tt(EQUAL (FLATTEN (GOPHER X)) (FLATTEN X))}, \hskip.3em
 the literal 
 \\\mbox{~~~}\LINEnomath{%
 \mbox{\tt(NOT (EQUAL (FLATTEN (GOPHER \math{t''})) (FLATTEN \math{t''})))}}
 \\is added to \nlbmath{T[t]}
 in case of \math t \nolinebreak being of the form 
 \mbox{\tt(GOPHER \math{t''})}, \hskip.3em
 but not 
 in case of \math t \nolinebreak being of the form 
 \mbox{\tt(FLATTEN \math{t''})} \hskip.1em
 where the first occurrence of {\tt FLATTEN} is not 
 removed by the generalization. 
 See \litspageref{156\f}\ of \cite{bm} for the details.%
}
\subsubsection{Elimination of Irrelevance in\/ \THM}\label
{subsubsection Elimination of Irrelevance in}%
\index{elimination of irrelevance|(}%
\THM\ includes another waterfall stage not in the \mbox{\PURELISPTP,}
the elimination of irrelevant literals.
This is the last transformation 
before we \nolinebreak come to 
``induction\closequotefullstop
Like generalization,
this stage may turn a valid clause into an invalid one. 
The main reason for taking this risk is that
the subsequent heuristic procedures for induction
assume all literals to be relevant:
irrelevant literals may
suggest inappropriate induction \index{induction schemes}schemes 
which may result in a failure of the induction proof.
Moreover, if all literals seem to be irrelevant,
then the goal is probably invalid and
we should not do a costly induction but just fail immediately.\footnote{%
 See \litspageref{160\f}\ of \cite{bm} for a typical example of this.%
}

Let us call two literals {\em connected}\/ if 
there is a variable that occurs in both of them.
Consider the partition of a clause into its equivalence classes \wrt\
the reflexive and transitive closure of connectedness.
If we have more than one equivalence class in a clause,
this is an alarm signal for irrelevance:
if the original clause is valid,
then a sub-clause consisting only of the literals of 
one of these equivalence classes must be valid as well.
This is a consequence of the logical equivalence
of \hskip.2em\bigmathnlb{\forall x\stopq\inpit{A\hskip.07em\tightoder B}}{} 
with \bigmathnlb{A\oder\forall x\stopq B}, \hskip.3em
provided that 
\math x does not occur in \nlbmaths A. \hskip.3em
Then we \nolinebreak should remove one of the irrelevant equivalence classes
after the other from the original clause.
To this end, 
\THM\ has two heuristic tests for irrelevance.\begin{enumerate}\noitem\item{\em
An equivalence class of literals is irrelevant if it does not
contain any properly \mbox{recursive} function symbol.} \hskip.4em
Based on the assumption 
that the previous stages of the \mbox{waterfall} are sufficiently powerful 
to prove clauses composed only of 
\index{constructor function symbols}%
constructor functions (\ie\nolinebreak\ 
\index{shells}%
shells and base objects) \hskip.1em
and functions with explicit 
\mbox{(\ie\ non-recursive)}
\mbox{definitions,} \hskip.2em
the justification for this heuristic test is the following: \hskip.3em
If \nolinebreak the clause of the equivalence class were valid,
then the previous stages of the waterfall should already have established 
the validity of this equivalence class.\item{\em
An equivalence class of literals is irrelevant if it consists of 
only one literal and if this literal is the application 
of a properly recursive function to a list of distinct variables.} \hskip.3em
Based on the assumption 
that the soft typing rules are sufficiently powerful
and that the user has not defined a tautological, 
but tricky predicate,\footnote{%
 This assumption is critical because it often occurs
 that updated program code contains recursive predicates
 that are actually trivially true, but very tricky.
 See \litsectref{3.2} of \cite{wirthcardinal} for 
 such an example. 
 Moreover,
 users sometimes supply such predicates 
 to suggest a particular induction ordering.
 For example,
 if we want to supply the function \sqrtindordsymbol{} 
 of \sectref{subsubsection Conclusion on} to \THM,
 then we have to provide a complete definition,
 typically given by setting \sqrtindordsymbol{} to be {\tt T}
 in all other cases.
 Luckily, such nonsense functions will typically not 
 occur in any proof.%
}
the justification for this heuristic test is the following:
The bit-vector of this literal must contain the 
singleton type of \hskip.1em {\tt F}
(containing only the term {\tt F}, 
 \cfnlb\ \sectref{subsubsection simplification in}); \hskip.3em
otherwise the validity of the literal and the clause would have
been recognized by the stage ``simplification\closequotefullstopextraspace
This means that {\tt F} is most probably a possible value
for some combination of arguments.%
\index{elimination of irrelevance|)}%
\noitem\end{enumerate}
\pagebreak
\subsubsection{Induction in\/ \THM\ as compared to the\/ \PURELISPTP}\label
{subsubsection Induction in}%
\index{recursion analysis|(}%
As we have seen in 
\sectref{subsubsection Induction of pure lisp theorem prover}, \hskip.1em
the 
\index{recursion analysis}%
{\em recursion analysis}\/ 
in the \PURELISPTP\
is only rudimentary.
Indeed,
the whole information on the body of the recursive function
definitions comes out of the 
poor\footnote{%
 See the discussion  
 in \sectref{subsubsection Conclusion on the pure lisp theorem prover}
 on \examref{example induction rule} 
 from \sectref{section example induction rule}.%
} 
feedback of the ``evaluation''
procedure of the simplification stage 
of the \PURELISPTP\@. \hskip.3em
Roughly speaking,
this information consists only in the two facts\begin{enumerate}\noitem\item 
that a destructor symbol occurring as
an argument of the recursive function call in the body 
is not removed by the ``evaluation'' procedure 
in the context of the current goal
and in the local environment, \hskip.1em
and\noitem\item 
that it is not possible to derive 
that this recursive function call is unreachable
in this context and environment.\noitem\end{enumerate}
In \THM, however,
the first part of 
\index{recursion analysis}%
recursion analysis is done at 
{\em definition time}, \hskip.1em
\ie\ at the time the function is defined, 
and applied at {\em proof time}, 
\ie\ at the time the \inductionrule\ produces the base and step cases.
Surprisingly, 
there is no reachability analysis for the recursive calls in
this second part of the 
\index{recursion analysis}%
recursion analysis in \THM\@.
While the information in \lititemref 1 is thoroughly improved 
as compared to the \PURELISPTP, 
the information in \lititemref 2 is partly weaker because
all recursive function calls are assumed to be reachable during
\index{recursion analysis}%
recursion analysis.
The overwhelming success of \THM\ means that the heuristic
decision to abandon reachability analysis in \THM\ was appropriate.\footnote{%
 \majorheadroom
 Note that in most cases the step formula of the reachable cases
 works somehow 
 in \THM,
 as long as no better step case was canceled 
 because of  
 unreachable step cases, which, of course, are trivial to prove,
 simply because their condition is false.
 Moreover, 
 note that, contrary to {\em\descenteinfinie} which can 
 get along
 with 
 the first part of 
 \index{recursion analysis}%
 recursion analysis alone,
 the heuristics of \index{induction!explicit}explicit induction have to guess
 the induction steps eagerly,
 which is always a fault-prone procedure,
 to \nolinebreak be \nolinebreak corrected by additional induction proofs,
 as we have seen in \examref{example second proof of (+3)}
 of \nlbsectref{section example second proof of (+3)}.%
}

\subsubsection{Induction Templates generated by Definition-Time Recursion Analysis}\label
{subsubsection Induction Templates}%
The first part of 
\index{recursion analysis}%
recursion analysis in \THM\ 
consists 
of
a termination analysis of
every recursive function
at the time of its definition.
The system does not only look for one 
termination proof that is sufficient for the admissibility of the 
function definition,
but 
\mbox{---~to be able} 
to generate a plenitude of sound sets of step formulas later~---
actually looks through all termination proofs in a finite
search space
and gathers from them all information required for
justifying the termination of the recursive function definition.
This information will later be used to guarantee 
the soundness 
and improve the feasibility of the step cases 
to be generated by the \inductionrule.

To this end,
\THM\ constructs valid \index{induction templates}induction templates very similar to 
our description in \sectref{subsection Termination}\@.\footnotemark\ \hskip.4em
Let us approach the idea of a valid \index{induction templates}induction template 
with some typical examples,
which are actually the templates for the constructor-style
examples of \nlbsectref{subsection Termination}, \hskip.1em
but now for the 
\index{destructor style}%
destructor-style definitions of \lessymbol\ and \acksymbol,
because only 
\index{destructor style}%
destructor-style definitions are admissible in \THM\@.

\pagebreak
\label{section example induction template two measured subsets}%
\begin{example}%
[Two Induction Templates with different Measured Positions]%
\label{example induction template two measured subsets}%
\mbox{}\\
For the ordering predicate \lessymbol\ 
as defined by \inpit{\lessymbol 1'\mbox{--}3'}
in \examref{example merging} of 
\sectref{section example merging}, \hskip.2em
we get two \index{induction templates}induction templates with the 
sets of \index{measured positions}measured positions \math{\{1\}} and \nlbmaths{\{2\}}, respectively,
both for
the \wellfounded\ ordering \nlbmaths{\lambda x,y\stopq\inpit{\lespp x y
\tightequal\truepp}}. \hskip.5em
The first template has the weight term \nlbmath{\inpit 1} \hskip.1em
and the \index{relational descriptions}relational description 
\\\LINEmaths{
\displayset{\displaytrip
{\lespp x y}
{\{\lespp{\ppp x}{\ppp y}\}}
{\{x\tightnotequal\zeropp\}}}}.
\\The second one has the weight term \nlbmath{\inpit 2} \hskip.1em
and the \index{relational descriptions}relational description 
\\\phantom\qed\LINEmaths{
\displayset{\displaytrip{\lespp x y}
{\{\lespp{\ppp x}{\ppp y}\}}
{\{y\tightnotequal\zeropp\}}}}.\qed
\footnotetext{%
 Those parts of the condition of the equation that 
 contain the new function symbol \nlbmath{f} must be ignored
 in the case conditions of the \index{induction templates}induction template
 because the definition of the function \nlbmath f is admitted in \THM\ only 
 {\em after}\/ it has passed the termination proof.\par
 That \THM\ ignores the governing conditions that contain the new 
 function symbol \nlbmath{f} \hskip.1em is described in the \nth 2\,paragraph
 on \litspageref{165} of \cite{bm}. \hskip.3em
 Moreover, 
 an example for this is the definition of {\tt OCCUR} on 
 \litspageref{166} of \cite{bm}.
 \par
 After one successful termination proof, however,
 the function can be admitted in \THM, \hskip.1em
 and then these conditions could actually be
 admitted in the templates.
 So the actual reason 
 why \THM\ ignores these conditions in the templates is that 
 it generates the templates with the help of previously proved
 {\em induction lemmas}, \hskip.1em
 which, of course, cannot contain the new function \nolinebreak yet.%
}%
\end{example}%
\begin{example}[One Induction Template with Two Measured Positions]%
\label{example non-singleton measured subset}%
\mbox{}\\
For the \ackermannfunction\ \acksymbol\ 
as defined by \inpit{\acksymbol 1'\mbox{--}3'}
in \examref{example induction rule} of 
\sectref{section example induction rule}, \hskip.2em
we \nolinebreak get only one appropriate \index{induction templates}induction template.
The set of its \index{measured positions}measured positions is \nlbmaths{\{1,2\}}, \
because of the weight function \cnspp{\inpit 1}{\cnspp{\inpit 2}\nilpp} 
\hskip.2em
(in \THM\ actually: \mbox{\tt(CONS \math x \math y)}) \hskip.1em
in the \wellfounded\ lexicographic ordering 
\\\LINEmaths{\lambda l,k\stopq\inpit{\lexlimlespp l k{\spp{\spp{\spp\zeropp}}}
\tightequal\truepp}}. 
\\The \index{relational descriptions}relational description has two elements: \hskip.3em
For the equation \inpit{\acksymbol 2'} \hskip.1em we get
\\\LINEmaths{\displaytrip
{\ackpp x y}{\{\ackpp{\ppp x}{\spp\zeropp}\}}{\{x\tightnotequal\zeropp\}}},
\\and for the equation \inpit{\acksymbol 3'} \hskip.1em we get
\\\phantom\qed\LINEmaths{\displaytrip
{\ackpp x y}
{\{\ackpp{x}{\ppp y}\comma\ackpp{\ppp x}{\ackpp{x}{\ppp y}}\}}
{\{x\tightnotequal\zeropp\comma y\tightnotequal\zeropp\}}}.\qed\end{example}

\halftop\halftop\noindent
To find valid \index{induction templates}induction templates automatically by exhaustive search, \hskip.1em
\THM\ allows the user to tag certain theorems as
{\em``induction lemmas\closequotefullstop}
An induction lemma consists of the application
of a \wellfounded\ relation to two terms with the same
top function symbol \nlbmaths w, 
playing the \role\ of the weight term; 
\hskip.2em
{plus} a condition without extra variables,
which is used to generate the case conditions of the \index{induction templates}induction template.
Moreover, 
the arguments of the application of 
\nlbmath{w} occurring as the second argument of the \wellfounded\ relation 
must be distinct variables in \nolinebreak\THM,
mirroring the left-hand side of its function definitions in
\index{destructor style}%
destructor style.

Certain induction lemmas are generated
with each 
\index{shells}%
shell declaration.
Such an induction lemma generated for the shell {\tt ADD1}, 
which is roughly
\par\noindent\LINEnomath
{\mbox
{\tt(LESSP (COUNT (SUB1 X)) (COUNT X)) \antiimplies\ (NOT (ZEROP X))},}
\par\noindent 
suffices 
for generating the two templates of 
\examref{example induction template two measured subsets}. \hskip.3em
Note that {\tt COUNT}, \hskip.1em
playing the \role\ of \nlbmath{w} here, \hskip.2em 
is a special function in \THM, \hskip.1em
which is generically extended by every 
\index{shells}%
shell declaration
in an object-oriented style for the elements of the new shell. \hskip.2em
On the natural numbers here, \hskip.2em
{\tt COUNT} is the identity.
On other 
\index{shells}%
shells, \hskip.2em
{\tt COUNT} is defined similar to our function 
\sizesymbol\ from \sectref{subsection Standard Data Types}.\footnotemark
\pagebreak
\footnotetext{%
 For more details on the 
 \index{recursion analysis}%
 recursion analysis a definition time in \THM, \hskip.1em
 see \litspageref{180\ff}\ of \cite{bm}.
}%
\subsubsection{Proof-Time Recursion Analysis in\/ \THM}\label
{subsubsection Proof-Time Recursion Analysis in}
\halftop\halftop\noindent
The \inductionrule\ uses the information from the \index{induction templates}induction templates
as follows: For each recursive function 
occurring in the input formula,
all {\em applicable}\/ \index{induction templates}induction templates are retrieved and
turned into {\em induction schemes} as described in 
\nlbsectref{subsection Induction Schemes}. \hskip.2em
Any induction \index{induction schemes}scheme 
that is {\em subsumed}\/ \hskip.1em 
by another one is deleted after 
adding its \index{hitting ratio}hitting ratio to the one of the other. \hskip.2em
The remaining \index{induction schemes}schemes are {\em merged}\/ 
into new ones with a higher 
\index{hitting ratio}hitting ratio,
and finally,
after the {\em flawed}\/ \index{induction schemes}schemes are deleted,  
the \index{induction schemes}scheme with the highest\footnote{%
 \majorheadroom
 This part of the heuristics is made perspicuous
 in \examref{example induction schemes constructor style}
 of \sectref{section example induction schemes constructor style}.
} 
\index{hitting ratio}hitting ratio will be used by the \inductionrule\ 
to generate the base and step cases.
\halftop\begin{example}[Applicable Induction Templates]\label
{example Applicable Induction Templates}%
\hfill
\\Let us consider the conjecture \inpit{\acksymbol 4} from
\sectref{subsection Mathematical Induction and the Natural Numbers}.
From the three \index{induction templates}induction templates of \examrefs
{example induction template two measured subsets}
{example non-singleton measured subset},
only the second one of \examref{example induction template two measured subsets}
is not applicable because the second position of \lessymbol\
(which is the only \index{measured positions}measured position of that template) \hskip.1em
is \index{changeable positions}changeable, but filled in \inpit{\acksymbol 4} \hskip.1em
by the non-variable \nlbmaths{\ackpp x y}.%
\getittotheright\qed\end{example}

\halftop\halftop\halftop\noindent
From the 
\index{destructor style}%
destructor-style definitions 
\inpit{\lessymbol 1'\mbox{--}3'} (\cfnlb\ \examref{example merging})
and \inpit{\acksymbol 1'\mbox{--}3'} (\cfnlb\ \examref{example induction rule}),
\hskip.2em
we have generated two 
\index{induction templates}induction templates 
applicable to 
\\[+.3ex]\noindent\math{\begin{array}{@{}l@{~~~~~~~}l@{\ =\ }l}
  (\acksymbol 4)
 &\lespp y{\ackpp x y}
 &\truepp
\\\end{array}}%
\\[+.3ex]They yield the two induction \index{induction schemes}schemes of 
\examref{example induction schemes}. 
See also \examref{example induction schemes constructor style}
for the single induction \index{induction schemes}scheme for the constructor-style
definitions \inpit{\lessymbol 1\mbox{--}3} 
and \inpit{\acksymbol 1\mbox{--}3}.
\halftop
\begin{example}[Induction Schemes]\label{example induction schemes}%
\mbox{}\par\noindent
The \index{induction templates}induction template for \lessymbol\ of 
\examref{example induction template two measured subsets}
that is applicable to \inpit{\acksymbol 4} 
according to 
\examref{example Applicable Induction Templates} \hskip.1em
and
whose \index{relational descriptions}relational description 
contains only the triple
\\[+.5ex]
\LINEmaths{\displaytrip
{\lespp x y}
{\{\lespp{\ppp x}{\ppp y}\}}
{\{x\tightnotequal\zeropp\}}}{}
\\[+.5ex]
yields the induction \index{induction schemes}scheme with 
\index{position sets}position set \nlbmath{\{1.1\}} \hskip.1em
(\ie\ left-hand side of first literal in \nlbmath{\inpit{\acksymbol 4}});
\hskip.4em
the \index{step-case descriptions}step-case description is 
\hskip.3em\displayset{\!\!\!\!\displaytrip
{\!\!\!\domres\id{\{x,y\}}}
{\{\mu_1\}}
{\{y\tightnotequal\zeropp\}\!\!\!}\!\!\!\!}, \hskip.5em
where \nolinebreak\hskip.1em\maths{\mu_1=\{
x\tight\mapsto x}, \maths{y\tight\mapsto\ppp y
\}}; \hskip.5em
the set of \index{induction variables}induction variables 
is \nlbmath{\{y\}}; \hskip.3em
and the \index{hitting ratio}hitting ratio \nolinebreak is \nlbmaths{{1\over 2}}. \hskip.3em
\par
This can be seen as follows: 
The substitution called \nlbmath\xi\ in the
discussion of \nlbsectref{subsection Induction Schemes} \hskip.1em
can be chosen to be the identity substitution 
\nlbmath{\domres\id{\{x,y\}}}
on \nlbmath{\{x,y\}} \hskip.1em
because 
the first element of the triple does not contain any constructors.
This is always the case for 
\index{induction templates}induction templates for 
\index{destructor style}%
destructor-style definitions 
such as \nlbmaths{\inpit{\lessymbol 1'\mbox{--}3'}}. \hskip.3em
The substitution called \nlbmath\sigma\
in \sectref{subsection Induction Schemes}
(which has to match the first element of the triple to the term 
\nlbmaths{\inpit{\acksymbol 4}/1.1}, \hskip.2em
\ie\ the term at the 
position\,\math{1.1} in \inpit{\acksymbol 4})
is \nlbmaths{\sigma=\{x\tight\mapsto y\comma y\tight\mapsto
\ackpp x y\}}. \hskip.5em
So the constraints for \nlbmath{\mu_1} 
(which tries to match 
\math{\inpit{\acksymbol 4}/1.1} to the \math\sigma-instance of the
second element of the triple)
are: \bigmaths{y\mu_1\tightequal\ppp y}{}
for the first (measured) position of \nlbmaths\lessymbol; \hskip.5em
and
\bigmaths{\ackpp x y\mu_1\tightequal\ppp{\ackpp x y}}{}
for the second (unmeasured) position,
which cannot be achieved and is skipped. \hskip.2em
This results in a \index{hitting ratio}hitting ratio of only \nlbmaths{{1\over 2}}. \hskip.3em
The single \index{measured positions}measured position\,\nlbmath 1 of the 
\index{induction templates}induction template 
results in the induction 
variable \nlbmaths{\inpit{\acksymbol 4}/1.1.1=y}.
\pagebreak
\par
The template for \acksymbol\ of \examref
{example non-singleton measured subset}
yields an induction \index{induction schemes}scheme with the 
\index{position sets}position set 
\nlbmaths{\{1.1.2\}}, \hskip.2em
and the set of \index{induction variables}induction variables \nlbmaths{\{x,y\}}. \hskip.3em
The triple
\\
\LINEmaths{\displaytrip
{\ackpp x y}
{\{\ackpp{\ppp x}{\spp\zeropp}\}}
{\{x\tightnotequal\zeropp\}}}{}
\\
(generated by the equation \inpit{\acksymbol 2'}) \hskip.2em
is replaced with
\hskip.3em
\displaytrip{\!\!\domres\id{\{x,y\}}}
{\{\mu'_{1,1}\}}
{\{x\tightnotequal\zeropp\}\!\!}, \hskip.3em
where \maths{\mu'_{1,1}=
\{x\tight\mapsto\ppp x\comma y\tight\mapsto\spp\zeropp\}}. 
\hskip.3em
The triple
\\
\LINEmath{
\displaytrip
{\ackpp x y}
{\{\ackpp{x}{\ppp y}\comma\ackpp{\ppp x}{\ackpp{x}{\ppp y}}\}}
{\{x\tightnotequal\zeropp\comma y\tightnotequal\zeropp\}}}{}
\\
(generated by \inpit{\acksymbol 3'}) \hskip.1em
is replaced with
\bigmaths
{\displaytrip{\!\!\domres\id{\{x,y\}}}
{\{\mu'_{2,1}\comma\mu'_{2,2}\}}
{\{x\tightnotequal\zeropp\comma y\tightnotequal\zeropp\}\!\!}},
where \hskip.2em\maths{\mu'_{2,1}=
\{x\tight\mapsto x\comma y\tight\mapsto\ppp y\}}, \hskip.2em
and \hskip.2em\nlbmaths{\mu'_{2,2}=
\{x\tight\mapsto\ppp x\comma y\tight\mapsto\ackpp{x}{\ppp y}\}}. \hskip.3em
\par
This can be seen as follows:
The substitution called \nlbmath\sigma\
in the above discussion is
{\domres\id{\{x,y\}}} \hskip.1em in both cases, \hskip.3em
and so the constraints for the (measured) positions are
\bigmaths{x\mu'_{1,1}\boldequal\ppp x},
\maths{y\mu'_{1,1}\boldequal\spp\zeropp}; \hskip.3em
\bigmaths{x\mu'_{2,1}\boldequal x},
\maths{y\mu'_{2,1}\boldequal\ppp y}; \hskip.3em
\bigmaths{x\mu'_{2,2}\boldequal\ppp x},
\maths{y\mu'_{2,2}\boldequal\ackpp{x}{\ppp y}}.
\par
As all six constraints are satisfied,
the \index{hitting ratio}hitting ratio is \hskip.2em\nlbmaths{{6\over 6}=1}.
\getittotheright\qed\end{example}
\par\halftop\halftop\noindent 
An induction \index{induction schemes}scheme that is either {\em subsumed by}\/ 
or {\em merged into}\/ another induction \index{induction schemes}scheme 
adds its \index{hitting ratio}hitting ratio and sets of 
positions and \index{induction variables}induction variables to those of the other's, 
respectively, and then it is deleted.

The most important case of subsumption 
are \index{induction schemes}schemes that are
identical except for their \index{position sets}position sets,
where 
---~no matter which \index{induction schemes}scheme is deleted~---
the result is the same.
The more general case 
of proper subsumption
occurs when the subsumer provides the essential structure
of the subsumee, but not vice versa.

Merging and proper subsumption of \index{induction schemes}schemes 
---~seen as binary algebraic operations~---
are not commutative, however, because the second argument 
inherits the \wellfoundedness\ guarantee alone and
somehow 
absorbs the first argument, and so the result for swapped
arguments is often undefined.

More precisely, subsumption is given if
the \index{step-case descriptions}step-case description of the first induction \index{induction schemes}scheme
can be injectively mapped to 
the \index{step-case descriptions}step-case description of the second one,
such that 
(using the notation of \nlbsectref
{section example induction schemes constructor style}
and \examref{example induction schemes}), \hskip.2em
for each step \nolinebreak 
case \trip\id{\setwith{\mu_j}{j\tightin J}}C \hskip.2em
mapped to \hskip.1em
\trip\id{\setwith{\mu'_j}{j\tightin J\tight\uplus J'}}{C'}, \hskip.2em
we have \bigmaths{C\subseteq C'},
and the set of substitutions \mbox{\math{\setwith{\mu_j}{j\tightin J}}{}}
can be injectively\footnote{\label{note injectivity 1}%
 From a logical viewpoint, 
 it is not clear why this second injectivity requirement 
 is found here, 
 just as in different (but equivalent) form 
 in \cite[\p\,191, \nth 1\,paragraph]{bm}. \hskip.3em
 (The first injectivity requirement
  may prevent us from choosing an induction ordering
  that is too small,
  \cfnlb\ \sectref{subsubsection Conclusion on}.) \hskip.4em
 An omission of the second requirement would just 
 admit a term of the subsumer to have multiple subterms of the 
 subsumee, which seems reasonable.
 Nevertheless, as pointed out in \sectref{subsubsection Conclusion on},
 only practical testing of the heuristics is what matters here. \hskip.3em
 \mbox{See also \noteref{note injectivity 2}.}%
} 
mapped to \mbox{\setwith{\mu'_j}{j\tightin J\uplus J'}}
(\wrog\ say \math{\mu_i} to \math{\mu'_i} for \math{i\tightin J}), \hskip.2em
such that,  \hskip.1em
\mbox{for each \math{j\in J} and} \math{x\in\DOM{\mu_j}}: \hskip.4em
\mbox{\math{x\tightin\DOM{\mu'_j}};} \hskip.5em
\mbox{\math{x\mu_j\boldequal x}} \hskip.3em
implies \hskip.2em\math{x\mu'_j\boldequal x}; \hskip.4em
and \math{x\mu_j} is a subterm \nolinebreak of \nlbmaths{x\mu'_j}.%
\notop\halftop\begin{example}[Subsumption of Induction Schemes]\label
{example Subsumption}\mbox{}\majorheadroom\\
In \examref{example induction schemes}, \hskip.1em
the induction \index{induction schemes}scheme for \lessymbol\ 
is subsumed by the induction \index{induction schemes}scheme for \acksymbol, \hskip.1em
because we can map the only element of the \index{step-case descriptions}step-case description of 
the former to the second element of the \index{step-case descriptions}step-case description of latter:
the case condition \bigmaths{\{y\tightnotequal\zeropp\}}{} 
is a subset
of the case condition \bigmaths{\{x\tightnotequal\zeropp\comma
y\tightnotequal\zeropp\}},
and we have \bigmaths{\mu_1=\mu'_{2,1}}.
So the former \index{induction schemes}scheme is deleted and the \index{induction schemes}scheme
for \acksymbol\ is updated to have the \index{position sets}position set 
\nlbmaths{\{1.1\comma 1.1.2\}}{} \hskip.1em
\hskip.1em and the \index{hitting ratio}hitting ratio 
\nlbmaths{{3\over 2}}.\vspace*{-1.099ex}
\getittotheright\qed\end{example}

\vfill\pagebreak

\halftop\noindent
In \examref{example merging} of 
\sectref{subsubsection Induction of pure lisp theorem prover} 
we have already seen a rudimentary, but pretty successful kind of 
{\em merging of suggested step cases}\/ in the \PURELISPTP\@. \hskip.3em
As \nolinebreak\THM\ additionally has induction \index{induction schemes}schemes, \hskip.1em
it applies a more sophisticated {\em merging of induction schemes}\/ instead.

Two substitutions \math{\mu_1}
and \math{\mu_2} are {\em\opt{non-trivially} mergeable}\/ if
\mbox{\math{x\mu_1\tightequal x\mu_2}}
for each \math{x\in\DOM{\mu_1}\cap\DOM{\mu_2}} \hskip.2em
\opt{and there is a \math{y\in\DOM{\mu_1}\cap\DOM{\mu_2}} with
\math{y\mu_1\tightnotequal y}}.

Two triples \trip{\domres\id{V_1}}{A_1}{C_1}
and \trip{\domres\id{V_2}}{A_2}{C_2}
of two \index{step-case descriptions}step-case descriptions of two induction \index{induction schemes}schemes,
each with 
domain \math{V_k=\DOM{\mu_k}} for all \math{\mu_k\in A_k}
\mbox{(for \math{k\in\{1,2\}}),} \hskip.2em
are {\em\opt{non-trivially} mergeable}\/ if
for each \math{\mu_1\in A_1} there is a \math{\mu_2\in A_2}
such that \math{\mu_1} and \math{\mu_2} are 
\opt{non-trivially} mergeable.
The result of their merging is \hskip.1em
\displaytrip{\domres\id{V_1\cup V_2}}{m(A_1,A_2)}{C_1\tightcup C_2}, \hskip.4em
where \math{m(A_1,A_2)} is the set containing all
substitutions \math{\mu_1\cup\mu_2} with 
\math{\mu_1\in A_1} and \math{\mu_2\in A_2} 
such that \math{\mu_1} and \math{\mu_2}
are mergeable
\aswellas\ all substitutions \math{\domres\id{V_1\setminus V_2}\cup\mu_2}
with \math{\mu_2\in A_2} for which there is no substitution
\math{\mu_1\in A_1} such that \math{\mu_1} and \math{\mu_2}
are mergeable.
\\\indent
Two induction \index{induction schemes}schemes are {\em mergeable}\/
if the \index{step-case descriptions}step-case description of the first induction \index{induction schemes}scheme
can be injectively\footnotemark\
mapped to 
the \index{step-case descriptions}step-case description of the second one,
such that each argument and its image are non-trivially mergeable.
The \index{step-case descriptions}step-case description of the induction \index{induction schemes}scheme
that results from {\em merging the first induction scheme into the second}\/ 
contains the merging of all mergeable triples of the 
\index{step-case descriptions}step-case descriptions of first and second induction \index{induction schemes}scheme,
respectively.%
\par\halftop\halftop\halftop\indent
Finally, 
we have to describe what it means that an induction \index{induction schemes}scheme is {\em flawed}.
This simply is the case if
---~after merging is completed~---
the intersection of its \index{induction variables}induction variables with the (common) domain
of the substitutions of the \index{step-case descriptions}step-case description of another remaining
induction \index{induction schemes}scheme is non-empty.

If an induction \index{induction schemes}scheme is flawed 
by 
another one that cannot be merged
with \nolinebreak it, \hskip.1em
this indicates that
an induction on it 
will probably result in a permanent clash between 
the induction conclusion and the available induction hypotheses 
at some occurrences of the 
\index{induction variables}induction variables.\footnotemark%
\figuremerging{h}\pagebreak
\label{section example merging 2}%
\begin{example}[Merging and Flawedness of Induction Schemes]\label
{example merging 2}\mbox{}%
\par\noindent
\addtocounter{footnote}{-1}\footnotetext{\label{note injectivity 2}%
 From a logical viewpoint, 
 it is again not clear why an injectivity requirement 
 is found here, 
 just as in different (but equivalent) form 
 in \cite[\p\,193, \nth 1\,paragraph]{bm}. \hskip.3em
 An \nolinebreak \mbox{omission} of the injectivity requirement would admit 
 to define merging as a commutative \mbox{associative} operation.
 Nevertheless, \hskip.1em
 as pointed out in \sectref{subsubsection Conclusion on}, 
 \hskip.2em
 only practical testing of the heuristics is what matters here. \hskip.3em
 See also \noteref{note injectivity 1}.%
}\addtocounter{footnote}{1}\footnotetext{%
 \majorheadroom
 See \litspageref{194\f}\ of \cite{bm}
 for a short further discussion and a nice example.%
}%
Let us reconsider merging in the proof of lemma \inpit{\lessymbol 7} 
\wrt\ the definition of \nlbmath\lessymbol\ via
\nlbmaths{\inpit{\lessymbol 1'\mbox{--}3'}}, \hskip.1em
just as we did in \examref{example merging}. \hskip.3em
Let us abbreviate \mbox{\math{p\tightequal\truepp}{}} \hskip.1em
with \maths p, \hskip.1em
just as in our very first proof of lemma \inpit{\lessymbol 7} in
\examref{example first proof of (less7)}, \hskip.1em
and also following the \LISP\ style of \THM\@. \hskip.3em
Simplification reduces \inpit{\lessymbol 7} first to the clause
\\[+.5ex]\noindent\math{\begin{array}{@{}l@{~~~~~~}l}
  \inpit{\lessymbol 7'}
 &\lespp{x}{\ppp z}
  \comma
  \neg\lespp x y
  \comma
  \neg\lespp y z
  \comma
  z\tightequal\zeropp
\\\end{array}}
\\[+.5ex]\noindent
Then the \boyermoorewaterfall\ sends this clause through 
three rounds of reduction
between 
\index{destructor elimination}%
destructor elimination and 
simplification as
discussed at the end of \nlbsectref{subsubsection Destructor Elimination in},
\hskip.2em
finally returning again to \inpit{\lessymbol 7'}, \hskip.1em
but now with all its 
variables marked as being introduced by 
\index{destructor elimination}%
destructor elimination, \hskip.1em
which prevents looping by blocking further 
\index{destructor elimination}%
destructor elimination.%

Note that the marked variables refer actually to the predecessors of the
values of the original lemma \inpit{\lessymbol 7'}, 
and that 
these three rounds of reduction
already include all that is required for the entire induction proof,
such that {\em\descenteinfinie}\/ would now conclude the proof with 
an induction-hypothesis application.
This most nicely illustrates the crucial 
similarity between recursion and induction,
which \boyer\ and \moore\ ``exploit'' \ldots\ 
``or, rather, contrived\closequotefullstopnospace\footnote{%
 \majorheadroom
 \Cf\ \cite[\p\,163, last paragraph]{bm}.%
}%

The proof by \index{induction!explicit}explicit induction in \THM, however,
now just starts to compute induction \index{induction schemes}schemes.
The two \index{induction templates}induction templates for \nlbmath\lessymbol\ found in 
\examref{example induction template two measured subsets} \hskip.1em
are applicable five times, resulting in the induction \index{induction schemes}schemes\,1--5
in \figuref{figure induction schemes}.

From the domains of the substitutions in the 
\index{step-case descriptions}step-case descriptions, \hskip.1em
it is obvious that
---~among \index{induction schemes}schemes\,1--5~---
only the two pairs of \index{induction schemes}schemes~2~and~3 \aswellas\ 4~and~5 
are candidates for subsumption,
\hskip.1em
which is not given here, however, 
because the case conditions of these two pairs of \index{induction schemes}schemes
are not subsets of each other. 

Nevertheless, these pairs of \index{induction schemes}schemes merge, resulting in the 
\index{induction schemes}schemes~6~and~7, respectively, 
which merge again, resulting in \index{induction schemes}scheme\,8\@.

Now only the \index{induction schemes}schemes~1~and~8 remain.
As each of them has \math x as an \index{induction variables}induction variable,
both \index{induction schemes}schemes would be flawed if they could not be merged.

It does not matter that the \index{induction schemes}scheme\,1 is subsumed by scheme\,8
simply 
because the phase of subsumption is already over; \hskip.3em
but they are also mergeable, 
actually with the same result as subsumption would have,
namely the \index{induction schemes}scheme\,9, 
which admits us 
to prove the generic step-case formula it describes
without further induction, and so \THM\ achieves the crucial task
of heuristic anticipation of an appropriate induction hypotheses, \hskip.1em
just as well as the \PURELISPTP.\footnote{%
 \majorheadroom
 The base cases show no improvement to the proof
 with the \PURELISPTP\ in \examref{example merging} 
 and a further additional, but
 also negligible overhead is 
 the preceding 
 reduction
 from \inpit{\lessymbol 7} over 
 \inpit{\lessymbol 7'} 
 to a version of \inpit{\lessymbol 7'} 
 with marked variables.%
}%
\getittotheright\qed
\index{recursion analysis|)}%
\pagebreak\end{example}
\subsubsection{Conclusion on \THM}\label
{subsubsection Conclusion on}%
Logicians reading on \THM\ may ask themselves many questions
such as: Why is merging of induction \index{induction schemes}schemes
---~seen as a binary algebraic operation~---
not \mbox{realized} to satisfy the constraint of associativity,
so that the result of merging become independent of the order
of the operations? \hskip.3em
Why does merging not admit the subterm-property in the same way 
as subsumption of induction \index{induction schemes}schemes does?
Why do some of the injectivity requirements\footnote{%
 \Cfnlb\ \noterefs{note injectivity 1}{note injectivity 2}.%
}
of subsumption and
mergeability lack a meaningful justification, and how can it be that
they do not matter?

The answer is trivial, although it is easily overlooked:
The part of the automation of induction 
we have discussed in this section on \THM,
belongs mostly to the field of heuristics and not in the field of logics.
Therefore, 
the final judgment cannot come from logical and intellectual
adequacy and comprehensibility
---~which are not much more applicable here than in the field of neural nets
    for instance~---
but must come from complete testing with a huge and growing corpus of 
example theorems. 
\hskip.1em
A modification of an operation, 
say merging of induction \index{induction schemes}schemes,
that may have some practical advantages for some examples
or admit humans some insight or understanding,
can be accepted
\onlyif\ it admits us to run,
as efficiently as before,
all the lemmas that could be 
automatically proved with the system before.
All \nolinebreak in \nolinebreak all, 
logical and formal considerations may help us to 
find new heuristics, 
but they cannot play any \role\ in their evaluation.\footnote{%
 \majorheadroom
 While \walthername\ is well aware of the primacy of testing in
 \makeaciteoftwo{waltherLPAR92}{waltherIJCAI93},
 this awareness is not reflected in the sloppy language of
 the most interesting papers 
 \cite{stevens-rational-reconstruction}
 and \mbox{\cite{bundy-recursion-analysis}:} \hskip.2em
 \hskip.3em
 Heuristics cannot be ``bugged'' or ``have serious flaws\closequotecomma
 unless this would mean that they turn out to be inferior to others \wrt\ a 
 standard corpus.
 A ``rational reconstruction'' or a 
 ``meta-theoretic analysis'' may 
 help to guess even superior heuristics,
 but 
 they 
 may 
 not have any epistemological value {\it per se}.%
}
\par\halftop\halftop\indent
Moreover,
it is remarkable that the \wellfounded\ relation that is expressed 
by the subsuming induction \index{induction schemes}scheme is smaller than that 
expressed by the subsumed one,
and the relation expressed by a merged \index{induction schemes}scheme is typically smaller
than those expressed by the original ones.
This means that the newly generated induction \index{induction schemes}schemes 
do not represent a more powerful
induction ordering (say, in terms of \noetherian\ induction), \hskip.1em 
but actually achieve an improvement \wrt\ the eager instantiation of
the induction hypothesis (both for a direct proof and for generalization), 
and provide case conditions that further
a successful generalization without further case analysis.

\halftop\halftop\indent
Since the end of the 1970s until today,
\THM\ has set the standard for \index{induction!explicit}explicit induction; \hskip.3em
moreover, \hskip.1em
\THM\ and its successors \NQTHM\ and \ACLTWO\ 
have given many researchers a hard time 
trying to demonstrate weaknesses 
of their \index{induction!explicit}explicit-induction heuristics, \hskip.1em
because examples carefully devised to fail with 
certain steps of the construction of 
induction \index{induction schemes}schemes 
(or other stages of the waterfall) \hskip.1em
tend to end up with alternative proofs
not imagined before.

Restricted to the mechanization of
the selection of an appropriate induction \index{induction schemes}scheme for 
\index{induction!explicit}explicit induction,
no significant 
overall
progress has been seen beyond \THM\ 
and we do not expect any for the future.
A heuristic approach that has to anticipate appropriate induction steps
with a lookahead of one individual rewrite step for 
each recursive function occurring in the input formula 
cannot go much further than the carefully developed and 
exhaustively tested \index{induction!explicit}explicit-induction heuristics 
of \THM\@.

\halftop\halftop\indent
Working with \THM\ (or \NQTHM) \hskip.1em
for the first time will always fascinate 
informaticians and mathematicians,
simply because it helps to save more time
with the standard everyday inductive proof work
than it takes, 
and the system often comes up with completely un\-expected proofs.
Mathematicians, however, should be warned that the less trivial mathematical
proofs that require some creativity and would deserve to be explicated in a
mathematics lecture, 
will require some hints, 
especially if the induction ordering
is not a combination of the termination orderings of the 
given function definitions. 
This is already the case for the simple proofs of the lemma
on the irrationality of the
square root of two, 
simply because the induction orderings of the
typical proofs exist only under the assumption that the lemma is wrong.
To make \THM\ find the standard proof,
the user has to define a function such as the following one:
\par\halftop\noindent\math{\begin{array}{@{}l@{~~~~~}l@{}}
  \inpit{\sqrtindordsymbol{}{}1}
 &\sqrtindordpp{}x y
\\
 &\mbox{~~}=\andsymbol(
   {\sqrtindordpp{}y{\divpp x{\spp{\spp\zeropp}}}},
\\
 &\mbox{~~~~~~}\andsymbol(
    {\sqrtindordpp{}
       {\timesppnoparentheses{\spp{\spp\zeropp}}{\minuspp x y}}
       {\minusppnoparentheses{\timespp{\spp{\spp\zeropp}}y}x}},
\\
 &\mbox{~~~~~~~~~~~\,}\hskip.07em{\sqrtindordpp{}
       {\minusppnoparentheses{\timespp{\spp{\spp\zeropp}}y}x}
       {\minusppnoparentheses x y}}))
\\
 &\mbox{~~~~~~~~~~~~~~~~~~~~~~~~~~~~~~~~~~~~~~~~~~~}
{\nottight{\nottight\antiimplies}}\timesppnoparentheses x x
\tightequal\timesppnoparentheses{\spp{\spp\zeropp}}{\timesppnoparentheses y y}
\nottight\und y\tightnotequal\zeropp
\\\end{array}}
\par\halftop\noindent
Note that the condition of \inpit{\sqrtindordsymbol{}{}1} cannot be fulfilled.
The three different occurrences of \sqrtindordsymbol{} on the right-hand 
side of the \pnc\ equation 
become immediately clear from \figuref{figure square root}.
Actually,
any single one of these occurrences is sufficient for a
proof of the irrationality lemma with \THM, \hskip.1em
provided that we give the hint that the \index{induction templates}induction templates of 
\sqrtindordsymbol{} should be used for computing the induction \index{induction schemes}schemes,
in spite of the fact that \sqrtindordsymbol{} does not occur in the lemma.%
\THMindexend

\begin{figure}[h]
\yestop\yestop\noindent\LINEnomath{\math{
\begin{array}[b]{@{}r@{\,}l@{\,}l@{\,}l@{\,}l@{\,}l@{\,}l@{\,}l@{\,}l@{}}x
 &=
 &\overline{A E}
\\\headroom
  y
 &=
 &\overline{F E}
 &=
 &\overline{B E}
 &=
 &\overline{A F}
\\\headroom
  \divpp x{\spp{\spp\zeropp}}
 &=
 &\overline{A C}
 &=
 &\overline{C F}
\\\headroom
  \minusppnoparentheses x y
 &=
 &\overline{A B}
 &=
 &\overline{B D}
 &=
 &\overline{B G}
 &=
 &\overline{G F}
\\\headroom
  \timesppnoparentheses{\spp{\spp\zeropp}}{\minuspp x y}
 &=
 &\overline{A D}
\\\headroom
  \minusppnoparentheses{\timespp{\spp{\spp\zeropp}}y}x
 &=
 &\overline{A G}
 &=
 &\overline{G D}
\\[-.9ex]
\\\end{array}}\hfill
{\begin{picture}(155,123)(-50,-10)
\put(100,0){\arc{200}{3.14152654}{4.71238898}}
\drawline(0,0)(100,0)
\drawline(0,0)(0,100)
\drawline(100,0)(0,100)
\drawline(0,0)(50,50)
\drawline(0,41.42135624)(29.28932188,70.71067812)
\drawline(0,41.42135624)(58.57864376,41.42135624)
\dashline{5}(29.28932188,70.71067812)(58.57864376,100)
\dashline{5}(0,100)(100,100)
\dashline{5}(58.57864376,41.42135624)(58.57864376,100)
\dashline{5}(100,0)(100,100)
\dashline{5}(50,50)(100,100)
\dashline{5}(0,41.42135624)(-29.28932188,70.71067812)
\dashline{5}(0,100)(-50,50)
\dashline{5}(0,0)(-50,50)
\dottedline[.]{2}(0,41.42135624)(100,0)
\put(-5,103){\math A}
\put(34,67){\math B}
\put(47,54){\math C}
\put(-5,-10){\math F}
\put(97,-10){\math E}
\put(61,39){\math D}
\put(-9,33){\math G}
\end{picture}}}\caption
{Four possibilities to descend with rational representations of 
\protect\nlbmath{\sqrt 2}:\label{figure square root}}
\par\getittotheright
{From the triangle with 
 right 
 angle at 
  \protect\nlbmath F \protect\hskip.1em to those at \protect\nlbmaths C, 
  \protect\nlbmaths G, or \protect\nlbmaths B.%
  \hskip 3.848em
  \mbox{}}
\end{figure}

\vfill\pagebreak

\subsection{\NQTHM}\label{subsection NQTHM}
Subsequent theorem provers by \boyer\ and \moore\ did not add much to 
the selection of an appropriate induction \index{induction schemes}scheme.
While both \NQTHM\ and \ACLTWO\ have been very
influential in theorem proving, their inductive heuristics are nearly the
same 
as
those in \THM\ and their waterfalls have quite similar structures.
Since we are concerned with the history of the mechanization of induction, 
we just sketch developments since 1979.
\\\indent
The one change from \THM\ to \NQTHM\ that most directly affected the
inductions carried out by the system is the abandonment of fixed
lexicographic relations on natural numbers as the only available well-founded
relations. \hskip.3em 
\vspace*{-1.5ex}%
\NQTHM\ introduces a formal representation of the ordinals up to
\nlbmaths{\varepsilon_0}, \hskip.1em
\ie\ up to \nlbmaths{\omega^{\omega^{\iddots}}}, \hskip.1em
and assumes that the
``less than'' relation on such ordinals is well-founded. \hskip.2em
This did not
change the induction heuristics themselves, 
it just allowed the admission of
more complex function definitions and the justification of more sophisticated
\index{induction templates}induction templates.
\\\indent
After the publication 
of \cite{bm} describing \THM, \hskip.1em
\boyer\ and
\moore\ turned to the question of providing limited support for higher-order
functions in their first-order setting.  This had two very practical
motivations.  One was to allow the user to extend the prover by defining and
mechanically verifying new proof procedures in the pure \LISP\ dialect
supported by \THM.  The other was to allow the user the convenience of \LISP's
``map functions'' and {\tt{LOOP}} facility.  
Both required formally defining
the semantics of the logical language in the logic, 
\ie\ axiomatizing the evaluation function {\tt EVAL}\@. \hskip.3em  
Ultimately this resulted in the provision of
{\em{metafunctions}} \cite{boyer-moore-1981-authors} and the non-constructive
``value-and-cost'' function \nolinebreak {\tt V\&C\$} \hskip.1em
\cite{bmeval-1988}, \hskip.2em
which were
provided as part of the \NQTHM\ system described in
\makeaciteoftwo{boyermoore}{boyermooresecondedition}. \hskip.3em
\\\indent
The most important
side-effect of these additions, however, 
is under the hood; \hskip.3em
\boyer\ and
\moore\ contrived to make the representation of 
constructor ground terms in the logic be
identical to their representation as constants 
in
its underlying implementation language \LISP: \hskip.3em
integers
are represented directly as \mbox{\LISP\ integers;} \hskip.3em
\mbox{for instance,} \hskip.2em
\spp{\spp{\spp\zeropp}} \hskip.1em
is represented by the machine-oriented internal \LISP\ representation
of {\tt 3}, \hskip.1em
instead of the previous
\mbox{\tt (ADD1 (ADD1 (ADD1 (ZERO))))}. \hskip.5em  
Symbols and list
structures are embedded this way \aswell, \hskip.1em
so that they can can profit from the very efficient representation
of these basic data types in \LISP\@. \hskip.3em
It thus also became possible to represent
symbolic machine states containing actual assembly code or the parse trees of
actual programs in the logic of \NQTHM\@. \hskip.3em
Metafunctions were put to good use canonicalizing symbolic
state expressions. \hskip.2em
The exploration of formal operational semantics with
\NQTHM\ blossomed.
\\\indent
In addition, 
\NQTHM\ adds a rational 
\index{linear arithmetic}%
linear-arithmetic\footnotemark\ 
decision procedure to
the simplification stage of the waterfall \cite{boyer-moore-1988}, \hskip.1em
reducing the amount of user interaction necessary to prove arithmetic
theorems. \hskip.2em
The incompleteness of the procedure 
when operating on terms beyond the 
\index{linear arithmetic}%
linear fragment
is of little practical importance since induction is available 
(and often automatic).%
\\\indent
With \NQTHM\ it became possible to formalize and verify
problems beyond the scope of \THM, \hskip.2em
such as
the correctness of a netlist implementing the
instruction-set architecture of a microprocessor 
\mbox{\cite{hunt-1985},}
\hskip.2em
\goedelsfirstincompletenesstheorem,\footnotemark\ \hskip.2em
the verified hard- and software stack of Computational
Logic, Inc., relating a fabricated microprocessor design through an
assembler, linker, loader, several compilers, and an operating system to
simple verified application programs,\footnotemark\
and the verification of the
Berkeley {\sc C} String Library.\footnotemark\ \hskip.3em
Many more examples are listed in \cite{boyermooresecondedition}.%
\pagebreak
\subsection{\ACLTWO}\label{subsection ACLTWO}%
\halftop\noindent
\addtocounter{footnote}{-3}%
\footnotetext{\label{note presburger}%
  \index{linear arithmetic}%
  Linear arithmetic is traditionally called 
  ``\presburgerarithmetic'' after \presburgername\ 
  (actually: ``\presburgertruename'')
  \presburgerlifetime; \hskip.2em
  \cfnlb\ 
  \cite{presburger}, 
  \cite{presburger-remarks-translation}, 
  \cite{presburger-life}.%
}%
\addtocounter{footnote}{1}%
\footnotetext{\label{note shankar}%
 \majorheadroom
 \Cfnlb\ \cite{shankar-1994}. \hskip.3em
 In \cite[\p\,xii]{shankar-1994} we read on this work with \NQTHM:%
 \footroom\notop
 \begin{quote}\headroom``%
 This theorem prover is known for its powerful heuristics for constructing
 proofs by induction while making clever use of previously proved lemmas.
 The \boyermooretheoremprover\ did not discover proofs of the incompleteness
 theorem but merely checked a detailed but fairly high-level proof containing
 over 2000 definitions and lemmas leading to the main theorems.
 These definitions and lemmas were constructed through a process of interaction
 with the theorem prover which was able to automatically prove a large number
 of nontrivial lemmas. By thus proving a well-chosen sequence of lemmas,
 the theorem prover is actually used as a {\em proof checker}\/ rather than a
 theorem prover.\\
 If we exclude the time spent thinking, planning, and writing about the proof,
 the verification of the incompleteness theorem occupied about eighteen months
 of effort with the theorem prover.%
 ''\notop\end{quote}%
}%
\addtocounter{footnote}{1}%
\footnotetext{%
 \majorheadroom
 \Cf\ \makeaciteoftwo{moore-1989-1}{moore-1989-2},
 \cite{cli-1989-1}, \cite{hunt-1989}, \cite{young-1989}, \cite{bevier-1989}.%
}%
\addtocounter{footnote}{1}%
\footnotetext{%
 \majorheadroom
 Via verification of its 
 {\tt{gcc}}-generated 
 Motorola MC68020
 machine code 
 \cite{c-string-lib-1996}.%
}%
Because of the pervasive change in 
the representation of constants,
the \LISP\ subset
supported by \NQTHM\ is exponentially more efficient than the \LISP s
supported by \THM\ and the \PURELISPTP.  
It is still too inefficient,
however:
Emerging applications of \NQTHM\ 
in the late 1980s 
included models of commercial
microprocessors; users wished to run their models on industrial test 
\nolinebreak suites.
The root cause of the inefficiency was that ground execution in \NQTHM\ was
done by a purpose-built interpreter implemented by \boyer\ and \moore.  
To \nolinebreak reach competitive speeds, 
it would have been necessary to build a good compiler and
full runtime 
system
for the \LISP\ subset axiomatized in \NQTHM\@. \hskip.1em  
Instead, \mbox{in August\,1989,}
less than a year after the publication of \cite{boyermoore} describing
\NQTHM, \hskip.2em
\boyer\ and \moore\ decided to axiomatize a practical subset of
\COMMONLISP\ \cite{commonlisp}, the then-emerging standard 
\LISP\nolinebreak\hspace*{-.12em}\nolinebreak, 
and to build an
\NQTHM-like theorem prover for \nolinebreak it. 
To demonstrate that the subset was a
practical programming language, they decided to code the theorem prover
applicatively in that subset. \hskip.2em
Thus, \ACLTWO\ was born.
\par\halftop\indent
\boyer\ left Computational Logic, \Inc, (CLI) \hskip.1em
and returned to his duties at
the \unitexasaustin\
in 1989, 
while \moore\ resigned his tenure and stayed at CLI\@. \hskip.2em
This meant \moore\ was working full-time on \ACLTWO,
whereas \boyer\ was working on it only at night.  
\kaufmannname\ \kaufmannlifetime, \hskip.1em
who had worked with
\boyer\ and \moore\ since the mid-1980s on \NQTHM\ and had joined them at
CLI, was invited to join the \ACLTWO\ project.
By the mid-1990s, \hskip.1em
\boyer\ requested that his name be removed as an author of \ACLTWO\ because
he no longer knew every line of code.
\par\halftop\indent
The only major change to inductive reasoning 
introduced by \ACLTWO\ was the further
refinement of the \index{induction templates}induction templates 
computed at definition time.  \hskip.1em
While \NQTHM\
built the case analysis from the case conditions ``governing'' 
the recursive calls, 
\ACLTWO\
uses the more restrictive notion of the tests ``ruling'' the recursive calls.
\hskip.1em
Compare the definition of {\em governors}\/ on \litspageref{180} of 
\cite{boyermooresecondedition}
to the definition of {\em rulers}\/ on \litspageref{90} of \cite{ACLTWO}.

\ACLTWO\ represents a major step,
however, 
toward \boyer\ and \moore's dream
of a {\em computational logic}\/ because it is a theorem prover for a
practical programming language.  Because it is so used, {\em scaling}\/ its
algorithms and heuristics to deal with enormous models and the formulas they
generate has been a major concern, as has been the efficiency of ground
execution.  
Moreover,
it also added many other proof techniques including
congruence-based contextual rewriting, additional decision procedures,
disjunctive search (meaning the waterfall no longer has just one pool but may
generate several, one of which must be ``emptied'' to succeed), \hskip.1em
and many
features made possible by the fact that the system code and state is visible
to the logic and the user.

Among the landmark applications of \ACLTWO\ are the verification of a
Motorola digital signal processor 
\cite{brock-hunt-1999} \hskip.1em and of
the floating-point division microcode for the 
AMD K5$^{\textsc{tm}}$ microprocessor 
\cite{moore-lynch-kaufmann-1998}, \hskip.2em
the routine
verification of all elementary floating point arithmetic 
on the AMD Athlon$^{\textsc{tm}}$
\cite{russinoff-1998}, \hskip.2em
the certification of the Rockwell Collins AAMP7G$^{\textsc{tm}}$ 
for multi-level secure
applications by the US National Security Agency based on the \ACLTWO\ proofs
\cite{aamp7g-2005}, \hskip.2em
and the integration of \ACLTWO\ into the work-flow of 
Centaur Technology, Inc., a major
manufacturer of X86 microprocessors 
\cite{hunt-swords-2009}. \hskip.3em
Some of this
work was done several years
before the publications appeared because the early use of formal methods was
considered proprietary. \hskip.2em
For example, the work for \cite{brock-hunt-1999} was
completed in\,\,1994, \hskip.1em
and that for \cite{moore-lynch-kaufmann-1998} \hskip.15em
in\,\,1995.

In most industrial applications of \ACLTWO,  
induction is not used in every proof.
Many of the proofs involve huge intermediate \formulae,
some requiring megabytes of storage simply to
represent, let alone simplify.
Almost all the proofs,
however,
depend on lemmas that require induction to prove.  

To be successful,
\ACLTWO\ must be good at both induction and simplification and 
{\em integrate}\/ them
seamlessly in a well-engineered system, \hskip.1em
so that the user can state and prove
in a single system all the theorems needed.  

\ACLTWO\nolinebreak\ is most relevant to
the historiography
of inductive theorem proving \mbox{because} it
demonstrates that the induction heuristics and the waterfall provide such
an
integration in ways that can be scaled to industrial-strength applications.%

\par\halftop\indent
\ACLTWO\ and, by extension, inductive theorem proving, 
have changed the way microprocessors 
and low-level critical software are designed.  
Proof of correctness, or at least
proof of some important system properties, is now a possibility.%
\begin{sloppypar}
\boyer, \moore, and \kaufmann\ 
were awarded the 2005 ACM Software Systems Award
for ``the \boyermooreTheoremProver'':\begin{quote}
``The \boyermooreTheoremProver\ is a highly engineered and effective 
formal-methods tool that pioneered the automation of proofs by induction, 
and now
provides fully automatic or human-guided verification of critical computing
systems. 
The latest version of the system, \ACLTWO, \hskip.2em
is the only
simulation/verification system that provides a standard modeling language
and industrial-strength model simulation in a unified framework. This
technology is truly remarkable in that simulation is comparable to C in
performance, but runs inside a theorem prover that verifies properties by
mathematical proof. 
\ACLTWO\ is used in industry by AMD, \hskip.1em
IBM, \hskip.1em
and Rockwell-Collins, \hskip.1em
among others.''\footnotemark
\end{quote}\end{sloppypar}
\pagebreak
\subsection{Further Historically Important Explicit-Induction Systems}\label
{subsection Further Most Noteworthy Explicit-Induction Systems}%

\noindent
Explicit induction is nowadays applied in many theorem proving systems,
such as 
\ISABELLEHOL, 
\COQ, 
\PVS,
and
\ISAPLANNER,
to name just a few.
We cannot treat all of these systems in \englishdiesespapier.
Thus, in this section, we sketch only those systems that provided crucial 
contributions to the history of the automation of mathematical induction.

\subsubsection{\RRL}\label{section RRL}%
\noindent
\footnotetext{%
 For the complete text of the citation of \boyer, \moore, and \kaufmann\ see 
 \url{http://awards.acm.org/citation.cfm?id=4797627&aw=149}.%
}%
\RRL, \hskip.1em
the {\em Rewrite Rule Laboratory}\/ \cite{rrl}, \hskip.2em
was initiated in 1982
and showed its main activity during its first dozen years. \hskip.1em
\RRL\ is a system for proving
the viability of many techniques related to term rewriting. \hskip.3em
Besides other forms of induction, 
\RRL\ includes {\em cover-set induction}, 
which has eager induction-hypothesis generation,
but is restricted to syntactic term orderings.

\subsubsection{\INKA}\label{section INKA}%

The \INKA\ project and the development of the \INKA\
\underline{in}duction systems
began at the University of \underline{Ka}rl\esi ruhe
at the beginning of the 1980s. \hskip.3em
It became part of the 
\SFB\ \oldsfbshort\ ``\AI\closequotecommasmallextraspace
which started in\,1985 and
was financed by the German Research Community (DFG) \hskip.1em
to overcome a backwardness in artificial intelligence in Germany 
of 
more than a 
decade
compared to the research in \EB\ and in the 
\nolinebreak US\@. \hskip.3em

\begin{sloppypar}
The \INKA\ systems were 
based on the concepts of \citet{bm} \hskip.1em
and proved the executability of several 
new concepts, \hskip.15em
but they were never competitive with their contemporary
\boyermooretheoremprovers,\footnote{\label{note induction contest}%
 \INKA\,5.0 \cite{inkafuenf}, however, 
 was competitive in speed with \NQTHM. \hskip.3em
 This can roughly be concluded from the results of the 
 inductive theorem proving contest at the \thesixteenthCADEninetynine\
 (the design of which is described in \cite{inductioncontest}), \hskip.1em
 where the following systems competed with each other
 (in interaction with the following humans): 
 \NQTHM\ ({\namefont Laurence Pierre}),
 \INKA\,5.0 (\huttername),
 \OYSTERCLAM\ (\bundyname),
 and a first prototype of \QUODLIBET\ (\kuehlername). \hskip.3em
 Only \OYSTERCLAM\ turned out to be significantly 
 slower than the other systems,
 but all participating systems would have been left far behind \ACLTWO\ 
 if it had participated.%
}
\hskip.2em
and the \mbox{development} of \INKA\ was discontinued 
in the year\,\,2000.%
\end{sloppypar}

\begin{sloppypar}
Three \INKA\ system descriptions were presented at the \CADEshort\ conference
series: \ \mbox{\cite{inka},} \ \cite{inkanext}, \ \cite{inkafuenf}.%
\end{sloppypar}



 Besides interfaces to users and other systems, and the integration of logics, 
 specifications, and results of other theorem provers, 
 the essentially induction-relevant additions of \INKA\ 
 as compared to the system described in \cite{bm}
 are the following: \hskip.3em
 In \nolinebreak\cite{inka}, \hskip.2em
 there is an existential quantification 
 where the system tries to find 
 witnesses for the existentially quantified variables
 by interactive program synthesis. \hskip.3em
 In \nolinebreak\cite{hutter-cade-nancy}, \hskip.2em
 there is synthesis of induction orderings by rippling
 (\cfnlb\ \sectref{subsection Rippling}).%

\begin{sloppypar}
A lot of most interesting work on 
\index{induction!explicit}explicit induction was realized
along the line of the \INKA\ systems: \hskip.2em
We have to mention here \walthername's \waltherlifetime\ 
elegant treatment of {\em automated termination proofs}\/ for
recursive function definitions 
\makeaciteoftwo{walthertermination}{walthertermination2}, \hskip.2em
and his theoretically outstanding work on the {\em generation of step cases}\/
with eager induction-hypothesis generation 
\makeaciteoftwo{waltherLPAR92}{waltherIJCAI93}. \hskip.5em
Moreover, 
there is \huttername's \hutterlifetime\ \hskip.1em
further development 
of 
\index{rippling}{\em rippling}
(\cf\,\sectref{subsection Rippling}), \hskip.2em
and
\protzenname's \protzenlifetime\ 
profound work on {\em patching of faulty conjectures}\/
and on breaking out of the imagined cage of 
\index{induction!explicit}explicit induction by 
\index{induction!lazy}{\em``lazy induction''}\/
\mbox{\makeaciteofthree{protzenlazy}{protzendiss}{protzenpatching}.}%
\end{sloppypar}

\subsubsection{\OYSTERCLAM}
The \OYSTERCLAM\ system was developed at the \uniEB\ in the late \nolinebreak
1980s\footnote{%
 The system description \cite{oysterclam} of \OYSTERCLAM\ 
 appeared already in summer
 1990 at the \CADEshort\ conference series
 (with a submission in winter\,1989/1990); \hskip.2em
 so the development must have started before the 1990s,
 contrary to what is stated in \litsectref{11.4} of \cite{bundy-survey}.%
} 
and the 1990s by a 
large team led by \bundyname\fullstopnospace\footnote{%
 \majorheadroom
 For \bundyname\ see also \noteref{note people and departments}.%
} \par
\OYSTER\ is a reimplementation of \NUPRL\ \cite{nuprl-book}, \hskip.2em
a proof editor for \loef\ constructive type theory 
with rules for 
\structuralinductionindex%
structural induction in the style of \peanoindex\peano\
---~a logic that is not well-suited for inductive proof search, as discussed in 
\nlbsectref{subsection The Standard High-Level Method of Mathematical Induction}.
\hskip.4em
\OYSTER\ is based on tactics with specifications in a meta-level language 
which provides a complete representation of the object level, \hskip.1em
but with a search space much better suited for inductive proof search.
\par
\CLAM\ is a proof planner (\cfnlb\ \sectref{subsection Proof Planning}) 
\hskip.1em which guides \OYSTER, \hskip.1em
based on proof search in the meta-language, \hskip.1em 
which includes
\index{rippling}rippling (\cfnlb\ \sectref{subsection Rippling}).

\OYSTERCLAM\ is the slowest system explicitly mentioned 
in this article.\arXivfootnotemarkref{note induction contest} \hskip.3em
One reason for this inefficiency is its constructive object-level logic. 
\hskip.3em
Its successor systems, however, are much faster.\footnote{%
 \majorheadroom
 One of the much faster successor systems of \OYSTERCLAM\ 
 under further development is \mbox\ISAPLANNER, \hskip.1em
 which is based on \ISABELLE\ \cite{isabellesevenhundred}. \hskip.3em
 See \cite{DF03-CADE-19} and
 \cite{proofplanningsystems} for early publications on \ISAPLANNER.%
}

In its line of development, \hskip.2em
\OYSTERCLAM\ proved the viability of several most important new concepts:
\begin{itemize}\noitem\item
Among the approaches that more or less address
theorem proving in general, \hskip.1em
we have to mention 
\index{rippling}%
{\em rippling} (\cfnlb\ \sectref{subsection Rippling}) \hskip.1em 
and a {\em productive use of failure}\/ for the 
suggestion of crucial new lemmas.\footnote{%
 \majorheadroom
 \Cfnlb\ \cite{failure-guide-induction}. \hskip.3em
 Moreover, 
 see our discussion on the particular theoretical relevance of 
 finding new lemmas in 
 mathematical induction in
 \sectref{subsection Proof-Theoretical Peculiarities of Mathematical Induction}.
 \hskip.4em
 Furthermore, 
 note that the practical relevance of finding new lemmas 
 addresses the efficiency of theorem proving in general, \hskip.2em 
 as described in \noterefss
 {note no difference in practice}{note practice one}{note practice two}
 of \sectref
 {subsection Proof-Theoretical Peculiarities of Mathematical Induction}.%
}\item\sloppy
A most interesting approach that addresses the core of the automation of 
inductive theorem proving and that deserves further development 
is the extension of 
\index{recursion analysis}%
\mbox{recursion analysis} to 
\index{ripple analysis}%
{\em ripple analysis}.\footnote{%
 \majorheadroom
 Ripple analysis is sketched already in 
 \cite[\litsectref 7]{bundy-recursion-analysis} \hskip.1em
 and nicely presented in {\cite[\litsectref{7.10}]{bundy-survey}.}%
}\index{induction!explicit|)}
\end{itemize}
\vfill\pagebreak
\section{Alternative Approaches Besides Explicit Induction}\label
{section Alternative Approaches to the Automation of Induction}%
In this section we will discuss the approaches to the automation 
of mathematical induction that do not strictly follow the 
method of \index{induction!explicit}explicit induction as we have described it.
In \nolinebreak general,
these approaches are not disjoint from 
\index{induction!explicit}explicit induction.
To the contrary, 
\index{proof planning}%
{\em proof planning}\/ and 
\index{rippling}%
{\em rippling}\/ have until now been
applied mostly to systems more or less based on \index{induction!explicit}explicit induction,
but they are not exclusively related to induction and 
they are not following \boyermoore's method of \index{induction!explicit}explicit induction
in every detail.
Even systems for 
\index{induction!implicit}{\em implicit induction}\/ may include many features
of \index{induction!explicit}explicit induction and some of them actually do, such as \RRL\ 
(\cfnlb\ \sectref{section RRL})
and \QUODLIBET\ (\cfnlb\ \sectref{section QUODLIBET}).

\subsection{Proof Planning}\label{subsection Proof Planning}%
\index{proof planning}%
Suggestions on how to overcome an envisioned
dead end in automated theorem proving 
were summarized in the end of the 1980s
under the keyword {\em proof planning}. 
Besides its human-science aspects\commanospace\footnote{%
 \Cfnlb\ \cite{science-of-reasoning}.%
}
the main idea\footnote{%
 \Cfnlb\ \cite{bundy-inductive-proof-planning},
 \cite{proofplanningsystems}.}
of proof planning is to extend a theorem-proving system
---~on top of the {\em low-level search space}\/ of the logic calculus
of a proof checker~---
with a {\em higher-level search space},
which is typically smaller or better organized \wrt\ searching,
more abstract, and 
more human-oriented.

The 
extensive and sophisticated subject 
of proof planning is not especially related to induction,
but addresses automated theorem proving in general.
We \nolinebreak cannot cover it here 
and have to refer the reader to 
the standard publications on the subject.\footnote{
 In addition to 
 \makeaciteoftwo{bundy-inductive-proof-planning}{science-of-reasoning} and
 \cite{proofplanningsystems}, see also
 \cite{dodi-diss},
 \cite{proof_planning_with_multiple_strategies},
 \cite{automatic-learning-proof-planning}, 
 and the references there.%
}

\subsection{Rippling}\label
{subsection Rippling}%
\index{rippling}%
{\em Rippling} is a technique for augmenting rewrite rules with 
information that helps to find a way to rewrite one expression ({\em goal}\/)
into another ({\em target}\/), \hskip.2em
more precisely to reduce the difference between the goal and the target
by rewriting the goal. 

Although rippling is not restricted to inductive theorem proving,
it was first used by \aubinname\footnote{%
 The verb ``to ripple up'' is used in \litsectrefs{3.2}{3.4} of 
 \cite{aubin-1976} \hskip.2em
 ---~not as a technical term, \hskip.1em
 but just as an informal term for motivating some
 heuristics. \hskip.3em
 The formalizers of 
 \index{rippling}%
 rippling 
 give explicit credit to \citet{aubin-1976} for their inspiration 
 in \cite[\litsectref{1.10}, \p\,21]{rippling-book}, \hskip.2em
 although \aubin\ does not mention the term at any other place 
 in his publications \makeaciteoftwo{aubin-1976}{aubin-1979}. \hskip.3em
 Note, 
 however,
 that instead of today's name ``rippling out\closequotecomma
 \aubin\ actually used ``rippling up\closequotefullstopnospace%
}
in the context of the description of 
heuristics for 
the automation of mathematical induction
and found most of its applications there. \hskip.3em
The leading developers and formalizers of the technique are
\bundyname, \huttername, \basinname, \harmelenname, and \irelandname.

We \nolinebreak 
have already mentioned 
\index{rippling}%
rippling 
in 
\sectref{subsection Further Most Noteworthy Explicit-Induction Systems}
several times,
but this huge and well-documented area of research cannot be covered here,
and we have to refer the reader to the 
monograph \cite{rippling-book}.\footnotemark
\par\pagebreak
\footnotetext{%
 Historically important are also the following publications on 
 \index{rippling}%
 rippling:
 \cite{hutter-rippling}, 
 \cite{rippling},
 \cite{failure-guide-induction},
 \cite{rippling-calculus}.%
}%
Let us explain here, 
however,
why 
\index{rippling}%
rippling can be most helpful 
in the automation of simple inductive proofs.

Roughly speaking,
the remarkable 
success in proving {\em simple}\/ theorems by induction automatically,
can be explained as follows:
If we look upon the task of proving a theorem 
as reducing it to a tautology,
then we have more heuristic guidance when we know that we probably 
have to do it by mathematical induction: 
Tautologies can have arbitrary subformulas,
but the induction hypothesis we are going to apply can restrict the search
space tremendously. 

In a cartoon of \bundyname's,
the original theorem is pictured 
as a zigzagged mountainscape and the reduced theorem after the 
unfolding of recursive operators 
according to 
\index{recursion analysis}%
recursion analysis (goal) \hskip.1em
is pictured
as the reflection of the mountainscape on the surface of a lake with ripples.
\hskip.2em
To apply the induction hypothesis (target), \hskip.1em
instead of the uninformed search for an arbitrary tautology, 
we have to {\em get rid of the ripples}\/ to be able to apply an instance of
the theorem as induction hypothesis to the mountainscape
mirrored by the calmed surface of the 
lake.

A crucial advantage of 
\index{rippling}%
rippling in the area of automated induction
is that it can also be used to suggest missing lemmas as described in
 \cite{failure-guide-induction}.

Until today,
\index{rippling}%
rippling was applied to the automation of induction
only within \index{induction!explicit}explicit induction,
whereas it is clearly not limited to 
\index{induction!explicit}explicit induction,
and we actually expect it to be more useful in areas of
automated theorem proving with bigger search spaces and, 
in \nolinebreak particular, 
in {\it\descenteinfinie}. 
\subsection{Implicit Induction}%
\index{induction!implicit}%
The further approaches to mechanize 
mathematical induction
{\em not}\/ subsumed by \index{induction!explicit}explicit induction,
however,
are united under the name 
\index{induction!implicit}%
``implicit induction\closequotefullstopnospace

Triggered\footnote{%
 Although it is obvious that in the relatively small community
 of artificial intelligence  and computer science in the 1970s,
 the success of \cite{bm} triggered the publication of papers
 on induction in the term rewriting community,
 we can document the influence of \boyer\ and \moore's work
 here only with the following facts: 
 \makeaciteoftwo{boyer-moore-1975}{bm}
 are both cited in \cite{huethullotinductionlessinduction}.
 \cite{boyer-moore-1977} is cited in 
 \cite{musserinductionlessinduction} as one of the 
 ``important sources of inspiration\closequotefullstop
 Moreover, 
 \citet{inductionlessinduction1} constitutively 
 refers to a personal communication 
 with \boyername\ in 1979. \hskip.3em
 Finally,
 \citet{gogueninductionlessinduction} avoids a direct reference to
 \boyer\ and \moore, but cites only the \PhDthesis\ \cite{aubin-1976}
 of \aubinname, 
 following their work in \EB.%
}
by the success of \citet{bm},
publication
on these alternative approaches started 
already in the year \nolinebreak 1980
in purely equational theories.\footnote{%
 \Cfnlb\
 \cite{gogueninductionlessinduction},
 \cite{huethullotinductionlessinduction},
 \cite{inductionlessinduction1},
 \cite{musserinductionlessinduction}.}
A \nolinebreak sequence of papers on technical improvements\footnote{%
 \Cfnlb\
 \cite{inductionlesshistc},
 \cite{inductionlesshista},
 \cite{inductionlesshistb},
 \cite{inductionlesshistd}.}
was topped by \cite{bachmair}, \hskip.1em
which gave rise to a hope to develop 
the method into practical usefulness,
although it was still restricted to purely equational theories. \hskip.2em
Inspired by this \mbox{paper,}
in the late 1980s and the first half of the 1990s 
several researchers tried to understand more clearly
what 
\index{induction!implicit}%
implicit induction means from a theoretical point of view 
and whether it could be 
useful in practice\@.\footnotemark
\par\pagebreak
\footnotetext{%
 \Cfnlb\ \eg\
 \cite{ZKK88},
 \cite{rrl},
 \cite{bevers&lewi},
 \cite{reddy},
 \cite{unicom},
 \cite{GaSt},
 \cite{spike},
 \cite{padawitzjsc}.
}%
While it is generally accepted that \cite{bachmair} is about 
\index{induction!implicit}%
implicit induction and
\cite{bm} is about \index{induction!explicit}explicit induction, 
there are the following three different viewpoints on what the 
essential aspect of 
\index{induction!implicit}%
implicit induction actually is.\begin{description}
\item[Proof by Consistency:\protect\footnotemark\ ]%
\footnotetext{%
 The name 
 \index{proof by consistency}%
 ``proof by consistency'' was coined in the title of \cite{kapur1},
 which is the later published forerunner 
 of its outstanding improved version \cite{kapur2}.}
Systems for 
\index{proof by consistency}%
proof by consistency run
some \KNUTHBENDIX\footnote{%
 See \UNICOM\ \cite{unicom} for such a system, 
 following \cite{bachmair} with several improvements. \hskip.1em
 See \cite{KB70} for the \KNUTHBENDIX\ completion procedure.%
} 
or superposition\footnote{%
 See \cite{GaSt} for such a system.%
}
completion procedure.
A proof attempt is successful
when the prover has drawn all necessary inferences and stops
without having detected any 
\index{consistency}inconsistency.
\\
\index{proof by consistency}Proof by consistency 
typically produces a huge number of irrelevant inferences 
under which the ones relevant for establishing the 
induction steps can hardly be made explicit. \hskip.3em
\index{proof by consistency}Proof by consistency 
has shown to perform far worse than 
any other known form of mechanizing mathematical induction;
\hskip.2em
mainly because it requires the generation of far 
too many superfluous inferences. \hskip.3em
Moreover, 
the runs are typically infinite, 
and the admissibility conditions are too restrictive
for most applications. 
\\Roughly speaking,  
the conceptual flaw in 
\index{proof by consistency}%
proof by consistency is that, 
instead of finding a sufficient set of reasonable inferences,
the research follows the idea of 
ruling out as many irrelevant inferences as \nolinebreak possible.%
\item[Implicit Induction Ordering: ] \ \sloppy
In the early implicit-induction systems,\footnote{%
 See \cite{unicom} and \cite{GaSt} for such systems.%
} \hskip.2em
\mbox{induction} proceeds over a syntactical term ordering,
which typically cannot be made explicit in the sense
that there would be some predicate term in the logical syntax
that denotes this ordering 
in the intended models of the specification. 
The \nolinebreak semantical orderings of \index{induction!explicit}explicit induction, 
however,
cannot depend 
on the precise syntactical term structure of a weight \nlbmaths w, \hskip.2em
but only on the value of \math w under an evaluation in the intended models.
\\
Contrary to rudimentary inference systems
that turned out to be more or less useless in practice
(such as the one of \cite{bachmair} for inductive completion
 in unconditional equational specifications), \hskip.2em
more powerful human-oriented inference systems
(such as the one of \QUODLIBET) \hskip.1em
are considerably restrained by the constraint to be sound also for
induction orderings that depend on the precise 
syntactical structure of terms (beyond their values).%
\footnote{%
 This soundness constraint,
 which was still observed in \cite{wirthdiss},
 was dropped during the further development of \QUODLIBET\ in
 \cite{kuehlerdiss}, 
 because it turned out to be unintuitive and superfluous.%
}
\\
The early 
\index{induction!implicit}%
implicit-induction systems needed such 
sophisticated term orderings\commanospace\footnotemark\
\hskip.2em
because they started from the induction conclusion and every inference step
reduced the formulas \wrt\ the induction ordering again and again,
but an application of an induction hypothesis 
was admissible to greater formulas only.
This deterioration of the ordering information with every inference step
was overcome by the introduction of 
explicit weight terms in 
\cite{wirthbecker}, \hskip.1em
which 
obviate the former need for syntactical term orderings as induction orderings.%
\par\footnotetext{%
 \Cf\ \eg\ \cite{bachmair}, 
 \makeaciteoftwo{SR--88--12}{simplificationorderings},
 \cite{geser-improved-general-path-order}.%
}%
\item[\DescenteInfinie\ (``Lazy Induction''): ]
\index{induction!lazy}%
Contrary to \index{induction!explicit}explicit induction,
where induction is introduced into 
an otherwise merely deductive inference system
only by the explicit application of induction axioms
in the \inductionrule,
the cyclic arguments and their 
\wellfoundedness\
in 
\index{induction!implicit}%
implicit induction 
need not be confined to single inference \nolinebreak steps.\footnote{%
 For this reason, 
 the funny name 
 \index{induction!inductionless}``inductionless induction''
 was originally coined for 
 \index{induction!implicit}%
 implicit induction in the titles of
 \makeaciteoftwo
 {inductionlessinduction1} 
 {inductionlessinduction2} 
 as a short form  for 
 ``induction without \inductionrule\closequotefullstop
 See also the title of 
 \cite{gogueninductionlessinduction} for a similar phrase.%
} 
The \inductionrule\ of \index{induction!explicit}explicit induction 
generates all induction hypotheses in a single inference step.
To \nolinebreak the contrary, in 
\index{induction!implicit}%
implicit induction,
the inference system ``knows'' 
what an induction hypothesis is, 
\ie\ it includes inference rules that provide or apply induction hypotheses,
given that certain ordering conditions resulting from these applications
can be met by an induction ordering. \hskip.3em
Because this aspect of 
\index{induction!implicit}%
implicit induction can facilitate the human-oriented
induction method described in 
\nlbsectref{subsection The Standard High-Level Method of Mathematical Induction},
\hskip.2em
the name {\em\descenteinfinie}
was coined for it (\cfnlb\ \sectref{subsection DescenteInfinie}). \hskip.3em
Researchers introduced to this aspect 
by \nolinebreak\cite{protzenlazy}
(entitled ``Lazy Generation of Induction Hypotheses'')
sometimes speak of \index{induction!lazy}``{lazy induction}'' 
instead of {\em\descenteinfinie}.\end{description}
The entire handbook article \cite{comonAR} 
(with corrections in \cite{zombie})
is dedicated to the two aspects of 
\index{proof by consistency}%
{\em proof by consistency}\/ and 
{\em implicit induction orderings}. \hskip.3em
Today,
however,
the interest in these two aspects 
tends to be  
historical or
theoretical,
especially because these aspects can hardly be combined with
\index{induction!explicit}explicit induction.

To the contrary, {\it\descenteinfinie}\/
synergetically combines with \index{induction!explicit}explicit induction,
as \nolinebreak witnessed by the \QUODLIBET\ system, 
which we will discuss in \nlbsectref{section QUODLIBET}.
\subsection{\QUODLIBET}\label{section QUODLIBET}
\begin{sloppypar}%
In the last years of the \SFB\ \oldsfbshort\ ``\AI''
(\cfnlb\ \sectref{section INKA}),
\hskip.2em
after extensive experiments with several inductive theorem proving systems,%
\footnote{%
 \majorheadroom
 \Cf\ \cite{kuehler-master}.%
} 
\hskip.2em
such as the \index{induction!explicit}explicit-induction systems 
\NQTHM\ (\cfnlb\ \sectref{subsection NQTHM}) \hskip.1em
and \INKA\
(\cfnlb\ \sectref{section INKA}),
\hskip.2em
the 
\index{induction!implicit}%
implicit-induction system \UNICOM\ \cite{unicom}, \hskip.2em
and the mixed system \RRL\
(\cfnlb\ \sectref{section RRL}),
\hskip.3em
\mbox\wirthname\ \wirthlifetime\ and
\mbox\kuehlername\ \kuehlerlifetime\
came to the conclusion that 
\mbox{---~in spite of the} excellent interaction concept of \UNICOM\footnote{%
 \majorheadroom
 For the assessment of \UNICOM's interaction concept see 
 \cite[\p\,134\ff]{kuehler-master}.%
}~---
\mbox{\em\descenteinfinie}\/ was actually the only aspect  
of 
\index{induction!implicit}%
implicit induction that deserved further investigation. \hskip.2em
Moreover, 
the coding of recursive functions in {\em unconditional}\/ equations
in \UNICOM\
turned out to be most inadequate for inductive theorem proving in practice,
where \pnc\ equations were in demand for specification, \hskip.2em
\aswellas\ 
clausal logic for theorem proving.\footnote{%
 \majorheadroom
 See \cite[\pp\,134,\,138]{kuehler-master}.%
}%
\end{sloppypar}

Therefore,
a new system had to be 
created%
, which was given the name 
\QUODLIBET\ (Latin for ``as you like it''), \hskip.3em 
because it 
should enable its users to avoid overspecification 
by admitting partial function specifications, \hskip.1em
and to execute proofs whose crucial proof steps 
mirror exactly the intended ones.\footnotemark\pagebreak\par\footnotetext{%
 We cannot claim that \QUODLIBET\ 
 is actually able to execute proofs whose crucial proof steps 
 mirror exactly the ones intended by its human users,
 simply because this was not scientifically investigated
 in terms of cognitive psychology.
 Users, however, considered it to be more appropriate that 
 other systems in this aspect, 
 mostly due to the direct 
 support for partial and mutually recursive function specification,
 \cfnlb\ \cite{loechner-lpo}. \hskip.3em
 Moreover, 
 the four dozen elementary 
 rules of \QUODLIBET's inference machine were designed to mirror
 the way human's organize their proofs 
 (\cfnlb\ \cite{wirthdiss}, \cite{kuehlerdiss}); \hskip.2em
 so a \nolinebreak user has to deal with one natural inference step where
 \OYSTER\ may have hundreds of intuitionistic steps.
 The appropriateness of \QUODLIBET's calculus for interchanging information 
 with humans deteriorated, however, 
 after adding inference rules for the efficient implementation
 of \presburgerarithmetic, as we will explain below.
 Note that the calculus is only the lowest logic level 
 a user of a theorem-proving system may have to deal with; 
 from our experience with many such systems we came to the firm conviction,
 however,
 that the automation of proof search 
 will always fail on the lowest logic level from time to time,
 such that human-oriented state-of-the-art logic calculi
 are essential for the acceptance of automated, interactive theorem provers
 by their users.%
}

A concept for
partial function specification instead of the totality requirement of \index{induction!explicit}explicit 
induction was easily obtained by elaborating the first part of 
\cite{wirth-master} \hskip.1em
into the framework for \pnc\ rewrite systems
of \cite{wgjsc}. \hskip.3em
After inventing 
\index{constructor variables}%
constructor variables in \cite{wgkp}, \hskip.1em
the monotonicity of validity \wrt\ consistent extension of the 
partial specifications was easily achieved \cite{wgcade}, \hskip.2em
so that the induction proofs did not have to be re-done after 
such an extension of a partially defined function.

Although the efficiently decidable 
\index{confluence}confluence criterion that defines 
admissibility of function definitions in \QUODLIBET\
and guarantees their (object-level) 
\index{consistency}%
consistency 
(\cfnlb\ \sectref{subsection Confluence}) 
was very hard to prove and 
was presented completely and in an appropriate form not before \cite{wirth-jsc},
\hskip.2em
the essential admissibility requirements were already clear 
in\,1996.\footnotemark\par\footnotetext{%
 \majorheadroom
 See \cite{kwspec} \hskip.1em
 for the first publication of the object-level 
 \index{consistency}%
 consistency of 
 the specifications that are admissible and supported with strong induction
 heuristics in \QUODLIBET.
 \mbox{In \cite{kwspec},} \hskip.1em
 a huge proof from the original 1995 edition of
 \cite{wirthconfluence} guaranteed the 
 \index{consistency}%
 consistency. \hskip.3em
 Moreover, 
 the most relevant and appropriate one of the seven inductive validities of 
 \cite{wgcade} is chosen for \QUODLIBET\ in \cite{kwspec}
 (no longer the initial or free models typical for 
  \index{induction!implicit}%
  implicit induction!).%
}%
The weak admissibility conditions of \QUODLIBET\
{---~mutually} recursive functions, possibly partially defined 
because of missing cases or non-termination~---
are of practical importance. \hskip.1em
Although humans can code mutually recursive functions into 
non-mutually recursive functions,\footnotemark\ 
they will hardly be able to understand complicated formulas 
where these encodings occur, and so they will have severe problems
in assisting the proving system in the construction of hard proofs.
Partiality due to non-termination essentially occurs in interpreters
with undecidable domains. \hskip.2em
Partiality due to missing cases of the definition 
can often be avoided by overspecification in theory, \hskip.2em
but not in practice where the unintended results of overspecification
may complicate matters considerably. 

For instance, \mbox\loechnername\ \loechnerlifetime\
(a \nolinebreak user, not a developer of \QUODLIBET) \hskip.1em
concludes in 
{\cite[\p\,76]{loechner-lpo}:}\notop\halftop\begin
{quote}``%
The translation of the different specifications into the input language of the 
inductive theorem prover \QUODLIBET\ \cite{quodlibet-cade} was straightforward. 
We later realized that this is difficult or impossible
with several other inductive provers as these have problems 
with mutual recursive
functions and partiality%
'' \ldots\end{quote}%
\pagebreak\par\indent
\footnotetext{%
 See the first paragraph of 
 \sectref{section example Applicable Induction Templates constructor style}.%
}%
Based on 
the {\em\descenteinfinie}\/ inference system for clausal first-order logic 
of 
\cite{SR--95--15}\commanospace\footnote{%
 \majorheadroom
 Later improvements of this inference system are found in 
 \cite{wirthdiss},
 \cite{kuehlerdiss},
 and
 \cite{samoa-phd}.%
} \hskip.3em
the 
system development
of \QUODLIBET\ in \COMMONLISP\
(\cf\,\sectref{subsection ACLTWO}), \hskip.2em
mostly by \kuehler\
and \samoaname\ \samoalifetime, \hskip.2em
lasted from\,1995 to\,\,2006\@. \hskip.3em
The system was described and demonstrated at the 
\thenineteenthCADEthree\ \cite{quodlibet-cade}. \hskip.2em
\mbox{The extension} of the 
{\em\descenteinfinie}\/ inference systems of \QUODLIBET\ to the full \opt{modal}
higher-order logic of \makeaciteoftwo{wirthcardinal}{SR--2011--01} \hskip.1em
has not been implemented yet.

\halftop\halftop\halftop\indent
To the best of our knowledge,
\QUODLIBET\ is the first theorem prover whose proof state is an and-or-tree
(of clauses); \hskip.2em 
actually, 
a forest of such trees, 
so that in a mutual induction proof each conjecture providing induction
hypotheses has its own tree \cite{kuehlerdiss}. \hskip.3em
An extension 
of the 
\index{recursion analysis}%
recursion analysis of \cite{bm} 
for constructor-style
specifications (\cfnlb\ \sectref{subsection Termination}) \hskip.1em
was developed by writing and testing tactics in 
\QUODLIBET's \Pascal-like\footnotemark\
meta-language \QML\ \cite{kuehlerdiss}.
To achieve an acceptable run-time performance 
(but not competitive with \ACLTWO, of course), \hskip.2em
\QML\ tactics are compiled before execution.\footnotetext{%
 \majorheadroom
 See \cite{wirth-pascal} for the programming language \Pascal.
 The critical decision for an imperative instead of a functional 
 tactics language
 turned out to be most appropriate during the ten years of using \QML.%
}%
\par
In principle, termination proofs are not required,
simply because termination is not an admissibility requirement in \QUODLIBET.
\hskip.3em
Instead,
definition-time 
\index{recursion analysis}%
recursion analysis uses
induction lemmas (\cfnlb\ \sectref{subsubsection Induction Templates})
to prove lemmas on function domains by induction.\footnote{%
 \majorheadroom
 While domain lemmas for totally defined functions are usually found without
 interaction and total functions do not provide relevant overhead
 in \QUODLIBET, \hskip.1em
 the \nolinebreak user often has to help in case of partial function
 definitions by providing {\em domain lemmas}\/ such as
 \par\noindent\LINEmaths{
 \DEF\dloncepp x l\comma \mbppp x l\tightnotequal\truepp},
 \par\noindent for \dloncesymbol\ 
 defined via \inpit{\dloncesymbol 1\mbox{--}2} of
 \sectref{subsection Standard Data Types}.%
}
\par\indent
At proof time, 
\index{recursion analysis}%
recursion analysis is used by the standard tactic
only to determine the \index{induction variables}induction variables 
from the \index{induction templates}induction templates:
As seen in \examref{example first proof of (less7)} 
of \nlbsectref{section example first proof of (less7)}
\wrt\ the strengthened transitivity of \lessymbol\
(as compared to the \index{induction!explicit}explicit-induction proof in 
 \examref{example second proof of (less7)}
 of \sectref{section example second proof of (less7)} 
 and \examref{example merging 2}
 of \sectref{section example merging 2}),
\hskip.25em
subsumption and merging of \index{induction schemes}schemes are not required
in {\it\descenteinfinie}.\footnote{%
 \majorheadroom
 Although it is not a must and not part of the standard tactic,
 induction hypotheses may be generated eagerly in \QUODLIBET\ 
 to enhance generalization as 
 in \examref{example proof of (ack4)} 
 of \sectref{section example proof of (ack4)}, \hskip.1em
 in which case subsumption and merging of induction \index{induction schemes}schemes as 
 described in \sectref{subsubsection Proof-Time Recursion Analysis in} \hskip.1em
 are required. \hskip.3em
 Moreover, the concept of flawed induction \index{induction schemes}schemes of \QUODLIBET\
 (taken over from \THM\ \aswell, 
  \cf\ \nlbsectref{subsubsection Proof-Time Recursion Analysis in})
 depends on the mergeability of \index{induction schemes}schemes.
 Furthermore, \QUODLIBET\ actually applies some merging techniques
 to plan case analyses optimized for induction 
 \cite[\litsectref{8.3.3}]{kuehlerdiss}.
 The question why \QUODLIBET\ 
 adopts 
 the great ideas of 
 \index{recursion analysis}%
 recursion analysis
 from \THM,
 but does not follow them precisely, has 
 two answers:
 First, it was necessary to extend the heuristics of \THM\ to deal with
 constructor-style definitions.
 The second answer was already given in 
 \sectref{subsubsection Conclusion on}: 
 Testing is the only judge on heuristics.%
}

\pagebreak
\par\halftop\halftop\indent
A considerable speed-up of \QUODLIBET\ 
and an extension of its automatically provable theorems
was achieved by \samoa\ during his \PhD\ work with the
system in 2004--2006\@. \hskip.3em
He developed a marking concept for the tagging of rewrite lemmas
(\cfnlb\ \sectref{subsubsection simplification in}), \hskip.1em
where the elements of a clause can be marked as Forbidden, Mandatory, 
Obligatory, and Generous, to control the recursive relief of conditions
in contextual rewriting \makeaciteoftwo{samoa-phd}{jancl}. \hskip.3em
Moreover, a very simple, but most effective reuse mechanism
analyzes during a proof attempt whether it actually establishes a proof
of some sub-clause, and uses this knowledge to crop conjunctive branches
that do not contribute to the actual goal \cite{samoa-phd}. \hskip.3em
Finally,
an even closer integration of 
\index{linear arithmetic}%
linear arithmetic
(\cfnlb\ \noteref{note presburger})
with excellent results
\makeaciteoftwo{samoacalculemus}{samoa-phd} questioned one of the
basic principles of \QUODLIBET, namely the idea that the prover
does not try to be clever, but stops early if there is no progress
visible, and presents the human user the proof state in a nice
graphical tree representation:
The expanded highly-optimized formulation of arithmetic
by means of special functions for the decidable fragment
of \presburgerarithmetic\
results in clauses that do not easily admit human inspection anymore.
We did not find means to overcome this,
because we did not find a way to fold theses clauses to achieve
a human-oriented higher level of abstraction. 

\halftop\indent
\QUODLIBET\ is, of course, able to do all\footnote{%
 These three {\em\descenteinfinie}\/ proofs are presented as
 \examrefs{example first proof of (+3)}{example first proof of (less7)} of
 \nlbsectref{section example first proof of (+3)}, \hskip.1em
 and 
 \examref{example proof of (ack4)} of
 \nlbsectref{section example proof of (ack4)}.%
} 
{\em\descenteinfinie}\/ 
proofs of our examples automatically. \hskip.3em
Moreover, \QUODLIBET\ finds all proofs for the irrationality 
of the square root of two indicated in \figuref{figure square root}
(sketched in \sectref{subsubsection Conclusion on})
automatically and without explicit hints on the induction ordering
(say, via newly defined nonsensical functions,
 such as the one given in \inpit{\sqrtindordsymbol{}{}1}
of \sectref{subsubsection Conclusion on}) \hskip.1em
--- provided that the required lemmas are available.

\halftop\indent
All in all,
\QUODLIBET\ has proved that 
{\em\descenteinfinie} 
(\index{induction!lazy}``lazy induction'')
goes well together with \index{induction!explicit}explicit induction
and that we have reason to hope that eager induction-hypotheses generation
can be overcome for theorems with difficult induction proofs, 
sacrificing neither efficiency nor the 
usefulness of the excellent
heuristic knowledge developed in \index{induction!explicit}explicit induction.
Why {\em\descenteinfinie}\/ and human-orientedness 
should remain on the agenda for induction in \maslong s
is explained in the manifesto \cite{wirth-manifesto-ljigpl}. 
\vfill\pagebreak
\yestop\yestop\section{Lessons Learned}\label
{section Lessons Learned}

\noindent
What lessons can we draw from the history of 
the automation of 
induction?

\begin{itemize}\noitem\item
Do not be too inclined to follow the current fads.  
Choose a hard problem,
give thought to the ``right'' foundations, and then pursue its solution with
patience and perseverance.\noitem\item
Another piece of oft-repeated advice to the young researcher: start simply.

From the standpoint of formalizing microprocessors, investing in a theorem
prover supporting only {\tt NIL} and {\tt CONS} is clearly inadequate.
\hskip.05em  
From the standpoint of understanding induction and simplification, 
however,
it presents virtually all the problems, \hskip.1em  
and its successors then gradually refined
and elaborated the techniques.
The four key provers discussed here 
---~the \PURELISPTP, \THM, \NQTHM, and \ACLTWO~--- 
are clearly ``of~a~kind\closequotefullstopextraspace  
The lessons learned from one tool
directly informed the design of the next.\noitem\item
If you are interested in building an inductive theorem prover, do not
make the mistake of focusing merely on an induction principle and the heuristics
for controlling it.  A successful inductive theorem prover must be able to 
simplify and generalize.  Ideally, it 
should 
be able to invent new concepts to
express inductively provably theorems.\noitem\item
If theorems and proofs are simple and obvious for humans, 
a good automatic theorem
prover ought not to struggle with them.  
If it takes a lot of time and machinery to
prove obvious theorems, then truly interesting theorems are out of reach.\noitem\item
Do not be too eager to add features that break old ones.  
Instead,
truly \mbox{explore} 
the extent to which new problems can be formalized within the
\mbox{existing} framework so as to exploit the power of the existing system.  

Had \boyer\ and \moore\ adopted higher-order logic initially or 
attempted to solve the
\mbox{problem}
solely by exhaustive searching in a general purpose logic calculus,
the discovery of many powerful techniques would have
been delayed.\noitem\item
We strongly recommend collecting all your successful proofs into a regression
suite and re-running your improved provers on this suite regularly. \hskip.2em
It \nolinebreak is remarkably easy to ``improve'' a theorem prover 
such that it discovers
a new
proof at the cost of failing to re-discover old ones.  

The \ACLTWO\ regression
suite,
which is used
as the acceptance test that any suggested
possible improvement has to pass,
contains over 90,000 {\tt DEFTHM} commands, \ie\ conjectures 
to be proved.
\hskip.2em
It is an invaluable resource to \kaufmann\ and \moore\ 
when
they explore
new heuristics.\noitem\item
Finally, \hskip.1em
\boyer\ and \moore\ 
did not give names
to their provers before \ACLTWO, \hskip.1em
and so they became most commonly known under the name 
{\em the \boyermooretheoremprover}. \hskip.4em

So here is some advice to young researchers who want to become well-known:
Build a good system, but do not give it a name,
so that people have to attach your name to it!%
\pagebreak\end{itemize}
\halftop\yestop\section{Conclusion}\label
{section Conclusion}%

\begin{quote}\halftop
``%
One of the reasons our theorem prover is successful
is that we trick the user into telling us the proof.
And the best example of that, that I know, is:
If \nolinebreak you \nolinebreak want 
to prove that there exists a prime factorization
---~that is to say a list of primes whose product is any given number~---
then the way you state it is:
You define a function that takes a natural number and 
delivers a list of primes, and then you prove that it does that.
And, 
of course,
the definition of that function is everybody else's proof.
The absence of quantifiers and the focus on constructive,
you know, 
recursive definitions
forces people to do the work.
And so then,
when the theorem prover proves it,
they say `Oh what wonderful theorem prover!',
without even realizing they sweated bullets
to express
the theorem in that impoverished logic.''\end{quote} 
said \moore,
and \boyer\ agreed laughingly.\footnote{\cite{boyer-moore-2012}.}\\\mbox{}
\par
\appendix
\addcontentsline{toc}{section}{Acknowledgments}
\begin{ack}\begin{quote}\noindent\sloppy
We would like to thank
\acerbiname,
\barnername,
\boyerswifename,
\boyername,
\bundyname,
\goldsteinname,
\gramlichname,
\huntname,
\huttername,
\kaufmannname,
\kuehlername,
\madlenername, 
\mbox{\mooreswifename,}
\padawitzname,
\rezaeinamenoindex, 
\samoaname,
and
{\namefont Judith Stengel}.

In particular, we are obliged to our second reader 
\bundyname\ for his most careful reading
and long long list of flaws and elaborate
suggestions that improved this article considerably.

As our other second reader \gramlichname\ \gramlichlifetime, \hskip.2em
a gifted teacher, a most creative and scrutinous researcher, and a true friend,
\hskip.1em 
passed away 
\mbox{---~much} too early~--- before dawn on June\,\,3, 2014, \hskip.1em
we would like to dedicate this article to him.
Besides the automation of mathematical induction, {\namefont\bernhard}'s 
main focus was on confluence and termination of term rewriting systems,
where he proved many of the hardest theorems of the 
area \cite{gramlich-phdthesis96}.

\end{quote}\end{ack}
\nocite{writing-mathematics,Boy71,boyer-moore-fast-string-searching,boyer-moore-1977,boyer-moore-proof-checking-RSA,boyer-moore-turing-completeness-lisp,boyer-moore-halting-problem,boyer-moore-1985,BM90,boyer-moore-shostak,boyer-yu-1992,moore-1979,herbrand-handbook,wirth-heijenoort%
}
\vfill\pagebreak
\halftop
\addcontentsline{toc}{section}{References}
\catcode`\@=11
\renewcommand\@openbib@code{%
      \advance\leftmargin\bibindent
      \itemindent -\bibindent
      \listparindent \itemindent
      \parsep 10pt
      \itemsep -6.25pt
      \baselineskip 0pt
      }%
\catcode`\@=12
\footnotesize
\bibliography{herbrandbib}
\vfill\cleardoublepage
\addcontentsline{toc}{section}{Index}
\index{Axiom of Choice|see{choice, Axiom of Choice}}%
\index{Axiom of Structural Induction|see{induction, structural}}%
\index{Theorem of Noetherian Induction|see{induction, Noetherian}}%
\index{structural induction|see{induction, structural}}%
\index{Noetherian induction|see{induction, Noetherian}}%
\index{induction!descente infinie|see{descente infinie}}%
\index{Presburger Arithmetic|see{linear arithmetic}}%
\printindex
\end{document}

%% file: pdf.bbl
\begin{thebibliography}{9}

\bibitem[\protect\citeauthoryear{Abrahams \bgroup\&al.\egroup
  }{1980}]{seventhPOPLeighty}
Paul~W. Abrahams, Richard~J. Lipton, and Stephen~R. Bourne, editors.
\newblock {\em Conference Record of the \nth 7\,\POPLname, Las Vegas (NV),
  1980}. \acmpress, 1980.
\newblock \url{http://dl.acm.org/citation.cfm?id=567446}.

\bibitem[\protect\citeauthoryear{Acerbi{\protect\acerbiindex}}{2000}]{plato-in%
duction}
Fabio Acerbi{\protect\acerbiindex}.
\newblock {\plato: Parmenides 149a7--c3\@. A} proof by complete induction?
\newblock {\em Archive for History of Exact Sciences}, 55:57--76, 2000.

\bibitem[\protect\citeauthoryear{Ackermann\protect\ackermannindex}{1928}]{acke%
rmann-1928a}
Wilhelm Ackermann\protect\ackermannindex.
\newblock {Zum \hilbert schen Au\fbi au der reellen Zahlen}.
\newblock {\em {\MathematischeAnnalen}}, 99:118--133, 1928.
\newblock Received \Jan\,20, 1927.

\bibitem[\protect\citeauthoryear{Ackermann\protect\ackermannindex}{1940}]{acke%
rmann-consistency-of-arithmetic}
Wilhelm Ackermann\protect\ackermannindex.
\newblock {Zur Wider\-spruch\esi frei\-heit der Zahlen\-theorie}.
\newblock {\em {\MathematischeAnnalen}}, 117:163--194, 1940.
\newblock Received \Aug\,15, 1939.

\bibitem[\protect\citeauthoryear{A{\"\i}t-Kaci \bgroup\&\ \egroup
  Nivat}{1989}]{kaci-nivat}
Hassan A{\"\i}t-Kaci and Maurice Nivat, editors.
\newblock {\em {\Proc\ of the Colloquium on Resolution of Equations in
  Algebraic Structures (CREAS), Lakeway (TX), 1987}}. \academicpress, 1989.

\bibitem[\protect\citeauthoryear{Anon}{2005}]{aamp7g-2005}
Anon.
\newblock {Advanced Architecture MicroProcessor 7 Government (AAMP7G)}
  microprocessor.
\newblock Rockwell Collins, Inc. \WWW\ only:
  \url{http://www.rockwellcollins.com/sitecore/content/Data/Products/Informati%
on_Assurance/Cryptography/AAMP7G_Microprocessor.aspx}, 2005.

\bibitem[\protect\citeauthoryear{Armando \bgroup\&al.\egroup
  }{2008}]{fourthIJCAReight}
Alessandro Armando, Peter Baumgartner, and Gilles Dowek, editors.
\newblock {\em {\thefourthIJCAReight}}, number 5195 in Lecture Notes in
  Artificial Intelligence. {\springerverlag}, 2008.

\bibitem[\protect\citeauthoryear{Aubin{\protectedaubinindex}}{1976}]{aubin-197%
6}
Raymond Aubin{\protectedaubinindex}.
\newblock {\em Mechanizing Structural Induction}.
\newblock PhD thesis, \uniEBshort, 1976.
\newblock Short version is \cite{aubin-1979}.
  \url{http://hdl.handle.net/1842/6649}.

\bibitem[\protect\citeauthoryear{Aubin{\protectedaubinindex}}{1979}]{aubin-197%
9}
Raymond Aubin{\protectedaubinindex}.
\newblock {Mechanizing Structural Induction --- Part\,I: Formal System.
  Part\,II: Strategies}.
\newblock {\em \tcsname}, 9:329--345+347--362, 1979.
\newblock Received March (Part\,I) and November (Part\,II) 1977, \rev\
  \Mar\,1978. Long version is \cite{aubin-1976}.

\bibitem[\protect\citeauthoryear{Autexier{\protectedautexierindex}
  \bgroup\&al.\egroup }{1999}]{inkafuenf}
Serge Autexier{\protectedautexierindex}, Dieter Hutter{\protectedhutterindex},
  Heiko Mantel, and Axel Schairer.
\newblock System description: {\INKA\,5.0} -- a logical voyager.
\newblock 1999.
\newblock In \cite[\PP{207}{211}]{sixteenthCADEninetynine}.

\bibitem[\protect\citeauthoryear{Autexier{\protectedautexierindex}}{2005}]{ser%
getableau}
Serge Autexier{\protectedautexierindex}.
\newblock On the dynamic increase of multiplicities in matrix proof methods for
  classical higher-order logic.
\newblock 2005.
\newblock In \cite[\PP{48}{62}]{fourteenthTABLEAUfive}.

\bibitem[\protect\citeauthoryear{Avenhaus \bgroup\&al.\egroup
  }{2003}]{quodlibet-cade}
J{\"u}rgen Avenhaus, Ulrich K{\"u}hler{\protectedkuehlerindex}, Tobias
  Schmidt-Samoa{\protectedsamoaindex}, and Claus-Peter
  Wirth{\protectedwirthindex}.
\newblock How to prove inductive theorems? {\QUODLIBET}!
\newblock 2003.
\newblock In \cite[\PP{328}{333}]{nineteenthCADEthree}, \www\url{/p/quodlibet}.

\bibitem[\protect\citeauthoryear{Baader}{2003}]{nineteenthCADEthree}
Franz Baader, editor.
\newblock {\em {\thenineteenthCADEthree}}, number 2741 in Lecture Notes in
  Artificial Intelligence. {\springerverlag}, 2003.

\bibitem[\protect\citeauthoryear{Baaz \bgroup\&\ \egroup
  Leitsch}{1995}]{baazleitschcolllog}
Matthias Baaz and Alexander Leitsch.
\newblock Methods of functional extension.
\newblock {\em Collegium Logicum --- Annals of the {\goedelname\ Society}},
  1:87--122, 1995.

\bibitem[\protect\citeauthoryear{Bachmair \bgroup\&al.\egroup
  }{1992}]{Hotz-Festschrift}
Leo Bachmair, Harald Ganzinger, and Wolfgang~J. Paul, editors.
\newblock {\em {Informatik -- Festschrift zum 60.\,Geburt\esi tag von
  {\namefont G\ue nter Hotz}}}.
\newblock \teubnerverlag, 1992.

\bibitem[\protect\citeauthoryear{Bachmair}{1988}]{bachmair}
Leo Bachmair.
\newblock Proof by consistency in equational theories.
\newblock 1988.
\newblock In \cite[\PP{228}{233}]{lics3}.

\bibitem[\protect\citeauthoryear{Bajscy}{1993}]{ijcai13}
Ruzena Bajscy, editor.
\newblock {\em {\Proc\ \nth{13} \IJCAIname\ (\IJCAI), Chambery (France)}}.
  \morgankaufmann, 1993.
\newblock \url{http://ijcai.org/Past%20Proceedings}.

\bibitem[\protect\citeauthoryear{Barendregt}{1981}]{lambda-calculus-1st}
Hen(dri)k~P. Barendregt.
\newblock {\em The Lambda Calculus --- Its Syntax and Semantics}.
\newblock Number 103 in Studies in Logic and the Foundations of Mathematics.
  \northholland, 1981.
\newblock \nth 1\,\edn\ (final\,\rev\,\edn\ is \cite{lambda-calculus-final}).

\bibitem[\protect\citeauthoryear{Barendregt}{2012}]{lambda-calculus-final}
Hen(dri)k~P. Barendregt.
\newblock {\em The Lambda Calculus --- Its Syntax and Semantics}.
\newblock Number~40 in Studies in Logic. \collegepublications, 2012.
\newblock \nth 6\,\rev\,\edn\ (\nth 1\,\edn\ is \cite{lambda-calculus-1st}).

\bibitem[\protect\citeauthoryear{Barner{\protectedbarnerindex}}{2001a}]{fermat%
slife-english-2}
Klaus Barner{\protectedbarnerindex}.
\newblock {\fermatname\ (1601?--1665) --- His life beside mathematics}.
\newblock {\em European Mathematical Society Newsletter},
  43\,(\Dec\,2001):12--16, 2001.
\newblock Long version in German is \cite{fermatslife}.
  \url{www.ems-ph.org/journals/newsletter/pdf/2001-12-42.pdf}.

\bibitem[\protect\citeauthoryear{Barner{\protectedbarnerindex}}{2001b}]{fermat%
slife}
Klaus Barner{\protectedbarnerindex}.
\newblock {Da\es\ Leben \fermat\es}.
\newblock {\em \dmvmitteilungenname}, 3/2001:12--26, 2001.
\newblock Extensions in \cite{fermat-birth-date}. Short versions in English are
  \makeaciteoftwo{fermatslife-english-1}{fermatslife-english-2}.

\bibitem[\protect\citeauthoryear{Barner{\protectedbarnerindex}}{2001c}]{fermat%
slife-english-1}
Klaus Barner{\protectedbarnerindex}.
\newblock How old did {\fermat} become?
\newblock {\em NTM Internationale Zeitschrift f\ue r Geschichte und Ethik der
  Naturwissenschaften, Technik und Medizin, Neue Serie, ISSN\,00366978},
  9:209--228, 2001.
\newblock Long version in German is \cite{fermatslife}. New results on the
  subject in \cite{fermat-birth-date}.

\bibitem[\protect\citeauthoryear{Barner{\protectedbarnerindex}}{2007}]{fermat-%
birth-date}
Klaus Barner{\protectedbarnerindex}.
\newblock {Neue\es\ zu \fermat\es\ Geburt\esi datum}.
\newblock {\em \dmvmitteilungenname}, 15:12--14, 2007.
\newblock \bibcrit{Further support for the results of
  \cite{fermatslife-english-1}, narrowing down \fermat's birth date from 1607/8
  to \Nov\,1607}.

\bibitem[\protect\citeauthoryear{Basin{\protectedbasinindex} \bgroup\&\ \egroup
  Walsh}{1996}]{rippling-calculus}
David Basin{\protectedbasinindex} and Toby Walsh.
\newblock A calculus for and termination of rippling.
\newblock {\em \jarname}, 16:147--180, 1996.

\bibitem[\protect\citeauthoryear{Becker}{1965}]{becker-griechische}
Oscar Becker, editor.
\newblock {\em {Zur Geschichte der griechischen Mathematik}}.
\newblock \wissenschaftlichebuchgesellschaftdarmstadt, 1965.

\bibitem[\protect\citeauthoryear{Beckert}{2005}]{fourteenthTABLEAUfive}
Bernhard Beckert, editor.
\newblock {\em {\thefourteenthTABLEAUfive}}, number 3702 in Lecture Notes in
  Artificial Intelligence. {\springerverlag}, 2005.

\bibitem[\protect\citeauthoryear{Bell \bgroup\&\ \egroup
  Thayer}{1976}]{waterfall}
Thomas~E. Bell and T.~A. Thayer.
\newblock Software requirements: Are they really a problem?
\newblock 1976.
\newblock In \cite[\PP{61}{68}]{secondICSEseventysix},
  \url{http://pdf.aminer.org/000/361/405/software_requirements_are_they_really%
_a_problem.pdf}.

\bibitem[\protect\citeauthoryear{Benz{\-}m{\ue}ller{\protectedbenzmuellerindex}
  \bgroup\&al.\egroup }{2008}]{C26}
{\mbox{Ch}}ristoph Benz{\-}m{\ue}ller{\protectedbenzmuellerindex}, Frank
  Theiss, Lawrence~C. Paulson, and Arnaud Fietz{\-}ke.
\newblock {\LEOII} --- a cooperative automatic theorem prover for higher-order
  logic.
\newblock 2008.
\newblock In \cite[\PP{162}{170}]{fourthIJCAReight}.

\bibitem[\protect\citeauthoryear{Berka \bgroup\&\ \egroup
  Kreiser}{1973}]{logiktexte}
Karel Berka and Lothar Kreiser, editors.
\newblock {\em {Logik-Texte -- Kommentierte Au\esi wahl zur Geschichte der
  modernen Logik}}.
\newblock \akademieverlag, 1973.
\newblock \nth 2\,\rev\,\edn\ (\nth 1\,\edn\,1971; \nth
  4\,\rev\,\rev\,\edn\,1986).

\bibitem[\protect\citeauthoryear{Bevers \bgroup\&\ \egroup
  Lewi}{1990}]{bevers&lewi}
Eddy Bevers and Johan Lewi.
\newblock Proof by consistency in conditional equational theories.
\newblock \Tech\ Report CW 102, \Dept\ \Comp\ \Sci, K. U. Leuven, 1990.
\newblock \Rev\,\Jul\,1990. Short version in
  \cite[\PP{194}{205}]{secondCTRSninety}.

\bibitem[\protect\citeauthoryear{Bevier \bgroup\&al.\egroup
  }{1989}]{cli-1989-1}
William~R. Bevier, Warren~A. Hunt{\protectedhuntindex}, J~Strother
  Moore{\protect\mooreindex}, and William~D. Young.
\newblock An approach to systems verification.
\newblock {\em \jarname}, 5:411--428, 1989.

\bibitem[\protect\citeauthoryear{Bevier}{1989}]{bevier-1989}
William~R. Bevier.
\newblock Kit and the short stack.
\newblock {\em \jarname}, 5:519--530, 1989.

\bibitem[\protect\citeauthoryear{Bibel \bgroup\&\ \egroup
  Kowalski{\protectedkowalskiindex}}{1980}]{fifthCADEeighty}
Wolfgang Bibel and Robert~A. Kowalski{\protectedkowalskiindex}, editors.
\newblock {\em {\thefifthCADEeighty}}, number~87 in Lecture Notes in Computer
  Science. {\springerverlag}, 1980.

\bibitem[\protect\citeauthoryear{Biundo \bgroup\&al.\egroup }{1986}]{inka}
Susanne Biundo, Birgit Hummel, Dieter Hutter{\protectedhutterindex}, and
  {\mbox{Ch}}ristoph Walther{\protectedwaltherindex}.
\newblock The {\KA} inductive theorem proving system.
\newblock 1986.
\newblock In \cite[\PP{673}{675}]{eighthCADEeightysix}.

\bibitem[\protect\citeauthoryear{Bledsoe{\protect\bledsoeindex} \bgroup\&\
  \egroup Loveland}{1984}]{after-25-years}
W.~W. Bledsoe{\protect\bledsoeindex} and Donald~W. Loveland, editors.
\newblock {\em Automated Theorem Proving: After 25 Years}.
\newblock Number~29 in Contemporary Mathematics. \AMSname, Providence (RI),
  1984.
\newblock \Proc\ of the Special Session on Automatic Theorem Proving,
  \nth{89}\,Annual Meeting of the \AMSname, Denver (CO), \Jan\,1983.

\bibitem[\protect\citeauthoryear{Bledsoe{\protect\bledsoeindex}
  \bgroup\&al.\egroup }{1971}]{BBH72-short}
W.~W. Bledsoe{\protect\bledsoeindex}, Robert~S. Boyer{\protect\boyerindex}, and
  William~H. Henneman.
\newblock Computer proofs of limit theorems.
\newblock 1971.
\newblock In \cite[\PP{586}{600}]{ijcai2}. Long version is \cite{BBH72}.

\bibitem[\protect\citeauthoryear{Bledsoe{\protect\bledsoeindex}
  \bgroup\&al.\egroup }{1972}]{BBH72}
W.~W. Bledsoe{\protect\bledsoeindex}, Robert~S. Boyer{\protect\boyerindex}, and
  William~H. Henneman.
\newblock Computer proofs of limit theorems.
\newblock {\em \artificialintelligencename}, 3:27--60, 1972.
\newblock Short version is \cite{BBH72-short}.

\bibitem[\protect\citeauthoryear{Bledsoe{\protect\bledsoeindex}}{1971}]{Ble71}
W.~W. Bledsoe{\protect\bledsoeindex}.
\newblock Splitting and reduction heuristics in automatic theorem proving.
\newblock {\em \artificialintelligencename}, 2:55--77, 1971.

\bibitem[\protect\citeauthoryear{Bouajjani \bgroup\&\ \egroup
  Maler}{2009}]{twentyfirstCAVnine}
Ahmed Bouajjani and Oded Maler, editors.
\newblock {\em \Proc\ \nth{21} \Int\ \Conf\ on Computer Aided Verification
  (CAV), Grenoble (France), 2009}, volume 5643 of {\em Lecture Notes in
  Computer Science}. {\springerverlag}, 2009.

\bibitem[\protect\citeauthoryear{Bouhoula \bgroup\&\ \egroup
  Rusinowitch}{1995}]{spike}
Adel Bouhoula and Micha{\"e}l Rusinowitch.
\newblock Implicit induction in conditional theories.
\newblock {\em \jarname}, 14:189--235, 1995.

\bibitem[\protect\citeauthoryear{Bourbaki{\protectedbourbakiindex}}{1939}]{bou%
rbaki-set-theory-results-1st-edn}
Nicola{\es} Bourbaki{\protectedbourbakiindex}.
\newblock {\em {\'El\'ements des Math\'ematique --- Livre\,1: Th\'eorie des
  Ensembles. Fascicule De R\'esultats}}.
\newblock Number 846 in \hermannparisseries. \hermannparis, 1939.
\newblock \nth 1\,\edn, \PPcount{VIII\,+\,50}. Review is \cite{churchbourbaki}.
  \nth 2 \rev\,\extd\,\edn\ is \cite{bourbaki-set-theory-results-2nd-edn}.

\bibitem[\protect\citeauthoryear{Bourbaki{\protectedbourbakiindex}}{1951}]{bou%
rbaki-set-theory-results-2nd-edn}
Nicola{\es} Bourbaki{\protectedbourbakiindex}.
\newblock {\em {\'El\'ements des Math\'ematique --- Livre\,1: Th\'eorie des
  Ensembles. Fascicule De R\'esultats}}.
\newblock Number 846-1141 in \hermannparisseries. \hermannparis, 1951.
\newblock \nth 2\,\rev\,\extd\,\edn\,of
  \cite{bourbaki-set-theory-results-1st-edn}. \nth 3 \rev\,\extd\,\edn\ is
  \cite{bourbaki-set-theory-results-3rd-edn}.

\bibitem[\protect\citeauthoryear{Bourbaki{\protectedbourbakiindex}}{1954}]{bou%
rbaki-set-theory-chapter-1-2-1st-edn}
Nicola{\es} Bourbaki{\protectedbourbakiindex}.
\newblock {\em {\'El\'ements des Math\'ematique --- Livre\,1: Th\'eorie des
  Ensembles. Chapitre\,I\,\&\,II}}.
\newblock Number 1212 in \hermannparisseries. \hermannparis, 1954.
\newblock \nth 1\,\edn. \nth 2\,\rev\,\edn\ is
  \cite{bourbaki-set-theory-chapter-1-2-2nd-edn}.

\bibitem[\protect\citeauthoryear{Bourbaki{\protectedbourbakiindex}}{1956}]{bou%
rbaki-set-theory-chapter-3-1st-edn}
Nicola{\es} Bourbaki{\protectedbourbakiindex}.
\newblock {\em {\'El\'ements des Math\'ematique --- Livre\,1: Th\'eorie des
  Ensembles. Chapitre\,III}}.
\newblock Number 1243 in \hermannparisseries. \hermannparis, 1956.
\newblock \nth 1\,\edn,
  \PPcount{II\,+\,119\,+\,4\,(mode\,d'emploi)\,+\,23\,(errata\,{no.}\,6)}. \nth
  2\,\rev\,\extd\,\edn\ is \cite{bourbaki-set-theory-chapter-3-2nd-edn}.

\bibitem[\protect\citeauthoryear{Bourbaki{\protectedbourbakiindex}}{1958a}]{bo%
urbaki-set-theory-chapter-4-1st-edn}
Nicola{\es} Bourbaki{\protectedbourbakiindex}.
\newblock {\em {\'El\'ements des Math\'ematique --- Livre\,1: Th\'eorie des
  Ensembles. Chapitre\,IV}}.
\newblock Number 1258 in \hermannparisseries. \hermannparis, 1958.
\newblock \nth 1\,\edn. \nth 2\,\rev\,\extd\,\edn\ is
  \cite{bourbaki-set-theory-chapter-4-2nd-edn}.

\bibitem[\protect\citeauthoryear{Bourbaki{\protectedbourbakiindex}}{1958b}]{bo%
urbaki-set-theory-results-3rd-edn}
Nicola{\es} Bourbaki{\protectedbourbakiindex}.
\newblock {\em {\'El\'ements des Math\'ematique --- Livre\,1: Th\'eorie des
  Ensembles. Fascicule De R\'esultats}}.
\newblock Number 1141 in \hermannparisseries. \hermannparis, 1958.
\newblock \nth 3\,\rev\,\extd\,\edn\,of
  \cite{bourbaki-set-theory-results-2nd-edn}. \nth 4 \rev\,\extd\,\edn\ is
  \cite{bourbaki-set-theory-results-4th-edn}.

\bibitem[\protect\citeauthoryear{Bourbaki{\protectedbourbakiindex}}{1960}]{bou%
rbaki-set-theory-chapter-1-2-2nd-edn}
Nicola{\es} Bourbaki{\protectedbourbakiindex}.
\newblock {\em {\'El\'ements des Math\'ematique --- Livre\,1: Th\'eorie des
  Ensembles. Chapitre\,I\,\&\,II}}.
\newblock Number 1212 in \hermannparisseries. \hermannparis, 1960.
\newblock \nth 2\,\rev\,\extd\,\edn\,of
  \cite{bourbaki-set-theory-chapter-1-2-1st-edn}. \nth 3\,\rev\,\edn\ is
  \cite{bourbaki-set-theory-chapter-1-2-3rd-edn}.

\bibitem[\protect\citeauthoryear{Bourbaki{\protectedbourbakiindex}}{1964}]{bou%
rbaki-set-theory-results-4th-edn}
Nicola{\es} Bourbaki{\protectedbourbakiindex}.
\newblock {\em {\'El\'ements des Math\'ematique --- Livre\,1: Th\'eorie des
  Ensembles. Fascicule De R\'esultats}}.
\newblock Number 1141 in \hermannparisseries. \hermannparis, 1964.
\newblock \nth 4\,\rev\,\extd\,\edn\,of
  \cite{bourbaki-set-theory-results-3rd-edn}. \nth 5 \rev\,\extd\,\edn\ is
  \cite{bourbaki-set-theory-results-5th-edn}.

\bibitem[\protect\citeauthoryear{Bourbaki{\protectedbourbakiindex}}{1966a}]{bo%
urbaki-set-theory-chapter-4-2nd-edn}
Nicola{\es} Bourbaki{\protectedbourbakiindex}.
\newblock {\em {\'El\'ements des Math\'ematique --- Livre\,1: Th\'eorie des
  Ensembles. Chapitre\,IV}}.
\newblock Number 1258 in \hermannparisseries. \hermannparis, 1966.
\newblock \nth 2\,\rev\,\extd\,\edn\,of
  \cite{bourbaki-set-theory-chapter-4-1st-edn}. English translation in
  \cite{bourbaki-english}.

\bibitem[\protect\citeauthoryear{Bourbaki{\protectedbourbakiindex}}{1966b}]{bo%
urbaki-set-theory-chapter-1-2-3rd-edn}
Nicola{\es} Bourbaki{\protectedbourbakiindex}.
\newblock {\em {\'El\'ements des Math\'ematique --- Livre\,1: Th\'eorie des
  Ensembles. Chapitres\,I\,\&\,II}}.
\newblock Number 1212 in \hermannparisseries. \hermannparis, 1966.
\newblock \nth 3\,\rev\,\edn\,of
  \cite{bourbaki-set-theory-chapter-1-2-2nd-edn},
  \PPcount{VI\,+\,143\,+\,7\,(errata\,{no.}\,13)}. English translation in
  \cite{bourbaki-english}.

\bibitem[\protect\citeauthoryear{Bourbaki{\protectedbourbakiindex}}{1967}]{bou%
rbaki-set-theory-chapter-3-2nd-edn}
Nicola{\es} Bourbaki{\protectedbourbakiindex}.
\newblock {\em {\'El\'ements des Math\'ematique --- Livre\,1: Th\'eorie des
  Ensembles. Chapitre\,III}}.
\newblock Number 1243 in \hermannparisseries. \hermannparis, 1967.
\newblock \nth 2\,\rev\,\extd\,\edn\,of
  \cite{bourbaki-set-theory-chapter-3-1st-edn},
  \PPcount{151\,+\,7\,(errata\,{no.}\,13)}. \nth 3\,\rev\,\edn\ results from
  executing these errata. English translation in \cite{bourbaki-english}.

\bibitem[\protect\citeauthoryear{Bourbaki{\protectedbourbakiindex}}{1968a}]{bo%
urbaki-english}
Nicola{\es} Bourbaki{\protectedbourbakiindex}.
\newblock {\em Elements of Mathematics --- Theory of Sets}.
\newblock \hermannparisseries. \hermannparis, jointly published with
  \mbox{AdiWes} International Series in Mathematics, \addisonwesley, 1968.
\newblock English translation of
  \makeaciteoffour{bourbaki-set-theory-chapter-1-2-3rd-edn}{bourbaki-set-theor%
y-chapter-3-2nd-edn}{bourbaki-set-theory-chapter-4-2nd-edn}{bourbaki-set-theor%
y-results-5th-edn}.

\bibitem[\protect\citeauthoryear{Bourbaki{\protectedbourbakiindex}}{1968b}]{bo%
urbaki-set-theory-results-5th-edn}
Nicola{\es} Bourbaki{\protectedbourbakiindex}.
\newblock {\em {\'El\'ements des Math\'ematique --- Livre\,1: Th\'eorie des
  Ensembles. Fascicule De R\'esultats}}.
\newblock Number 1141 in \hermannparisseries. \hermannparis, 1968.
\newblock \nth 5\,\rev\,\extd\,\edn\,of
  \cite{bourbaki-set-theory-results-4th-edn}. English translation in
  \cite{bourbaki-english}.

\bibitem[\protect\citeauthoryear{Boyer{\protect\boyerindex} \bgroup\&\ \egroup
  Moore{\protect\mooreindex}}{1971}]{boyer-moore-1971}
Robert~S. Boyer{\protect\boyerindex} and J~Strother Moore{\protect\mooreindex}.
\newblock The sharing of structure in resolution programs.
\newblock Memo~47, \uniEBshort, \Dept\ of Computational Logic, 1971.
\newblock \PPcount{II\,+\,24}. Revised version is \cite{boyer-moore-1972}.

\bibitem[\protect\citeauthoryear{Boyer{\protect\boyerindex} \bgroup\&\ \egroup
  Moore{\protect\mooreindex}}{1972}]{boyer-moore-1972}
Robert~S. Boyer{\protect\boyerindex} and J~Strother Moore{\protect\mooreindex}.
\newblock The sharing of structure in theorem-proving programs.
\newblock 1972.
\newblock In \cite[\PP{101}{116}]{machine-intelligence-7}.

\bibitem[\protect\citeauthoryear{Boyer{\protect\boyerindex} \bgroup\&\ \egroup
  Moore{\protect\mooreindex}}{1973}]{boyer-moore-1973}
Robert~S. Boyer{\protect\boyerindex} and J~Strother Moore{\protect\mooreindex}.
\newblock Proving theorems about {LISP} functions.
\newblock 1973.
\newblock In \cite[\PP{486}{493}]{ijcai3}.
  \url{http://ijcai.org/Past%20Proceedings/IJCAI-73/PDF/053.pdf}. \Rev\
  version, \extd\ with a section ``Failures\closequotecomma is
  \cite{boyer-moore-1975}.

\bibitem[\protect\citeauthoryear{Boyer{\protect\boyerindex} \bgroup\&\ \egroup
  Moore{\protect\mooreindex}}{1975}]{boyer-moore-1975}
Robert~S. Boyer{\protect\boyerindex} and J~Strother Moore{\protect\mooreindex}.
\newblock Proving theorems about {LISP} functions.
\newblock {\em {\jacmname}}, 22:129--144, 1975.
\newblock \Rev\,\extd\,\edn\,of \cite{boyer-moore-1973}. Received \Sep\,1973,
  \Rev\ \Apr\,1974.

\bibitem[\protect\citeauthoryear{Boyer{\protect\boyerindex\bmfssapindex}
  \bgroup\&\ \egroup
  Moore{\protect\mooreindex}}{1977a}]{boyer-moore-fast-string-searching}
Robert~S. Boyer{\protect\boyerindex\bmfssapindex} and J~Strother
  Moore{\protect\mooreindex}.
\newblock A fast string searching algorithm.
\newblock {\em \commacmname}, 20:762--772, 1977.
\newblock \url{http://doi.acm.org/10.1145/359842.359859}.

\bibitem[\protect\citeauthoryear{Boyer{\protect\boyerindex} \bgroup\&\ \egroup
  Moore{\protect\mooreindex}}{1977b}]{boyer-moore-1977}
Robert~S. Boyer{\protect\boyerindex} and J~Strother Moore{\protect\mooreindex}.
\newblock A lemma driven automatic theorem prover for recursive function
  theory.
\newblock 1977.
\newblock In \cite[\Vol\,I, \PP{511}{519}]{ijcai5}.
  \url{http://ijcai.org/Past%20Proceedings/IJCAI-77-VOL1/PDF/089.pdf}.

\bibitem[\protect\citeauthoryear{Boyer{\protect\boyerindex} \bgroup\&\ \egroup
  Moore{\protect\mooreindex}}{1979}]{bm}
Robert~S. Boyer{\protect\boyerindex} and J~Strother Moore{\protect\mooreindex}.
\newblock {\em A Computational Logic}.
\newblock \academicpress, 1979.
\newblock \url{http://www.cs.utexas.edu/users/boyer/acl.text}.

\bibitem[\protect\citeauthoryear{Boyer{\protect\boyerindex} \bgroup\&\ \egroup
  Moore{\protect\mooreindex}}{1981a}]{boyer-moore-1981-editors}
Robert~S. Boyer{\protect\boyerindex} and J~Strother Moore{\protect\mooreindex},
  editors.
\newblock {\em The Correctness Problem in Computer Science}.
\newblock \academicpress, 1981.

\bibitem[\protect\citeauthoryear{Boyer{\protect\boyerindex} \bgroup\&\ \egroup
  Moore{\protect\mooreindex}}{1981b}]{boyer-moore-1981-authors}
Robert~S. Boyer{\protect\boyerindex} and J~Strother Moore{\protect\mooreindex}.
\newblock Metafunctions: Proving them correct and using them efficiently as new
  proof procedures.
\newblock 1981.
\newblock In \cite[\PP{103}{184}]{boyer-moore-1981-editors}.

\bibitem[\protect\citeauthoryear{Boyer{\protect\boyerindex} \bgroup\&\ \egroup
  Moore{\protect\mooreindex}}{1984a}]{boyer-moore-turing-completeness-lisp}
Robert~S. Boyer{\protect\boyerindex} and J~Strother Moore{\protect\mooreindex}.
\newblock A mechanical proof of the {\turing} completeness of pure {\LISP}.
\newblock 1984.
\newblock In \cite[\PP{133}{167}]{after-25-years}.

\bibitem[\protect\citeauthoryear{Boyer{\protect\boyerindex} \bgroup\&\ \egroup
  Moore{\protect\mooreindex}}{1984b}]{boyer-moore-halting-problem}
Robert~S. Boyer{\protect\boyerindex} and J~Strother Moore{\protect\mooreindex}.
\newblock A mechanical proof of the unsolvability of the halting problem.
\newblock {\em \jacmname}, 31:441--458, 1984.

\bibitem[\protect\citeauthoryear{Boyer{\protect\boyerindex} \bgroup\&\ \egroup
  Moore{\protect\mooreindex}}{1984c}]{boyer-moore-proof-checking-RSA}
Robert~S. Boyer{\protect\boyerindex} and J~Strother Moore{\protect\mooreindex}.
\newblock Proof checking the {RSA} public key encryption algorithm.
\newblock {\em {\americanmathematicalmonthlyname}}, 91:181--189, 1984.

\bibitem[\protect\citeauthoryear{Boyer{\protect\boyerindex} \bgroup\&\ \egroup
  Moore{\protect\mooreindex}}{1985}]{boyer-moore-1985}
Robert~S. Boyer{\protect\boyerindex} and J~Strother Moore{\protect\mooreindex}.
\newblock Program verification.
\newblock {\em \jarname}, 1:17--23, 1985.

\bibitem[\protect\citeauthoryear{Boyer{\protect\boyerindex} \bgroup\&\ \egroup
  Moore{\protect\mooreindex}}{1987}]{bmeval-1987}
Robert~S. Boyer{\protect\boyerindex} and J~Strother Moore{\protect\mooreindex}.
\newblock The addition of bounded quantification and partial functions to a
  computational logic and its theorem prover.
\newblock Technical Report ICSCA-CMP-52, \Inst\ for Computing Science and
  Computing Applications, \unitexasaustin, 1987.
\newblock Printed \Jan\,1987\@. Also published as
  \makeaciteoftwo{bmeval-1988}{bmeval-1989}.

\bibitem[\protect\citeauthoryear{Boyer{\protect\boyerindex} \bgroup\&\ \egroup
  Moore{\protect\mooreindex}}{1988a}]{bmeval-1988}
Robert~S. Boyer{\protect\boyerindex} and J~Strother Moore{\protect\mooreindex}.
\newblock The addition of bounded quantification and partial functions to a
  computational logic and its theorem prover.
\newblock {\em {\jarname}}, 4:117--172, 1988.
\newblock Received \Feb\,11,\,1987. Also pubished as
  \makeaciteoftwo{bmeval-1987}{bmeval-1989}.

\bibitem[\protect\citeauthoryear{Boyer{\protect\boyerindex} \bgroup\&\ \egroup
  Moore{\protect\mooreindex}}{1988b}]{boyermoore}
Robert~S. Boyer{\protect\boyerindex} and J~Strother Moore{\protect\mooreindex}.
\newblock {\em A Computational Logic Handbook}.
\newblock Number~23 in Perspectives in Computing. \academicpress, 1988.
\newblock \nth 2\,\rev\,\extd\,\edn\ is \cite{boyermooresecondedition}.

\bibitem[\protect\citeauthoryear{Boyer{\protect\boyerindex} \bgroup\&\ \egroup
  Moore{\protect\mooreindex}}{1988c}]{boyer-moore-1988}
Robert~S. Boyer{\protect\boyerindex} and J~Strother Moore{\protect\mooreindex}.
\newblock Integrating decision procedures into heuristic theorem provers: A
  case study of {\index{linear arithmetic}}linear arithmetic.
\newblock 1988.
\newblock In \cite[\PP{83}{124}]{machine-intelligence-11}.

\bibitem[\protect\citeauthoryear{Boyer{\protect\boyerindex} \bgroup\&\ \egroup
  Moore{\protect\mooreindex}}{1989}]{bmeval-1989}
Robert~S. Boyer{\protect\boyerindex} and J~Strother Moore{\protect\mooreindex}.
\newblock The addition of bounded quantification and partial functions to a
  computational logic and its theorem prover.
\newblock 1989.
\newblock In \cite[\PP{95}{145}]{broy-1989} (received \Jan\,1988). Also
  published as \makeaciteoftwo{bmeval-1987}{bmeval-1988}.

\bibitem[\protect\citeauthoryear{Boyer{\protect\boyerindex} \bgroup\&\ \egroup
  Moore{\protect\mooreindex}}{1990}]{BM90}
Robert~S. Boyer{\protect\boyerindex} and J~Strother Moore{\protect\mooreindex}.
\newblock A theorem prover for a computational logic.
\newblock 1990.
\newblock In \cite[\PP{1}{15}]{tenthCADEninety}.

\bibitem[\protect\citeauthoryear{Boyer{\protect\boyerindex} \bgroup\&\ \egroup
  Moore{\protect\mooreindex}}{1998}]{boyermooresecondedition}
Robert~S. Boyer{\protect\boyerindex} and J~Strother Moore{\protect\mooreindex}.
\newblock {\em A Computational Logic Handbook}.
\newblock International Series in Formal Methods. \academicpress, 1998.
\newblock \nth 2\,\rev\,\extd\,\edn\,of \cite{boyermoore}, \rev\ to work with
  \NQTHM--1992, a new version of \NQTHM.

\bibitem[\protect\citeauthoryear{Boyer{\protect\boyerindex} \bgroup\&\ \egroup
  Yu}{1992}]{boyer-yu-1992}
Robert~S. Boyer{\protect\boyerindex} and Yuan Yu.
\newblock Automated correctness proofs of machine code programs for a
  commercial microprocessor.
\newblock 1992.
\newblock In \cite[416--430]{eleventhCADEninetytwo}.

\bibitem[\protect\citeauthoryear{Boyer{\protect\boyerindex} \bgroup\&\ \egroup
  Yu}{1996}]{c-string-lib-1996}
Robert~S. Boyer{\protect\boyerindex} and Yuan Yu.
\newblock Automated proofs of object code for a widely used microprocessor.
\newblock {\em \jacmname}, 43:166--192, 1996.

\bibitem[\protect\citeauthoryear{Boyer{\protect\boyerindex} \bgroup\&al.\egroup
  }{1973}]{boyer-moore-davies-1973}
Robert~S. Boyer{\protect\boyerindex}, D.~Julian~M. Davies, and J~Strother
  Moore{\protect\mooreindex}.
\newblock The 77-editor.
\newblock Memo~62, \uniEBshort, \Dept\ of Computational Logic, 1973.

\bibitem[\protect\citeauthoryear{Boyer{\protect\boyerindex} \bgroup\&al.\egroup
  }{1976}]{boyer-moore-shostak}
Robert~S. Boyer{\protect\boyerindex}, J~Strother Moore{\protect\mooreindex},
  and Robert~E. Shostak.
\newblock Primitive recursive program transformations.
\newblock 1976.
\newblock In \cite[\PP{171}{174}]{thirdPOPLseventysix}.
  \url{http://doi.acm.org/10.1145/800168.811550}.

\bibitem[\protect\citeauthoryear{Boyer{\protect\boyerindex}}{1971}]{Boy71}
Robert~S. Boyer{\protect\boyerindex}.
\newblock {\em Locking: a restriction of resolution}.
\newblock PhD thesis, \unitexasaustin, 1971.

\bibitem[\protect\citeauthoryear{Boyer{\protect\boyerindex}}{2012}]{boyer-2012}
Robert~S. Boyer{\protect\boyerindex}.
\newblock {\EMAIL\ to \wirthname, \Nov\,19,\hskip-.28em}.
\newblock 2012.

\bibitem[\protect\citeauthoryear{Brock \bgroup\&\ \egroup
  Hunt{\protectedhuntindex}}{1999}]{brock-hunt-1999}
Bishop Brock and Warren~A. Hunt{\protectedhuntindex}.
\newblock Formal analysis of the {Motorola CAP DSP}.
\newblock 1999.
\newblock In \cite[\PP{81}{116}]{industrial-strength-1999}.

\bibitem[\protect\citeauthoryear{Brotherston \bgroup\&\ \egroup
  Simpson}{2007}]{DBLP:conf/lics/BrotherstonS07}
James Brotherston and Alex Simpson.
\newblock Complete sequent calculi for induction and infinite descent.
\newblock 2007.
\newblock In \cite[\PP{51}{62?}]{lics22}. Thoroughly \rev\ version in
  \cite{brotherston-cut-elimination}.

\bibitem[\protect\citeauthoryear{Brotherston \bgroup\&\ \egroup
  Simpson}{2011}]{brotherston-cut-elimination}
James Brotherston and Alex Simpson.
\newblock Sequent calculi for induction and infinite descent.
\newblock {\em {\jlcname}}, 21:1177--1216, 2011.
\newblock Thoroughly \rev\ version of \cite{DBLP:conf/tableaux/Brotherston05}
  and \cite{DBLP:conf/lics/BrotherstonS07}. Received \Apr\,3, 2009. Published
  online \Sep\,30, 2010, \url{http://dx.doi.org/10.1093/logcom/exq052}.

\bibitem[\protect\citeauthoryear{Brotherston}{2005}]{DBLP:conf/tableaux/Brothe%
rston05}
James Brotherston.
\newblock Cyclic proofs for first-order logic with inductive definitions.
\newblock 2005.
\newblock In \cite[\PP{78}{92}]{fourteenthTABLEAUfive}. Thoroughly \rev\
  version in \cite{brotherston-cut-elimination}.

\bibitem[\protect\citeauthoryear{Brown}{2012}]{satallax}
Chad~E. Brown.
\newblock {\SATALLAX}: An automatic higher-order prover.
\newblock 2012.
\newblock In \cite[\PP{111}{117}]{sixthIJCARtwelve}.

\bibitem[\protect\citeauthoryear{Broy}{1989}]{broy-1989}
Manfred Broy, editor.
\newblock {\em Constructive Methods in Computing Science}, number {F\,55} in
  NATO ASI Series. {\springerverlag}, 1989.

\bibitem[\protect\citeauthoryear{Buch \bgroup\&\ \egroup
  Hillenbrand{\protectedhillenbrandindex}}{1996}]{waldmeister}
Armin Buch and Thomas Hillenbrand{\protectedhillenbrandindex}.
\newblock {\em {\WALDMEISTER}: Development of a High Performance
  Completion-Based Theorem Prover}.
\newblock {SEKI-Report SR--96--01 (ISSN 1860--5931)}. {SEKI Publications},
  \FBinfshort, \uniKLshort, 1996.
\newblock \url{agent.informatik.uni-kl.de/seki/1996/Buch.SR-96-01.ps.gz}.

\bibitem[\protect\citeauthoryear{Bundy{\protectedbundyindex}
  \bgroup\&al.\egroup }{1989}]{bundy-recursion-analysis}
Alan Bundy{\protectedbundyindex}, Frank~van Harmelen{\protectedharmelenindex},
  Jane Hesketh, Alan Smaill, and Andrew Stevens.
\newblock A rational reconstruction and extension of recursion analysis.
\newblock 1989.
\newblock In \cite[\PP{359}{365}]{ijcai11}.

\bibitem[\protect\citeauthoryear{Bundy{\protectedbundyindex}
  \bgroup\&al.\egroup }{1990}]{oysterclam}
Alan Bundy{\protectedbundyindex}, Frank~van Harmelen{\protectedharmelenindex},
  Christian Horn, and Alan Smaill.
\newblock The {\OYSTERCLAM} system.
\newblock 1990.
\newblock In \cite[\PP{647}{648}]{tenthCADEninety}.

\bibitem[\protect\citeauthoryear{Bundy{\protectedbundyindex}
  \bgroup\&al.\egroup }{1991}]{rippling}
Alan Bundy{\protectedbundyindex}, Andrew Stevens, Frank~van
  Harmelen{\protectedharmelenindex}, Andrew Ireland{\protectedirelandindex},
  and Alan Smaill.
\newblock {\em Rippling: A Heuristic for Guiding Inductive Proofs}.
\newblock 1991.
\newblock \daireport{567}\@. Also in
  \artificialintelligenceprintyear{1993}{62}{185}{253}.

\bibitem[\protect\citeauthoryear{Bundy{\protectedbundyindex}
  \bgroup\&al.\egroup }{2005}]{rippling-book}
Alan Bundy{\protectedbundyindex}, Dieter Hutter{\protectedhutterindex}, David
  Basin{\protectedbasinindex}, and Andrew Ireland{\protectedirelandindex}.
\newblock {\em Rippling: Meta-Level Guidance for Mathematical Reasoning}.
\newblock \cambridgeunipress, 2005.

\bibitem[\protect\citeauthoryear{Bundy{\protectedbundyindex}}{1988}]{bundy-ind%
uctive-proof-planning}
Alan Bundy{\protectedbundyindex}.
\newblock {\em The use of Explicit Plans to Guide Inductive Proofs}.
\newblock 1988.
\newblock \daireport{349}\@. Short version in
  \cite[\PP{111}{120}]{ninthCADEeightyeight}.

\bibitem[\protect\citeauthoryear{Bundy{\protectedbundyindex}}{1989}]{science-o%
f-reasoning}
Alan Bundy{\protectedbundyindex}.
\newblock {\em A Science of Reasoning}.
\newblock 1989.
\newblock \daireport{445}\@. Also in \cite[\PP{178}{198}]{honor-robinson}.

\bibitem[\protect\citeauthoryear{Bundy{\protectedbundyindex}}{1994}]{twelvethC%
ADEninetyfour}
Alan Bundy{\protectedbundyindex}, editor.
\newblock {\em {\thetwelvethCADEninetyfour}}, number 814 in Lecture Notes in
  Artificial Intelligence. {\springerverlag}, 1994.

\bibitem[\protect\citeauthoryear{Bundy{\protectedbundyindex}}{1999}]{bundy-sur%
vey}
Alan Bundy{\protectedbundyindex}.
\newblock {\em The Automation of Proof by Mathematical Induction}.
\newblock Informatics Research Report \Numero\,2, Division of Informatics,
  \uniEBshort, 1999.
\newblock Also in \cite[\Vol\,1, \PP{845}{911}]{HandbookAR}.

\bibitem[\protect\citeauthoryear{Burstall{\protectedburstallindex}
  \bgroup\&al.\egroup }{1971}]{pop2}
Rod~M. Burstall{\protectedburstallindex}, John~S. Collins, and Robin~J.
  Popplestone.
\newblock {\em Programming in\/ {\sc POP--2}}.
\newblock \uniEBshort\ Press, 1971.

\bibitem[\protect\citeauthoryear{Burstall{\protectedburstallindex}}{1969}]{bur%
stall-1969}
Rod~M. Burstall{\protectedburstallindex}.
\newblock Proving properties of programs by structural induction.
\newblock {\em The Computer Journal}, 12:48--51, 1969.
\newblock Received \Apr\,1968, \rev\,\Aug\,1968.

\bibitem[\protect\citeauthoryear{Bussey}{1917}]{maurolycus}
W.~H. Bussey.
\newblock The origin of mathematical induction.
\newblock {\em {\americanmathematicalmonthlyname}}, XXIV:199--207, 1917.

\bibitem[\protect\citeauthoryear{Bussotti}{2006}]{From-Fermat-to-Gauss}
Paolo Bussotti.
\newblock {\em {From \fermat\ to \gauss: indefinite descent and methods of
  reduction in number theory}}.
\newblock Number~55 in Algorismus. \Dr\ Erwin Rauner Verlag, \Augsburg, 2006.

\bibitem[\protect\citeauthoryear{Cajori}{1918}]{cajori-1918}
Florian Cajori.
\newblock Origin of the name ``mathematical induction".
\newblock {\em {\americanmathematicalmonthlyname}}, 25:197--201, 1918.

\bibitem[\protect\citeauthoryear{Church}{1946}]{churchbourbaki}
Alonzo Church.
\newblock {Review of \cite{bourbaki-set-theory-results-1st-edn}}.
\newblock {\em \jslname}, 11:91, 1946.

\bibitem[\protect\citeauthoryear{Clocksin \bgroup\&\ \egroup
  Mellish}{2003}]{Prolog}
William~F. Clocksin and Christopher~S. Mellish.
\newblock {\em {Programming in \PROLOG}}.
\newblock {\springerverlag}, 2003.
\newblock \nth 5\,\edn\ (\nth 1\,\edn\,1981).

\bibitem[\protect\citeauthoryear{Cohn}{1965}]{cohn-1965}
Paul~Moritz Cohn.
\newblock {\em Universal Algebra}.
\newblock Harper \& Row, \NewYork, 1965.
\newblock \nth 1\,\edn. \nth 2\,\rev\,\edn\ is \cite{cohn-1981}.

\bibitem[\protect\citeauthoryear{Cohn}{1981}]{cohn-1981}
Paul~Moritz Cohn.
\newblock {\em Universal Algebra}.
\newblock Number~6 in Mathematics and Its Applications. \DReidelpublishing,
  1981.
\newblock \nth 2\,\rev\,\edn\ (\nth 1\,\edn\ is \cite{cohn-1965}).

\bibitem[\protect\citeauthoryear{Comon}{1997}]{eighthRTAninetyseven}
Hubert Comon, editor.
\newblock {\em {\theeighthRTAninetyseven}}, number 1232 in Lecture Notes in
  Computer Science. {\springerverlag}, 1997.

\bibitem[\protect\citeauthoryear{Comon}{2001}]{comonAR}
Hubert Comon.
\newblock Inductionless induction.
\newblock 2001.
\newblock In \cite[\Vol\,I, \PP{913}{970}]{HandbookAR}.

\bibitem[\protect\citeauthoryear{Constable \bgroup\&al.\egroup
  }{1985}]{nuprl-book}
Robert~L. Constable, Stuart~F. Allen, H.~M. Bromly, W.~R. Cleaveland, J.~F.
  Cremer, R.~W. Harper, Douglas~J. Howe, T.~B. Knoblock, N.~P. Mendler,
  P.~Panangaden, James~T. Sasaki, and Scott~F. Smith.
\newblock {\em Implementing Mathematics with {the\/ \NUPRL} Proof Development
  System}.
\newblock \PrenticeHall, 1985.
\newblock \url{http://www.nuprl.org/book}.

\bibitem[\protect\citeauthoryear{Cooper}{1971}]{ijcai2}
D.~C. Cooper, editor.
\newblock {\em {\Proc\ \nth 2 \IJCAIname\ (\IJCAI), \Sep\,1971, Imperial
  College, \London}}. \morgankaufmannwithoutadds, Los Altos (CA), 1971.
\newblock
  \url{http://ijcai.org/Past%20Proceedings/IJCAI-1971/CONTENT/content.htm}.

\bibitem[\protect\citeauthoryear{DAC}{2001}]{dac/2001}
{\em \Proc\ \nth{38} Design Automation Conference (DAC), Las Vegas (NV), 2001}.
  \acmpress, 2001.

\bibitem[\protect\citeauthoryear{Darlington}{1968}]{darlington-1968}
Jared~L. Darlington.
\newblock Automated theorem proving with equality substitutions and
  mathematical induction.
\newblock 1968.
\newblock In \cite[\PP{113}{127}]{machine-intelligence-3}.

\bibitem[\protect\citeauthoryear{Davis}{2009}]{davis-2009}
Jared Davis.
\newblock {\em A Self-Verifying Theorem Prover}.
\newblock PhD thesis, \unitexasaustin, 2009.

\bibitem[\protect\citeauthoryear{Dedekind{\protect\dedekindindex}}{1888}]{dede%
kind-1888}
Richard Dedekind{\protect\dedekindindex}.
\newblock {\em {Was sind und was sollen die Zahlen}}.
\newblock \vieweg, 1888.
\newblock Also in \cite[\Vol\,3, \PP{335}{391}]{dedekind-1930-32}. Also in
  \cite{dedekind-1969}.

\bibitem[\protect\citeauthoryear{Dedekind{\protect\dedekindindex}}{1930--32}]{%
dedekind-1930-32}
Richard Dedekind{\protect\dedekindindex}.
\newblock {\em {Gesammelte mathematische Werke}}.
\newblock \vieweg, 1930--32.
\newblock \Ed\ by \frickename, \noethername, and {\namefont \Oe ystein Ore}.

\bibitem[\protect\citeauthoryear{Dedekind{\protect\dedekindindex}}{1969}]{dede%
kind-1969}
Richard Dedekind{\protect\dedekindindex}.
\newblock {\em {Wa\es\ sind und wa\es\ sollen die Zahlen? Stetigkeit und
  irrationale Zahlen}}.
\newblock Friedrich Vieweg und Sohn, Braunschweig, 1969.

\bibitem[\protect\citeauthoryear{Dennis \bgroup\&al.\egroup
  }{2005}]{proofplanningsystems}
Louise~A. Dennis, Mateja Jamnik, and Martin Pollet.
\newblock On the comparison of proof planning systems {\LAMBDACLAM, \OMEGA\
  and\/ \ISAPLANNER}.
\newblock {\em {\ENTCSname}}, 151:93--110, 2005.

\bibitem[\protect\citeauthoryear{Dershowitz \bgroup\&\ \egroup
  Jouannaud}{1990}]{term-rewriting-one}
Nachum Dershowitz and Jean-Pierre Jouannaud.
\newblock Rewrite systems.
\newblock 1990.
\newblock In \cite[\Vol\,B, \PP{243}{320}]{handbook-tcs}.

\bibitem[\protect\citeauthoryear{Dershowitz \bgroup\&\ \egroup
  Lindenstrauss}{1995}]{fourthCTRSninetyfour}
Nachum Dershowitz and Naomi Lindenstrauss, editors.
\newblock {\em {\thefourthCTRSninetyfour}}, number 968 in Lecture Notes in
  Computer Science, 1995.

\bibitem[\protect\citeauthoryear{Dershowitz}{1989}]{thirdRTAeightynine}
Nachum Dershowitz, editor.
\newblock {\em {\thethirdRTAeightynine}}, number 355 in Lecture Notes in
  Computer Science. {\springerverlag}, 1989.

\bibitem[\protect\citeauthoryear{Dietrich}{2011}]{dodi-diss}
Dominik Dietrich.
\newblock {\em Assertion Level Proof Planning with Compiled Strategies}.
\newblock Optimus Verlag, Alexander Mostafa, \Goettingen, 2011.
\newblock \PhDthesis, \Dept\ Informatics, \addressuniSBshort.

\bibitem[\protect\citeauthoryear{Dixon \bgroup\&\ \egroup
  Fleuriot}{2003}]{DF03-CADE-19}
Lucas Dixon and Jacques Fleuriot.
\newblock {\ISAPLANNER}: A prototype proof planner in {\ISABELLE}.
\newblock 2003.
\newblock In \cite[\PP{279}{283}]{nineteenthCADEthree}.

\bibitem[\protect\citeauthoryear{Euclid{\protect\euclidindex}}{\ca\,300\,\BC}]%
{elements}
\ignorespaces Euclid{\protect\euclidindex} of~Alexandria.
\newblock {\em Elements}.
\newblock \ca\,300\,\BC.
\newblock Web version without the figures:
  \url{http://www.perseus.tufts.edu/hopper/text?doc=Perseus:text:1999.01.0085}.
  \ English translation: \heathname\ (\ed). {\em The Thirteen Books of\/
  \euclid's Elements}. \cambridgeunipress, 1908; web version without the
  figures:
  \url{http://www.perseus.tufts.edu/hopper/text?doc=Perseus:text:1999.01.0086}.
  English web version (\incl\ figures): {\namefont D. E. Joyce} (\ed). {\em
  \euclid's Elements}.
  \url{http://aleph0.clarku.edu/~djoyce/java/elements/elements.html}, \Dept\
  \Math\ \& \Comp\ \Sci, Clark \Univ, Worcester (MA).

\bibitem[\protect\citeauthoryear{Fermat}{1891\ff}]{fermat-oeuvres}
Pierre Fermat.
\newblock {\em \OE uvres de \fermat}.
\newblock {\gauthiervillars}, 1891\ff.
\newblock \Ed\ by {\namefont Paul Tannery}, {\namefont Charles Henry}.

\bibitem[\protect\citeauthoryear{Fitting}{1990}]{Fitting90}
Melvin Fitting.
\newblock {\em First-order logic and automated theorem proving}.
\newblock {\springerverlag}, 1990.
\newblock \nth 1\,\edn\ (\nth 2\,\rev\,\edn\ is \cite{fitting}).

\bibitem[\protect\citeauthoryear{Fitting}{1996}]{fitting}
Melvin Fitting.
\newblock {\em First-order logic and automated theorem proving}.
\newblock {\springerverlag}, 1996.
\newblock \nth 2\,\rev\,\edn\ (\nth 1\,\edn\ is \cite{Fitting90}).

\bibitem[\protect\citeauthoryear{FOCS}{1980}]{focs21}
{\em {\Proc\ \nth{21} \FOCSname, Syracuse, 1980}}. \ieeepress, 1980.
\newblock \url{http://ieee-focs.org/}.

\bibitem[\protect\citeauthoryear{Fowler}{1994}]{Greeks-induction}
David Fowler.
\newblock Could the {G}reeks have used mathematical induction? {D}id they use
  it?
\newblock {\em \physisname}, XXXI(1):253--265, 1994.

\bibitem[\protect\citeauthoryear{Freudenthal}{1953}]{freudenthal}
Hans Freudenthal.
\newblock {Zur Geschichte der voll\-st\ae n\-digen Induktion}.
\newblock {\em Archives Internationales d'Histoire des Sciences}, 6:17--37,
  1953.

\bibitem[\protect\citeauthoryear{Fribourg}{1986}]{inductionlesshistb}
Laurent Fribourg.
\newblock A strong restriction of the inductive completion procedure.
\newblock 1986.
\newblock In \cite[\PP{105}{116}]{thirteenthICALPeightysix}. Also in
  \jscprintyear{1989}{8}{253}{276}.

\bibitem[\protect\citeauthoryear{Fries{\protectedfriesindex}}{1822}]{fries-vol%
lstaendige-induction}
Jakob~Friedrich Fries{\protectedfriesindex}.
\newblock {\em {Die mathematische Naturphilosophie nach philosophischer Methode
  bearbeitet -- Ein Versuch}}.
\newblock Christian Friedrich Winter, \Heidelberg, 1822.

\bibitem[\protect\citeauthoryear{Fritz}{1945}]{hippa}
Kurt~von Fritz.
\newblock The discovery of incommensurability by {\hippasosname}.
\newblock {\em \annalsofmathematicsname}, 46:242--264, 1945.
\newblock German translation: {\em Die Entdeckung der Inkommensurabilit\ae t
  durch \hippasosnamegerman} in \cite[\PP{271}{308}]{becker-griechische}.

\bibitem[\protect\citeauthoryear{Fuchi \bgroup\&\ \egroup
  Kott}{1988}]{future-generation}
Kazuhiro Fuchi and Laurent Kott, editors.
\newblock {\em {Programming of Future Generation Computers II: \Proc\ of the
  \nth 2 Franco-Japanese Symposium}}. {\northholland}, 1988.

\bibitem[\protect\citeauthoryear{Gabbay{\protectedgabbayindex} \bgroup\&\
  \egroup Woods}{2004\ff}]{handbook-of-the-history-of-logic}
Dov Gabbay{\protectedgabbayindex} and John Woods, editors.
\newblock {\em Handbook of the History of Logic}.
\newblock \northholland, 2004\ff.

\bibitem[\protect\citeauthoryear{Gabbay{\protectedgabbayindex}
  \bgroup\&al.\egroup }{1994}]{handbooklogicailpvol2}
Dov Gabbay{\protectedgabbayindex}, {\mbox{Ch}}ristopher~John Hogger, and
  J.~Alan Robinson{\protectedrobinsonindex}, editors.
\newblock {\em Handbook of Logic in Artificial Intelligence and Logic
  Programming. {\Vol\,2}: Deduction Methodologies}.
\newblock \oxfordunipress, 1994.

\bibitem[\protect\citeauthoryear{Ganzinger \bgroup\&\ \egroup
  Stuber}{1992}]{GaSt}
Harald Ganzinger and J{\"u}rgen Stuber.
\newblock {\ITP\ by Consistency for First-Order Clauses}.
\newblock 1992.
\newblock In \cite[\PP{441}{462}]{Hotz-Festschrift}. Also in
  \cite[\PP{226}{241}]{thirdCTRSninetytwo}.

\bibitem[\protect\citeauthoryear{Ganzinger}{1996}]{seventhRTAninetysix}
Harald Ganzinger, editor.
\newblock {\em {\theseventhRTAninetysix}}, number 1103 in Lecture Notes in
  Computer Science. {\springerverlag}, 1996.

\bibitem[\protect\citeauthoryear{Ganzinger}{1999}]{sixteenthCADEninetynine}
Harald Ganzinger, editor.
\newblock {\em {\thesixteenthCADEninetynine}}, number 1632 in Lecture Notes in
  Artificial Intelligence. {\springerverlag}, 1999.

\bibitem[\protect\citeauthoryear{Gentzen}{1935}]{gentzen}
Gerhard Gentzen.
\newblock {Untersuchungen \ue ber das logische Schlie\sz en}.
\newblock {\em Mathematische Zeitschrift}, 39:176--210,405--431, 1935.
\newblock Also in \cite[\PP{192}{253}]{logiktexte}. {English} translation in
  \cite{gentzen-collected}.

\bibitem[\protect\citeauthoryear{Gentzen}{1969}]{gentzen-collected}
Gerhard Gentzen.
\newblock {\em The Collected Papers of \gentzenname}.
\newblock {\northholland}, 1969.
\newblock \Ed\ by \szaboname.

\bibitem[\protect\citeauthoryear{Geser}{1995}]{geseraxiomofchoice}
Alfons Geser.
\newblock A principle of non-wellfounded induction.
\newblock 1995.
\newblock In
  \cite[\PP{117}{124}]{Kolloquium-Programmiersprachen-und-Grundlagen-der-Progr%
ammierung}.

\bibitem[\protect\citeauthoryear{Geser}{1996}]{geser-improved-general-path-ord%
er}
Alfons Geser.
\newblock An improved general path order.
\newblock {\em \aaeccname}, 7:469--511, 1996.

\bibitem[\protect\citeauthoryear{Gillman}{1987}]{writing-mathematics}
Leonard Gillman.
\newblock {\em Writing Mathematics Well}.
\newblock The Mathematical Association of America, 1987.

\bibitem[\protect\citeauthoryear{G{\"o}bel}{1985}]{inductionlesshistc}
Richard G{\"o}bel.
\newblock Completion of globally finite term rewriting systems for inductive
  proofs.
\newblock 1985.
\newblock In \cite[\PP{101}{110}]{gwai9}.

\bibitem[\protect\citeauthoryear{G{\oe
  de{\protectedgoedelindex}l}}{1931}]{goedel}
Kurt G{\oe de{\protectedgoedelindex}l}.
\newblock {\Ue ber formal unentscheidbare S\ae tze der \PM\ und verwandter
  Sy\esi teme I}.
\newblock {\em {\monatsheftempname}}, 38:173--198, 1931.
\newblock With {English} translation also in \cite[\Vol\,I,
  \PP{145}{195}]{goedelcollected}. {English} translation also in
  \cite[\PP{596}{616}]{heijenoort-source-book} and in \cite{goedel-meltzer}.

\bibitem[\protect\citeauthoryear{G{\oe
  de{\protectedgoedelindex}l}}{1962}]{goedel-meltzer}
Kurt G{\oe de{\protectedgoedelindex}l}.
\newblock {\em On formally undecidable propositions of {\PM} and related
  systems}.
\newblock Basic Books, \NewYork, 1962.
\newblock English translation of \cite{goedel} by \meltzername. With an
  introduction by {\namefont R. B. Braithwaite}. \nth 2\,\edn\ by Dover
  Publications, 1992.

\bibitem[\protect\citeauthoryear{G{\oe
  de{\protectedgoedelindex}l}}{1986ff.}]{goedelcollected}
Kurt G{\oe de{\protectedgoedelindex}l}.
\newblock {\em Collected Works}.
\newblock {\oxfordunipress}, 1986ff.
\newblock \Ed\ by \fefermanname, \dawsonname\protect\dawsonindex,
  \goldfarbname, \heijenoortname, \kleenename, \parsonsname, \siegname,
  \etalabbrev.

\bibitem[\protect\citeauthoryear{Goguen}{1980}]{gogueninductionlessinduction}
Joseph Goguen.
\newblock How to prove algebraic inductive hypotheses without induction.
\newblock 1980.
\newblock In \cite[\PP{356}{373}]{fifthCADEeighty}.

\bibitem[\protect\citeauthoryear{Goldstein{\protectedgoldsteinindex}}{2008}]{c%
atherine-goldstein-fermat-priceton-companion-math-2008}
Catherine Goldstein{\protectedgoldsteinindex}.
\newblock {\fermatname}.
\newblock 2008.
\newblock In \cite[\litsectref{VI.12},
  \PP{740}{741}]{priceton-companion-math-2008}.

\bibitem[\protect\citeauthoryear{Gordon{\protectedgordonindex}}{2000}]{fromLCF%
toHOL}
Mike J.~C. Gordon{\protectedgordonindex}.
\newblock From {\LCF} to {\HOL}: a short history.
\newblock 2000.
\newblock In \cite[\PP{169}{186}]{honorrobinmilner}.
  \url{http://www.cl.cam.ac.uk/~mjcg/papers/HolHistory.pdf}.

\bibitem[\protect\citeauthoryear{Gore \bgroup\&al.\egroup
  }{2001}]{firstIJCARone}
Rajeev Gore, Alexander Leitsch, and Tobias Nipkow, editors.
\newblock {\em {\thefirstIJCARone}}, number 2083 in Lecture Notes in Artificial
  Intelligence. {\springerverlag}, 2001.

\bibitem[\protect\citeauthoryear{Gowers \bgroup\&al.\egroup
  }{2008}]{priceton-companion-math-2008}
Timothy Gowers, June Barrow-Green, and Imre Leader, editors.
\newblock {\em The {\Princetonnostate} Companion to Mathematics}.
\newblock \princetonunipress, 2008.

\bibitem[\protect\citeauthoryear{Graham \bgroup\&al.\egroup
  }{1976}]{thirdPOPLseventysix}
Susan~L. Graham, Robert~M. Graham, Michael~A. Harrison, William~I. Grosky, and
  Jeffrey~D. Ullman, editors.
\newblock {\em Conference Record of the \nth 3\,\POPLname, Atlanta (GA),
  \Jan\,1976}. \acmpress, 1976.
\newblock \url{http://dl.acm.org/citation.cfm?id=800168}.

\bibitem[\protect\citeauthoryear{Gram{\-}li{\protectedgramlichindex}ch
  \bgroup\&\ \egroup Lindner}{1991}]{unicom}
Bernhard Gram{\-}li{\protectedgramlichindex}ch and Wolfgang Lindner.
\newblock {\em A Guide to {\UNICOM}, an Inductive Theorem Prover Based on
  Rewriting and Completion Techniques}.
\newblock {SEKI-Report SR--91--17 (ISSN 1860--5931)}. {SEKI Publications},
  \FBinfshort, \uniKLshort, 1991.
\newblock
  \url{http://agent.informatik.uni-kl.de/seki/1991/Lindner.SR-91-17.ps.gz}.

\bibitem[\protect\citeauthoryear{Gram{\-}li{\protectedgramlichindex}ch
  \bgroup\&\ \egroup Wirth{\protectedwirthindex}}{1996}]{gwrta}
Bernhard Gram{\-}li{\protectedgramlichindex}ch and Claus-Peter
  Wirth{\protectedwirthindex}.
\newblock Confluence of terminating conditional term rewriting systems
  revisited.
\newblock 1996.
\newblock In \cite[\PP{245}{259}]{seventhRTAninetysix}.

\bibitem[\protect\citeauthoryear{Gram{\-}li{\protectedgramlichindex}ch
  \bgroup\&al.\egroup }{2012}]{sixthIJCARtwelve}
Bernhard Gram{\-}li{\protectedgramlichindex}ch, Dale~A. Miller, and Uli
  Sattler, editors.
\newblock {\em {\thesixthIJCARtwelve}}, number 7364 in Lecture Notes in
  Artificial Intelligence. {\springerverlag}, 2012.

\bibitem[\protect\citeauthoryear{Gram{\-}li{\protectedgramlichindex}ch}{1996}]%
{gramlich-phdthesis96}
Bernhard Gram{\-}li{\protectedgramlichindex}ch.
\newblock {\em Termination and Confluence Properties of Structured Rewrite
  Systems}.
\newblock PhD thesis, \FBinfshort, \uniKLshort, 1996.
\newblock \url{www.logic.at/staff/gramlich/papers/thesis96.pdf}.
  \PPcount{x+217}.

\bibitem[\protect\citeauthoryear{Hayes \bgroup\&al.\egroup
  }{1988}]{machine-intelligence-11}
Jean~E. Hayes, Donald Michie{\protectedmichieindex}, and Judith Richards,
  editors.
\newblock {\em {Proceedings of the \nth{11}\,Annual Machine Intelligence
  Workshop (Machine Intelligence\,11), \Univ\ Strathclyde, \Glasgow, 1985}}.
  \clarendonpress, 1988.
\newblock
  \url{aitopics.org/sites/default/files/classic/Machine_Intelligence_11/Machin%
e_Intelligence_v.11.pdf}.

\bibitem[\protect\citeauthoryear{Heijenoort{\protectedheijenoortindex}}{1971}]%
{heijenoort-source-book}
Jean~van Heijenoort{\protectedheijenoortindex}.
\newblock {\em From {\frege} to {\goedel}: A Source Book in Mathematical Logic,
  1879--1931}.
\newblock {\harvardunipress}, 1971.
\newblock {\nth 2\,\rev\ \edn\ (\nth 1\,\edn\,1967)}.

\bibitem[\protect\citeauthoryear{Herbelin}{2009}]{Coq1}
Hugo Herbelin, editor.
\newblock {\em The {\nth 1} {\COQ} Workshop}. {\Inst\ f\ue r Informatik, \Tech\
  \Univ\ \Muenchen}, 2009.
\newblock TUM-I0919,
  \url{http://www.lix.polytechnique.fr/coq/files/coq-workshop-TUM-I0919.pdf}.

\bibitem[\protect\citeauthoryear{Hilbe{\protectedhilbertindex}rt \bgroup\&\
  \egroup
  Bernays\protect\bernaysindex}{1934}]{grundlagen-first-edition-volume-one}
David Hilbe{\protectedhilbertindex}rt and Paul Bernays\protect\bernaysindex.
\newblock {\em {Die Grundlagen der Mathematik --- Erster Band}}.
\newblock Number~XL in {Die Grundlehren der Mathematischen Wissenschaften in
  Einzeldarstellungen}. {\springerverlag}, 1934.
\newblock \nth 1\,\edn\ (\nth 2\,\edn\ is
  \cite{grundlagen-second-edition-volume-one}). English translation is
  \makeaciteoftwo{grundlagen-german-english-edition-volume-one-one}{grundlagen%
-german-english-edition-volume-one-two}.

\bibitem[\protect\citeauthoryear{Hilbe{\protectedhilbertindex}rt \bgroup\&\
  \egroup
  Bernays\protect\bernaysindex}{1939}]{grundlagen-first-edition-volume-two}
David Hilbe{\protectedhilbertindex}rt and Paul Bernays\protect\bernaysindex.
\newblock {\em {Die Grundlagen der Mathematik --- Zweiter Band}}.
\newblock Number~L in {Die Grundlehren der Mathematischen Wissenschaften in
  Einzeldarstellungen}. {\springerverlag}, 1939.
\newblock \nth 1\,\edn\ (\nth 2\,\edn\ is
  \cite{grundlagen-second-edition-volume-two}).

\bibitem[\protect\citeauthoryear{Hilbe{\protectedhilbertindex}rt \bgroup\&\
  \egroup
  Bernays\protect\bernaysindex}{1968}]{grundlagen-second-edition-volume-one}
David Hilbe{\protectedhilbertindex}rt and Paul Bernays\protect\bernaysindex.
\newblock {\em {Die Grundlagen der Mathematik~I}}.
\newblock Number~40 in {Die Grundlehren der Mathematischen Wissenschaften in
  Einzeldarstellungen}. {\springerverlag}, 1968.
\newblock \nth 2\,\rev\,\edn\,of \cite{grundlagen-first-edition-volume-one}.
  English translation is
  \makeaciteoftwo{grundlagen-german-english-edition-volume-one-one}{grundlagen%
-german-english-edition-volume-one-two}.

\bibitem[\protect\citeauthoryear{Hilbe{\protectedhilbertindex}rt \bgroup\&\
  \egroup
  Bernays\protect\bernaysindex}{1970}]{grundlagen-second-edition-volume-two}
David Hilbe{\protectedhilbertindex}rt and Paul Bernays\protect\bernaysindex.
\newblock {\em {Die Grundlagen der Mathematik~II}}.
\newblock Number~50 in {Die Grundlehren der Mathematischen Wissenschaften in
  Einzeldarstellungen}. {\springerverlag}, 1970.
\newblock \nth 2\,\rev\,\edn\,of \cite{grundlagen-first-edition-volume-two}.

\bibitem[\protect\citeauthoryear{Hilbe{\protectedhilbertindex}rt \bgroup\&\
  \egroup
  Bernays\protect\bernaysindex}{2013a}]{grundlagen-german-english-edition-volu%
me-one-one}
David Hilbe{\protectedhilbertindex}rt and Paul Bernays\protect\bernaysindex.
\newblock {\em {Grundlagen der Mathematik~I --- Foundations of Mathematics~I,
  Part\,A: \whatisinVolIPartA}}.
\newblock \url{http://wirth.bplaced.net/p/hilbertbernays}, 2013.
\newblock Thoroughly \rev\,\nth 2\,\edn\ (\nth 1\,\edn\ \collegepublications,
  2011). First English translation and bilingual facsimile \edn\ of the \nth 2
  German \edn\ \cite{grundlagen-second-edition-volume-one}, \incl\ the
  annotation and translation of all differences of the \nth 1 German \edn\
  \cite{grundlagen-first-edition-volume-one}. Translated and commented by
  \wirthname. \Ed\ by \wirthname, \siekmannname, \michaelgabbayname,
  \gabbayname. Advisory Board: \siegname\ (chair), \anellisname, \awodeyname,
  \baazname, \buchholzname, \buldtname, \kahlename, \mancosuname, \parsonsname,
  \peckhausname, \taitname, \tappname, \zachname.

\bibitem[\protect\citeauthoryear{Hilbe{\protectedhilbertindex}rt \bgroup\&\
  \egroup
  Bernays\protect\bernaysindex}{2013b}]{grundlagen-german-english-edition-volu%
me-one-two}
David Hilbe{\protectedhilbertindex}rt and Paul Bernays\protect\bernaysindex.
\newblock {\em {Grundlagen der Mathematik~I --- Foundations of Mathematics~I,
  Part\,B: \whatisinVolIPartB}}.
\newblock \url{http://wirth.bplaced.net/p/hilbertbernays}, 2013.
\newblock Thoroughly \rev\,\nth 2\,\edn. First English translation and
  bilingual facsimile \edn\ of the \nth 2 German \edn\
  \cite{grundlagen-second-edition-volume-one}, \incl\ the annotation and
  translation of all deleted texts of the \nth 1 German \edn\
  \cite{grundlagen-first-edition-volume-one}. Translated and commented by
  \wirthname. \Ed\ by \wirthname, \siekmannname, \michaelgabbayname,
  \gabbayname. Advisory Board: \siegname\ (chair), \anellisname, \awodeyname,
  \baazname, \buchholzname, \buldtname, \kahlename, \mancosuname, \parsonsname,
  \peckhausname, \taitname, \tappname, \zachname.

\bibitem[\protect\citeauthoryear{Hillenbrand{\protectedhillenbrandindex}
  \bgroup\&\ \egroup L{\oe}chner{\protectedloechnerindex}}{2002}]{HL02}
Thomas Hillenbrand{\protectedhillenbrandindex} and Bernd
  L{\oe}chner{\protectedloechnerindex}.
\newblock The next {\WALDMEISTER} loop.
\newblock 2002.
\newblock In \cite[\PP{486}{500}]{eightteenthCADEtwo}.
  \url{http://www.waldmeister.org}.

\bibitem[\protect\citeauthoryear{Hinchey \bgroup\&\ \egroup
  Bowen}{1999}]{industrial-strength-1999}
Michael~G. Hinchey and Jonathan~P. Bowen, editors.
\newblock {\em Industrial-Strength Formal Methods in Practice}.
\newblock Formal Approaches to Computing and Information Technology (FACIT).
  {\springerverlag}, 1999.

\bibitem[\protect\citeauthoryear{Hobson \bgroup\&\ \egroup
  Love}{1913}]{5-int-congress-mathematicians-1912}
E.~W. Hobson and A.~E.~H. Love, editors.
\newblock {\em {\Proc\ \nth 5 \Int\ Congress of Mathematicians, \Cambridge,
  Aug\,22--28, 1912}}. \cambridgeunipress, 1913.
\newblock \url{http://gallica.bnf.fr/ark:/12148/bpt6k99444q}.

\bibitem[\protect\citeauthoryear{Howard \bgroup\&\ \egroup
  Rubin}{1998}]{weakaxiomofchoice}
Paul Howard and Jean~E. Rubin.
\newblock {\em Consequences of the Axiom of Choice}.
\newblock American Math.\ Society, 1998.

\bibitem[\protect\citeauthoryear{Hudlak \bgroup\&al.\egroup }{1999}]{haskell}
Paul Hudlak, John Peterson, and Joseph~H. Fasel.
\newblock A gentle introduction to {\HASKELL}.
\newblock Web only: \url{http://www.haskell.org/tutorial}, 1999.

\bibitem[\protect\citeauthoryear{Huet \bgroup\&\ \egroup
  Hullot}{1980}]{huethullotinductionlessinduction}
G{\'e}rard Huet and Jean-Marie Hullot.
\newblock Proofs by induction in equational theories with constructors.
\newblock 1980.
\newblock In \cite[\PP{96}{107}]{focs21}. Also in
  \jcssprintyear{1982}{25}{239}{266}.

\bibitem[\protect\citeauthoryear{Huet}{1980}]{huet}
G{\'e}rard Huet.
\newblock Confluent reductions: Abstract properties and applications to term
  rewriting systems.
\newblock {\em \jacmname}, 27:797--821, 1980.

\bibitem[\protect\citeauthoryear{Hunt{\protectedhuntindex} \bgroup\&\ \egroup
  Swords}{2009}]{hunt-swords-2009}
Warren~A. Hunt{\protectedhuntindex} and Sol Swords.
\newblock Centaur technology media unit verification.
\newblock 2009.
\newblock In \cite[\PP{353}{367}]{twentyfirstCAVnine}.

\bibitem[\protect\citeauthoryear{Hunt{\protectedhuntindex}}{1985}]{hunt-1985}
Warren~A. Hunt{\protectedhuntindex}.
\newblock {\em {FM8501}: A Verified Microprocessor}.
\newblock PhD thesis, \unitexasaustin, 1985.
\newblock Also published as \cite{hunt-1994}.

\bibitem[\protect\citeauthoryear{Hunt{\protectedhuntindex}}{1989}]{hunt-1989}
Warren~A. Hunt{\protectedhuntindex}.
\newblock Microprocessor design verification.
\newblock {\em \jarname}, 5:429--460, 1989.

\bibitem[\protect\citeauthoryear{Hunt{\protectedhuntindex}}{1994}]{hunt-1994}
Warren~A. Hunt{\protectedhuntindex}.
\newblock {\em {FM8501}: A Verified Microprocessor}.
\newblock Number 795 in Lecture Notes in Artificial Intelligence.
  {\springerverlag}, 1994.
\newblock Originally published as \cite{hunt-1985}.

\bibitem[\protect\citeauthoryear{Hutter{\protectedhutterindex} \bgroup\&\
  \egroup Bundy{\protectedbundyindex}}{1999}]{inductioncontest}
Dieter Hutter{\protectedhutterindex} and Alan Bundy{\protectedbundyindex}.
\newblock The design of the {CADE-16 Inductive Theorem Prover Contest}.
\newblock 1999.
\newblock In \cite[\PP{374}{377}]{sixteenthCADEninetynine}.

\bibitem[\protect\citeauthoryear{Hutter{\protectedhutterindex} \bgroup\&\
  \egroup Sengler}{1996}]{inkanext}
Dieter Hutter{\protectedhutterindex} and Claus Sengler.
\newblock {\INKA:} the next generation.
\newblock 1996.
\newblock In \cite[\PP{288}{292}]{thirteenthCADEninetysix}.

\bibitem[\protect\citeauthoryear{Hutter{\protectedhutterindex} \bgroup\&\
  \egroup Stephan}{2005}]{siekmann-60}
Dieter Hutter{\protectedhutterindex} and Werner Stephan, editors.
\newblock {\em Mechanizing Mathematical Reasoning: Essays in Honor of\/
  {\siekmannname} on the Occasion of His {\nth{60}\,}Birthday}.
\newblock Number 2605 in Lecture Notes in Artificial Intelligence.
  {\springerverlag}, 2005.

\bibitem[\protect\citeauthoryear{Hutter{\protectedhutterindex}}{1990}]{hutter-%
rippling}
Dieter Hutter{\protectedhutterindex}.
\newblock Guiding inductive proofs.
\newblock 1990.
\newblock In \cite[\PP{147}{161}]{tenthCADEninety}.

\bibitem[\protect\citeauthoryear{Hutter{\protectedhutterindex}}{1994}]{hutter-%
cade-nancy}
Dieter Hutter{\protectedhutterindex}.
\newblock Synthesis of induction orderings for existence proofs.
\newblock 1994.
\newblock In \cite[\PP{29}{41}]{twelvethCADEninetyfour}.

\bibitem[\protect\citeauthoryear{Ireland{\protectedirelandindex} \bgroup\&\
  \egroup Bundy{\protectedbundyindex}}{1994}]{failure-guide-induction}
Andrew Ireland{\protectedirelandindex} and Alan Bundy{\protectedbundyindex}.
\newblock {\em Productive Use of Failure in Inductive Proof}.
\newblock 1994.
\newblock \daireport{716}\@. Also in: \jarprintyear{1996}{16}{79}{111}.

\bibitem[\protect\citeauthoryear{Jamnik \bgroup\&al.\egroup
  }{2003}]{automatic-learning-proof-planning}
Mateja Jamnik, Manfred Kerber, Martin Pollet, and {\mbox{Ch}}ristoph
  Benz{\-}m{\ue}ller{\protectedbenzmuellerindex}.
\newblock Automatic learning of proof methods in proof planning.
\newblock {\em {\ljigplname}}, 11:647--673, 2003.

\bibitem[\protect\citeauthoryear{Jouannaud \bgroup\&\ \egroup
  Kounalis}{1986}]{inductionlesshista}
Jean-Pierre Jouannaud and Emmanu{\ewithtrema}l Kounalis.
\newblock Automatic proofs by induction in equational theories without
  constructors.
\newblock 1986.
\newblock In \cite[\PP{358}{366}]{lics1}. Also in
  \informationandcomputationprintyear{1989}{82}{1}{33}, 1989.

\bibitem[\protect\citeauthoryear{Kaplan \bgroup\&\ \egroup
  Jouannaud}{1988}]{firstCTRSeightyseven}
St{\'e}phane Kaplan and Jean-Pierre Jouannaud, editors.
\newblock {\em {\thefirstCTRSeightyseven}}, number 308 in Lecture Notes in
  Computer Science, 1988.

\bibitem[\protect\citeauthoryear{Kaplan \bgroup\&\ \egroup
  Okada}{1991}]{secondCTRSninety}
St{\'e}phane Kaplan and Mitsuhiro Okada, editors.
\newblock {\em {\thesecondCTRSninety}}, number 516 in Lecture Notes in Computer
  Science, 1991.

\bibitem[\protect\citeauthoryear{Kapur \bgroup\&\ \egroup
  Musser}{1986}]{kapur2}
Deepak Kapur and David~R. Musser.
\newblock Inductive reasoning with incomplete specifications.
\newblock 1986.
\newblock In \cite[\PP{367}{377}]{lics1}.

\bibitem[\protect\citeauthoryear{Kapur \bgroup\&\ \egroup
  Musser}{1987}]{kapur1}
Deepak Kapur and David~R. Musser.
\newblock Proof by consistency.
\newblock {\em \artificialintelligencename}, 31:125--157, 1987.

\bibitem[\protect\citeauthoryear{Kapur \bgroup\&\ \egroup
  Subramaniam}{1996}]{mutualexplicitinduction}
Deepak Kapur and Mahadevan Subramaniam.
\newblock Automating induction over mutually recursive functions.
\newblock 1996.
\newblock In \cite[\PP{117}{131}]{fifthAMASTninetysix}.

\bibitem[\protect\citeauthoryear{Kapur \bgroup\&\ \egroup Zhang}{1989}]{rrl}
Deepak Kapur and Han{\-}tao Zhang.
\newblock An overview of {Rewrite Rule Laboratory~(\/\RRL)}.
\newblock 1989.
\newblock In \cite[\PP{559}{563}]{thirdRTAeightynine}. Journal version is
  \cite{Kapur95}.

\bibitem[\protect\citeauthoryear{Kapur \bgroup\&\ \egroup
  Zhang}{1995}]{Kapur95}
Deepak Kapur and Han{\-}tao Zhang.
\newblock An overview of {Rewrite Rule Laboratory~(\/\RRL)}.
\newblock {\em Computers and Mathematics with Applications}, 29(2):91--114,
  1995.

\bibitem[\protect\citeauthoryear{Kapur}{1992}]{eleventhCADEninetytwo}
Deepak Kapur, editor.
\newblock {\em {\theeleventhCADEninetytwo}}, number 607 in Lecture Notes in
  Artificial Intelligence. {\springerverlag}, 1992.

\bibitem[\protect\citeauthoryear{Katz}{1998}]{katz-history}
Victor~J. Katz.
\newblock {\em A History of Mathematics: An Introduction}.
\newblock \addisonwesley, 1998.
\newblock \nth 2 \edn.

\bibitem[\protect\citeauthoryear{Kaufmann{\protectedkaufmannindex}
  \bgroup\&al.\egroup }{2000a}]{ACLTWO-CASESTUDIES}
Matt Kaufmann{\protectedkaufmannindex}, Panagiotis Manolios, and J~Strother
  Moore{\protect\mooreindex}, editors.
\newblock {\em Computer-Aided Reasoning: {\ACLTWO} Case Studies}.
\newblock Number~4 in Advances in Formal Methods. \kluwer, 2000.
\newblock With a foreword from the series editor {\namefont Mike Hinchey}.

\bibitem[\protect\citeauthoryear{Kaufmann{\protectedkaufmannindex}
  \bgroup\&al.\egroup }{2000b}]{ACLTWO}
Matt Kaufmann{\protectedkaufmannindex}, Panagiotis Manolios, and J~Strother
  Moore{\protect\mooreindex}.
\newblock {\em Computer-Aided Reasoning: An Approach}.
\newblock Number~3 in Advances in Formal Methods. \kluwer, 2000.
\newblock With a foreword from the series editor {\namefont Mike Hinchey}.

\bibitem[\protect\citeauthoryear{Knuth \bgroup\&\ \egroup Bendix}{1970}]{KB70}
Donald~E Knuth and Peter~B. Bendix.
\newblock Simple word problems in universal algebra.
\newblock 1970.
\newblock In \cite[\PP{263}{297}]{leech-1970}.

\bibitem[\protect\citeauthoryear{Kodratoff}{1988}]{eighthECAIeightyeight}
Yves Kodratoff, editor.
\newblock {\em \Proc\ \nth 8 \ECAIname\ (ECAI)}. \pitman, 1988.

\bibitem[\protect\citeauthoryear{Kott}{1986}]{thirteenthICALPeightysix}
Laurent Kott, editor.
\newblock {\em {\nth{13} \ICALPname, Rennes (France)}}, number 226 in Lecture
  Notes in Computer Science. {\springerverlag}, 1986.

\bibitem[\protect\citeauthoryear{Kowalski{\protectedkowalskiindex}}{1974}]{Kow%
74}
Robert~A. Kowalski{\protectedkowalskiindex}.
\newblock Predicate logic as a programming language.
\newblock 1974.
\newblock In \cite[\PP{569}{574}]{IFIP-1974}.

\bibitem[\protect\citeauthoryear{Kowalski{\protectedkowalskiindex}}{1988}]{kow%
alski-1988}
Robert~A. Kowalski{\protectedkowalskiindex}.
\newblock The early years of logic programming.
\newblock {\em \commacmname}, 31:38--43, 1988.

\bibitem[\protect\citeauthoryear{Kreisel}{1965}]{induction-no-cut}
Georg Kreisel.
\newblock Mathematical logic.
\newblock 1965.
\newblock {In \cite[\Vol\,III, \PP{95}{195}]{saaty}}.

\bibitem[\protect\citeauthoryear{K{\"u}chlin}{1989}]{inductionlesshistd}
Wolfgang K{\"u}chlin.
\newblock Inductive completion by ground proof transformation.
\newblock 1989.
\newblock In \cite[\Vol\,2, \PP{211}{244}]{kaci-nivat}.

\bibitem[\protect\citeauthoryear{K{\"u}hler{\protectedkuehlerindex} \bgroup\&\
  \egroup Wirth{\protectedwirthindex}}{1996}]{kwspec}
Ulrich K{\"u}hler{\protectedkuehlerindex} and Claus-Peter
  Wirth{\protectedwirthindex}.
\newblock {\em Conditional Equational Specifications of Data Types with Partial
  Operations for Inductive Theorem Proving}.
\newblock {SEKI-Report SR--1996--11 (ISSN 1437--4447)}. {SEKI Publications},
  \FBinfshort, \uniKLshort, 1996.
\newblock \PPcount{24}, \www\url{/p/rta97}. Short version is \cite{kwspec2}.

\bibitem[\protect\citeauthoryear{K{\"u}hler{\protectedkuehlerindex} \bgroup\&\
  \egroup Wirth{\protectedwirthindex}}{1997}]{kwspec2}
Ulrich K{\"u}hler{\protectedkuehlerindex} and Claus-Peter
  Wirth{\protectedwirthindex}.
\newblock Conditional equational specifications of data types with partial
  operations for inductive theorem proving.
\newblock 1997.
\newblock In \cite[\PP{38}{52}]{eighthRTAninetyseven}. Extended version is
  \cite{kwspec}.

\bibitem[\protect\citeauthoryear{K{\"u}hler{\protectedkuehlerindex}}{1991}]{ku%
ehler-master}
Ulrich K{\"u}hler{\protectedkuehlerindex}.
\newblock {Ein funktionaler und struktureller Vergleich verschiedener
  Induktion\esi beweiser}.
\newblock (English translation of title: ``A functional and structural
  comparsion of several inductive theorem-proving systems'' (\INKA, LP (Larch
  Prover), \NQTHM, \RRL, \UNICOM)). \PPcount{vi+143}, \Diplomarbeit,
  \FBinfshort, \uniKLshort, 1991.

\bibitem[\protect\citeauthoryear{K{\"u}hler{\protectedkuehlerindex}}{2000}]{ku%
ehlerdiss}
Ulrich K{\"u}hler{\protectedkuehlerindex}.
\newblock {\em A Tactic-Based Inductive Theorem Prover for Data Types with
  Partial Operations}.
\newblock \infixverlag, 2000.
\newblock \PhDthesis, \uniKLshort, {ISBN 1586031287}, \www\url{/p/kuehlerdiss}.

\bibitem[\protect\citeauthoryear{Lambert}{1764}]{lambert-1764}
Johann~Heinrich Lambert.
\newblock {\em {Neues Organon oder Gedanken \ue ber die Erforschung und
  Bezeichnung des Wahren und dessen Unterscheidung von Irrthum und Schein.}}
\newblock Johann Wendler, \Leipzig, 1764.
\newblock {\Vol\,I\ (Dianoiologie oder die Lehre von den Gesetzen de\es\
  Denken\es, Alethiologie oder Lehre von der Wahrheit)
  (\url{http://books.google.de/books/about/Neues_Organon_oder_Gedanken_Uber_di%
e_Erf.html?id=ViS3XCuJEw8C}) \& \Vol\,II (Semiotik oder Lehre von der
  Bezeichnung der Gedanken und Dinge, Ph\ae nomenologie oder Lehre von dem
  Schein)
  (\url{http://books.google.de/books/about/Neues_Organon_oder_Gedanken_%C3%BCb%
er_die_Er.html?id=X8UAAAAAcAAj})\@. \ Facsimile reprint by \olmsverlag, 1965,
  with a German introduction by \hanswernerarndtname}.

\bibitem[\protect\citeauthoryear{Lankford}{1980}]{inductionlessinduction1}
Dalla{\es}~S. Lankford.
\newblock Some remarks on inductionless induction.
\newblock Memo MTP-11, \Math\ \Dept, Louisiana \Tech\ \Univ, Ruston (LA), 1980.

\bibitem[\protect\citeauthoryear{Lankford}{1981}]{inductionlessinduction2}
Dalla{\es}~S. Lankford.
\newblock A simple explanation of inductionless induction.
\newblock Memo MTP-14, \Math\ \Dept, Louisiana \Tech\ \Univ, Ruston (LA), 1981.

\bibitem[\protect\citeauthoryear{Lassez \bgroup\&\ \egroup
  Plotkin}{1991}]{honor-robinson}
Jean-Louis Lassez and Gordon~D. Plotkin, editors.
\newblock {\em Computational Logic --- Essays in Honor of\/ {\robinsonname}}.
\newblock \mitpress, 1991.

\bibitem[\protect\citeauthoryear{Leech}{1970}]{leech-1970}
John Leech, editor.
\newblock {\em {Computational Word Problems in Abstract Algebra --- \Proc\ of a
  \Conf\ held at \Oxford, under the auspices of the Science Research Council,
  Atlas Computer Laboratory, \nth{29}\,\Aug\ to \nth 2\,\Sep\,\,1967}}.
  \pergamonpress, 1970.
\newblock With a foreword by {\namefont J. Howlett}.

\bibitem[\protect\citeauthoryear{Leeuwen}{1990}]{handbook-tcs}
Jan~van Leeuwen, editor.
\newblock {\em Handbook of Theoretical Computer \Sci}.
\newblock \mitpress, 1990.

\bibitem[\protect\citeauthoryear{LICS}{1986}]{lics1}
{\em {\Proc\ \nth 1 \LICSname, \CambridgeMassachusetts, 1986}}. \ieeepress,
  1986.
\newblock \licsarchive\url{1986}.

\bibitem[\protect\citeauthoryear{LICS}{1988}]{lics3}
{\em {\Proc\ \nth 3 \LICSname, Edinburgh, 1988}}. \ieeepress, 1988.
\newblock \licsarchive\url{1988}.

\bibitem[\protect\citeauthoryear{LICS}{2007}]{lics22}
{\em {\Proc\ \nth{22} \LICSname, \Breslau, 2007}}. \ieeepress, 2007.
\newblock \licsarchive\url{2007}.

\bibitem[\protect\citeauthoryear{L{\oe}chner{\protectedloechnerindex}}{2006}]{%
loechner-lpo}
Bernd L{\oe}chner{\protectedloechnerindex}.
\newblock Things to know when implementing {LPO}.
\newblock {\em \artificialintelligencetools}, 15:53--79, 2006.

\bibitem[\protect\citeauthoryear{Lusk \bgroup\&\ \egroup
  Overbeek}{1988}]{ninthCADEeightyeight}
Ewing Lusk and Ross Overbeek, editors.
\newblock {\em {\theninthCADEeightyeight}}, number 310 in Lecture Notes in
  Artificial Intelligence. {\springerverlag}, 1988.

\bibitem[\protect\citeauthoryear{Mahoney}{1994}]{fermat-career}
Michael~Sean Mahoney.
\newblock {\em {The Mathematical Career of\/ \fermatnoblename\ 1601--1665}}.
\newblock \princetonunipress, 1994.
\newblock \nth 2\,\rev\,\edn\ (\nth 1\,\edn\,1973).

\bibitem[\protect\citeauthoryear{Marchisotto \bgroup\&\ \egroup
  Smit{\protect\index{Smith, James T.}}h}{2007}]{smith-pieri}
Elena~Anne Marchisotto and James~T. Smit{\protect\index{Smith, James T.}}h.
\newblock {\em The Legacy of {\pieriname} in Geometry and Arithmetic}.
\newblock \birkhaeuser, 2007.

\bibitem[\protect\citeauthoryear{Margaria}{1995}]{Kolloquium-Programmiersprach%
en-und-Grundlagen-der-Programmierung}
Tiziana Margaria, editor.
\newblock {\em {Kolloquium Programmiersprachen und Grundlagen der
  Programmierung}}, 1995.
\newblock \Tech\ Report MIP--9519, \Univ\ Passau.

\bibitem[\protect\citeauthoryear{McCarthy \bgroup\&al.\egroup }{1965}]{LISP}
John McCarthy, Paul~W. Abrahams, D.~J. Edwards, T.~P. Hart, and M.~I. Levin.
\newblock {\em {LISP 1.5} Programmer's Manual}.
\newblock \mitpress, 1965.

\bibitem[\protect\citeauthoryear{McRobbie \bgroup\&\ \egroup
  Slaney}{1996}]{thirteenthCADEninetysix}
Michael~A. McRobbie and John~K. Slaney, editors.
\newblock {\em {\thethirteenthCADEninetysix}}, number 1104 in Lecture Notes in
  Artificial Intelligence. {\springerverlag}, 1996.

\bibitem[\protect\citeauthoryear{Melis \bgroup\&al.\egroup
  }{2008}]{proof_planning_with_multiple_strategies}
Erica Melis, Andreas Meier, and J{\"o}rg Siek{\-}mann{\protect\siekmannindex}.
\newblock Proof planning with multiple strategies.
\newblock {\em \artificialintelligencename}, 172:656--684, 2008.
\newblock Received \May\,2, 2006. Published online \Nov\,22, 2007.
  \url{http://dx.doi.org/10.1016/j.artint.2007.11.004}.

\bibitem[\protect\citeauthoryear{Meltzer{\protectedmeltzerindex} \bgroup\&\
  \egroup Michie{\protectedmichieindex}}{1972}]{machine-intelligence-7}
Bernard Meltzer{\protectedmeltzerindex} and Donald
  Michie{\protectedmichieindex}, editors.
\newblock {\em {Proceedings of the \nth 7\,Annual Machine Intelligence Workshop
  (Machine Intelligence\,7), \EB, 1971}}. \uniEBshort\ Press, 1972.
\newblock
  \url{http://aitopics.org/sites/default/files/classic/Machine%20Intelligence%%
203/Machine%20Intelligence%20v3.pdf}.

\bibitem[\protect\citeauthoryear{Meltzer{\protectedmeltzerindex}}{1975}]{meltz%
er-1975}
Bernard Meltzer{\protectedmeltzerindex}.
\newblock {Department of A.I. -- Univ.\ of \EB}.
\newblock {\em ACM SIGART Bulletin}, 50:5, 1975.

\bibitem[\protect\citeauthoryear{Michie{\protectedmichieindex}}{1968}]{machine%
-intelligence-3}
Donald Michie{\protectedmichieindex}, editor.
\newblock {\em {Proceedings of the \nth 3\,Annual Machine Intelligence Workshop
  (Machine Intelligence\,3), \EB, 1967}}. \uniEBshort\ Press, 1968.
\newblock
  \url{http://aitopics.org/sites/default/files/classic/Machine%20Intelligence%%
203/Machine%20Intelligence%20v3.pdf}.

\bibitem[\protect\citeauthoryear{Milner}{1972}]{milner-1972}
Robin Milner.
\newblock Logic for computable functions --- description of a machine
  interpretation.
\newblock Technical Report Memo\,AIM--169, STAN--CS--72--288, \Dept\ \CS,
  Stanford University, 1972.
\newblock
  \url{ftp://reports.stanford.edu/pub/cstr/reports/cs/tr/72/288/CS-TR-72-288.p%
df}.

\bibitem[\protect\citeauthoryear{Moore{\protect\mooreindex} \bgroup\&al.\egroup
  }{1998}]{moore-lynch-kaufmann-1998}
J~Strother Moore{\protect\mooreindex}, Thomas Lynch, and Matt
  Kaufmann{\protectedkaufmannindex}.
\newblock A mechanically checked proof of the correctness of the kernel of the
  {AMD5K86} floating point division algorithm.
\newblock {\em IEEE Transactions on Computers}, 47:913--926, 1998.

\bibitem[\protect\citeauthoryear{Moore{\protect\mooreindex}}{1973}]{moore-1973}
J~Strother Moore{\protect\mooreindex}.
\newblock {\em Computational Logic: Structure Sharing and Proof of Program
  Properties}.
\newblock PhD thesis, \Dept\ \AI, \uniEBshort, 1973.
\newblock \url{http://hdl.handle.net/1842/2245}.

\bibitem[\protect\citeauthoryear{Moore{\protect\mooreindex}}{1975a}]{moore-197%
5}
J~Strother Moore{\protect\mooreindex}.
\newblock Introducing iteration into the {\protect\PURELISPTP}.
\newblock Technical Report CSL\,74--3, Xerox, Palo Alto Research Center, 3333
  Coyote Hill Rd., Palo Alto (CA), 1975.
\newblock \PPcount{ii+37}, Received \Dec\,1974, \rev\,\Mar\,1975. Short version
  is \cite{moore-1975-short}.

\bibitem[\protect\citeauthoryear{Moore{\protect\mooreindex}}{1975b}]{moore-197%
5-short}
J~Strother Moore{\protect\mooreindex}.
\newblock Introducing iteration into the {\protect\PURELISPTP}.
\newblock {\em \ieeetranssename}, 1:328--338, 1975.
\newblock \url{http://doi.ieeecomputersociety.org/10.1109/TSE.1975.6312857}.
  Long version is \cite{moore-1975}.

\bibitem[\protect\citeauthoryear{Moore{\protect\mooreindex}}{1979}]{moore-1979}
J~Strother Moore{\protect\mooreindex}.
\newblock A mechanical proof of the termination of {\takeuti's} function.
\newblock {\em Information Processing Letters}, 9:176--181, 1979.
\newblock Received \Jul\,13, 1979. \Rev\ \Sep\,5, 1979.
  \url{http://dx.doi.org/10.1016/0020-0190(79)90063-2}.

\bibitem[\protect\citeauthoryear{Moore{\protect\mooreindex}}{1981}]{moore-1981}
J~Strother Moore{\protect\mooreindex}.
\newblock Text editing primitives --- the {TXDT} package.
\newblock Technical Report CSL\,81--2, Xerox, Palo Alto Research Center, 3333
  Coyote Hill Rd., Palo Alto (CA), 1981.

\bibitem[\protect\citeauthoryear{Moore{\protect\mooreindex}}{1989a}]{moore-198%
9-2}
J~Strother Moore{\protect\mooreindex}.
\newblock A mechanically verified language implementation.
\newblock {\em \jarname}, 5:461--492, 1989.

\bibitem[\protect\citeauthoryear{Moore{\protect\mooreindex}}{1989b}]{moore-198%
9-1}
J~Strother Moore{\protect\mooreindex}.
\newblock System verification.
\newblock {\em \jarname}, 5:409--410, 1989.

\bibitem[\protect\citeauthoryear{Moskewicz \bgroup\&al.\egroup }{2001}]{chaff}
Matthew~W. Moskewicz, Conor~F. Madigan, Ying Zhao, Lintao Zhang, and Sharad
  Malik.
\newblock {\sc Chaff}: Engineering an efficient {SAT} solver.
\newblock 2001.
\newblock In \cite[\PP{530}{535}]{dac/2001}.

\bibitem[\protect\citeauthoryear{Musser}{1980}]{musserinductionlessinduction}
David~R. Musser.
\newblock On proving inductive properties of abstract data types.
\newblock 1980.
\newblock In \cite[\PP{154}{162}]{seventhPOPLeighty}.
  \url{http://dl.acm.org/citation.cfm?id=567461}.

\bibitem[\protect\citeauthoryear{Nilsson}{1973}]{ijcai3}
Nils~J. Nilsson, editor.
\newblock {\em {\Proc\ \nth 3 \IJCAIname\ (\IJCAI), \StanfordCA}}. Stanford
  Research Institute, Publications \Dept, \StanfordCA, 1973.
\newblock
  \url{http://ijcai.org/Past%20Proceedings/IJCAI-73/CONTENT/content.htm}.

\bibitem[\protect\citeauthoryear{Odifreddi}{1990}]{odifreddi}
Piergiorgio Odifreddi, editor.
\newblock {\em Logic and Computer Science}.
\newblock \academicpress, 1990.

\bibitem[\protect\citeauthoryear{Padawitz}{1996}]{padawitzjsc}
Peter Padawitz.
\newblock Inductive theorem proving for design specifications.
\newblock {\em \jscname}, 21:41--99, 1996.

\bibitem[\protect\citeauthoryear{Padoa{\protectedpadoaindex}}{1913}]{padoa-191%
3}
Alessandro Padoa{\protectedpadoaindex}.
\newblock {La valeur et les r\^oles du principe d'induction math\'ematique}.
\newblock 1913.
\newblock In \cite[\PP{471}{479}]{5-int-congress-mathematicians-1912}.

\bibitem[\protect\citeauthoryear{Pascal}{1954}]{pascal}
Blaise Pascal.
\newblock {\em {\OE}uvres Compl{\`e}tes}.
\newblock Gallimard, Paris, 1954.
\newblock \Ed\ by {\namefont Jacques Chevalier}.

\bibitem[\protect\citeauthoryear{Paulson}{1990}]{isabellesevenhundred}
Lawrence~C. Paulson.
\newblock {\ISABELLE}: The next 700 theorem provers.
\newblock 1990.
\newblock In \cite[\PP{361}{386}]{odifreddi}.

\bibitem[\protect\citeauthoryear{Paulson}{1996}]{Pau96}
Lawrence~C. Paulson.
\newblock {\em {\ml} for the Working Programmer}.
\newblock \cambridgeunipress, 1996.
\newblock \nth 2\,\edn\ (\nth 1\,\edn\,1991).

\bibitem[\protect\citeauthoryear{Peano{\protect\peanoindex}}{1889}]{peanonovam%
ethodo}
Guiseppe Peano{\protect\peanoindex}.
\newblock {\em Arithmetices principia, novo methodo exposita}.
\newblock Fratelli Bocca, \Torino, 1889.

\bibitem[\protect\citeauthoryear{P{\'e}ter{\protectedpeterindex}}{1951}]{peter%
-1951}
R{\'o}sza P{\'e}ter{\protectedpeterindex}.
\newblock {\em Rekursive Funktionen}.
\newblock Akad.~Kiad\'o, \Budapest, 1951.

\bibitem[\protect\citeauthoryear{Pieri{\protect\pieriindex}}{1908}]{pieri}
Mario Pieri{\protect\pieriindex}.
\newblock Sopra gli assiomi aritmetici.
\newblock {\em Il Bollettino delle seduta della Accademia Gioenia di Scienze
  Naturali in Catania}, Series\,2, 1--2:26--30, 1908.
\newblock Written \Dec\,1907. Received \Jan\,8, 1908. English translation {\em
  On the Axioms of Arithmetic}\/ in \cite[\litsectref{4.2},
  \PP{308}{313}]{smith-pieri}.

\bibitem[\protect\citeauthoryear{Plotkin \bgroup\&al.\egroup
  }{2000}]{honorrobinmilner}
Gordon~D. Plotkin, Colin Stirling, and Mads Tofte, editors.
\newblock {\em Proof, Language, and Interaction, Essays in Honour of
  {\milnername}}.
\newblock \mitpress, 2000.

\bibitem[\protect\citeauthoryear{Presburger{\protect\presburgerindex}}{1930}]{%
presburger}
Moj{\zwithdot}esz Presburger{\protect\presburgerindex}.
\newblock {\Ue ber die Vollst\ae ndigkeit eine\es\ gewissen Sy\esi tem\es\ der
  Arithmetik ganzer Zahlen, in welchem die Addition als einzige Operation
  hervortritt}.
\newblock In {\em Sprawozdanie z I Kongresu metematyk\'ow kraj\'ow s\l
  owianskich, Warszawa 1929 (Comptes-rendus du \frenchnth 1 Congr\`es des
  Math\'ematiciens des Pays Slaves, Varsovie 1929)}, pages 92--101+395, 1930.
\newblock Remarks and {E}nglish translation in
  \cite{presburger-remarks-translation}.

\bibitem[\protect\citeauthoryear{Protzen{\protectedprotzenindex}}{1994}]{protz%
enlazy}
Martin Protzen{\protectedprotzenindex}.
\newblock Lazy generation of induction hypotheses.
\newblock 1994.
\newblock In \cite[\PP{42}{56}]{twelvethCADEninetyfour}.

\bibitem[\protect\citeauthoryear{Protzen{\protectedprotzenindex}}{1995}]{protz%
endiss}
Martin Protzen{\protectedprotzenindex}.
\newblock {\em Lazy Generation of Induction Hypotheses and Patching Faulty
  Conjectures}.
\newblock \infixverlag, 1995.
\newblock \PhDthesis.

\bibitem[\protect\citeauthoryear{Protzen{\protectedprotzenindex}}{1996}]{protz%
enpatching}
Martin Protzen{\protectedprotzenindex}.
\newblock Patching faulty conjectures.
\newblock 1996.
\newblock In \cite[\PP{77}{91}]{thirteenthCADEninetysix}.

\bibitem[\protect\citeauthoryear{Rabinovitch}{1970}]{Ravinovitch-Gershon}
Nachum~L. Rabinovitch.
\newblock Rabbi {\gersonname} and the origins of mathematical induction.
\newblock {\em Archive for History of Exact Sciences}, 6:237--248, 1970.
\newblock Received \Jan\,12, 1970.

\bibitem[\protect\citeauthoryear{Reddy}{1977}]{ijcai5}
Ray Reddy, editor.
\newblock {\em {\Proc\ \nth 5 \IJCAIname\ (\IJCAI), \CambridgeMassachusetts}}.
  \Dept\ of \CS, \CMU, \CambridgeMassachusetts, 1977.
\newblock \url{http://ijcai.org/Past%20Proceedings}.

\bibitem[\protect\citeauthoryear{Reddy}{1990}]{reddy}
Uday~S. Reddy.
\newblock Term rewriting induction.
\newblock 1990.
\newblock \cite[\PP{162}{177}]{tenthCADEninety}.

\bibitem[\protect\citeauthoryear{Riazanov \bgroup\&\ \egroup
  Voronkov}{2001}]{vampire01}
Alexander Riazanov and Andrei Voronkov.
\newblock Vampire~1.1 (system description).
\newblock 2001.
\newblock In \cite[\PP{376}{380}]{firstIJCARone}.

\bibitem[\protect\citeauthoryear{Robinson{\protectedrobinsonindex} \bgroup\&\
  \egroup Voronkow}{2001}]{HandbookAR}
J.~Alan Robinson{\protectedrobinsonindex} and Andrei Voronkow, editors.
\newblock {\em Handbook of Automated Reasoning}.
\newblock {\elsevier}, 2001.

\bibitem[\protect\citeauthoryear{Rosenfeld}{1974}]{IFIP-1974}
Jack~L. Rosenfeld, editor.
\newblock {\em {\Proc\ of the Congress of the \Int\ Federation for Information
  Processing (IFIP), Stockholm (Sweden), \Aug\,5--10, 1974}}. \northholland,
  1974.

\bibitem[\protect\citeauthoryear{Rubin \bgroup\&\ \egroup
  Rubin}{1985}]{axiomofchoice}
Herman Rubin and Jean~E. Rubin.
\newblock {\em Equivalents of the {Axiom of Choice}}.
\newblock {\northholland}, 1985.
\newblock \nth 2\,\rev\,\edn\ (\nth 1\,\edn\,1963).

\bibitem[\protect\citeauthoryear{Rusinowitch \bgroup\&\ \egroup
  Remy}{1993}]{thirdCTRSninetytwo}
Micha{\"e}l Rusinowitch and Jean-Luc Remy, editors.
\newblock {\em {\thethirdCTRSninetytwo}}, number 656 in Lecture Notes in
  Computer Science, 1993.

\bibitem[\protect\citeauthoryear{Russinoff}{1998}]{russinoff-1998}
David~M. Russinoff.
\newblock A mechanically checked proof of {IEEE} compliance of a
  register-transfer-level specification of the {AMD-K7} floating-point
  multiplication, division, and square root instructions.
\newblock {\em London Mathematical Society Journal of Computation and
  Mathematics}, 1:148--200, 1998.

\bibitem[\protect\citeauthoryear{Saaty}{1965}]{saaty}
T.~L. Saaty, editor.
\newblock {\em Lectures on Modern Mathematics}.
\newblock \wiley, 1965.

\bibitem[\protect\citeauthoryear{Schmidt-Samoa{\protectedsamoaindex}}{2006a}]{%
samoacalculemus}
Tobias Schmidt-Samoa{\protectedsamoaindex}.
\newblock An even closer integration of {\index{linear arithmetic}}linear
  arithmetic into inductive theorem proving.
\newblock {\em {\ENTCSname}}, 151:3--20, 2006.
\newblock \www\url{/p/evencloser},
  \url{http://dx.doi.org/10.1016/j.entcs.2005.11.020}.

\bibitem[\protect\citeauthoryear{Schmidt-Samoa{\protectedsamoaindex}}{2006b}]{%
samoa-phd}
Tobias Schmidt-Samoa{\protectedsamoaindex}.
\newblock {\em Flexible Heuristic Control for Combining Automation and
  User-Interaction in Inductive Theorem Proving}.
\newblock PhD thesis, \uniKLshort, 2006.
\newblock \www\url{/p/samoadiss}.

\bibitem[\protect\citeauthoryear{Schmidt-Samoa{\protectedsamoaindex}}{2006c}]{%
jancl}
Tobias Schmidt-Samoa{\protectedsamoaindex}.
\newblock Flexible heuristics for simplification with conditional lemmas by
  marking formulas as forbidden, mandatory, obligatory, and generous.
\newblock {\em \janclname}, 16:209--239, 2006.
\newblock \url{http://dx.doi.org/10.3166/jancl.16.208-239}.

\bibitem[\protect\citeauthoryear{Schoenfield}{1967}]{schoenfield}
Joseph~R. Schoenfield.
\newblock {\em Mathematical Logic}.
\newblock \addisonwesley, 1967.

\bibitem[\protect\citeauthoryear{Scott{\protectedscottindex}}{1993}]{scott-LCF}
Dana~S. Scott{\protectedscottindex}.
\newblock A type-theoretical alternative to {ISWIM}, {CUCH}, {OWHY}.
\newblock {\em \tcsname}, 121:411--440, 1993.
\newblock Annotated version of a manuscript from the year\,1969.
  \url{www.cs.cmu.edu/~kw/scans/scott93tcs.pdf}.

\bibitem[\protect\citeauthoryear{Shankar{\protectedshankarindex}}{1986}]{shank%
ar-1986}
Natarajan Shankar{\protectedshankarindex}.
\newblock {\em Proof-checking Metamathematics}.
\newblock PhD thesis, \unitexasaustin, 1986.
\newblock Thoroughly revised version is \cite{shankar-1994}.

\bibitem[\protect\citeauthoryear{Shankar{\protectedshankarindex}}{1988}]{churc%
h-rosser-bm}
Natarajan Shankar{\protectedshankarindex}.
\newblock A mechanical proof of the {\churchrosser\ theorem}.
\newblock {\em \jacmname}, 35:475--522, 1988.
\newblock Received \May\,1985, \rev\,\Aug\,1987\@. See also \litchapref 6 in
  \cite{shankar-1994}.

\bibitem[\protect\citeauthoryear{Shankar{\protectedshankarindex}}{1994}]{shank%
ar-1994}
Natarajan Shankar{\protectedshankarindex}.
\newblock {\em Metamathematics, Machines, and {\goedel}'s Proof}.
\newblock \cambridgeunipress, 1994.
\newblock Originally published as \cite{shankar-1986}. Paperback reprint\,1997.

\bibitem[\protect\citeauthoryear{Siek{\-}mann{\protect\siekmannindex}}{1986}]{%
eighthCADEeightysix}
J{\"o}rg Siek{\-}mann{\protect\siekmannindex}, editor.
\newblock {\em {\theeighthCADEeightysix}}, number 230 in Lecture Notes in
  Artificial Intelligence. {\springerverlag}, 1986.

\bibitem[\protect\citeauthoryear{Sridharan}{1989}]{ijcai11}
N.~S. Sridharan, editor.
\newblock {\em {\Proc\ \nth{11} \IJCAIname\ (\IJCAI), Detroit (MI)}}.
  \morgankaufmann, 1989.
\newblock \url{http://ijcai.org/Past%20Proceedings}.

\bibitem[\protect\citeauthoryear{Stansifer}{1984}]{presburger-remarks-translat%
ion}
Ryan Stansifer.
\newblock {\presburgerindex\presburger's Article on Integer Arithmetic: Remarks
  and Translation}.
\newblock Technical Report TR\,84--639, \Dept\ of \CS, Cornell \Univ, Ithaca
  (NY), 1984.
\newblock \url{http://hdl.handle.net/1813/6478}.

\bibitem[\protect\citeauthoryear{Steele}{1990}]{commonlisp}
Guy~L. Steele {\jr}.
\newblock {\em {\COMMONLISP} --- The Language}.
\newblock \digitalpress, 1990.
\newblock \nth 2\,\edn\ (\nth 1\,\edn\,1984).

\bibitem[\protect\citeauthoryear{Steinbach}{1988}]{SR--88--12}
Joachim Steinbach.
\newblock {\em Term Orderings With Status}.
\newblock {SEKI-Report SR--88--12 (ISSN 1437--4447)}. {SEKI Publications}, FB
  Informatik, Univ. Kaiserslautern, 1988.
\newblock \PPcount{57}, \urlsreightyeighttwelve.

\bibitem[\protect\citeauthoryear{Steinbach}{1995}]{simplificationorderings}
Joachim Steinbach.
\newblock Simplification orderings --- history of results.
\newblock {\em \fundamentainformaticaename}, 24:47--87, 1995.

\bibitem[\protect\citeauthoryear{Stevens}{1988}]{stevens-rational-reconstructi%
on}
Andrew Stevens.
\newblock {\em A Rational Reconstruction of {\boyer} and {\moore}'s Technique
  for Constructing Induction Formulas}.
\newblock 1988.
\newblock \daireport{360}\@. Also in
  \cite[\PP{565}{570}]{eighthECAIeightyeight}.

\bibitem[\protect\citeauthoryear{Stickel}{1990}]{tenthCADEninety}
Mark~E. Stickel, editor.
\newblock {\em {\thetenthCADEninety}}, number 449 in Lecture Notes in
  Artificial Intelligence. {\springerverlag}, 1990.

\bibitem[\protect\citeauthoryear{Stoyan}{1985}]{gwai9}
Herbert Stoyan, editor.
\newblock {\em {\nth 9 \GWAIname, Dassel (Germany), 1985}}, number 118 in
  Informatik-Fachberichte. {\springerverlag}, 1985.

\bibitem[\protect\citeauthoryear{Toyama}{1988}]{toyama}
Yoshihito Toyama.
\newblock Commutativity of term rewriting systems.
\newblock 1988.
\newblock In \cite[\PP{393}{407}]{future-generation}. \ Also in
  \cite{toyamadiss}.

\bibitem[\protect\citeauthoryear{Toyama}{1990}]{toyamadiss}
Yoshihito Toyama.
\newblock {\em {Term Rewriting Systems and the \churchrosser\ Property}}.
\newblock PhD thesis, Tohoku \Univ\ / Nippon Telegraph and Telephone
  Corporation, 1990.

\bibitem[\protect\citeauthoryear{Unguru}{1991}]{unguru-one}
Sabetai Unguru.
\newblock Greek mathematics and mathematical induction.
\newblock {\em \physisname}, XXVIII(2):273--289, 1991.

\bibitem[\protect\citeauthoryear{Verma}{2005?}]{simonyi-interview}
Shamit Verma.
\newblock Interview with \simonyiname.
\newblock \WWW\ only: \url{http://www.shamit.org/charles_simonyi.htm}, 2005?

\bibitem[\protect\citeauthoryear{Voicu \bgroup\&\ \egroup
  Li}{2009}]{voicu-li-di}
R{\u a}zvan Voicu and Mengran Li.
\newblock {\em\DescenteInfinie} proofs in {\COQ}.
\newblock 2009.
\newblock In \cite[\PP{73}{84}]{Coq1}.

\bibitem[\protect\citeauthoryear{Voronkov}{1992}]{thirdLPARninetytwo}
Andrei Voronkov, editor.
\newblock {\em \Proc\ \nth 3 \LPARname\ (\LPAR)}, number 624 in Lecture Notes
  in Artificial Intelligence. {\springerverlag}, 1992.

\bibitem[\protect\citeauthoryear{Voronkov}{2002}]{eightteenthCADEtwo}
Andrei Voronkov, editor.
\newblock {\em {\theeightteenthCADEtwo}}, number 2392 in Lecture Notes in
  Artificial Intelligence. {\springerverlag}, 2002.

\bibitem[\protect\citeauthoryear{Walther{\protectedwaltherindex}}{1988}]{walth%
ertermination}
{\mbox{Ch}}ristoph Walther{\protectedwaltherindex}.
\newblock Argument-bounded algorithms as a basis for automated termination
  proofs.
\newblock 1988.
\newblock In \cite[\PP{601}{622}]{ninthCADEeightyeight}.

\bibitem[\protect\citeauthoryear{Walther{\protectedwaltherindex}}{1992}]{walth%
erLPAR92}
{\mbox{Ch}}ristoph Walther{\protectedwaltherindex}.
\newblock Computing induction axioms.
\newblock 1992.
\newblock In \cite[\PP{381}{392}]{thirdLPARninetytwo}.

\bibitem[\protect\citeauthoryear{Walther{\protectedwaltherindex}}{1993}]{walth%
erIJCAI93}
{\mbox{Ch}}ristoph Walther{\protectedwaltherindex}.
\newblock Combining induction axioms by machine.
\newblock 1993.
\newblock In \cite[\PP{95}{101}]{ijcai13}.

\bibitem[\protect\citeauthoryear{Walther{\protectedwaltherindex}}{1994a}]{walt%
herhandbook}
{\mbox{Ch}}ristoph Walther{\protectedwaltherindex}.
\newblock Mathematical induction.
\newblock 1994.
\newblock In \cite[\PP{127}{228}]{handbooklogicailpvol2}.

\bibitem[\protect\citeauthoryear{Walther{\protectedwaltherindex}}{1994b}]{walt%
hertermination2}
{\mbox{Ch}}ristoph Walther{\protectedwaltherindex}.
\newblock On proving termination of algorithms by machine.
\newblock {\em \artificialintelligencename}, 71:101--157, 1994.

\bibitem[\protect\citeauthoryear{Wirsing \bgroup\&\ \egroup
  Nivat}{1996}]{fifthAMASTninetysix}
Martin Wirsing and Maurice Nivat, editors.
\newblock {\em \Proc\ \nth 5 \Int\ \Conf\ on Algebraic Methodology and Software
  Technology (AMAST), \Muenchen\ (Germany), 1996}, number 1101 in Lecture Notes
  in Computer Science. {\springerverlag}, 1996.

\bibitem[\protect\citeauthoryear{Wirth{\protectedwirthindex} \bgroup\&\ \egroup
  Becker}{1995}]{wirthbecker}
Claus-Peter Wirth{\protectedwirthindex} and Klaus Becker.
\newblock Abstract notions and inference systems for proofs by mathematical
  induction.
\newblock 1995.
\newblock In \cite[\PP{353}{373}]{fourthCTRSninetyfour}.

\bibitem[\protect\citeauthoryear{Wirth{\protectedwirthindex} \bgroup\&\ \egroup
  Gram{\-}li{\protectedgramlichindex}ch}{1994a}]{wgjsc}
Claus-Peter Wirth{\protectedwirthindex} and Bernhard
  Gram{\-}li{\protectedgramlichindex}ch.
\newblock A constructor-based approach to positive/negative-conditional
  equational specifications.
\newblock {\em {\jscname}}, 17:51--90, 1994.
\newblock \url{http://dx.doi.org/10.1006/jsco.1994.1004}, \www\url{/p/jsc94}.

\bibitem[\protect\citeauthoryear{Wirth{\protectedwirthindex} \bgroup\&\ \egroup
  Gram{\-}li{\protectedgramlichindex}ch}{1994b}]{wgcade}
Claus-Peter Wirth{\protectedwirthindex} and Bernhard
  Gram{\-}li{\protectedgramlichindex}ch.
\newblock On notions of inductive validity for first-order equational clauses.
\newblock 1994.
\newblock In \cite[\PP{162}{176}]{twelvethCADEninetyfour}, \www\url{/p/cade94}.

\bibitem[\protect\citeauthoryear{Wirth{\protectedwirthindex} \bgroup\&\ \egroup
  K{\"u}hler{\protectedkuehlerindex}}{1995}]{SR--95--15}
Claus-Peter Wirth{\protectedwirthindex} and Ulrich
  K{\"u}hler{\protectedkuehlerindex}.
\newblock {\em Inductive Theorem Proving in Theories Specified by \PNC\
  Equations}.
\newblock {SEKI-Report SR--95--15 (ISSN 1437--4447)}. {SEKI Publications},
  Univ.\ Kaiserslautern, 1995.
\newblock \PPcount{iv+126}.

\bibitem[\protect\citeauthoryear{Wirth{\protectedwirthindex}
  \bgroup\&al.\egroup }{1993}]{wgkp}
Claus-Peter Wirth{\protectedwirthindex}, Bernhard
  Gram{\-}li{\protectedgramlichindex}ch, Ulrich
  K{\"u}hler{\protectedkuehlerindex}, and Horst Prote.
\newblock {\em Constructor-Based Inductive Validity in
  Positive/Negative-Conditional Equational Specifications}.
\newblock {SEKI-Report SR--93--05~(SFB) (ISSN 1437--4447)}. {SEKI
  Publications}, \FBinfshort, \uniKLshort, 1993.
\newblock \PPcount{iv+58}, \urlsrninetythreedashzerofive. \Rev\,\extd\,\edn\ of
  \nth 1\,part is \cite{wgjsc}, \rev\,\edn\ of \nth 2\,part is \cite{wgcade}.

\bibitem[\protect\citeauthoryear{Wirth{\protectedwirthindex}
  \bgroup\&al.\egroup }{2009}]{herbrand-handbook}
Claus-Peter Wirth{\protectedwirthindex}, J{\"o}rg Siek{\-}mann,
  {\mbox{Ch}}ristoph Benz{\-}m{\ue}ller{\protectedbenzmuellerindex}, and Serge
  Autexier{\protectedautexierindex}.
\newblock \herbrandname: Life, logic, and automated deduction.
\newblock 2009.
\newblock In \cite[\Vol\,5: Logic from \russell\ to \church,
  \PP{195}{254}]{handbook-of-the-history-of-logic}.

\bibitem[\protect\citeauthoryear{Wirth}{1971}]{wirth-pascal}
Niklaus Wirth.
\newblock The programming language {\Pascal}.
\newblock {\em Acta Informatica}, 1:35--63, 1971.

\bibitem[\protect\citeauthoryear{Wirth{\protectedwirthindex}}{1991}]{wirth-mas%
ter}
Claus-Peter Wirth{\protectedwirthindex}.
\newblock Inductive theorem proving in theories specified by
  posi\-tive/ne\-ga\-tive-condi\-tional equations.
\newblock \Diplomarbeit, \FBinfshort, \uniKLshort, 1991.

\bibitem[\protect\citeauthoryear{Wirth{\protectedwirthindex}}{1997}]{wirthdiss}
Claus-Peter Wirth{\protectedwirthindex}.
\newblock {\em Positive/Negative-Conditional Equations: A Constructor-Based
  Framework for Specification and Inductive Theorem Proving}, volume~31 of {\em
  {Schriftenreihe Forschungsergebnisse zur Informatik}}.
\newblock {Verlag Dr.\ Kova\v c, Hamburg}, 1997.
\newblock \PhDthesis, \uniKLshort, {ISBN 386064551X}, {\www\url{/p/diss}}.

\bibitem[\protect\citeauthoryear{Wirth{\protectedwirthindex}}{2004}]{wirthcard%
inal}
Claus-Peter Wirth{\protectedwirthindex}.
\newblock {\DescenteInfinie\ + Deduction}.
\newblock {\em {\ljigplname}}, 12:1--96, 2004.
\newblock \www\url{/p/d}.

\bibitem[\protect\citeauthoryear{Wirth{\protectedwirthindex}}{2005a}]{zombie}
Claus-Peter Wirth{\protectedwirthindex}.
\newblock History and future of implicit and inductionless induction: Beware
  the old jade and the zombie!
\newblock 2005.
\newblock In \cite[\PP{192}{203}]{siekmann-60}, {\www\url{/p/zombie}}.

\bibitem[\protect\citeauthoryear{Wirth{\protectedwirthindex}}{2005b}]{wirthcon%
fluence}
Claus-Peter Wirth{\protectedwirthindex}.
\newblock {\em Syntactic Confluence Criteria for Positive/Negative-Conditional
  Term Rewriting Systems}.
\newblock {SEKI-Report SR--95--09 (ISSN 1437--4447)}. {SEKI Publications},
  Univ.\ Kaiserslautern, 2005.
\newblock \Rev\,\edn\ \Oct\,2005 (\nth 1\,\edn\,1995), \PPcount{ii+188},
  \url{http://arxiv.org/abs/0902.3614}.

\bibitem[\protect\citeauthoryear{Wirth{\protectedwirthindex}}{2006}]{nonpermut}
Claus-Peter Wirth{\protectedwirthindex}.
\newblock {\em {$\lim$$+$}, {$\delta^+$}, and Non-Permutability of
  {$\beta$}-Steps}.
\newblock {SEKI-Report SR--2005--01 (ISSN 1437--4447)}. {SEKI Publications},
  Saarland Univ., 2006.
\newblock \Rev\,\edn\ \Jul\,2006 (\nth 1\,\edn\,2005), \PPcount{ii+36},
  \url{http://arxiv.org/abs/0902.3635}. Thoroughly improved version is
  \cite{wirth-jsc-non-permut}.

\bibitem[\protect\citeauthoryear{Wirth{\protectedwirthindex}}{2009}]{wirth-jsc}
Claus-Peter Wirth{\protectedwirthindex}.
\newblock Shallow confluence of conditional term rewriting systems.
\newblock {\em {\jscname}}, 44:69--98, 2009.
\newblock \url{http://dx.doi.org/10.1016/j.jsc.2008.05.005}.

\bibitem[\protect\citeauthoryear{Wirth{\protectedwirthindex}}{2010a}]{swp20060%
1}
Claus-Peter Wirth{\protectedwirthindex}.
\newblock {\em Progress in Computer-Assisted Inductive Theorem Proving by
  Human-Orientedness and Descente Infinie?}
\newblock {SEKI-Working-Paper SWP--2006--01 (ISSN 1860--5931)}. {SEKI
  Publications}, Saarland Univ., 2010.
\newblock \Rev\,\edn\,Dec\,2010 (\nth 1\,\edn\,2006), \PPcount{ii+36},
  \url{http://arxiv.org/abs/0902.3294}.

\bibitem[\protect\citeauthoryear{Wirth{\protectedwirthindex}}{2010b}]{fermatsp%
roof}
Claus-Peter Wirth{\protectedwirthindex}.
\newblock {\em A Self-Contained and Easily Accessible Discussion of the Method
  of Descente Infinie and\/ {\fermat}'s Only Explicitly Known Proof by {\it
  Descente Infinie}}.
\newblock {SEKI-Working-Paper SWP--2006--02 (ISSN 1860--5931)}. {SEKI
  Publications}, {DFKI Bremen GmbH, Safe and Secure Cognitive Systems,
  Cartesium, Enrique Schmidt Str.\,5, D--28359 Bremen, Germany}, 2010.
\newblock \Rev\,\ed\,\Dec\,2010, \PPcount{ii+36},
  \url{http://arxiv.org/abs/0902.3623}.

\bibitem[\protect\citeauthoryear{Wirth{\protectedwirthindex}}{2012a}]{wirth-he%
ijenoort}
Claus-Peter Wirth{\protectedwirthindex}.
\newblock {\herbrandsfundamentaltheorem} in the eyes of {\heijenoortname}.
\newblock {\em {\logicauniversalisname}}, 6:485--520, 2012.
\newblock Received \Jan\,12, 2012. Published online \Jun\,22, 2012,
  \url{http://dx.doi.org/10.1007/s11787-012-0056-7}.

\bibitem[\protect\citeauthoryear{Wirth{\protectedwirthindex}}{2012b}]{wirth-js%
c-non-permut}
Claus-Peter Wirth{\protectedwirthindex}.
\newblock {\math{\lim\tight+}, \math{\delta^+}, and \NonPermutability\ of
  \math\beta-Steps}.
\newblock {\em {\jscname}}, 47:1109--1135, 2012.
\newblock Received \Jan\,18, 2011. Published online \Jul\,15, 2011,
  \url{http://dx.doi.org/10.1016/j.jsc.2011.12.035}. More funny version is
  \cite{nonpermut}.

\bibitem[\protect\citeauthoryear{Wirth{\protectedwirthindex}}{2012c}]{wirth-ma%
nifesto-ljigpl}
Claus-Peter Wirth{\protectedwirthindex}.
\newblock Human-oriented inductive theorem proving by descente infinie --- {a
  \nolinebreak manifesto}.
\newblock {\em {\ljigplname}}, 20:1046--1063, 2012.
\newblock Received \Jul\,11, 2011. Published online \Mar\,12, 2012,
  \url{http://dx.doi.org/10.1093/jigpal/jzr048}.

\bibitem[\protect\citeauthoryear{Wirth{\protectedwirthindex}}{2012d}]{boyer-mo%
ore-2012}
Claus-Peter Wirth{\protectedwirthindex}.
\newblock {Unpublished Interview of \boyername\ and \moorename\ at \boyer's
  place in Austin (TX) on Thursday, \Oct\,7}.
\newblock 2012.

\bibitem[\protect\citeauthoryear{Wirth{\protectedwirthindex}}{2013}]{SR--2011-%
-01}
Claus-Peter Wirth{\protectedwirthindex}.
\newblock {\em {A Simplified and Improved Free-Variable Framework for
  \hilbert's epsilon as an Operator of Indefinite Committed Choice}}.
\newblock {SEKI Report SR--2011--01 (ISSN 1437--4447)}. {SEKI Publications},
  {DFKI Bremen GmbH, Safe and Secure Cognitive Systems, Cartesium, Enrique
  Schmidt Str.\,5, D--28359 Bremen, Germany}, 2013.
\newblock \Rev\,\edn\ \Jan\,2013 (\nth 1\,\edn\,2011), \PPcount{ii+65},
  {\url{http://arxiv.org/abs/1104.2444}}.

\bibitem[\protect\citeauthoryear{Wolff}{1728}]{wolff-rationalis}
Christian Wolff.
\newblock {\em Philosophia rationalis sive Logica, methodo scientifica
  pertractata et ad usum scientiarium atque vitae aptata}.
\newblock Rengerische Buchhandlung, \FFM\ \& \Leipzig, 1728.
\newblock \nth 1\,\edn.

\bibitem[\protect\citeauthoryear{Wolff}{1740}]{wolff-rationalis-1740}
Christian Wolff.
\newblock {\em Philosophia rationalis sive Logica, methodo scientifica
  pertractata et ad usum scientiarium atque vitae aptata}.
\newblock Rengerische Buchhandlung, \FFM\ \& \Leipzig, 1740.
\newblock \nth 3\,\extd\,\edn\ of \cite{wolff-rationalis}\@. \ Facsimile
  reprint by \olmsverlag, 1983, with a French introduction by {\namefont\jean\
  \'Ecole}.

\bibitem[\protect\citeauthoryear{Yeh \bgroup\&\ \egroup
  Ramamoorthy}{1976}]{secondICSEseventysix}
Raymond~T. Yeh and C.~V. Ramamoorthy, editors.
\newblock {\em \Proc\ \nth 2 \Int\ \Conf\ on Software Engineering, San
  Francisco (CA), \Oct\,13--15, 1976}. \ieeeCSpress, 1976.
\newblock \url{http://dl.acm.org/citation.cfm?id=800253}.

\bibitem[\protect\citeauthoryear{Young}{1989}]{young-1989}
William~D. Young.
\newblock A mechanically verified code generator.
\newblock {\em \jarname}, 5:493--518, 1989.

\bibitem[\protect\citeauthoryear{Zhang \bgroup\&al.\egroup }{1988}]{ZKK88}
Han{\-}tao Zhang, Deepak Kapur, and Mukkai~S. Krishnamoorthy.
\newblock A mechanizable induction principle for equational specifications.
\newblock 1988.
\newblock In \cite[\PP{162}{181}]{ninthCADEeightyeight}.

\bibitem[\protect\citeauthoryear{Zygmunt}{1991}]{presburger-life}
Jan Zygmunt.
\newblock \presburgername: Life and work.
\newblock {\em History and Philosophy of Logic}, 12:211--223, 1991.

\end{thebibliography}
